%% file: iclr2020_conference.tex
\documentclass{article} % For LaTeX2e
\usepackage{iclr2020_conference,times}
\usepackage{textcomp}

\input{commands/math_commands.tex}

\usepackage{flexisym}
\usepackage{hyperref}
\usepackage{url}
\usepackage[pdftex]{graphicx}
\usepackage{caption}
\usepackage{subcaption}

\usepackage{pgfplots}
\pgfplotsset{compat=newest}
\usepgfplotslibrary{groupplots}
\usepgfplotslibrary{dateplot}
\newcommand{\quotes}[1]{``#1''}
\usepackage{pdflscape}

\def \effectiveAttnPlotHeight {4.5cm}
\def \tokenClassifierPlotHeight {4.5cm}
\def \contributionPlotsHeightFirst {4cm}

\def \localKernelPlotsHeight {4.8cm}

\def \tokenClassifierPlotHeightAppendix {5.5cm}

\title{On Identifiability in Transformers}

\iclrfinalcopy

\author{Gino Brunner$^{1*}$,  Yang Liu$^{2*}$, Dami\'an Pascual$^{1}$\thanks{Equal contribution with authors in alphabetical order. Yang Liu initiated the transformer models study, perceived and performed the study of attention identifiability and effective attention, i.e., Section 3 and Appendix A, and contributed to the token attribution discussions and calculations.} , Oliver Richter$^1$,\\ {\bf Massimiliano Ciaramita$^3$, Roger Wattenhofer$^1$}
 \\
Departments of~$^1$Electrical Engineering and Information Technology,~$^2$Computer Science\\
ETH Zurich, Switzerland\\
$^3$Google Research, Zurich, Switzerland \\
$^1$\texttt{\{brunnegi,dpascual,richtero,wattenhofer\}@ethz.ch},\\
$^2$\texttt{liu.yang@alumni.ethz.ch} \\
$^3$\texttt{massi@google.com} 
}

\newif\ifcomment\commenttrue

\ifcomment
\newcommand{\pinaforecomment}[3]{\colorbox{#1}{\parbox{.8\linewidth}{#2: #3}}}
\else
\newcommand{\pinaforecomment}[3]{}
\fi

\iclrfinalcopy % Uncomment for camera-ready version, but NOT for submission.
\begin{document}

\maketitle

\begin{abstract}
In this paper we delve deep in the Transformer architecture by investigating two of its core components: self-attention and contextual embeddings. In particular, we study the identifiability of attention weights and token embeddings, and the aggregation of context into hidden tokens. We show that, for sequences longer than the attention head dimension, attention weights are not identifiable. We propose \emph{effective attention} as a complementary tool for improving explanatory interpretations based on attention. Furthermore, we show that input tokens retain to a large degree their identity across the model. We also find evidence suggesting that identity information is mainly encoded in the angle of the embeddings and gradually decreases with depth. Finally, we demonstrate strong mixing of input information in the generation of contextual embeddings by means of a novel quantification method based on gradient attribution. Overall, we show that self-attention distributions are not directly interpretable and present tools to better understand and further investigate Transformer models.   
\end{abstract}

\input{sections/introduction}

\input{sections/background}

\input{sections/attention}

\input{sections/context.tex}

\input{sections/related.tex}

\input{sections/conclusion.tex}

\clearpage

\bibliography{iclr2020_conference}
\bibliographystyle{iclr2020_conference}

\newpage
\appendix

\input{sections/appendix.tex}

\end{document}

%% file: commands/math_commands.tex
\usepackage{amsmath,amsfonts,bm}

\def\eqref#1{equation~\ref{#1}}
\def\1{\bm{1}}

\def\vzero{{\bm{0}}}
\def\vone{{\bm{1}}}

\def\va{{\bm{a}}}

\def\ve{{\bm{e}}}

\def\vx{{\bm{x}}}

\def\mA{{\bm{A}}}
\def\mB{{\bm{B}}}
\def\mC{{\bm{C}}}

\def\mE{{\bm{E}}}

\def\mH{{\bm{H}}}

\def\mK{{\bm{K}}}

\def\mQ{{\bm{Q}}}

\def\mT{{\bm{T}}}

\def\mV{{\bm{V}}}
\def\mW{{\bm{W}}}
\def\mX{{\bm{X}}}

\DeclareMathAlphabet{\mathsfit}{\encodingdefault}{\sfdefault}{m}{sl}
\SetMathAlphabet{\mathsfit}{bold}{\encodingdefault}{\sfdefault}{bx}{n}

\newcommand{\R}{\mathbb{R}}

%% file: sections/introduction.tex
\section{Introduction}
In this paper we investigate neural models of language based on self-attention by concentrating on the concept of \emph{identifiability}. Intuitively, identifiability refers to the ability of a model to learn stable representations. This is arguably a desirable property, as it affects the replicability and interpretability of the model's predictions.
Concretely, we focus on two aspects of identifiability. The first is related to \emph{structural identifiability}~\citep{bellman-astrom-1970}: the theoretical possibility (a priori) to learn a unique optimal parameterization of a statistical model. From this perspective, we analyze the  identifiability of attention weights, what we call \emph{attention identifiability}, in the self-attention components of transformers~\citep{attentionIsAllYouNeed}, one of the most popular neural architectures for language encoding and decoding.
We also investigate \emph{token identifiability}, as the fine-grained, word-level mappings between input and output generated by the model.
The role of attention as a means of recovering input-output mappings, and various types of explanatory insights, is currently the focus of much research and depends to a significant extent on both types of identifiability. 

We contribute the following findings to the ongoing work:
With respect to attention indentifiability, in Section~\ref{sec:ai}, we show that -- under mild conditions with respect to input sequence length and attention head dimension -- the attention weights for a given input are not identifiable. This implies that there can be infinitely many different attention weights that yield the same output. This finding challenges the direct interpretability of attention distributions. As a supplement, we propose the concept of \emph{effective attention}, a diagnostic tool that examines attention weights for model explanations by removing the weight components that do not influence the model's predictions.

With respect to token identifiability, in Section~\ref{IdenTok}, we devise an experimental setting where we probe the hypothesis that contextual word embeddings maintain their identity as they pass through successive layers of a transformer. This is an assumption made in much current research, which has not received a clear validation yet.
Our findings give substance to this assumption, although it does not always hold in later layers. Furthermore, we show that the identity information is largely encoded in the angle of the embeddings and that it can be recovered by a nearest neighbour lookup after a learned linear mapping from hidden to input token space.

In Section~\ref{sec:attributionPart} we further investigate the contribution of all input tokens in the generation of the contextual embeddings in order to quantify the mixing of token and context information. We introduce \emph{Hidden Token Attribution}, a quantification method based on gradient attribution. We find that self-attention strongly mixes context and token contributions. Token contribution decreases monotonically with depth, but the corresponding token typically remains the largest individual contributor. We also find that, despite visible effects of long term dependencies, the context aggregated into the hidden embeddings is mostly local. We notice how, remarkably, this must be an effect of learning.

%% file: sections/background.tex
\section{Background on Transformers} 
The Transformer~\citep{attentionIsAllYouNeed} is currently the neural architecture of choice for natural language processing (NLP). At its core it consists of several multi-head self-attention layers. In these layers, every token of the input sequence attends to all other tokens by projecting its embedding to a query, key and value vector. 
Formally, let $\mQ\in\R^{d_s\times d_q}$ be the query matrix, $\mK\in\R^{d_s\times d_q}$ the key matrix and $\mV\in\R^{d_s\times d_v}$ the value matrix, where $d_s$ is the sequence length and $d_q$ and $d_v$ the dimension of query and value vectors, respectively. The output of an attention head is given by:
\begin{equation}\label{eq:attention}
    \text{Attention}(\mQ,\mK,\mV) = \mA\cdot \mV\qquad \text{with}\quad \mA = \text{softmax}\left(\frac{\mQ\mK^T}{\sqrt{d_q}}\right)
\end{equation} 
The attention matrix $\mA\in\mathbb{R}^{d_s\times d_s}$ calculates, for each token in the sequence, how much the computation of the hidden embedding at this sequence position should be influenced by each of the other (hidden) embeddings. 
Self-attention is a non-local operator, which means that at any layer a token can attend to all other tokens regardless of the distance in the input. Self-attention thus produces so-called \emph{contextual word embeddings}, as successive layers gradually aggregate contextual information into the embedding of the input word.

We focus on a Transformer model called BERT~\citep{BERT}, although our analysis can be easily extended to other models such as GPT, ~\citep{GPT,GPT2} RoBERTa~\citep{roberta}, XLNet ~\citep{xlnet}, or ALBERT~\citep{albert}.
BERT operates on input sequences of length $d_s$. We denote input tokens in the sentence as $\vx_i$, where $i\in [1,...,d_s]$. We use $\vx_i\in\R^d$ with embedding dimension $d$ to refer to the sum of the token-, segment- and position embeddings 
corresponding to the input word at position $i$. We denote the contextual embedding at position $i$ and layer $l$ as $\ve_i^l$. Lastly, we refer to the inputs and embeddings of all sequence positions as matrices $\mX$ and $\mE$, respectively, both in $\R^{d_s\times d}$.
For all experiments we use the pre-trained uncased BERT-Base model as provided by \cite{BERT}\footnote{\url{https://github.com/google-research/bert}}.

%% file: sections/attention.tex
\section{Attention Identifiability}
\label{sec:ai}
We begin with the identifiability analysis of self-attention weights.
Drawing an analogy with structural identifiability~\citep{bellman-astrom-1970}, we state that the attention weights of an attention head for a given input 
are \emph{identifiable}
if they can be uniquely determined from the head's output.\footnote{Cf. Appendix~\ref{app:structural_identif_background} for more background on indentifiability.}
We emphasize that attention weights are input dependent and \emph{not model parameters}. However, their identifiability affects the interpretability of the output, i.e., whether attention weights can provide the basis for explanatory insights on the model's predictions 
(cf.~\citet{attentionIsNotExplanation} and \citet{attentionisNotNotexplanation}). If attention is not identifiable, explanations based on attention may be unwarranted.

The output of a multi-head attention layer is the summation over each of the $h$ single head outputs (cf. Eq. \ref{eq:attention}) multiplied by the matrix $\mH \in \mathbb{R}^{d_v \times d}$ with reduced head dimension $d_v=d/h$,
\begin{align}
 \text{Attention}(\mQ,\mK, \mV)\mH
&=   \mA   \mE  \mW^{V} \mH 		\label{eq:mha:wvh}
=  \mA  \mT 
\end{align}

where $\mW^{V} \in \mathbb{R}^{d \times d_v}$ projects the embedding $\mE$ into the value matrix $\mV=\mE \mW^{V}$, and we define $\mT =\mE \mW^{V} \mH$. Here, the layer and head indices are omitted for simplicity, since the proof below is valid for each individual head and layer in Transformer models.
Intuitively, the head output is a linear combination of the $\mT$ vectors using the attention as weighting coefficients. If the sequence length, i.e. the number of weighting coefficient, is larger than the rank of $\mT$, attention weights are not uniquely determined from the head output; i.e., they include free variables. In other words, some of the $\mT$ rows are linear combinations of others.
We now prove, by analyzing the null space dimension of $\mT$, that attention weights are not identifiable using the head or final model output. 

\subsection{Upper Bound for $\text{rank}(T)$}
We first derive the upper bound of the rank of matrix $\mT=\mE \mW^{V}\mH$. Note that $\text{rank}(\mA\mB\mC)\leq \text{min} \left( \text{rank}(\mA), \text{rank}(\mB), \text{rank}(\mC) \right)$, therefore,
\begin{align}
\text{rank} \left(\mT\right) 
& \leq \text{min}\left(\text{rank}(\mE), \text{rank} (\mW^{V}), \text{rank}(\mH)\right)\\\nonumber
& \leq \text{min}(d_s, d, d,d_v, d_v, d) \\\nonumber
& = \text{min}\left(d_s,d_v \right).\nonumber
\end{align}
The second step holds since $\text{rank}(\mE) \leq \text{min}(d_s, d)$, $\text{rank}(\mW^{V}) \leq \text{min}(d, d_v)$ and $\text{rank}(\mH) \leq \text{min}(d_v, d)$.

\subsection{The null space of $T$}
\label{sect:null-t}
The (left) null space $\text{LN}(\mT)$ of $\mT$ describes all vectors that are mapped to the zero vector by $\mT$:
\begin{align}
 \text{LN}(\mT)=\{ \tilde{\vx}^T  \in \mathbb{R}^{1 \times d_s}| \tilde{\vx}^T  \mT = \vzero \} \label{eq:leftnull}
\end{align}
Its special property is that, for $\tilde{\mA} = [\tilde{\vx}_1, \tilde{\vx}_2, ..., \tilde{\vx}_{d_s}]^T$ where $\tilde{\vx}_i^T$ are vectors in this null space,
\begin{align}
(\mA+\tilde{\mA})\mT=\mA\mT. \label{eq:nullEq}
\end{align}
If the dimension of $\text{LN}(\mT)$ is not zero, there exist infinitely many attention weights $\mA+\tilde{\mA}$ yielding the exact same attention layer output and final model outputs.
By applying the Rank-Nullity theorem, the dimension of the null space is:
\begin{align}
\text{dim} ( \text{LN}(\mT) ) &= d_s-\text{rank} \left(\mT\right) \geq
d_s-\text{min}\left(d_s,d_v \right) %\\
=
\begin{cases}
d_s-d_v, & \text{if } d_s > d_v\\
0, & \text{otherwise}
\end{cases} \label{eq:null_dim}
\end{align}
here we make use of the facts that $\text{dim}(\text{LN}(\mT))=\text{dim}(\text{N}(\mT^T))$ and  $\text{rank}(\mT)=\text{rank}(\mT^T)$ where $\text{N}(\mT)$ represents the null space of a matrix $\mT$.
Equality holds if $\mE$, $\mW^V$ and $\mH$ are of full rank and their matrix product does not bring further rank reductions. 

Hence, when the sequence length is larger than the attention head dimension ($d_s>d_v$), self-attention is not unique. Furthermore, the null space dimension increases with the sequence length.
In the next section we show that the attention weights are also non-identifiable; i.e., the non-trivial null space of $\mT$ exists, even  when the weights are constrained within the probability simplex. 

\subsection{The null space with probability constraints}
Since $\mA$ is the result of a softmax operation, its rows are constrained within the probability simplex: $\mA\geq 0$ (element-wise), and $\mA\vone=\vone$, where $\vone \in \mathbb{R}^{d_s}$ is the vector of all ones. However, the derivation in Section~(\ref{sect:null-t}) does not take these constraints into account. It shows that $\mA$ is not unique, but it does not prove that alternative attention weights exist within the probability simplex, and thus that $\mA$ is not identifiable. Below, we show that $\tilde{\mA}$ exists in $\text{LN}(\mT)$ also when constraining the weights of $(\mA+\tilde{\mA})$ to the probability simplex.

For the row vectors from an alternative attention matrix $\mA+\tilde{\mA}$ to be valid distributions, we require, in addition to $\mA\geq 0$ (element-wise) and $\mA\vone=\vone$, that $\mA+\tilde{\mA}\geq 0$ (element-wise), and $\mA\vone+\tilde{\mA}\vone=\vone$. Furthermore, $\tilde{\mA}$ must be in the (left) null space of $\mT$. We formalize the three conditions below:
\begin{eqnarray}
\text{a})~\tilde{\mA}\mT=\vzero & \text{b})~\tilde{\mA}\vone=\vzero & \text{c})~\tilde{\mA}\geq-\mA\label{eq:constraints}
\end{eqnarray}
Conditions (\ref{eq:constraints}a) and (\ref{eq:constraints}b) can be combined as $\mA[\mT,\vone]=\vzero$, where $[\mT,\vone]$ is the augmented matrix resulting from adding a column vector of ones to $\mT$. Reusing the argument presented in Section~(\ref{sect:null-t}), the dimension of its (left) null space is: $\text{dim}(\text{LN}([\mT,\vone])) \geq \max(d_s-d_v-1, 0)$. Thus, for $d_s-d_v>1$, the null space of $[\mT,\vone]$ exists: it is a linear subspace of $\text{LN}(\mT)$.

We now prove that condition~(\ref{eq:constraints}c) can also be satisfied. We begin by providing an intuitive justification.
The condition restricts the space of $\tilde{\mA}$ from $\text{LN}([\mT,\vone])$ to be a bounded region which could be different for each row vector $\va=(a_1,a_2,\ldots)$ of $\mA$. The null space $\text{LN}([\mT,\vone])$ contains $\vzero$, defining a surface passing through the origin. Since $\va$ is a probability vector, resulting from a softmax transformation, each of its components is strictly positive, i.e., $\mA>0$ (element-wise). Hence, there exists $\epsilon > 0$ such that any point $\tilde{\va}$ in the sphere centered at the origin with radius $\epsilon$ will satisfy condition~(\ref{eq:constraints}c), $\tilde{\va}>-\va$. Crucially, this sphere intersects the null space, as they share the origin. Any point in this intersection satisfies all three conditions in~(\ref{eq:constraints}). 

Formally, the construction of the null space vector $\tilde{\va}$ for the alternative attention weights $\tilde{\va}+\va$ goes as follows. For a vector $\tilde{\va}=(\tilde{a}_1, \tilde{a}_2,...) \in \text{LN}([\mT,\vone])$, to ensure one of its negative components $\tilde{a}_i < 0$ satisfying condition~(\ref{eq:constraints}c) that $\tilde{a}_i \ge - a_i$, one could shrink its magnitude into $\lambda \tilde{\va}$ with $0 \leq \lambda \leq -a_i/\tilde{a}_i$, so $\lambda \tilde{a_i} \ge -a_i$. Considering all negative components $i$, the overall scaling factor is $\lambda_{max} = \min_{i \in \{i | \tilde{a}_i<0 \} } (-a_i/\tilde{a}_i)$ so that the direction from the origin $\{ \lambda \tilde{\va} | 0 \leq \lambda \leq \lambda_{max} \}$ satisfies condition~(\ref{eq:constraints}c). Here $\lambda_{max}$ is strictly greater than $0$ because $a_i>0$. Only when there exists an index $i$ that $\tilde{a}_i < 0$ and $a_i \approx 0$, then this particular null space direction $\tilde{\va}$ is highly confined. In the extreme case, where $\va$ is a one-hot distribution, the solution should be an $\tilde{\va}$ with only one negative component. If such an $\tilde{\va}$ does not exist in $\text{LN}([\mT,\vone])$, the solution collapses to the trivial single point $\tilde{\va}=\vzero$. However, in general, $\text{LN}([\mT,\vone])$ with probability constraints is non-trivial.

\subsection{Effective attention}
The non-identifiability of self-attention, due to the existence of the non-trivial null space of $\mT$, challenges the interpretability of attention weights. However, one can decompose attention weights $\mA$ into the component in the null space $\mA^\parallel$ and the component orthogonal to the null space $\mA^{\perp}$:
\begin{align}
   \mA\mT=(\mA^\parallel + \mA^\perp )\mT = \mA^\perp \mT
\end{align}
since $\mA^\parallel \in \text{LN}(\mT) \implies  \mA^\parallel \mT=\vzero$. 
Hence, we propose a novel concept named \emph{effective attention}, 
\begin{align}
\mA^\perp=\mA-\text{Projection}_{\text{LN}(T)} \mA, 
\end{align}
which is the part of the attention weights that actually affects the model output. The null space projection is calculated by projecting attention weights into the left null space basis, i.e., the associated left singular vectors.

Here the definition of effective attention uses $\text{LN}(\mT)$ instead of the null space $\text{LN}([\mT, 1])$ with probability constraints. As a consequence, effective attention is not guaranteed to be a probability distribution; e.g., some of the weights might be negative. One could define effective attention as the minimal norm alternative attention using $\text{LN}([\mT,\vone])$, or possibly constrain it within the probability simplex. However, in this case, the minimal norm alternative attention is not orthogonal to the null space $\text{LN}(\mT)$ anymore. It seems unclear how to interpret the minimal norm alternative attention. The reason being that the distinction between components that affect or do not affect the output computation\footnote{A key aspect of the concept of identifiability.} are defined with respect to $\text{LN}(\mT)$ and not with respect to $\text{LN}([\mT,\vone])$. In fact, there may be useful information in the sign of the effective attention components. Although the combination of value vectors in the transformer architecture uses weights in the probability simplex, these probability constraints on attention may not be necessary. Hence, we provide here the base version of an effective attention, and leave the investigation of other formulations for future research.

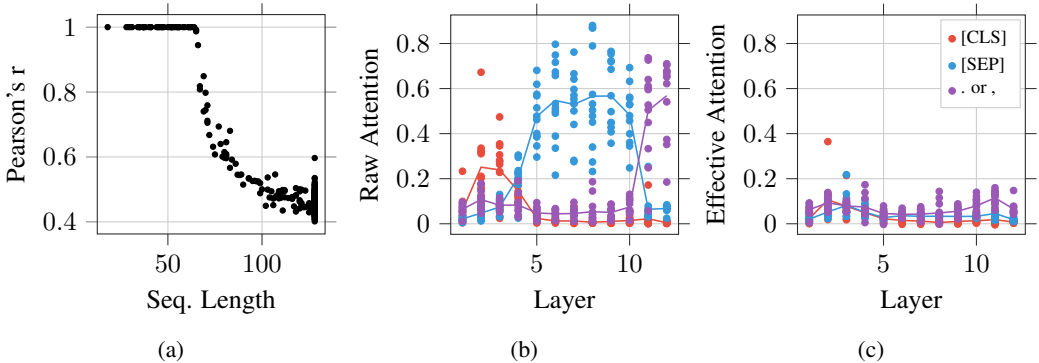
\begin{figure}[t]
\begin{subfigure}[t]{.33\linewidth}
\input{identifiability/Pearson.tex}
\caption{}
\label{subfig:effective_attn_pearson}
\end{subfigure}
\begin{subfigure}[t]{.33\linewidth}
\input{identifiability/avg_attn1_org.tex}
\caption{}
\label{subfig:effective_attn_raw}
\end{subfigure}%\hfill
\begin{subfigure}[t]{.33\linewidth}
\input{identifiability/avg_attn1.tex}
\caption{}
\label{subfig:effective_attn_effective}
\end{subfigure}%\hfill
\caption{
(a) Each point represents the Pearson correlation coefficient of effective attention and raw attention as a function of token length. (b) Raw attention vs. (c) effective attention, where each point represents the average (effective) attention of a given head to a token type. 
}
\label{fig:effective_attn}
\end{figure}

\subsection{Empirical evidence}
We conclude by providing some initial empirical evidence in support of the notion that effective attention can serve as a complementary diagnostic tool for examining how attention weights influence model outputs.

First, we show that effective attention can be detected, and can diverge significantly from raw attention. In Figure~\ref{subfig:effective_attn_pearson}, we illustrate how the Pearson correlation between effective and raw attention decreases with sequence length. We use the same Wikipedia samples as in ~\cite{whatBertLooksAt} with maximum sequence length 128. This result is in line with our theoretical finding in Eq.~\ref{eq:null_dim} that states an increase in the dimension of the null space with the sequence length. Given a bigger null space, more of the raw attention becomes irrelevant, yielding a lower correlation between effective and raw attention. Notice how, for sequences with fewer than $d_v=64$ tokens, the associated null space dimension is zero, and hence attention and effective attention are identical (Pearson correlation of value 1). This loss of correlation with increased sequence length questions the use of attention as explanation in practical models, where it is not uncommon to use large
sequence lengths. A few examples include: BERT for question answering~\citep{chris-a} and XL-Net~\citep{yang2019xlnet} with $d_s=512$, or document translation~\citep{junczysdowmunt:2019:WMT} with $d_s=1000$.

To illustrate the point further, Figure~\ref{fig:effective_attn} shows in (b) raw attention $\mA$ and (c) effective attention $\mA^\perp$, using again the data of~\cite{whatBertLooksAt}. We compute the average attention of BERT and compare it to the corresponding average effective attention. \cite{whatBertLooksAt} conclude that the [CLS] token attracts more attention in early layers, the [SEP] tokens attract more in middle layers, and periods and commas do so in deep layers. However, after a gradient based investigation, they propose that attention to the [SEP] token is generally a \quotes{no-op}. The effective attention weights suggest a more consistent pattern: while periods and commas seem to generally attract more attention than [CLS] and [SEP], the peak of [SEP] token observed by raw attention has disappeared. Effective attention provides an explanation: the [SEP] token peak is irrelevant to the computation of the output for middle layers; i.e., it is in the null space component of the corresponding attention vector.
The same arguments also hold for the sharp peak of raw attention on punctuation tokens between layers 10 and 12. An additional example showing similar results can be found in Appendix~\ref{app:additional_effective_attention_vs_raw_attention}. See also~Appendix~\ref{app:additional_effective_attention_depth}, where we discuss in more depth a case where effective attention would support interpretive conclusions that differ from those one might draw solely based on raw attention.
In conclusion, effective attention can help discover interesting interactions encoded in the attention weights which may be otherwise obfuscated by the null attention.

%% file: identifiability/Pearson.tex
% This file was created by tikzplotlib v0.8.2.
\begin{tikzpicture}

\begin{axis}[
axis line style={white!15.0!black},
legend cell align={left},
legend style={draw=white!80.0!black},
tick align=outside,
tick pos=both,
width=\linewidth,
height=\effectiveAttnPlotHeight,
x grid style={white!80.0!black},
xlabel={Seq. Length},
xmajorgrids,
xtick style={color=white!15.0!black},
y grid style={white!80.0!black},
ylabel={Pearson's r},
ymajorgrids,
xmin=12.4943092715402, xmax=133.50569072846,
ymin=0.362720575263314, ymax=1.03869160532082,
]
\addplot [only marks, draw=black,mark size=1pt, fill=black, colormap/viridis]
table{%
x                      y
128 0.481822345695529
128 0.458645872109814
128 0.453864382850211
52 0.999999999999562
128 0.44078135203427
128 0.458744324488732
128 0.491713339201286
128 0.473755484444118
128 0.43640100174647
128 0.471860177767164
128 0.447561963797584
128 0.444707226340922
128 0.455863634263244
128 0.457839990888376
128 0.472578981916148
128 0.437484940853265
128 0.456853590893249
128 0.430242330622271
128 0.463858250676322
128 0.454639439811896
128 0.466564054802905
128 0.441093697080691
37 0.999999999999578
128 0.441306366404058
128 0.453590215349346
48 0.999999999999601
128 0.453305680550177
128 0.441463535893623
128 0.465091920535428
128 0.450751019091155
111 0.498332451515408
128 0.468801955706453
128 0.454803784586214
128 0.47925442111211
128 0.473050886137429
128 0.47212944509492
128 0.434168756182502
128 0.458745410264215
128 0.430287871290948
128 0.445010084692598
128 0.455761814695417
93 0.533267619007398
128 0.453228897720169
128 0.460630460456436
112 0.497808609145996
128 0.463808179007711
128 0.43462449864452
128 0.470926350141198
128 0.596765613454139
128 0.465972113441355
128 0.453108864640288
128 0.447515717639843
128 0.430232027234319
128 0.466000380163644
128 0.4432815365329
128 0.456440635325892
128 0.443198194194406
128 0.461969189372779
128 0.458656242107878
128 0.440843585058284
95 0.525852564263858
128 0.438256557517088
128 0.46328166337582
45 0.999999999999616
128 0.44855541528904
128 0.470937455291603
128 0.432419826468903
128 0.44278170800846
128 0.459197214009814
128 0.468145435221231
128 0.463082359921666
128 0.499931304479796
128 0.450831838846484
128 0.522447096592407
128 0.450724009126416
77 0.693734711160028
128 0.490803108647615
128 0.447633703550769
128 0.445997935613247
128 0.459989381388626
128 0.442278558027045
128 0.44241590537898
128 0.477068899089314
128 0.441014513865883
128 0.43359806784465
128 0.461472518214637
110 0.474135687621098
88 0.544295563222775
120 0.441350764556236
128 0.448925614249275
128 0.471033210068813
128 0.462523241200814
128 0.450586485443722
128 0.460709558528688
128 0.4303404788441
128 0.455726120973275
128 0.440869710868574
128 0.448550208985788
128 0.456992738020481
128 0.465810940756779
128 0.453784758303391
128 0.447210042532394
128 0.451286431090491
128 0.462894145493921
128 0.492073466315194
109 0.472217369319301
128 0.486996919401887
128 0.486778033970822
128 0.457146042483282
128 0.444871780823982
128 0.467058670485075
128 0.469785269638236
128 0.466524472856842
128 0.446594719770686
128 0.455507154977939
128 0.483673488095022
128 0.46227194124874
128 0.454892039350526
128 0.46058587336014
128 0.46712828854596
128 0.473843838761987
128 0.450628996999386
128 0.456237063002799
128 0.459151648026857
128 0.435572421771747
128 0.448448415137958
128 0.441209796640871
128 0.472458418318195
128 0.46211803771101
128 0.446382570442612
128 0.466532959797468
128 0.486183603759459
128 0.437122168720035
97 0.524979962186155
128 0.454824345733323
128 0.453701626886555
128 0.463983512274705
128 0.454827260239101
128 0.511979356632488
128 0.435128259091147
51 0.999999999731446
128 0.458693386445652
128 0.463619318626046
128 0.448429134977068
74 0.631486271980388
128 0.43305921231808
128 0.436629744228498
128 0.465443445946032
128 0.449080249842541
128 0.449214249307947
128 0.462190588395123
128 0.455658990483269
128 0.451732088452203
128 0.448179491099814
128 0.451714519804482
128 0.453860558562491
128 0.463568382962674
128 0.435331032157534
128 0.447818470711022
128 0.443552056549594
128 0.451784636702766
128 0.430649286757971
128 0.458339991882877
128 0.462513367668727
128 0.473754867733582
70 0.744616947933265
128 0.454119456335426
128 0.457075916307697
128 0.493390339193698
128 0.437316257299213
128 0.445673785165698
128 0.437557162500972
128 0.456436213204002
128 0.440742707835975
128 0.463144802203801
128 0.458495655657989
128 0.452490888309747
128 0.460441799964974
128 0.442385942158573
128 0.463481831001542
128 0.442519712507689
61 0.999999999999574
128 0.464587402762529
128 0.467906337919268
128 0.424644147024535
128 0.434371098873227
35 0.999999999999595
114 0.473400069462411
128 0.45758612120875
118 0.431803308484426
47 0.999999999999634
128 0.457410589808723
128 0.44473848368827
45 0.999999999999641
128 0.459689124640145
128 0.460853286282918
65 0.991611024523102
128 0.436529324161925
128 0.447186758440981
128 0.467786847776764
128 0.488917096667997
128 0.451845245690481
128 0.465293078649537
128 0.436045244205117
128 0.432928589201447
128 0.463523198968482
128 0.442225137978188
128 0.450384755681026
128 0.44367888190808
128 0.461405008401458
128 0.450273564261897
128 0.446762885534041
128 0.461223478148589
128 0.459553147937688
128 0.474508597044852
128 0.472773357320574
128 0.447791172152483
128 0.461867293302356
128 0.463125582993433
128 0.457610947687224
128 0.453982475872423
128 0.458029240080322
128 0.43958119467345
128 0.447985448963787
81 0.596208564688514
128 0.480578881434291
128 0.472487122584652
128 0.445586860587024
128 0.449327610754237
69 0.740136493745341
128 0.451617792816285
128 0.452836299785462
128 0.477135499922519
128 0.450187459104212
128 0.446823839728316
128 0.455922876159245
128 0.44377554844208
87 0.545342279954669
128 0.466753231866846
69 0.848882345336129
128 0.45717850408824
128 0.441861884411369
128 0.50906531883288
128 0.460158989039457
128 0.437823500443291
128 0.454996183361866
128 0.437690256879264
128 0.439701490376979
128 0.466753837932014
128 0.40682441989561
128 0.454357776141002
128 0.448584578599287
121 0.499194800113136
128 0.442839384426275
128 0.463300085124299
128 0.460054791594466
123 0.496339623553829
128 0.447465809168184
128 0.437026392238122
128 0.440349425655249
128 0.455414489801637
128 0.458850089035523
128 0.48898782967559
128 0.441881561716046
128 0.452370996886753
128 0.473319658410953
128 0.45441738415438
62 0.999999999999547
128 0.44764456163175
128 0.446210578761073
128 0.474762587924144
46 0.99999999999962
128 0.468128287747528
128 0.44292351565847
128 0.448153164272507
128 0.458046610012012
128 0.437477279131485
106 0.471688398070637
128 0.446663983746635
128 0.434482661407968
116 0.471024296215026
81 0.608948511979529
128 0.442804238328079
128 0.459317967003011
128 0.465442247652539
47 0.999999999999604
128 0.450905086963357
128 0.455259358689678
128 0.435197961010378
128 0.468590244255366
128 0.469742017077472
128 0.44902647916312
128 0.477840839126718
79 0.601043817779582
128 0.46875828354611
128 0.45115491007152
128 0.466295999576036
128 0.442786208216631
128 0.474389622113969
128 0.445542949810478
128 0.471647249294592
52 0.99999999999957
128 0.448330816371777
128 0.454754063561748
128 0.461138620575163
128 0.435318933335299
128 0.479452993277701
128 0.487466328384139
128 0.450798482461736
128 0.463196134113819
128 0.439532658532521
120 0.500300210285777
128 0.454788434911652
128 0.476505352124739
128 0.45998610420142
128 0.448737401962795
128 0.454541555065211
128 0.440591781533581
128 0.438180276288881
128 0.445522252301534
128 0.436289645352017
128 0.448922026459427
128 0.498484306197622
128 0.455325853037836
128 0.451917813266697
64 0.999833799700425
128 0.46462039120936
128 0.444025918576025
89 0.545858046335258
128 0.44695696503871
128 0.446035566077249
128 0.451865855610454
128 0.448099628616216
50 0.999999999999575
128 0.459295681324519
128 0.439385696428855
128 0.443545531182968
128 0.416918625879792
128 0.465730880004813
128 0.472508899403547
128 0.467324706000283
128 0.495442929533335
128 0.468031411007462
128 0.476561704185717
128 0.454085301815823
128 0.468677448220865
118 0.504011173498805
96 0.526619034123908
128 0.451176197738216
128 0.466383337958792
123 0.445591780652803
128 0.442544524843684
128 0.453523879363659
128 0.431086721949089
128 0.460681275693933
128 0.454286315183604
128 0.452535527565328
128 0.444878392082744
128 0.46512260030007
107 0.546020546498552
128 0.47047534495545
128 0.462388496619686
128 0.46138688612649
128 0.454888420403763
128 0.463430735253555
128 0.439207215529865
128 0.458364622543313
128 0.439457748991281
128 0.516521978765833
128 0.453971887197303
128 0.433530806123541
128 0.446214297902896
128 0.456354759392541
55 0.999999974032698
128 0.436259881161931
128 0.477207654380178
128 0.449350696406288
128 0.441760656992478
112 0.494889742201172
128 0.446576158824955
128 0.464959313545891
103 0.499359496758915
128 0.461203432027789
128 0.44575987999814
128 0.446984479644928
128 0.470621637012594
128 0.471177489145388
128 0.47303314284212
128 0.442148800224352
128 0.461497279712135
128 0.453063980755651
128 0.439127660262407
128 0.401412180584368
128 0.419226877304369
128 0.458103566712137
128 0.433006421703612
128 0.463826866454953
128 0.469307070166995
128 0.456740499834908
125 0.457442839999638
128 0.481935436346425
103 0.536833797119112
128 0.445558722212658
128 0.460023120158073
128 0.457013771017481
124 0.447229062793849
128 0.457692986466351
128 0.461139949239433
128 0.462370224461169
128 0.456815966964205
128 0.47385250600197
128 0.449132883645802
128 0.437875303937154
128 0.45475162238342
128 0.447705360367474
128 0.45485031456102
128 0.430940258588441
128 0.46805346905131
128 0.433796904019438
128 0.457391848904147
128 0.451611974405539
128 0.466591908091257
128 0.427161475704159
128 0.460375476440469
128 0.46888433122167
120 0.493804965285644
78 0.639945175651355
128 0.460346353543303
128 0.469811178483869
128 0.468588966472299
128 0.439495153727602
128 0.45410653566091
128 0.444244229536686
128 0.454242672083986
128 0.435517399321914
128 0.457019099953135
128 0.416309750473793
128 0.449428288900468
128 0.439690090443535
128 0.462664980206253
128 0.460051473883313
83 0.566414915885495
128 0.452960092306815
110 0.488250779004967
128 0.44527787933204
128 0.427747075423731
128 0.457644152607601
128 0.441508244014733
128 0.448702493876387
128 0.468941314264895
128 0.447702684528654
128 0.43367225734307
128 0.469330473331638
128 0.454025675959949
128 0.454553501355374
128 0.437137555150949
128 0.449541658900247
128 0.458812995483055
128 0.464471449458008
128 0.444897134665537
128 0.473193567294114
128 0.451780834791001
128 0.470615192091905
128 0.440430671577245
128 0.430981957793112
128 0.454723878066508
40 0.999999999999586
128 0.46918581908468
128 0.43639340118934
128 0.427952517042169
128 0.465881032089552
58 0.999999999999566
108 0.482008823992616
128 0.448441128234637
128 0.467480846093219
128 0.455124734404569
128 0.479010097667036
128 0.460113447266826
128 0.463115357550511
128 0.450254391578199
128 0.472847945830732
128 0.466340282716536
128 0.451783625411637
128 0.470232680242798
128 0.432914624491703
128 0.455476891336656
128 0.437578693360852
41 0.999999999999582
128 0.446389411940326
128 0.435634098495772
128 0.457765832322556
128 0.449275654755789
128 0.451753113319161
128 0.471053627100478
128 0.467730202527694
128 0.459641390363569
128 0.464000300331896
128 0.448903126135299
128 0.462071659251945
128 0.44590269745034
106 0.451523028578525
128 0.462999100230938
128 0.462651606189969
128 0.435627399802389
128 0.441736800735571
128 0.452878937956219
128 0.438165522797121
128 0.441784099089924
128 0.440694809928191
120 0.462211663459605
128 0.470289063290776
83 0.68039518670693
128 0.43575475764523
128 0.441718711522771
128 0.460391639679056
128 0.479749450222156
128 0.452958559003714
128 0.408977294083287
80 0.615025952537899
128 0.475557810378309
128 0.446138982177106
105 0.494110809736758
128 0.457247892550116
128 0.468316323091907
128 0.44215110051937
128 0.456794377971815
128 0.455603775582648
128 0.420679041890634
128 0.461467007514089
128 0.463061381076399
128 0.446309356882489
53 0.99999999999955
81 0.602737104152695
128 0.446915076374629
128 0.450413206387563
128 0.445531113419368
128 0.447282969483255
128 0.450647864586138
128 0.460641943012673
118 0.481705671896182
128 0.446164373362698
128 0.441928956059764
128 0.463528028748315
128 0.456687388880991
128 0.480584033051881
128 0.535134532449229
128 0.449775851425166
128 0.455570657148205
18 0.99999999999977
65 0.987020885532121
128 0.480410779720461
128 0.487670919161583
128 0.430121782583461
128 0.442586159948027
128 0.455244615754508
128 0.44439307377002
128 0.448543436884299
128 0.452886710194334
128 0.442610903312004
128 0.433585246643741
128 0.449801086957528
128 0.466891200170959
128 0.467041148329252
128 0.505628505541019
41 0.999999999999629
128 0.515273183597254
128 0.462762557613999
128 0.473510830670981
128 0.482872932686146
128 0.457521012515441
128 0.459828433681598
128 0.461093378413411
128 0.495354133751206
128 0.436405800967117
102 0.449461525362251
128 0.452683888054658
128 0.462093804391193
128 0.432136833088132
128 0.47066367652384
128 0.467180975706944
128 0.440551128261486
35 0.999999999999611
128 0.463197131191136
125 0.429201886580821
52 0.999999999999518
128 0.43805622740305
128 0.444821144837993
128 0.474963733495682
128 0.45019480823641
128 0.467870485886255
128 0.449297801698827
128 0.478846430457364
128 0.453693289542979
128 0.449707073570561
128 0.466244385700982
81 0.645426348079201
56 0.999999999999584
128 0.474628716415427
128 0.440544634349291
128 0.468685716160004
128 0.449944339718462
128 0.447237768392474
117 0.447572335822983
128 0.436281371865736
128 0.431579914931976
128 0.504344612549723
44 0.999999999999624
128 0.466060985503539
120 0.466505812331528
128 0.459478526285668
128 0.449524724973192
128 0.450223086375662
128 0.474528249677584
116 0.476553007506895
128 0.458731014745688
128 0.472382770047826
128 0.441288655131187
128 0.457572808859152
128 0.457024495096274
128 0.469906449113734
128 0.476974167105414
128 0.4500330016827
128 0.487940253573236
128 0.446498578846323
128 0.444511313692277
128 0.437800748723294
128 0.433787440589707
116 0.472149328841457
34 0.999999999999596
128 0.468385858149943
128 0.481551837345588
128 0.428562445469392
59 0.999999997862582
128 0.451147984546972
128 0.439889403152586
128 0.438677917712121
128 0.447740712679291
128 0.476259593698774
128 0.468900788902285
128 0.457121311364527
128 0.473170541193568
128 0.478036850084588
128 0.465099764269287
128 0.463842203495782
128 0.434716187151829
128 0.473970856624374
128 0.443838470933788
128 0.445415468221976
128 0.480678710441689
128 0.445319290237884
128 0.456368352882921
128 0.449098643077193
128 0.468158140115637
128 0.438212765137337
37 0.999999999999609
128 0.446846744194343
128 0.437704747204745
128 0.474629969955842
128 0.453748864718776
37 0.999999999999592
128 0.441481167031227
128 0.437663584781819
128 0.440498350819612
128 0.447916719091482
128 0.473329477522707
128 0.464592161055991
128 0.465131411277262
67 0.818016510792989
128 0.435779007479488
128 0.44070866216887
128 0.487021286204044
128 0.459456531159822
128 0.409039459833102
118 0.481390775956719
128 0.457575540852015
128 0.449047496590972
128 0.44359440041068
128 0.447571371869018
128 0.436061214091222
128 0.457280892713521
128 0.451644059740973
128 0.468033065635245
38 0.999999999999568
128 0.479499476406059
54 0.999999999999566
128 0.447547835934124
128 0.431643144914465
128 0.45802233535455
70 0.797857764316765
128 0.482309685467512
128 0.475861658531881
128 0.44113857045676
128 0.460248223824965
128 0.464390381760989
128 0.46898828715984
128 0.448796044025057
128 0.452465809950341
128 0.446617163268511
128 0.443255628801518
128 0.452945014848928
128 0.45164923697541
55 0.999999999999563
99 0.51919234663971
128 0.461585339222049
128 0.487514664406528
121 0.45138365016402
128 0.464927760064142
128 0.448224513445795
128 0.458483797409754
128 0.464571570273425
112 0.475582760691967
128 0.444429175655414
128 0.462458820969988
128 0.473695210298655
128 0.439490859682472
128 0.448307119361031
99 0.488181378226269
128 0.453377711764289
128 0.436453713186794
128 0.481885654905006
100 0.500210596947448
128 0.44380925404194
111 0.434998387659076
128 0.441050236933666
128 0.445596725900221
128 0.471720563389406
128 0.453674223446898
128 0.440701025606175
128 0.456869130931376
128 0.456196178326025
128 0.453472979112488
128 0.433504096216625
128 0.45397462573969
128 0.44785924406336
128 0.439748901997498
128 0.425822922227901
128 0.460207610096584
128 0.485289512618354
128 0.469361574143756
128 0.474811283781758
128 0.448652112117774
128 0.45384480766519
128 0.445565130940991
128 0.459755859651779
128 0.456578364868531
128 0.456607242228412
128 0.457993613498086
128 0.446931932069797
128 0.457131113285932
128 0.443489256596475
128 0.431587300703737
128 0.445508054580321
128 0.478468812702437
128 0.456539982446136
128 0.42309042668286
71 0.705392850285739
128 0.450450613240445
30 0.999999999999705
128 0.448246812599331
128 0.45007796769562
98 0.524554316164848
128 0.429944201052542
128 0.450237619197196
128 0.440977908043318
93 0.534507545115304
128 0.461040494270756
128 0.455339960544134
128 0.444745079181624
128 0.414801279874441
128 0.444665997202713
128 0.459000692963282
128 0.456077926662525
128 0.441595279374942
128 0.451197017048346
128 0.449616122873936
128 0.438783184008781
128 0.472471346162958
128 0.436211949925931
128 0.431917454414322
128 0.460821338744466
128 0.448348386211777
128 0.444533712013981
128 0.482080009474438
128 0.472241394588691
128 0.461812464175593
128 0.434440900015673
128 0.442444252649263
128 0.451484104288495
128 0.46791183294173
128 0.453469868835845
128 0.461829310197874
84 0.591799377878511
128 0.456157817812039
128 0.457596086006001
128 0.45832577374589
128 0.458996687911081
128 0.448148009044725
128 0.475016110496617
128 0.440932774072055
71 0.758943444021283
128 0.443221005998813
128 0.444476785502304
128 0.468493248195328
62 0.999999999999508
128 0.458281480020441
128 0.440331916027837
128 0.50005340542963
128 0.447347998281145
128 0.514005571777049
128 0.455724909086385
128 0.420896385554342
128 0.470074791465031
66 0.944503114734181
128 0.427931245408275
128 0.453195540658357
128 0.453941434911643
128 0.464898964752115
86 0.579070941437552
128 0.43878231957996
128 0.427526236023164
128 0.442982917052045
128 0.452088075895911
128 0.450098630282196
128 0.457588511532247
128 0.453557130607449
128 0.442595068033708
128 0.470395974794432
128 0.479656951032169
128 0.435805801162737
128 0.444945976708399
128 0.437172671203353
128 0.471232029846501
71 0.711400820663041
128 0.441688093303195
128 0.473947810688694
128 0.453496138064257
128 0.483031961652669
128 0.439331959253686
128 0.444572567113037
128 0.461773087220683
128 0.447716619137871
128 0.436548449556871
128 0.469601695651581
128 0.430134266816828
128 0.438956584727479
128 0.4496790252035
128 0.46189661309294
128 0.433314803210308
128 0.433198991441098
128 0.445637730300546
128 0.469347705397719
128 0.457272727947758
128 0.43405644613446
128 0.463107226840283
128 0.458954690126933
128 0.456419714307107
63 0.999999999999512
109 0.4944785606572
128 0.448524386083411
128 0.452387191505702
128 0.48017893332715
128 0.465848326673635
128 0.460252664517084
128 0.458815895118304
128 0.533340171916954
128 0.457450834680848
128 0.433512757777497
128 0.432901114967486
29 0.999999999999729
128 0.460173009479984
67 0.80809775650214
128 0.47444395335473
128 0.451841552183614
128 0.456717552413256
128 0.45507940619258
128 0.488671082304663
128 0.44857335670521
128 0.457138823168515
128 0.455667137633514
128 0.429373469305312
128 0.476865435552813
128 0.455001251322607
128 0.417768612608971
128 0.426405457323025
128 0.439942149841418
128 0.44738460928237
128 0.456561432121095
128 0.461765193036772
104 0.475204932326339
128 0.466185258453822
128 0.458184691997579
128 0.459855311159446
128 0.42619038976778
128 0.449040954650466
128 0.463703545887555
122 0.444974418694285
128 0.444392857581222
128 0.442853155347964
121 0.457941770220217
128 0.451013697260214
128 0.450466547630443
80 0.597659455949583
128 0.461050816268021
128 0.442279047487886
128 0.441068782319566
128 0.446797816386915
128 0.440259833816473
126 0.442684321265902
128 0.432946284504191
128 0.451887210236417
128 0.451409583922657
128 0.461740773327735
128 0.445155971778826
128 0.447401871070183
128 0.446029293841774
128 0.441248100970309
128 0.45650623572509
128 0.440650095984018
61 0.999999999999542
128 0.457037854078489
128 0.455027569293599
128 0.448094939526782
128 0.450916443953159
128 0.467219486504121
56 0.999999999999587
128 0.474516348752631
128 0.455903880962983
128 0.466389881682241
128 0.446402802786425
128 0.457336224444419
128 0.46867166798645
128 0.447405698827075
128 0.443372080433355
128 0.477602041702532
128 0.498956779880464
128 0.473416679997714
128 0.465620354089109
90 0.514855883533668
128 0.466282447461104
128 0.447171636034947
128 0.441503534603459
128 0.467396545249945
128 0.446926135955899
128 0.443078767959638
128 0.426222440833271
128 0.441518557521392
128 0.45104846677131
128 0.448858366356053
128 0.464197774817081
107 0.476128971229347
128 0.461425250312439
128 0.436060815119404
128 0.444310771530316
128 0.421615637580348
128 0.488688731743259
128 0.446681682133285
128 0.450949687288875
128 0.458719206723071
128 0.444043285044003
128 0.49752549062408
128 0.436845422147627
28 0.999999999999729
128 0.449212498751365
128 0.433573884889557
128 0.448113660147514
128 0.473706297941993
72 0.667660148094148
128 0.455891350999003
128 0.441902719565848
128 0.448521936447225
128 0.451651460293567
128 0.476384077256651
128 0.42705917298463
128 0.459644321740343
128 0.45254782410346
128 0.440741414349386
75 0.60778594711303
128 0.44766343064404
128 0.428466857395536
128 0.441290774541008
128 0.436129562220271
128 0.429795981458245
128 0.467510247393489
128 0.481133784076049
128 0.455030646160975
128 0.440837551998391
128 0.454867879941179
128 0.441977782344197
31 0.999999999999648
128 0.457309394314501
128 0.480079354654676
128 0.459251476265393
};
\end{axis}

\end{tikzpicture}

%% file: identifiability/avg_attn1_org.tex
% This file was created by tikzplotlib v0.8.2.
\begin{tikzpicture}

\definecolor{color0}{rgb}{0.905882352941176,0.298039215686275,0.235294117647059}
\definecolor{color1}{rgb}{0.203921568627451,0.596078431372549,0.858823529411765}
\definecolor{color2}{rgb}{0.607843137254902,0.349019607843137,0.713725490196078}

\begin{axis}[
axis line style={white!15.0!black},
legend cell align={left},
legend style={draw=white!80.0!black},
tick align=outside,
tick pos=both,
width=\linewidth,
height=\effectiveAttnPlotHeight,
x grid style={white!80.0!black},
xlabel={Layer},
xmajorgrids,
xmin=0.366240817709798, xmax=12.6337591822902,
xtick style={color=white!15.0!black},
y grid style={white!80.0!black},
ylabel={Raw Attention},
ymajorgrids,
ymin=-0.0468060428997225, ymax=0.926731513806353,
ytick style={color=white!15.0!black}
]
\addplot [only marks, mark size=1.25pt, draw=color0, fill=color0, colormap={mymap}{[1pt]
 rgb(0pt)=(0.01060815,0.01808215,0.10018654);
  rgb(1pt)=(0.01428972,0.02048237,0.10374486);
  rgb(2pt)=(0.01831941,0.0229766,0.10738511);
  rgb(3pt)=(0.02275049,0.02554464,0.11108639);
  rgb(4pt)=(0.02759119,0.02818316,0.11483751);
  rgb(5pt)=(0.03285175,0.03088792,0.11863035);
  rgb(6pt)=(0.03853466,0.03365771,0.12245873);
  rgb(7pt)=(0.04447016,0.03648425,0.12631831);
  rgb(8pt)=(0.05032105,0.03936808,0.13020508);
  rgb(9pt)=(0.05611171,0.04224835,0.13411624);
  rgb(10pt)=(0.0618531,0.04504866,0.13804929);
  rgb(11pt)=(0.06755457,0.04778179,0.14200206);
  rgb(12pt)=(0.0732236,0.05045047,0.14597263);
  rgb(13pt)=(0.0788708,0.05305461,0.14995981);
  rgb(14pt)=(0.08450105,0.05559631,0.15396203);
  rgb(15pt)=(0.09011319,0.05808059,0.15797687);
  rgb(16pt)=(0.09572396,0.06050127,0.16200507);
  rgb(17pt)=(0.10132312,0.06286782,0.16604287);
  rgb(18pt)=(0.10692823,0.06517224,0.17009175);
  rgb(19pt)=(0.1125315,0.06742194,0.17414848);
  rgb(20pt)=(0.11813947,0.06961499,0.17821272);
  rgb(21pt)=(0.12375803,0.07174938,0.18228425);
  rgb(22pt)=(0.12938228,0.07383015,0.18636053);
  rgb(23pt)=(0.13501631,0.07585609,0.19044109);
  rgb(24pt)=(0.14066867,0.0778224,0.19452676);
  rgb(25pt)=(0.14633406,0.07973393,0.1986151);
  rgb(26pt)=(0.15201338,0.08159108,0.20270523);
  rgb(27pt)=(0.15770877,0.08339312,0.20679668);
  rgb(28pt)=(0.16342174,0.0851396,0.21088893);
  rgb(29pt)=(0.16915387,0.08682996,0.21498104);
  rgb(30pt)=(0.17489524,0.08848235,0.2190294);
  rgb(31pt)=(0.18065495,0.09009031,0.22303512);
  rgb(32pt)=(0.18643324,0.09165431,0.22699705);
  rgb(33pt)=(0.19223028,0.09317479,0.23091409);
  rgb(34pt)=(0.19804623,0.09465217,0.23478512);
  rgb(35pt)=(0.20388117,0.09608689,0.23860907);
  rgb(36pt)=(0.20973515,0.09747934,0.24238489);
  rgb(37pt)=(0.21560818,0.09882993,0.24611154);
  rgb(38pt)=(0.22150014,0.10013944,0.2497868);
  rgb(39pt)=(0.22741085,0.10140876,0.25340813);
  rgb(40pt)=(0.23334047,0.10263737,0.25697736);
  rgb(41pt)=(0.23928891,0.10382562,0.2604936);
  rgb(42pt)=(0.24525608,0.10497384,0.26395596);
  rgb(43pt)=(0.25124182,0.10608236,0.26736359);
  rgb(44pt)=(0.25724602,0.10715148,0.27071569);
  rgb(45pt)=(0.26326851,0.1081815,0.27401148);
  rgb(46pt)=(0.26930915,0.1091727,0.2772502);
  rgb(47pt)=(0.27536766,0.11012568,0.28043021);
  rgb(48pt)=(0.28144375,0.11104133,0.2835489);
  rgb(49pt)=(0.2875374,0.11191896,0.28660853);
  rgb(50pt)=(0.29364846,0.11275876,0.2896085);
  rgb(51pt)=(0.29977678,0.11356089,0.29254823);
  rgb(52pt)=(0.30592213,0.11432553,0.29542718);
  rgb(53pt)=(0.31208435,0.11505284,0.29824485);
  rgb(54pt)=(0.31826327,0.1157429,0.30100076);
  rgb(55pt)=(0.32445869,0.11639585,0.30369448);
  rgb(56pt)=(0.33067031,0.11701189,0.30632563);
  rgb(57pt)=(0.33689808,0.11759095,0.3088938);
  rgb(58pt)=(0.34314168,0.11813362,0.31139721);
  rgb(59pt)=(0.34940101,0.11863987,0.3138355);
  rgb(60pt)=(0.355676,0.11910909,0.31620996);
  rgb(61pt)=(0.36196644,0.1195413,0.31852037);
  rgb(62pt)=(0.36827206,0.11993653,0.32076656);
  rgb(63pt)=(0.37459292,0.12029443,0.32294825);
  rgb(64pt)=(0.38092887,0.12061482,0.32506528);
  rgb(65pt)=(0.38727975,0.12089756,0.3271175);
  rgb(66pt)=(0.39364518,0.12114272,0.32910494);
  rgb(67pt)=(0.40002537,0.12134964,0.33102734);
  rgb(68pt)=(0.40642019,0.12151801,0.33288464);
  rgb(69pt)=(0.41282936,0.12164769,0.33467689);
  rgb(70pt)=(0.41925278,0.12173833,0.33640407);
  rgb(71pt)=(0.42569057,0.12178916,0.33806605);
  rgb(72pt)=(0.43214263,0.12179973,0.33966284);
  rgb(73pt)=(0.43860848,0.12177004,0.34119475);
  rgb(74pt)=(0.44508855,0.12169883,0.34266151);
  rgb(75pt)=(0.45158266,0.12158557,0.34406324);
  rgb(76pt)=(0.45809049,0.12142996,0.34540024);
  rgb(77pt)=(0.46461238,0.12123063,0.34667231);
  rgb(78pt)=(0.47114798,0.12098721,0.34787978);
  rgb(79pt)=(0.47769736,0.12069864,0.34902273);
  rgb(80pt)=(0.48426077,0.12036349,0.35010104);
  rgb(81pt)=(0.49083761,0.11998161,0.35111537);
  rgb(82pt)=(0.49742847,0.11955087,0.35206533);
  rgb(83pt)=(0.50403286,0.11907081,0.35295152);
  rgb(84pt)=(0.51065109,0.11853959,0.35377385);
  rgb(85pt)=(0.51728314,0.1179558,0.35453252);
  rgb(86pt)=(0.52392883,0.11731817,0.35522789);
  rgb(87pt)=(0.53058853,0.11662445,0.35585982);
  rgb(88pt)=(0.53726173,0.11587369,0.35642903);
  rgb(89pt)=(0.54394898,0.11506307,0.35693521);
  rgb(90pt)=(0.5506426,0.11420757,0.35737863);
  rgb(91pt)=(0.55734473,0.11330456,0.35775059);
  rgb(92pt)=(0.56405586,0.11235265,0.35804813);
  rgb(93pt)=(0.57077365,0.11135597,0.35827146);
  rgb(94pt)=(0.5774991,0.11031233,0.35841679);
  rgb(95pt)=(0.58422945,0.10922707,0.35848469);
  rgb(96pt)=(0.59096382,0.10810205,0.35847347);
  rgb(97pt)=(0.59770215,0.10693774,0.35838029);
  rgb(98pt)=(0.60444226,0.10573912,0.35820487);
  rgb(99pt)=(0.61118304,0.10450943,0.35794557);
  rgb(100pt)=(0.61792306,0.10325288,0.35760108);
  rgb(101pt)=(0.62466162,0.10197244,0.35716891);
  rgb(102pt)=(0.63139686,0.10067417,0.35664819);
  rgb(103pt)=(0.63812122,0.09938212,0.35603757);
  rgb(104pt)=(0.64483795,0.0980891,0.35533555);
  rgb(105pt)=(0.65154562,0.09680192,0.35454107);
  rgb(106pt)=(0.65824241,0.09552918,0.3536529);
  rgb(107pt)=(0.66492652,0.09428017,0.3526697);
  rgb(108pt)=(0.67159578,0.09306598,0.35159077);
  rgb(109pt)=(0.67824099,0.09192342,0.3504148);
  rgb(110pt)=(0.684863,0.09085633,0.34914061);
  rgb(111pt)=(0.69146268,0.0898675,0.34776864);
  rgb(112pt)=(0.69803757,0.08897226,0.3462986);
  rgb(113pt)=(0.70457834,0.0882129,0.34473046);
  rgb(114pt)=(0.71108138,0.08761223,0.3430635);
  rgb(115pt)=(0.7175507,0.08716212,0.34129974);
  rgb(116pt)=(0.72398193,0.08688725,0.33943958);
  rgb(117pt)=(0.73035829,0.0868623,0.33748452);
  rgb(118pt)=(0.73669146,0.08704683,0.33543669);
  rgb(119pt)=(0.74297501,0.08747196,0.33329799);
  rgb(120pt)=(0.74919318,0.08820542,0.33107204);
  rgb(121pt)=(0.75535825,0.08919792,0.32876184);
  rgb(122pt)=(0.76145589,0.09050716,0.32637117);
  rgb(123pt)=(0.76748424,0.09213602,0.32390525);
  rgb(124pt)=(0.77344838,0.09405684,0.32136808);
  rgb(125pt)=(0.77932641,0.09634794,0.31876642);
  rgb(126pt)=(0.78513609,0.09892473,0.31610488);
  rgb(127pt)=(0.79085854,0.10184672,0.313391);
  rgb(128pt)=(0.7965014,0.10506637,0.31063031);
  rgb(129pt)=(0.80205987,0.10858333,0.30783);
  rgb(130pt)=(0.80752799,0.11239964,0.30499738);
  rgb(131pt)=(0.81291606,0.11645784,0.30213802);
  rgb(132pt)=(0.81820481,0.12080606,0.29926105);
  rgb(133pt)=(0.82341472,0.12535343,0.2963705);
  rgb(134pt)=(0.82852822,0.13014118,0.29347474);
  rgb(135pt)=(0.83355779,0.13511035,0.29057852);
  rgb(136pt)=(0.83850183,0.14025098,0.2876878);
  rgb(137pt)=(0.84335441,0.14556683,0.28480819);
  rgb(138pt)=(0.84813096,0.15099892,0.281943);
  rgb(139pt)=(0.85281737,0.15657772,0.27909826);
  rgb(140pt)=(0.85742602,0.1622583,0.27627462);
  rgb(141pt)=(0.86196552,0.16801239,0.27346473);
  rgb(142pt)=(0.86641628,0.17387796,0.27070818);
  rgb(143pt)=(0.87079129,0.17982114,0.26797378);
  rgb(144pt)=(0.87507281,0.18587368,0.26529697);
  rgb(145pt)=(0.87925878,0.19203259,0.26268136);
  rgb(146pt)=(0.8833417,0.19830556,0.26014181);
  rgb(147pt)=(0.88731387,0.20469941,0.25769539);
  rgb(148pt)=(0.89116859,0.21121788,0.2553592);
  rgb(149pt)=(0.89490337,0.21785614,0.25314362);
  rgb(150pt)=(0.8985026,0.22463251,0.25108745);
  rgb(151pt)=(0.90197527,0.23152063,0.24918223);
  rgb(152pt)=(0.90530097,0.23854541,0.24748098);
  rgb(153pt)=(0.90848638,0.24568473,0.24598324);
  rgb(154pt)=(0.911533,0.25292623,0.24470258);
  rgb(155pt)=(0.9144225,0.26028902,0.24369359);
  rgb(156pt)=(0.91717106,0.26773821,0.24294137);
  rgb(157pt)=(0.91978131,0.27526191,0.24245973);
  rgb(158pt)=(0.92223947,0.28287251,0.24229568);
  rgb(159pt)=(0.92456587,0.29053388,0.24242622);
  rgb(160pt)=(0.92676657,0.29823282,0.24285536);
  rgb(161pt)=(0.92882964,0.30598085,0.24362274);
  rgb(162pt)=(0.93078135,0.31373977,0.24468803);
  rgb(163pt)=(0.93262051,0.3215093,0.24606461);
  rgb(164pt)=(0.93435067,0.32928362,0.24775328);
  rgb(165pt)=(0.93599076,0.33703942,0.24972157);
  rgb(166pt)=(0.93752831,0.34479177,0.25199928);
  rgb(167pt)=(0.93899289,0.35250734,0.25452808);
  rgb(168pt)=(0.94036561,0.36020899,0.25734661);
  rgb(169pt)=(0.94167588,0.36786594,0.2603949);
  rgb(170pt)=(0.94291042,0.37549479,0.26369821);
  rgb(171pt)=(0.94408513,0.3830811,0.26722004);
  rgb(172pt)=(0.94520419,0.39062329,0.27094924);
  rgb(173pt)=(0.94625977,0.39813168,0.27489742);
  rgb(174pt)=(0.94727016,0.4055909,0.27902322);
  rgb(175pt)=(0.94823505,0.41300424,0.28332283);
  rgb(176pt)=(0.94914549,0.42038251,0.28780969);
  rgb(177pt)=(0.95001704,0.42771398,0.29244728);
  rgb(178pt)=(0.95085121,0.43500005,0.29722817);
  rgb(179pt)=(0.95165009,0.44224144,0.30214494);
  rgb(180pt)=(0.9524044,0.44944853,0.3072105);
  rgb(181pt)=(0.95312556,0.45661389,0.31239776);
  rgb(182pt)=(0.95381595,0.46373781,0.31769923);
  rgb(183pt)=(0.95447591,0.47082238,0.32310953);
  rgb(184pt)=(0.95510255,0.47787236,0.32862553);
  rgb(185pt)=(0.95569679,0.48489115,0.33421404);
  rgb(186pt)=(0.95626788,0.49187351,0.33985601);
  rgb(187pt)=(0.95681685,0.49882008,0.34555431);
  rgb(188pt)=(0.9573439,0.50573243,0.35130912);
  rgb(189pt)=(0.95784842,0.51261283,0.35711942);
  rgb(190pt)=(0.95833051,0.51946267,0.36298589);
  rgb(191pt)=(0.95879054,0.52628305,0.36890904);
  rgb(192pt)=(0.95922872,0.53307513,0.3748895);
  rgb(193pt)=(0.95964538,0.53983991,0.38092784);
  rgb(194pt)=(0.96004345,0.54657593,0.3870292);
  rgb(195pt)=(0.96042097,0.55328624,0.39319057);
  rgb(196pt)=(0.96077819,0.55997184,0.39941173);
  rgb(197pt)=(0.9611152,0.5666337,0.40569343);
  rgb(198pt)=(0.96143273,0.57327231,0.41203603);
  rgb(199pt)=(0.96173392,0.57988594,0.41844491);
  rgb(200pt)=(0.96201757,0.58647675,0.42491751);
  rgb(201pt)=(0.96228344,0.59304598,0.43145271);
  rgb(202pt)=(0.96253168,0.5995944,0.43805131);
  rgb(203pt)=(0.96276513,0.60612062,0.44471698);
  rgb(204pt)=(0.96298491,0.6126247,0.45145074);
  rgb(205pt)=(0.96318967,0.61910879,0.45824902);
  rgb(206pt)=(0.96337949,0.6255736,0.46511271);
  rgb(207pt)=(0.96355923,0.63201624,0.47204746);
  rgb(208pt)=(0.96372785,0.63843852,0.47905028);
  rgb(209pt)=(0.96388426,0.64484214,0.4861196);
  rgb(210pt)=(0.96403203,0.65122535,0.4932578);
  rgb(211pt)=(0.96417332,0.65758729,0.50046894);
  rgb(212pt)=(0.9643063,0.66393045,0.5077467);
  rgb(213pt)=(0.96443322,0.67025402,0.51509334);
  rgb(214pt)=(0.96455845,0.67655564,0.52251447);
  rgb(215pt)=(0.96467922,0.68283846,0.53000231);
  rgb(216pt)=(0.96479861,0.68910113,0.53756026);
  rgb(217pt)=(0.96492035,0.69534192,0.5451917);
  rgb(218pt)=(0.96504223,0.7015636,0.5528892);
  rgb(219pt)=(0.96516917,0.70776351,0.5606593);
  rgb(220pt)=(0.96530224,0.71394212,0.56849894);
  rgb(221pt)=(0.96544032,0.72010124,0.57640375);
  rgb(222pt)=(0.96559206,0.72623592,0.58438387);
  rgb(223pt)=(0.96575293,0.73235058,0.59242739);
  rgb(224pt)=(0.96592829,0.73844258,0.60053991);
  rgb(225pt)=(0.96612013,0.74451182,0.60871954);
  rgb(226pt)=(0.96632832,0.75055966,0.61696136);
  rgb(227pt)=(0.96656022,0.75658231,0.62527295);
  rgb(228pt)=(0.96681185,0.76258381,0.63364277);
  rgb(229pt)=(0.96709183,0.76855969,0.64207921);
  rgb(230pt)=(0.96739773,0.77451297,0.65057302);
  rgb(231pt)=(0.96773482,0.78044149,0.65912731);
  rgb(232pt)=(0.96810471,0.78634563,0.66773889);
  rgb(233pt)=(0.96850919,0.79222565,0.6764046);
  rgb(234pt)=(0.96893132,0.79809112,0.68512266);
  rgb(235pt)=(0.96935926,0.80395415,0.69383201);
  rgb(236pt)=(0.9698028,0.80981139,0.70252255);
  rgb(237pt)=(0.97025511,0.81566605,0.71120296);
  rgb(238pt)=(0.97071849,0.82151775,0.71987163);
  rgb(239pt)=(0.97120159,0.82736371,0.72851999);
  rgb(240pt)=(0.97169389,0.83320847,0.73716071);
  rgb(241pt)=(0.97220061,0.83905052,0.74578903);
  rgb(242pt)=(0.97272597,0.84488881,0.75440141);
  rgb(243pt)=(0.97327085,0.85072354,0.76299805);
  rgb(244pt)=(0.97383206,0.85655639,0.77158353);
  rgb(245pt)=(0.97441222,0.86238689,0.78015619);
  rgb(246pt)=(0.97501782,0.86821321,0.78871034);
  rgb(247pt)=(0.97564391,0.87403763,0.79725261);
  rgb(248pt)=(0.97628674,0.87986189,0.8057883);
  rgb(249pt)=(0.97696114,0.88568129,0.81430324);
  rgb(250pt)=(0.97765722,0.89149971,0.82280948);
  rgb(251pt)=(0.97837585,0.89731727,0.83130786);
  rgb(252pt)=(0.97912374,0.90313207,0.83979337);
  rgb(253pt)=(0.979891,0.90894778,0.84827858);
  rgb(254pt)=(0.98067764,0.91476465,0.85676611);
  rgb(255pt)=(0.98137749,0.92061729,0.86536915)
}]
table{%
x                      y
1 0.0043554597115259
1 0.0140434276573501
1 0.233033753920295
1 0.0612251193839642
1 0.0333420371606915
1 0.0311900233057824
1 0.038179737841062
1 0.0380214909647225
1 0.00672543981503899
1 0.0574923044155954
1 0.0888121404596463
1 0.0991057590887079
2 0.150893556587811
2 0.321942530382363
2 0.335541190579131
2 0.32053429405597
2 0.31085609399073
2 0.202361708913909
2 0.672457263559375
2 0.210687423323507
2 0.126006242961989
2 0.176977678212634
2 0.159408276168467
2 0.0258753840562574
3 0.0242152336024283
3 0.354565184301253
3 0.474097082990665
3 0.281270854330086
3 0.345626392578116
3 0.263826417070589
3 0.0536316320828226
3 0.228198387207596
3 0.305778417755038
3 0.031013237208453
3 0.275472097374836
3 0.231703248234764
4 0.174461216076719
4 0.0966833327053631
4 0.12051378121267
4 0.110311226127369
4 0.0955299464533932
4 0.226528593867168
4 0.125858714657935
4 0.170247555640869
4 0.127912755837574
4 0.22758498274024
4 0.130961402814891
4 0.143519615669102
5 0.00503725149364125
5 0.00374978733443077
5 0.0384507844592434
5 0.0208279032973212
5 0.00901955623308388
5 0.0264408596567353
5 0.0215030633956964
5 0.0350918670501071
5 0.0288391139328089
5 0.0385997744464781
5 0.0350873303929574
5 0.0250413086824384
6 0.00687271819912277
6 0.0150500617254493
6 0.0108599105817176
6 0.0092197858828677
6 0.0261551002834874
6 0.0121847753943651
6 0.0172131952367762
6 0.0141749434330692
6 0.00495822483387332
6 0.00347079924174293
6 0.00808241151258855
6 0.0226247591944673
7 0.00547721316514681
7 0.00936986152273173
7 0.00405677801754196
7 0.00314594555606768
7 0.00142554959711452
7 0.00387163810705215
7 0.00827108664883001
7 0.00599795999847178
7 0.0784324989086093
7 0.00695206288334857
7 0.0108502814210045
7 0.010288832513725
8 0.00319041509716107
8 0.00416050271292946
8 0.00518183522611613
8 0.00281605828910781
8 0.00850335811045745
8 0.00712886684107193
8 0.00391228459046295
8 0.00430512953515084
8 0.0275596691764748
8 0.00839783366324481
8 0.0041466714221021
8 0.00419883478490291
9 0.0240252235098372
9 0.0167360742658696
9 0.00447911879444248
9 0.0129395208448405
9 0.00220585216211144
9 0.00245011592229034
9 0.00824054392552725
9 0.00335621252357335
9 0.022823358755564
9 0.0128260620292262
9 0.00230885740164678
9 0.0120330185848221
10 0.00562166944867096
10 0.0268875715826653
10 0.0173624186553562
10 0.0120855334232284
10 0.00390107816222841
10 0.00313259738853028
10 0.040387192934026
10 0.00237108986260561
10 0.0173638493610621
10 0.0065684122022492
10 0.0142336631966271
10 0.0257768903224732
11 0.0213284999746768
11 0.000631923987020593
11 0.0134832764089886
11 0.0145261938784669
11 0.00965750907316159
11 0.000678687347516543
11 0.00903840228803898
11 0.00959256616986269
11 0.00918648787434269
11 0.00118533853175528
11 0.171920388227558
11 0.00397227950396499
12 0.00609480659023726
12 0.00356934784982759
12 0.00652962022285356
12 0.00325373712115401
12 0.00939966321483203
12 0.0024574526584738
12 0.00984578064157758
12 0.00244130746216398
12 0.0022602774302829
12 0.0121493408125023
12 0.00220162671493061
12 0.0011022305393071
};
\addlegendentry{[CLS]}
\legend{}
\addplot [only marks, mark size=1.25pt,draw=color1, fill=color1, colormap={mymap}{[1pt]
 rgb(0pt)=(0.01060815,0.01808215,0.10018654);
  rgb(1pt)=(0.01428972,0.02048237,0.10374486);
  rgb(2pt)=(0.01831941,0.0229766,0.10738511);
  rgb(3pt)=(0.02275049,0.02554464,0.11108639);
  rgb(4pt)=(0.02759119,0.02818316,0.11483751);
  rgb(5pt)=(0.03285175,0.03088792,0.11863035);
  rgb(6pt)=(0.03853466,0.03365771,0.12245873);
  rgb(7pt)=(0.04447016,0.03648425,0.12631831);
  rgb(8pt)=(0.05032105,0.03936808,0.13020508);
  rgb(9pt)=(0.05611171,0.04224835,0.13411624);
  rgb(10pt)=(0.0618531,0.04504866,0.13804929);
  rgb(11pt)=(0.06755457,0.04778179,0.14200206);
  rgb(12pt)=(0.0732236,0.05045047,0.14597263);
  rgb(13pt)=(0.0788708,0.05305461,0.14995981);
  rgb(14pt)=(0.08450105,0.05559631,0.15396203);
  rgb(15pt)=(0.09011319,0.05808059,0.15797687);
  rgb(16pt)=(0.09572396,0.06050127,0.16200507);
  rgb(17pt)=(0.10132312,0.06286782,0.16604287);
  rgb(18pt)=(0.10692823,0.06517224,0.17009175);
  rgb(19pt)=(0.1125315,0.06742194,0.17414848);
  rgb(20pt)=(0.11813947,0.06961499,0.17821272);
  rgb(21pt)=(0.12375803,0.07174938,0.18228425);
  rgb(22pt)=(0.12938228,0.07383015,0.18636053);
  rgb(23pt)=(0.13501631,0.07585609,0.19044109);
  rgb(24pt)=(0.14066867,0.0778224,0.19452676);
  rgb(25pt)=(0.14633406,0.07973393,0.1986151);
  rgb(26pt)=(0.15201338,0.08159108,0.20270523);
  rgb(27pt)=(0.15770877,0.08339312,0.20679668);
  rgb(28pt)=(0.16342174,0.0851396,0.21088893);
  rgb(29pt)=(0.16915387,0.08682996,0.21498104);
  rgb(30pt)=(0.17489524,0.08848235,0.2190294);
  rgb(31pt)=(0.18065495,0.09009031,0.22303512);
  rgb(32pt)=(0.18643324,0.09165431,0.22699705);
  rgb(33pt)=(0.19223028,0.09317479,0.23091409);
  rgb(34pt)=(0.19804623,0.09465217,0.23478512);
  rgb(35pt)=(0.20388117,0.09608689,0.23860907);
  rgb(36pt)=(0.20973515,0.09747934,0.24238489);
  rgb(37pt)=(0.21560818,0.09882993,0.24611154);
  rgb(38pt)=(0.22150014,0.10013944,0.2497868);
  rgb(39pt)=(0.22741085,0.10140876,0.25340813);
  rgb(40pt)=(0.23334047,0.10263737,0.25697736);
  rgb(41pt)=(0.23928891,0.10382562,0.2604936);
  rgb(42pt)=(0.24525608,0.10497384,0.26395596);
  rgb(43pt)=(0.25124182,0.10608236,0.26736359);
  rgb(44pt)=(0.25724602,0.10715148,0.27071569);
  rgb(45pt)=(0.26326851,0.1081815,0.27401148);
  rgb(46pt)=(0.26930915,0.1091727,0.2772502);
  rgb(47pt)=(0.27536766,0.11012568,0.28043021);
  rgb(48pt)=(0.28144375,0.11104133,0.2835489);
  rgb(49pt)=(0.2875374,0.11191896,0.28660853);
  rgb(50pt)=(0.29364846,0.11275876,0.2896085);
  rgb(51pt)=(0.29977678,0.11356089,0.29254823);
  rgb(52pt)=(0.30592213,0.11432553,0.29542718);
  rgb(53pt)=(0.31208435,0.11505284,0.29824485);
  rgb(54pt)=(0.31826327,0.1157429,0.30100076);
  rgb(55pt)=(0.32445869,0.11639585,0.30369448);
  rgb(56pt)=(0.33067031,0.11701189,0.30632563);
  rgb(57pt)=(0.33689808,0.11759095,0.3088938);
  rgb(58pt)=(0.34314168,0.11813362,0.31139721);
  rgb(59pt)=(0.34940101,0.11863987,0.3138355);
  rgb(60pt)=(0.355676,0.11910909,0.31620996);
  rgb(61pt)=(0.36196644,0.1195413,0.31852037);
  rgb(62pt)=(0.36827206,0.11993653,0.32076656);
  rgb(63pt)=(0.37459292,0.12029443,0.32294825);
  rgb(64pt)=(0.38092887,0.12061482,0.32506528);
  rgb(65pt)=(0.38727975,0.12089756,0.3271175);
  rgb(66pt)=(0.39364518,0.12114272,0.32910494);
  rgb(67pt)=(0.40002537,0.12134964,0.33102734);
  rgb(68pt)=(0.40642019,0.12151801,0.33288464);
  rgb(69pt)=(0.41282936,0.12164769,0.33467689);
  rgb(70pt)=(0.41925278,0.12173833,0.33640407);
  rgb(71pt)=(0.42569057,0.12178916,0.33806605);
  rgb(72pt)=(0.43214263,0.12179973,0.33966284);
  rgb(73pt)=(0.43860848,0.12177004,0.34119475);
  rgb(74pt)=(0.44508855,0.12169883,0.34266151);
  rgb(75pt)=(0.45158266,0.12158557,0.34406324);
  rgb(76pt)=(0.45809049,0.12142996,0.34540024);
  rgb(77pt)=(0.46461238,0.12123063,0.34667231);
  rgb(78pt)=(0.47114798,0.12098721,0.34787978);
  rgb(79pt)=(0.47769736,0.12069864,0.34902273);
  rgb(80pt)=(0.48426077,0.12036349,0.35010104);
  rgb(81pt)=(0.49083761,0.11998161,0.35111537);
  rgb(82pt)=(0.49742847,0.11955087,0.35206533);
  rgb(83pt)=(0.50403286,0.11907081,0.35295152);
  rgb(84pt)=(0.51065109,0.11853959,0.35377385);
  rgb(85pt)=(0.51728314,0.1179558,0.35453252);
  rgb(86pt)=(0.52392883,0.11731817,0.35522789);
  rgb(87pt)=(0.53058853,0.11662445,0.35585982);
  rgb(88pt)=(0.53726173,0.11587369,0.35642903);
  rgb(89pt)=(0.54394898,0.11506307,0.35693521);
  rgb(90pt)=(0.5506426,0.11420757,0.35737863);
  rgb(91pt)=(0.55734473,0.11330456,0.35775059);
  rgb(92pt)=(0.56405586,0.11235265,0.35804813);
  rgb(93pt)=(0.57077365,0.11135597,0.35827146);
  rgb(94pt)=(0.5774991,0.11031233,0.35841679);
  rgb(95pt)=(0.58422945,0.10922707,0.35848469);
  rgb(96pt)=(0.59096382,0.10810205,0.35847347);
  rgb(97pt)=(0.59770215,0.10693774,0.35838029);
  rgb(98pt)=(0.60444226,0.10573912,0.35820487);
  rgb(99pt)=(0.61118304,0.10450943,0.35794557);
  rgb(100pt)=(0.61792306,0.10325288,0.35760108);
  rgb(101pt)=(0.62466162,0.10197244,0.35716891);
  rgb(102pt)=(0.63139686,0.10067417,0.35664819);
  rgb(103pt)=(0.63812122,0.09938212,0.35603757);
  rgb(104pt)=(0.64483795,0.0980891,0.35533555);
  rgb(105pt)=(0.65154562,0.09680192,0.35454107);
  rgb(106pt)=(0.65824241,0.09552918,0.3536529);
  rgb(107pt)=(0.66492652,0.09428017,0.3526697);
  rgb(108pt)=(0.67159578,0.09306598,0.35159077);
  rgb(109pt)=(0.67824099,0.09192342,0.3504148);
  rgb(110pt)=(0.684863,0.09085633,0.34914061);
  rgb(111pt)=(0.69146268,0.0898675,0.34776864);
  rgb(112pt)=(0.69803757,0.08897226,0.3462986);
  rgb(113pt)=(0.70457834,0.0882129,0.34473046);
  rgb(114pt)=(0.71108138,0.08761223,0.3430635);
  rgb(115pt)=(0.7175507,0.08716212,0.34129974);
  rgb(116pt)=(0.72398193,0.08688725,0.33943958);
  rgb(117pt)=(0.73035829,0.0868623,0.33748452);
  rgb(118pt)=(0.73669146,0.08704683,0.33543669);
  rgb(119pt)=(0.74297501,0.08747196,0.33329799);
  rgb(120pt)=(0.74919318,0.08820542,0.33107204);
  rgb(121pt)=(0.75535825,0.08919792,0.32876184);
  rgb(122pt)=(0.76145589,0.09050716,0.32637117);
  rgb(123pt)=(0.76748424,0.09213602,0.32390525);
  rgb(124pt)=(0.77344838,0.09405684,0.32136808);
  rgb(125pt)=(0.77932641,0.09634794,0.31876642);
  rgb(126pt)=(0.78513609,0.09892473,0.31610488);
  rgb(127pt)=(0.79085854,0.10184672,0.313391);
  rgb(128pt)=(0.7965014,0.10506637,0.31063031);
  rgb(129pt)=(0.80205987,0.10858333,0.30783);
  rgb(130pt)=(0.80752799,0.11239964,0.30499738);
  rgb(131pt)=(0.81291606,0.11645784,0.30213802);
  rgb(132pt)=(0.81820481,0.12080606,0.29926105);
  rgb(133pt)=(0.82341472,0.12535343,0.2963705);
  rgb(134pt)=(0.82852822,0.13014118,0.29347474);
  rgb(135pt)=(0.83355779,0.13511035,0.29057852);
  rgb(136pt)=(0.83850183,0.14025098,0.2876878);
  rgb(137pt)=(0.84335441,0.14556683,0.28480819);
  rgb(138pt)=(0.84813096,0.15099892,0.281943);
  rgb(139pt)=(0.85281737,0.15657772,0.27909826);
  rgb(140pt)=(0.85742602,0.1622583,0.27627462);
  rgb(141pt)=(0.86196552,0.16801239,0.27346473);
  rgb(142pt)=(0.86641628,0.17387796,0.27070818);
  rgb(143pt)=(0.87079129,0.17982114,0.26797378);
  rgb(144pt)=(0.87507281,0.18587368,0.26529697);
  rgb(145pt)=(0.87925878,0.19203259,0.26268136);
  rgb(146pt)=(0.8833417,0.19830556,0.26014181);
  rgb(147pt)=(0.88731387,0.20469941,0.25769539);
  rgb(148pt)=(0.89116859,0.21121788,0.2553592);
  rgb(149pt)=(0.89490337,0.21785614,0.25314362);
  rgb(150pt)=(0.8985026,0.22463251,0.25108745);
  rgb(151pt)=(0.90197527,0.23152063,0.24918223);
  rgb(152pt)=(0.90530097,0.23854541,0.24748098);
  rgb(153pt)=(0.90848638,0.24568473,0.24598324);
  rgb(154pt)=(0.911533,0.25292623,0.24470258);
  rgb(155pt)=(0.9144225,0.26028902,0.24369359);
  rgb(156pt)=(0.91717106,0.26773821,0.24294137);
  rgb(157pt)=(0.91978131,0.27526191,0.24245973);
  rgb(158pt)=(0.92223947,0.28287251,0.24229568);
  rgb(159pt)=(0.92456587,0.29053388,0.24242622);
  rgb(160pt)=(0.92676657,0.29823282,0.24285536);
  rgb(161pt)=(0.92882964,0.30598085,0.24362274);
  rgb(162pt)=(0.93078135,0.31373977,0.24468803);
  rgb(163pt)=(0.93262051,0.3215093,0.24606461);
  rgb(164pt)=(0.93435067,0.32928362,0.24775328);
  rgb(165pt)=(0.93599076,0.33703942,0.24972157);
  rgb(166pt)=(0.93752831,0.34479177,0.25199928);
  rgb(167pt)=(0.93899289,0.35250734,0.25452808);
  rgb(168pt)=(0.94036561,0.36020899,0.25734661);
  rgb(169pt)=(0.94167588,0.36786594,0.2603949);
  rgb(170pt)=(0.94291042,0.37549479,0.26369821);
  rgb(171pt)=(0.94408513,0.3830811,0.26722004);
  rgb(172pt)=(0.94520419,0.39062329,0.27094924);
  rgb(173pt)=(0.94625977,0.39813168,0.27489742);
  rgb(174pt)=(0.94727016,0.4055909,0.27902322);
  rgb(175pt)=(0.94823505,0.41300424,0.28332283);
  rgb(176pt)=(0.94914549,0.42038251,0.28780969);
  rgb(177pt)=(0.95001704,0.42771398,0.29244728);
  rgb(178pt)=(0.95085121,0.43500005,0.29722817);
  rgb(179pt)=(0.95165009,0.44224144,0.30214494);
  rgb(180pt)=(0.9524044,0.44944853,0.3072105);
  rgb(181pt)=(0.95312556,0.45661389,0.31239776);
  rgb(182pt)=(0.95381595,0.46373781,0.31769923);
  rgb(183pt)=(0.95447591,0.47082238,0.32310953);
  rgb(184pt)=(0.95510255,0.47787236,0.32862553);
  rgb(185pt)=(0.95569679,0.48489115,0.33421404);
  rgb(186pt)=(0.95626788,0.49187351,0.33985601);
  rgb(187pt)=(0.95681685,0.49882008,0.34555431);
  rgb(188pt)=(0.9573439,0.50573243,0.35130912);
  rgb(189pt)=(0.95784842,0.51261283,0.35711942);
  rgb(190pt)=(0.95833051,0.51946267,0.36298589);
  rgb(191pt)=(0.95879054,0.52628305,0.36890904);
  rgb(192pt)=(0.95922872,0.53307513,0.3748895);
  rgb(193pt)=(0.95964538,0.53983991,0.38092784);
  rgb(194pt)=(0.96004345,0.54657593,0.3870292);
  rgb(195pt)=(0.96042097,0.55328624,0.39319057);
  rgb(196pt)=(0.96077819,0.55997184,0.39941173);
  rgb(197pt)=(0.9611152,0.5666337,0.40569343);
  rgb(198pt)=(0.96143273,0.57327231,0.41203603);
  rgb(199pt)=(0.96173392,0.57988594,0.41844491);
  rgb(200pt)=(0.96201757,0.58647675,0.42491751);
  rgb(201pt)=(0.96228344,0.59304598,0.43145271);
  rgb(202pt)=(0.96253168,0.5995944,0.43805131);
  rgb(203pt)=(0.96276513,0.60612062,0.44471698);
  rgb(204pt)=(0.96298491,0.6126247,0.45145074);
  rgb(205pt)=(0.96318967,0.61910879,0.45824902);
  rgb(206pt)=(0.96337949,0.6255736,0.46511271);
  rgb(207pt)=(0.96355923,0.63201624,0.47204746);
  rgb(208pt)=(0.96372785,0.63843852,0.47905028);
  rgb(209pt)=(0.96388426,0.64484214,0.4861196);
  rgb(210pt)=(0.96403203,0.65122535,0.4932578);
  rgb(211pt)=(0.96417332,0.65758729,0.50046894);
  rgb(212pt)=(0.9643063,0.66393045,0.5077467);
  rgb(213pt)=(0.96443322,0.67025402,0.51509334);
  rgb(214pt)=(0.96455845,0.67655564,0.52251447);
  rgb(215pt)=(0.96467922,0.68283846,0.53000231);
  rgb(216pt)=(0.96479861,0.68910113,0.53756026);
  rgb(217pt)=(0.96492035,0.69534192,0.5451917);
  rgb(218pt)=(0.96504223,0.7015636,0.5528892);
  rgb(219pt)=(0.96516917,0.70776351,0.5606593);
  rgb(220pt)=(0.96530224,0.71394212,0.56849894);
  rgb(221pt)=(0.96544032,0.72010124,0.57640375);
  rgb(222pt)=(0.96559206,0.72623592,0.58438387);
  rgb(223pt)=(0.96575293,0.73235058,0.59242739);
  rgb(224pt)=(0.96592829,0.73844258,0.60053991);
  rgb(225pt)=(0.96612013,0.74451182,0.60871954);
  rgb(226pt)=(0.96632832,0.75055966,0.61696136);
  rgb(227pt)=(0.96656022,0.75658231,0.62527295);
  rgb(228pt)=(0.96681185,0.76258381,0.63364277);
  rgb(229pt)=(0.96709183,0.76855969,0.64207921);
  rgb(230pt)=(0.96739773,0.77451297,0.65057302);
  rgb(231pt)=(0.96773482,0.78044149,0.65912731);
  rgb(232pt)=(0.96810471,0.78634563,0.66773889);
  rgb(233pt)=(0.96850919,0.79222565,0.6764046);
  rgb(234pt)=(0.96893132,0.79809112,0.68512266);
  rgb(235pt)=(0.96935926,0.80395415,0.69383201);
  rgb(236pt)=(0.9698028,0.80981139,0.70252255);
  rgb(237pt)=(0.97025511,0.81566605,0.71120296);
  rgb(238pt)=(0.97071849,0.82151775,0.71987163);
  rgb(239pt)=(0.97120159,0.82736371,0.72851999);
  rgb(240pt)=(0.97169389,0.83320847,0.73716071);
  rgb(241pt)=(0.97220061,0.83905052,0.74578903);
  rgb(242pt)=(0.97272597,0.84488881,0.75440141);
  rgb(243pt)=(0.97327085,0.85072354,0.76299805);
  rgb(244pt)=(0.97383206,0.85655639,0.77158353);
  rgb(245pt)=(0.97441222,0.86238689,0.78015619);
  rgb(246pt)=(0.97501782,0.86821321,0.78871034);
  rgb(247pt)=(0.97564391,0.87403763,0.79725261);
  rgb(248pt)=(0.97628674,0.87986189,0.8057883);
  rgb(249pt)=(0.97696114,0.88568129,0.81430324);
  rgb(250pt)=(0.97765722,0.89149971,0.82280948);
  rgb(251pt)=(0.97837585,0.89731727,0.83130786);
  rgb(252pt)=(0.97912374,0.90313207,0.83979337);
  rgb(253pt)=(0.979891,0.90894778,0.84827858);
  rgb(254pt)=(0.98067764,0.91476465,0.85676611);
  rgb(255pt)=(0.98137749,0.92061729,0.86536915)
}]
table{%
x                      y
1 0.0119738167441038
1 0.00481528411251306
1 0.0231263766377224
1 0.0224372668198672
1 0.051463203653159
1 0.0143046016300551
1 0.0169490673215344
1 0.0363554725966033
1 0.0151173261554377
1 0.0260289647358895
1 0.0235791283180186
1 0.016881252207167
2 0.0710920303316474
2 0.03559855041082
2 0.0125914184704892
2 0.0322792238558419
2 0.0167999998477631
2 0.081433937097298
2 0.0594367157469151
2 0.0403876320624088
2 0.0655787768140855
2 0.0734267160975636
2 0.0393991824424279
2 0.0387818025578567
3 0.0331378118017053
3 0.129978503330233
3 0.101715696631371
3 0.0972758142426587
3 0.0677775265678753
3 0.0588455070373554
3 0.109924032030926
3 0.0605138938777986
3 0.083806461341291
3 0.0308633465899031
3 0.059832104701074
3 0.0611143541504942
4 0.29474368191961
4 0.173961574996547
4 0.216009231139509
4 0.22493942575395
4 0.305109621915698
4 0.0252543280108716
4 0.201958479066985
4 0.196368201744569
4 0.226569506187228
4 0.186723140629571
4 0.236096618695849
4 0.199308359500126
5 0.477996401867863
5 0.345469731445003
5 0.567727332864125
5 0.680942051306564
5 0.296093682326518
5 0.385097147238212
5 0.462416802530572
5 0.619188564443703
5 0.514631787673652
5 0.389562609784229
5 0.40049094361766
5 0.575036835981636
6 0.490589668210014
6 0.796797043067523
6 0.50014045572983
6 0.532520307113165
6 0.389067337932249
6 0.716510229006247
6 0.748769617946848
6 0.673862702159002
6 0.513139820103132
6 0.215780724380772
6 0.335638853971867
6 0.657874531666902
7 0.665735877456442
7 0.596669612513307
7 0.553016405081338
7 0.761969481877571
7 0.570127021418804
7 0.533284530920836
7 0.533026398065321
7 0.440916522408051
7 0.500811626129154
7 0.275650663955503
7 0.641879110382527
7 0.28153842533063
8 0.274233385394017
8 0.574355772760966
8 0.531984040917416
8 0.86870870711695
8 0.232054798743903
8 0.478668330099042
8 0.880102800678015
8 0.870024346555573
8 0.340659059615694
8 0.504021562183433
8 0.451065720725904
8 0.789308144233416
9 0.719154456146203
9 0.449014270716174
9 0.723153213098295
9 0.568350211544789
9 0.685758233950577
9 0.753525564871087
9 0.765265530213319
9 0.428042997793865
9 0.357513174123349
9 0.407880094377247
9 0.497593852939136
9 0.448874478140817
10 0.479505979480945
10 0.530040324910196
10 0.360808128737238
10 0.561603217268906
10 0.632381384240237
10 0.345224543794002
10 0.528821158338855
10 0.556751155978165
10 0.248017644497002
10 0.692935954872566
10 0.457048335278082
10 0.399455469730359
11 0.0657509695286996
11 0.0306600656048803
11 0.0487696602265842
11 0.0787291433458245
11 0.0393309960501332
11 0.0353961285507781
11 0.0331832412885423
11 0.0579462606274138
11 0.0571802594440903
11 0.036814966059787
11 0.254144189606076
11 0.0349844453409045
12 0.0797878156864304
12 0.0762801506817307
12 0.061293759325923
12 0.0842745410437579
12 0.0700812442477203
12 0.059497684844985
12 0.0803133313841579
12 0.0703264917206904
12 0.0245991261602268
12 0.0653855396775036
12 0.0745723013811915
12 0.068459028212871
};
\addlegendentry{[SEP]}
\legend{}
\addplot [only marks, draw=color2,mark size=1.25pt, fill=color2, colormap={mymap}{[1pt]
 rgb(0pt)=(0.01060815,0.01808215,0.10018654);
  rgb(1pt)=(0.01428972,0.02048237,0.10374486);
  rgb(2pt)=(0.01831941,0.0229766,0.10738511);
  rgb(3pt)=(0.02275049,0.02554464,0.11108639);
  rgb(4pt)=(0.02759119,0.02818316,0.11483751);
  rgb(5pt)=(0.03285175,0.03088792,0.11863035);
  rgb(6pt)=(0.03853466,0.03365771,0.12245873);
  rgb(7pt)=(0.04447016,0.03648425,0.12631831);
  rgb(8pt)=(0.05032105,0.03936808,0.13020508);
  rgb(9pt)=(0.05611171,0.04224835,0.13411624);
  rgb(10pt)=(0.0618531,0.04504866,0.13804929);
  rgb(11pt)=(0.06755457,0.04778179,0.14200206);
  rgb(12pt)=(0.0732236,0.05045047,0.14597263);
  rgb(13pt)=(0.0788708,0.05305461,0.14995981);
  rgb(14pt)=(0.08450105,0.05559631,0.15396203);
  rgb(15pt)=(0.09011319,0.05808059,0.15797687);
  rgb(16pt)=(0.09572396,0.06050127,0.16200507);
  rgb(17pt)=(0.10132312,0.06286782,0.16604287);
  rgb(18pt)=(0.10692823,0.06517224,0.17009175);
  rgb(19pt)=(0.1125315,0.06742194,0.17414848);
  rgb(20pt)=(0.11813947,0.06961499,0.17821272);
  rgb(21pt)=(0.12375803,0.07174938,0.18228425);
  rgb(22pt)=(0.12938228,0.07383015,0.18636053);
  rgb(23pt)=(0.13501631,0.07585609,0.19044109);
  rgb(24pt)=(0.14066867,0.0778224,0.19452676);
  rgb(25pt)=(0.14633406,0.07973393,0.1986151);
  rgb(26pt)=(0.15201338,0.08159108,0.20270523);
  rgb(27pt)=(0.15770877,0.08339312,0.20679668);
  rgb(28pt)=(0.16342174,0.0851396,0.21088893);
  rgb(29pt)=(0.16915387,0.08682996,0.21498104);
  rgb(30pt)=(0.17489524,0.08848235,0.2190294);
  rgb(31pt)=(0.18065495,0.09009031,0.22303512);
  rgb(32pt)=(0.18643324,0.09165431,0.22699705);
  rgb(33pt)=(0.19223028,0.09317479,0.23091409);
  rgb(34pt)=(0.19804623,0.09465217,0.23478512);
  rgb(35pt)=(0.20388117,0.09608689,0.23860907);
  rgb(36pt)=(0.20973515,0.09747934,0.24238489);
  rgb(37pt)=(0.21560818,0.09882993,0.24611154);
  rgb(38pt)=(0.22150014,0.10013944,0.2497868);
  rgb(39pt)=(0.22741085,0.10140876,0.25340813);
  rgb(40pt)=(0.23334047,0.10263737,0.25697736);
  rgb(41pt)=(0.23928891,0.10382562,0.2604936);
  rgb(42pt)=(0.24525608,0.10497384,0.26395596);
  rgb(43pt)=(0.25124182,0.10608236,0.26736359);
  rgb(44pt)=(0.25724602,0.10715148,0.27071569);
  rgb(45pt)=(0.26326851,0.1081815,0.27401148);
  rgb(46pt)=(0.26930915,0.1091727,0.2772502);
  rgb(47pt)=(0.27536766,0.11012568,0.28043021);
  rgb(48pt)=(0.28144375,0.11104133,0.2835489);
  rgb(49pt)=(0.2875374,0.11191896,0.28660853);
  rgb(50pt)=(0.29364846,0.11275876,0.2896085);
  rgb(51pt)=(0.29977678,0.11356089,0.29254823);
  rgb(52pt)=(0.30592213,0.11432553,0.29542718);
  rgb(53pt)=(0.31208435,0.11505284,0.29824485);
  rgb(54pt)=(0.31826327,0.1157429,0.30100076);
  rgb(55pt)=(0.32445869,0.11639585,0.30369448);
  rgb(56pt)=(0.33067031,0.11701189,0.30632563);
  rgb(57pt)=(0.33689808,0.11759095,0.3088938);
  rgb(58pt)=(0.34314168,0.11813362,0.31139721);
  rgb(59pt)=(0.34940101,0.11863987,0.3138355);
  rgb(60pt)=(0.355676,0.11910909,0.31620996);
  rgb(61pt)=(0.36196644,0.1195413,0.31852037);
  rgb(62pt)=(0.36827206,0.11993653,0.32076656);
  rgb(63pt)=(0.37459292,0.12029443,0.32294825);
  rgb(64pt)=(0.38092887,0.12061482,0.32506528);
  rgb(65pt)=(0.38727975,0.12089756,0.3271175);
  rgb(66pt)=(0.39364518,0.12114272,0.32910494);
  rgb(67pt)=(0.40002537,0.12134964,0.33102734);
  rgb(68pt)=(0.40642019,0.12151801,0.33288464);
  rgb(69pt)=(0.41282936,0.12164769,0.33467689);
  rgb(70pt)=(0.41925278,0.12173833,0.33640407);
  rgb(71pt)=(0.42569057,0.12178916,0.33806605);
  rgb(72pt)=(0.43214263,0.12179973,0.33966284);
  rgb(73pt)=(0.43860848,0.12177004,0.34119475);
  rgb(74pt)=(0.44508855,0.12169883,0.34266151);
  rgb(75pt)=(0.45158266,0.12158557,0.34406324);
  rgb(76pt)=(0.45809049,0.12142996,0.34540024);
  rgb(77pt)=(0.46461238,0.12123063,0.34667231);
  rgb(78pt)=(0.47114798,0.12098721,0.34787978);
  rgb(79pt)=(0.47769736,0.12069864,0.34902273);
  rgb(80pt)=(0.48426077,0.12036349,0.35010104);
  rgb(81pt)=(0.49083761,0.11998161,0.35111537);
  rgb(82pt)=(0.49742847,0.11955087,0.35206533);
  rgb(83pt)=(0.50403286,0.11907081,0.35295152);
  rgb(84pt)=(0.51065109,0.11853959,0.35377385);
  rgb(85pt)=(0.51728314,0.1179558,0.35453252);
  rgb(86pt)=(0.52392883,0.11731817,0.35522789);
  rgb(87pt)=(0.53058853,0.11662445,0.35585982);
  rgb(88pt)=(0.53726173,0.11587369,0.35642903);
  rgb(89pt)=(0.54394898,0.11506307,0.35693521);
  rgb(90pt)=(0.5506426,0.11420757,0.35737863);
  rgb(91pt)=(0.55734473,0.11330456,0.35775059);
  rgb(92pt)=(0.56405586,0.11235265,0.35804813);
  rgb(93pt)=(0.57077365,0.11135597,0.35827146);
  rgb(94pt)=(0.5774991,0.11031233,0.35841679);
  rgb(95pt)=(0.58422945,0.10922707,0.35848469);
  rgb(96pt)=(0.59096382,0.10810205,0.35847347);
  rgb(97pt)=(0.59770215,0.10693774,0.35838029);
  rgb(98pt)=(0.60444226,0.10573912,0.35820487);
  rgb(99pt)=(0.61118304,0.10450943,0.35794557);
  rgb(100pt)=(0.61792306,0.10325288,0.35760108);
  rgb(101pt)=(0.62466162,0.10197244,0.35716891);
  rgb(102pt)=(0.63139686,0.10067417,0.35664819);
  rgb(103pt)=(0.63812122,0.09938212,0.35603757);
  rgb(104pt)=(0.64483795,0.0980891,0.35533555);
  rgb(105pt)=(0.65154562,0.09680192,0.35454107);
  rgb(106pt)=(0.65824241,0.09552918,0.3536529);
  rgb(107pt)=(0.66492652,0.09428017,0.3526697);
  rgb(108pt)=(0.67159578,0.09306598,0.35159077);
  rgb(109pt)=(0.67824099,0.09192342,0.3504148);
  rgb(110pt)=(0.684863,0.09085633,0.34914061);
  rgb(111pt)=(0.69146268,0.0898675,0.34776864);
  rgb(112pt)=(0.69803757,0.08897226,0.3462986);
  rgb(113pt)=(0.70457834,0.0882129,0.34473046);
  rgb(114pt)=(0.71108138,0.08761223,0.3430635);
  rgb(115pt)=(0.7175507,0.08716212,0.34129974);
  rgb(116pt)=(0.72398193,0.08688725,0.33943958);
  rgb(117pt)=(0.73035829,0.0868623,0.33748452);
  rgb(118pt)=(0.73669146,0.08704683,0.33543669);
  rgb(119pt)=(0.74297501,0.08747196,0.33329799);
  rgb(120pt)=(0.74919318,0.08820542,0.33107204);
  rgb(121pt)=(0.75535825,0.08919792,0.32876184);
  rgb(122pt)=(0.76145589,0.09050716,0.32637117);
  rgb(123pt)=(0.76748424,0.09213602,0.32390525);
  rgb(124pt)=(0.77344838,0.09405684,0.32136808);
  rgb(125pt)=(0.77932641,0.09634794,0.31876642);
  rgb(126pt)=(0.78513609,0.09892473,0.31610488);
  rgb(127pt)=(0.79085854,0.10184672,0.313391);
  rgb(128pt)=(0.7965014,0.10506637,0.31063031);
  rgb(129pt)=(0.80205987,0.10858333,0.30783);
  rgb(130pt)=(0.80752799,0.11239964,0.30499738);
  rgb(131pt)=(0.81291606,0.11645784,0.30213802);
  rgb(132pt)=(0.81820481,0.12080606,0.29926105);
  rgb(133pt)=(0.82341472,0.12535343,0.2963705);
  rgb(134pt)=(0.82852822,0.13014118,0.29347474);
  rgb(135pt)=(0.83355779,0.13511035,0.29057852);
  rgb(136pt)=(0.83850183,0.14025098,0.2876878);
  rgb(137pt)=(0.84335441,0.14556683,0.28480819);
  rgb(138pt)=(0.84813096,0.15099892,0.281943);
  rgb(139pt)=(0.85281737,0.15657772,0.27909826);
  rgb(140pt)=(0.85742602,0.1622583,0.27627462);
  rgb(141pt)=(0.86196552,0.16801239,0.27346473);
  rgb(142pt)=(0.86641628,0.17387796,0.27070818);
  rgb(143pt)=(0.87079129,0.17982114,0.26797378);
  rgb(144pt)=(0.87507281,0.18587368,0.26529697);
  rgb(145pt)=(0.87925878,0.19203259,0.26268136);
  rgb(146pt)=(0.8833417,0.19830556,0.26014181);
  rgb(147pt)=(0.88731387,0.20469941,0.25769539);
  rgb(148pt)=(0.89116859,0.21121788,0.2553592);
  rgb(149pt)=(0.89490337,0.21785614,0.25314362);
  rgb(150pt)=(0.8985026,0.22463251,0.25108745);
  rgb(151pt)=(0.90197527,0.23152063,0.24918223);
  rgb(152pt)=(0.90530097,0.23854541,0.24748098);
  rgb(153pt)=(0.90848638,0.24568473,0.24598324);
  rgb(154pt)=(0.911533,0.25292623,0.24470258);
  rgb(155pt)=(0.9144225,0.26028902,0.24369359);
  rgb(156pt)=(0.91717106,0.26773821,0.24294137);
  rgb(157pt)=(0.91978131,0.27526191,0.24245973);
  rgb(158pt)=(0.92223947,0.28287251,0.24229568);
  rgb(159pt)=(0.92456587,0.29053388,0.24242622);
  rgb(160pt)=(0.92676657,0.29823282,0.24285536);
  rgb(161pt)=(0.92882964,0.30598085,0.24362274);
  rgb(162pt)=(0.93078135,0.31373977,0.24468803);
  rgb(163pt)=(0.93262051,0.3215093,0.24606461);
  rgb(164pt)=(0.93435067,0.32928362,0.24775328);
  rgb(165pt)=(0.93599076,0.33703942,0.24972157);
  rgb(166pt)=(0.93752831,0.34479177,0.25199928);
  rgb(167pt)=(0.93899289,0.35250734,0.25452808);
  rgb(168pt)=(0.94036561,0.36020899,0.25734661);
  rgb(169pt)=(0.94167588,0.36786594,0.2603949);
  rgb(170pt)=(0.94291042,0.37549479,0.26369821);
  rgb(171pt)=(0.94408513,0.3830811,0.26722004);
  rgb(172pt)=(0.94520419,0.39062329,0.27094924);
  rgb(173pt)=(0.94625977,0.39813168,0.27489742);
  rgb(174pt)=(0.94727016,0.4055909,0.27902322);
  rgb(175pt)=(0.94823505,0.41300424,0.28332283);
  rgb(176pt)=(0.94914549,0.42038251,0.28780969);
  rgb(177pt)=(0.95001704,0.42771398,0.29244728);
  rgb(178pt)=(0.95085121,0.43500005,0.29722817);
  rgb(179pt)=(0.95165009,0.44224144,0.30214494);
  rgb(180pt)=(0.9524044,0.44944853,0.3072105);
  rgb(181pt)=(0.95312556,0.45661389,0.31239776);
  rgb(182pt)=(0.95381595,0.46373781,0.31769923);
  rgb(183pt)=(0.95447591,0.47082238,0.32310953);
  rgb(184pt)=(0.95510255,0.47787236,0.32862553);
  rgb(185pt)=(0.95569679,0.48489115,0.33421404);
  rgb(186pt)=(0.95626788,0.49187351,0.33985601);
  rgb(187pt)=(0.95681685,0.49882008,0.34555431);
  rgb(188pt)=(0.9573439,0.50573243,0.35130912);
  rgb(189pt)=(0.95784842,0.51261283,0.35711942);
  rgb(190pt)=(0.95833051,0.51946267,0.36298589);
  rgb(191pt)=(0.95879054,0.52628305,0.36890904);
  rgb(192pt)=(0.95922872,0.53307513,0.3748895);
  rgb(193pt)=(0.95964538,0.53983991,0.38092784);
  rgb(194pt)=(0.96004345,0.54657593,0.3870292);
  rgb(195pt)=(0.96042097,0.55328624,0.39319057);
  rgb(196pt)=(0.96077819,0.55997184,0.39941173);
  rgb(197pt)=(0.9611152,0.5666337,0.40569343);
  rgb(198pt)=(0.96143273,0.57327231,0.41203603);
  rgb(199pt)=(0.96173392,0.57988594,0.41844491);
  rgb(200pt)=(0.96201757,0.58647675,0.42491751);
  rgb(201pt)=(0.96228344,0.59304598,0.43145271);
  rgb(202pt)=(0.96253168,0.5995944,0.43805131);
  rgb(203pt)=(0.96276513,0.60612062,0.44471698);
  rgb(204pt)=(0.96298491,0.6126247,0.45145074);
  rgb(205pt)=(0.96318967,0.61910879,0.45824902);
  rgb(206pt)=(0.96337949,0.6255736,0.46511271);
  rgb(207pt)=(0.96355923,0.63201624,0.47204746);
  rgb(208pt)=(0.96372785,0.63843852,0.47905028);
  rgb(209pt)=(0.96388426,0.64484214,0.4861196);
  rgb(210pt)=(0.96403203,0.65122535,0.4932578);
  rgb(211pt)=(0.96417332,0.65758729,0.50046894);
  rgb(212pt)=(0.9643063,0.66393045,0.5077467);
  rgb(213pt)=(0.96443322,0.67025402,0.51509334);
  rgb(214pt)=(0.96455845,0.67655564,0.52251447);
  rgb(215pt)=(0.96467922,0.68283846,0.53000231);
  rgb(216pt)=(0.96479861,0.68910113,0.53756026);
  rgb(217pt)=(0.96492035,0.69534192,0.5451917);
  rgb(218pt)=(0.96504223,0.7015636,0.5528892);
  rgb(219pt)=(0.96516917,0.70776351,0.5606593);
  rgb(220pt)=(0.96530224,0.71394212,0.56849894);
  rgb(221pt)=(0.96544032,0.72010124,0.57640375);
  rgb(222pt)=(0.96559206,0.72623592,0.58438387);
  rgb(223pt)=(0.96575293,0.73235058,0.59242739);
  rgb(224pt)=(0.96592829,0.73844258,0.60053991);
  rgb(225pt)=(0.96612013,0.74451182,0.60871954);
  rgb(226pt)=(0.96632832,0.75055966,0.61696136);
  rgb(227pt)=(0.96656022,0.75658231,0.62527295);
  rgb(228pt)=(0.96681185,0.76258381,0.63364277);
  rgb(229pt)=(0.96709183,0.76855969,0.64207921);
  rgb(230pt)=(0.96739773,0.77451297,0.65057302);
  rgb(231pt)=(0.96773482,0.78044149,0.65912731);
  rgb(232pt)=(0.96810471,0.78634563,0.66773889);
  rgb(233pt)=(0.96850919,0.79222565,0.6764046);
  rgb(234pt)=(0.96893132,0.79809112,0.68512266);
  rgb(235pt)=(0.96935926,0.80395415,0.69383201);
  rgb(236pt)=(0.9698028,0.80981139,0.70252255);
  rgb(237pt)=(0.97025511,0.81566605,0.71120296);
  rgb(238pt)=(0.97071849,0.82151775,0.71987163);
  rgb(239pt)=(0.97120159,0.82736371,0.72851999);
  rgb(240pt)=(0.97169389,0.83320847,0.73716071);
  rgb(241pt)=(0.97220061,0.83905052,0.74578903);
  rgb(242pt)=(0.97272597,0.84488881,0.75440141);
  rgb(243pt)=(0.97327085,0.85072354,0.76299805);
  rgb(244pt)=(0.97383206,0.85655639,0.77158353);
  rgb(245pt)=(0.97441222,0.86238689,0.78015619);
  rgb(246pt)=(0.97501782,0.86821321,0.78871034);
  rgb(247pt)=(0.97564391,0.87403763,0.79725261);
  rgb(248pt)=(0.97628674,0.87986189,0.8057883);
  rgb(249pt)=(0.97696114,0.88568129,0.81430324);
  rgb(250pt)=(0.97765722,0.89149971,0.82280948);
  rgb(251pt)=(0.97837585,0.89731727,0.83130786);
  rgb(252pt)=(0.97912374,0.90313207,0.83979337);
  rgb(253pt)=(0.979891,0.90894778,0.84827858);
  rgb(254pt)=(0.98067764,0.91476465,0.85676611);
  rgb(255pt)=(0.98137749,0.92061729,0.86536915)
}]
table{%
x                      y
1 0.0694071093790224
1 0.0520541335627373
1 0.0749061862695763
1 0.0905140691164552
1 0.0525831339349438
1 0.0961060921767074
1 0.0607145923816982
1 0.0802103642521346
1 0.048278572966307
1 0.076975850712439
1 0.0663592914097157
1 0.0113502460544728
2 0.178090832500965
2 0.0637495999313755
2 0.0916155097633389
2 0.120570361712856
2 0.0635760247283323
2 0.144110914129537
2 0.046962719348625
2 0.174774434892485
2 0.109403175629263
2 0.082353986733731
2 0.113027479433548
2 0.10147826122212
3 0.0818740620711453
3 0.0693364241282379
3 0.078188840051543
3 0.0840423692451687
3 0.0804929125591556
3 0.104633030636581
3 0.0879857857587302
3 0.0795327377061828
3 0.115078603192214
3 0.0819262787993445
3 0.0521339853124618
3 0.0765558778459546
4 0.087904876428756
4 0.0754169275158252
4 0.190578343929705
4 0.0964965910329071
4 0.0352710298426224
4 0.0585464180598197
4 0.136279690365421
4 0.0915637401858562
4 0.061363066937046
4 0.0375010319682762
4 0.0301167737459386
4 0.093094458207819
5 0.0117595077847952
5 0.0711778817839806
5 0.0739500679267873
5 0.036032579151352
5 0.0327391395813299
5 0.0576622210235658
5 0.0383965184405257
5 0.0160123040660227
5 0.0939353396777261
5 0.0664921379312179
5 0.0585414795440272
5 0.0234898017545367
6 0.0465886886886735
6 0.0138042084649365
6 0.0648033570569845
6 0.0606269881652864
6 0.0674396377287396
6 0.033632030550307
6 0.0193122852811275
6 0.0184310525239701
6 0.0456123721180841
6 0.0401001547383266
6 0.0640578931626986
6 0.0532080881815374
7 0.0412351126756896
7 0.0571718880501789
7 0.060353300359191
7 0.0162946871706587
7 0.011468278831247
7 0.015794066785597
7 0.032996070799274
7 0.0485156116617015
7 0.0714084607195384
7 0.0956456587580265
7 0.0351614824133905
7 0.0632762816044467
8 0.145275464942205
8 0.0402679072315664
8 0.0425173696595889
8 0.00983360263215849
8 0.0770795539234448
8 0.013203633186484
8 0.00889086419183201
8 0.0274963049803473
8 0.186474557085936
8 0.051876620668283
8 0.0156840308414676
8 0.0248822802416389
9 0.0357821143817261
9 0.0876247293246044
9 0.0221053150306931
9 0.0624879602725948
9 0.0630729910354731
9 0.014080273686604
9 0.0343671052037108
9 0.0631983955019552
9 0.0835384538761429
9 0.060483909523091
9 0.015914875352931
9 0.0633804100278985
10 0.061443723156622
10 0.0525752719065199
10 0.103259816138076
10 0.0892400342532074
10 0.0451102904760936
10 0.0514826258044847
10 0.130960496732464
10 0.0148652971924757
10 0.10922818391906
10 0.0569523500380393
10 0.056919334554782
10 0.116941315289236
11 0.248537876810013
11 0.726508729879419
11 0.603490370775057
11 0.449536331639774
11 0.640221093477359
11 0.736677980616262
11 0.595484684496044
11 0.527253634305819
11 0.333232957571832
11 0.611189165538348
11 0.0571906716745234
11 0.50532113546371
12 0.655328543586595
12 0.651889027872299
12 0.374240684440409
12 0.676021584812553
12 0.621984928951113
12 0.540986783370448
12 0.668891407080791
12 0.704489279967912
12 0.185295544142634
12 0.351889827272457
12 0.678131420908801
12 0.710478943500407
};
\addlegendentry{. or ,}
\legend{}
\addplot [semithick, color0, forget plot]
table {%
1 0.0587938911436985
2 0.251128470232679
3 0.239116515394721
4 0.145842760316941
5 0.0239740500312452
6 0.012572223793294
7 0.0123449756949703
8 0.00695845495409852
9 0.0103686632266459
10 0.0146409972116436
11 0.0221001294387795
12 0.00510876593817856
};
\addplot [semithick, color1, forget plot]
table {%
1 0.0219193134110059
2 0.0472338321445931
3 0.0745654210252238
4 0.207253514130043
5 0.476221157589978
6 0.547557607607296
7 0.52955213962829
8 0.566265555752027
9 0.567010506492905
10 0.482716108093879
11 0.0644075271394762
12 0.0679059178639324
};
\addplot [semithick, color2, forget plot]
table {%
1 0.0649549701846841
2 0.107476108335515
3 0.0826484089422266
4 0.0828444123516661
5 0.0483490815554889
6 0.043968063055056
7 0.0457767416524117
8 0.0536235157987461
9 0.0505030444347854
10 0.0740815616217551
11 0.502887052687347
12 0.568302331325535
};
\end{axis}

\end{tikzpicture}

%% file: identifiability/avg_attn1.tex
% This file was created by tikzplotlib v0.8.2.
\begin{tikzpicture}

\definecolor{color0}{rgb}{0.905882352941176,0.298039215686275,0.235294117647059}
\definecolor{color1}{rgb}{0.203921568627451,0.596078431372549,0.858823529411765}
\definecolor{color2}{rgb}{0.607843137254902,0.349019607843137,0.713725490196078}

\begin{axis}[
axis line style={white!15.0!black},
legend cell align={left},
legend style={draw=white!80.0!black,font=\scriptsize},
tick align=outside,
tick pos=both,
x grid style={white!80.0!black},
xlabel={Layer},
xmajorgrids,
width=\linewidth,
height=\effectiveAttnPlotHeight,
xmin=0.366240817709798, xmax=12.6337591822902,
xtick style={color=white!15.0!black},
y grid style={white!80.0!black},
ylabel={Effective Attention},
ymajorgrids,
ymin=-0.0468060428997225, ymax=0.926731513806353,
ytick style={color=white!15.0!black}
]
\addplot [only marks, draw=color0, mark size=1.25pt, fill=color0, colormap={mymap}{[1pt]
 rgb(0pt)=(0.01060815,0.01808215,0.10018654);
  rgb(1pt)=(0.01428972,0.02048237,0.10374486);
  rgb(2pt)=(0.01831941,0.0229766,0.10738511);
  rgb(3pt)=(0.02275049,0.02554464,0.11108639);
  rgb(4pt)=(0.02759119,0.02818316,0.11483751);
  rgb(5pt)=(0.03285175,0.03088792,0.11863035);
  rgb(6pt)=(0.03853466,0.03365771,0.12245873);
  rgb(7pt)=(0.04447016,0.03648425,0.12631831);
  rgb(8pt)=(0.05032105,0.03936808,0.13020508);
  rgb(9pt)=(0.05611171,0.04224835,0.13411624);
  rgb(10pt)=(0.0618531,0.04504866,0.13804929);
  rgb(11pt)=(0.06755457,0.04778179,0.14200206);
  rgb(12pt)=(0.0732236,0.05045047,0.14597263);
  rgb(13pt)=(0.0788708,0.05305461,0.14995981);
  rgb(14pt)=(0.08450105,0.05559631,0.15396203);
  rgb(15pt)=(0.09011319,0.05808059,0.15797687);
  rgb(16pt)=(0.09572396,0.06050127,0.16200507);
  rgb(17pt)=(0.10132312,0.06286782,0.16604287);
  rgb(18pt)=(0.10692823,0.06517224,0.17009175);
  rgb(19pt)=(0.1125315,0.06742194,0.17414848);
  rgb(20pt)=(0.11813947,0.06961499,0.17821272);
  rgb(21pt)=(0.12375803,0.07174938,0.18228425);
  rgb(22pt)=(0.12938228,0.07383015,0.18636053);
  rgb(23pt)=(0.13501631,0.07585609,0.19044109);
  rgb(24pt)=(0.14066867,0.0778224,0.19452676);
  rgb(25pt)=(0.14633406,0.07973393,0.1986151);
  rgb(26pt)=(0.15201338,0.08159108,0.20270523);
  rgb(27pt)=(0.15770877,0.08339312,0.20679668);
  rgb(28pt)=(0.16342174,0.0851396,0.21088893);
  rgb(29pt)=(0.16915387,0.08682996,0.21498104);
  rgb(30pt)=(0.17489524,0.08848235,0.2190294);
  rgb(31pt)=(0.18065495,0.09009031,0.22303512);
  rgb(32pt)=(0.18643324,0.09165431,0.22699705);
  rgb(33pt)=(0.19223028,0.09317479,0.23091409);
  rgb(34pt)=(0.19804623,0.09465217,0.23478512);
  rgb(35pt)=(0.20388117,0.09608689,0.23860907);
  rgb(36pt)=(0.20973515,0.09747934,0.24238489);
  rgb(37pt)=(0.21560818,0.09882993,0.24611154);
  rgb(38pt)=(0.22150014,0.10013944,0.2497868);
  rgb(39pt)=(0.22741085,0.10140876,0.25340813);
  rgb(40pt)=(0.23334047,0.10263737,0.25697736);
  rgb(41pt)=(0.23928891,0.10382562,0.2604936);
  rgb(42pt)=(0.24525608,0.10497384,0.26395596);
  rgb(43pt)=(0.25124182,0.10608236,0.26736359);
  rgb(44pt)=(0.25724602,0.10715148,0.27071569);
  rgb(45pt)=(0.26326851,0.1081815,0.27401148);
  rgb(46pt)=(0.26930915,0.1091727,0.2772502);
  rgb(47pt)=(0.27536766,0.11012568,0.28043021);
  rgb(48pt)=(0.28144375,0.11104133,0.2835489);
  rgb(49pt)=(0.2875374,0.11191896,0.28660853);
  rgb(50pt)=(0.29364846,0.11275876,0.2896085);
  rgb(51pt)=(0.29977678,0.11356089,0.29254823);
  rgb(52pt)=(0.30592213,0.11432553,0.29542718);
  rgb(53pt)=(0.31208435,0.11505284,0.29824485);
  rgb(54pt)=(0.31826327,0.1157429,0.30100076);
  rgb(55pt)=(0.32445869,0.11639585,0.30369448);
  rgb(56pt)=(0.33067031,0.11701189,0.30632563);
  rgb(57pt)=(0.33689808,0.11759095,0.3088938);
  rgb(58pt)=(0.34314168,0.11813362,0.31139721);
  rgb(59pt)=(0.34940101,0.11863987,0.3138355);
  rgb(60pt)=(0.355676,0.11910909,0.31620996);
  rgb(61pt)=(0.36196644,0.1195413,0.31852037);
  rgb(62pt)=(0.36827206,0.11993653,0.32076656);
  rgb(63pt)=(0.37459292,0.12029443,0.32294825);
  rgb(64pt)=(0.38092887,0.12061482,0.32506528);
  rgb(65pt)=(0.38727975,0.12089756,0.3271175);
  rgb(66pt)=(0.39364518,0.12114272,0.32910494);
  rgb(67pt)=(0.40002537,0.12134964,0.33102734);
  rgb(68pt)=(0.40642019,0.12151801,0.33288464);
  rgb(69pt)=(0.41282936,0.12164769,0.33467689);
  rgb(70pt)=(0.41925278,0.12173833,0.33640407);
  rgb(71pt)=(0.42569057,0.12178916,0.33806605);
  rgb(72pt)=(0.43214263,0.12179973,0.33966284);
  rgb(73pt)=(0.43860848,0.12177004,0.34119475);
  rgb(74pt)=(0.44508855,0.12169883,0.34266151);
  rgb(75pt)=(0.45158266,0.12158557,0.34406324);
  rgb(76pt)=(0.45809049,0.12142996,0.34540024);
  rgb(77pt)=(0.46461238,0.12123063,0.34667231);
  rgb(78pt)=(0.47114798,0.12098721,0.34787978);
  rgb(79pt)=(0.47769736,0.12069864,0.34902273);
  rgb(80pt)=(0.48426077,0.12036349,0.35010104);
  rgb(81pt)=(0.49083761,0.11998161,0.35111537);
  rgb(82pt)=(0.49742847,0.11955087,0.35206533);
  rgb(83pt)=(0.50403286,0.11907081,0.35295152);
  rgb(84pt)=(0.51065109,0.11853959,0.35377385);
  rgb(85pt)=(0.51728314,0.1179558,0.35453252);
  rgb(86pt)=(0.52392883,0.11731817,0.35522789);
  rgb(87pt)=(0.53058853,0.11662445,0.35585982);
  rgb(88pt)=(0.53726173,0.11587369,0.35642903);
  rgb(89pt)=(0.54394898,0.11506307,0.35693521);
  rgb(90pt)=(0.5506426,0.11420757,0.35737863);
  rgb(91pt)=(0.55734473,0.11330456,0.35775059);
  rgb(92pt)=(0.56405586,0.11235265,0.35804813);
  rgb(93pt)=(0.57077365,0.11135597,0.35827146);
  rgb(94pt)=(0.5774991,0.11031233,0.35841679);
  rgb(95pt)=(0.58422945,0.10922707,0.35848469);
  rgb(96pt)=(0.59096382,0.10810205,0.35847347);
  rgb(97pt)=(0.59770215,0.10693774,0.35838029);
  rgb(98pt)=(0.60444226,0.10573912,0.35820487);
  rgb(99pt)=(0.61118304,0.10450943,0.35794557);
  rgb(100pt)=(0.61792306,0.10325288,0.35760108);
  rgb(101pt)=(0.62466162,0.10197244,0.35716891);
  rgb(102pt)=(0.63139686,0.10067417,0.35664819);
  rgb(103pt)=(0.63812122,0.09938212,0.35603757);
  rgb(104pt)=(0.64483795,0.0980891,0.35533555);
  rgb(105pt)=(0.65154562,0.09680192,0.35454107);
  rgb(106pt)=(0.65824241,0.09552918,0.3536529);
  rgb(107pt)=(0.66492652,0.09428017,0.3526697);
  rgb(108pt)=(0.67159578,0.09306598,0.35159077);
  rgb(109pt)=(0.67824099,0.09192342,0.3504148);
  rgb(110pt)=(0.684863,0.09085633,0.34914061);
  rgb(111pt)=(0.69146268,0.0898675,0.34776864);
  rgb(112pt)=(0.69803757,0.08897226,0.3462986);
  rgb(113pt)=(0.70457834,0.0882129,0.34473046);
  rgb(114pt)=(0.71108138,0.08761223,0.3430635);
  rgb(115pt)=(0.7175507,0.08716212,0.34129974);
  rgb(116pt)=(0.72398193,0.08688725,0.33943958);
  rgb(117pt)=(0.73035829,0.0868623,0.33748452);
  rgb(118pt)=(0.73669146,0.08704683,0.33543669);
  rgb(119pt)=(0.74297501,0.08747196,0.33329799);
  rgb(120pt)=(0.74919318,0.08820542,0.33107204);
  rgb(121pt)=(0.75535825,0.08919792,0.32876184);
  rgb(122pt)=(0.76145589,0.09050716,0.32637117);
  rgb(123pt)=(0.76748424,0.09213602,0.32390525);
  rgb(124pt)=(0.77344838,0.09405684,0.32136808);
  rgb(125pt)=(0.77932641,0.09634794,0.31876642);
  rgb(126pt)=(0.78513609,0.09892473,0.31610488);
  rgb(127pt)=(0.79085854,0.10184672,0.313391);
  rgb(128pt)=(0.7965014,0.10506637,0.31063031);
  rgb(129pt)=(0.80205987,0.10858333,0.30783);
  rgb(130pt)=(0.80752799,0.11239964,0.30499738);
  rgb(131pt)=(0.81291606,0.11645784,0.30213802);
  rgb(132pt)=(0.81820481,0.12080606,0.29926105);
  rgb(133pt)=(0.82341472,0.12535343,0.2963705);
  rgb(134pt)=(0.82852822,0.13014118,0.29347474);
  rgb(135pt)=(0.83355779,0.13511035,0.29057852);
  rgb(136pt)=(0.83850183,0.14025098,0.2876878);
  rgb(137pt)=(0.84335441,0.14556683,0.28480819);
  rgb(138pt)=(0.84813096,0.15099892,0.281943);
  rgb(139pt)=(0.85281737,0.15657772,0.27909826);
  rgb(140pt)=(0.85742602,0.1622583,0.27627462);
  rgb(141pt)=(0.86196552,0.16801239,0.27346473);
  rgb(142pt)=(0.86641628,0.17387796,0.27070818);
  rgb(143pt)=(0.87079129,0.17982114,0.26797378);
  rgb(144pt)=(0.87507281,0.18587368,0.26529697);
  rgb(145pt)=(0.87925878,0.19203259,0.26268136);
  rgb(146pt)=(0.8833417,0.19830556,0.26014181);
  rgb(147pt)=(0.88731387,0.20469941,0.25769539);
  rgb(148pt)=(0.89116859,0.21121788,0.2553592);
  rgb(149pt)=(0.89490337,0.21785614,0.25314362);
  rgb(150pt)=(0.8985026,0.22463251,0.25108745);
  rgb(151pt)=(0.90197527,0.23152063,0.24918223);
  rgb(152pt)=(0.90530097,0.23854541,0.24748098);
  rgb(153pt)=(0.90848638,0.24568473,0.24598324);
  rgb(154pt)=(0.911533,0.25292623,0.24470258);
  rgb(155pt)=(0.9144225,0.26028902,0.24369359);
  rgb(156pt)=(0.91717106,0.26773821,0.24294137);
  rgb(157pt)=(0.91978131,0.27526191,0.24245973);
  rgb(158pt)=(0.92223947,0.28287251,0.24229568);
  rgb(159pt)=(0.92456587,0.29053388,0.24242622);
  rgb(160pt)=(0.92676657,0.29823282,0.24285536);
  rgb(161pt)=(0.92882964,0.30598085,0.24362274);
  rgb(162pt)=(0.93078135,0.31373977,0.24468803);
  rgb(163pt)=(0.93262051,0.3215093,0.24606461);
  rgb(164pt)=(0.93435067,0.32928362,0.24775328);
  rgb(165pt)=(0.93599076,0.33703942,0.24972157);
  rgb(166pt)=(0.93752831,0.34479177,0.25199928);
  rgb(167pt)=(0.93899289,0.35250734,0.25452808);
  rgb(168pt)=(0.94036561,0.36020899,0.25734661);
  rgb(169pt)=(0.94167588,0.36786594,0.2603949);
  rgb(170pt)=(0.94291042,0.37549479,0.26369821);
  rgb(171pt)=(0.94408513,0.3830811,0.26722004);
  rgb(172pt)=(0.94520419,0.39062329,0.27094924);
  rgb(173pt)=(0.94625977,0.39813168,0.27489742);
  rgb(174pt)=(0.94727016,0.4055909,0.27902322);
  rgb(175pt)=(0.94823505,0.41300424,0.28332283);
  rgb(176pt)=(0.94914549,0.42038251,0.28780969);
  rgb(177pt)=(0.95001704,0.42771398,0.29244728);
  rgb(178pt)=(0.95085121,0.43500005,0.29722817);
  rgb(179pt)=(0.95165009,0.44224144,0.30214494);
  rgb(180pt)=(0.9524044,0.44944853,0.3072105);
  rgb(181pt)=(0.95312556,0.45661389,0.31239776);
  rgb(182pt)=(0.95381595,0.46373781,0.31769923);
  rgb(183pt)=(0.95447591,0.47082238,0.32310953);
  rgb(184pt)=(0.95510255,0.47787236,0.32862553);
  rgb(185pt)=(0.95569679,0.48489115,0.33421404);
  rgb(186pt)=(0.95626788,0.49187351,0.33985601);
  rgb(187pt)=(0.95681685,0.49882008,0.34555431);
  rgb(188pt)=(0.9573439,0.50573243,0.35130912);
  rgb(189pt)=(0.95784842,0.51261283,0.35711942);
  rgb(190pt)=(0.95833051,0.51946267,0.36298589);
  rgb(191pt)=(0.95879054,0.52628305,0.36890904);
  rgb(192pt)=(0.95922872,0.53307513,0.3748895);
  rgb(193pt)=(0.95964538,0.53983991,0.38092784);
  rgb(194pt)=(0.96004345,0.54657593,0.3870292);
  rgb(195pt)=(0.96042097,0.55328624,0.39319057);
  rgb(196pt)=(0.96077819,0.55997184,0.39941173);
  rgb(197pt)=(0.9611152,0.5666337,0.40569343);
  rgb(198pt)=(0.96143273,0.57327231,0.41203603);
  rgb(199pt)=(0.96173392,0.57988594,0.41844491);
  rgb(200pt)=(0.96201757,0.58647675,0.42491751);
  rgb(201pt)=(0.96228344,0.59304598,0.43145271);
  rgb(202pt)=(0.96253168,0.5995944,0.43805131);
  rgb(203pt)=(0.96276513,0.60612062,0.44471698);
  rgb(204pt)=(0.96298491,0.6126247,0.45145074);
  rgb(205pt)=(0.96318967,0.61910879,0.45824902);
  rgb(206pt)=(0.96337949,0.6255736,0.46511271);
  rgb(207pt)=(0.96355923,0.63201624,0.47204746);
  rgb(208pt)=(0.96372785,0.63843852,0.47905028);
  rgb(209pt)=(0.96388426,0.64484214,0.4861196);
  rgb(210pt)=(0.96403203,0.65122535,0.4932578);
  rgb(211pt)=(0.96417332,0.65758729,0.50046894);
  rgb(212pt)=(0.9643063,0.66393045,0.5077467);
  rgb(213pt)=(0.96443322,0.67025402,0.51509334);
  rgb(214pt)=(0.96455845,0.67655564,0.52251447);
  rgb(215pt)=(0.96467922,0.68283846,0.53000231);
  rgb(216pt)=(0.96479861,0.68910113,0.53756026);
  rgb(217pt)=(0.96492035,0.69534192,0.5451917);
  rgb(218pt)=(0.96504223,0.7015636,0.5528892);
  rgb(219pt)=(0.96516917,0.70776351,0.5606593);
  rgb(220pt)=(0.96530224,0.71394212,0.56849894);
  rgb(221pt)=(0.96544032,0.72010124,0.57640375);
  rgb(222pt)=(0.96559206,0.72623592,0.58438387);
  rgb(223pt)=(0.96575293,0.73235058,0.59242739);
  rgb(224pt)=(0.96592829,0.73844258,0.60053991);
  rgb(225pt)=(0.96612013,0.74451182,0.60871954);
  rgb(226pt)=(0.96632832,0.75055966,0.61696136);
  rgb(227pt)=(0.96656022,0.75658231,0.62527295);
  rgb(228pt)=(0.96681185,0.76258381,0.63364277);
  rgb(229pt)=(0.96709183,0.76855969,0.64207921);
  rgb(230pt)=(0.96739773,0.77451297,0.65057302);
  rgb(231pt)=(0.96773482,0.78044149,0.65912731);
  rgb(232pt)=(0.96810471,0.78634563,0.66773889);
  rgb(233pt)=(0.96850919,0.79222565,0.6764046);
  rgb(234pt)=(0.96893132,0.79809112,0.68512266);
  rgb(235pt)=(0.96935926,0.80395415,0.69383201);
  rgb(236pt)=(0.9698028,0.80981139,0.70252255);
  rgb(237pt)=(0.97025511,0.81566605,0.71120296);
  rgb(238pt)=(0.97071849,0.82151775,0.71987163);
  rgb(239pt)=(0.97120159,0.82736371,0.72851999);
  rgb(240pt)=(0.97169389,0.83320847,0.73716071);
  rgb(241pt)=(0.97220061,0.83905052,0.74578903);
  rgb(242pt)=(0.97272597,0.84488881,0.75440141);
  rgb(243pt)=(0.97327085,0.85072354,0.76299805);
  rgb(244pt)=(0.97383206,0.85655639,0.77158353);
  rgb(245pt)=(0.97441222,0.86238689,0.78015619);
  rgb(246pt)=(0.97501782,0.86821321,0.78871034);
  rgb(247pt)=(0.97564391,0.87403763,0.79725261);
  rgb(248pt)=(0.97628674,0.87986189,0.8057883);
  rgb(249pt)=(0.97696114,0.88568129,0.81430324);
  rgb(250pt)=(0.97765722,0.89149971,0.82280948);
  rgb(251pt)=(0.97837585,0.89731727,0.83130786);
  rgb(252pt)=(0.97912374,0.90313207,0.83979337);
  rgb(253pt)=(0.979891,0.90894778,0.84827858);
  rgb(254pt)=(0.98067764,0.91476465,0.85676611);
  rgb(255pt)=(0.98137749,0.92061729,0.86536915)
}]
table{%
x                      y
1 0.000267230380010058
1 0.00254838226944823
1 0.0541544206995508
1 0.0379135813136304
1 0.0196095771462737
1 0.0102855852273538
1 0.021015918773525
1 0.0120139146638508
1 0.00270276968660798
1 0.0169317556616254
1 0.0539334696984598
1 0.041907253274646
2 0.0864085563241128
2 0.124065063809417
2 0.129070106637688
2 0.136294260158224
2 0.123594315878124
2 0.052674160149586
2 0.364579231312094
2 0.0630906126688538
2 0.0490663627103029
2 0.0777681750057255
2 0.052514262486864
2 0.0190870993775768
3 0.0179704657247327
3 0.0922962845045939
3 0.214294301642558
3 0.113759237346762
3 0.0862520200379948
3 0.0761258883516974
3 0.0502754150618
3 0.0607301091681606
3 0.0696604586624192
3 0.0233362324826734
3 0.0759427452080613
3 0.0546670908560776
4 0.088915399317486
4 0.0273130247717624
4 0.0381549151123194
4 0.0405934837244762
4 0.0343013187304422
4 0.0619971297120315
4 0.0426622132023031
4 0.0512612083321475
4 0.0487242143808312
4 0.0435678380482567
4 0.0449740986389384
4 0.0660621999348832
5 0.00318694993077699
5 0.00330328676156343
5 0.0225787991261754
5 0.0432979705147962
5 0.0101602136494002
5 0.0264026976904968
5 0.0190253399173828
5 0.0373398235693863
5 0.0220554550765683
5 0.0258907546246607
5 0.0306907756698188
5 0.0227426745437754
6 0.012150586601167
6 0.0165479451265792
6 0.00897779883390258
6 0.00979877129907956
6 0.0291639382726096
6 0.0182365470825328
6 0.0290707161925048
6 0.0125171770638365
6 0.00709385743961169
6 -0.00244689382628411
6 0.0107322403922184
6 0.0272778493729729
7 0.00796696587139475
7 0.00982686910294308
7 0.00725455688304851
7 -0.00251689107433363
7 0.00294781385209151
7 0.00448265964826889
7 0.0107221614197639
7 0.00511597454335754
7 0.0630470798774721
7 0.00628814424109146
7 0.0194569083811138
7 0.00959417505047612
8 0.000159917508387501
8 0.00442522192845665
8 0.00478453658783868
8 -0.00122116278212932
8 0.00812513227133751
8 0.00661523731125828
8 -0.00018143162675171
8 0.0107547851286745
8 0.0181208153943856
8 0.0117412673259346
8 0.0020454428387318
8 0.0036304405479202
9 0.0129340225289717
9 0.0142105134988156
9 0.0111588529068304
9 0.0087750869331439
9 -0.000949229876769547
9 0.00329185345792549
9 0.0251507869241151
9 0.00386894501988991
9 0.0182837621107782
9 0.0140797993100206
9 0.00120601765160939
9 0.0128256374906759
10 0.0107182960526879
10 0.0225343477735682
10 0.0145510097556643
10 0.00685735491118507
10 0.00608608597872184
10 0.00420732440599847
10 0.0381328796566384
10 0.00315900919049161
10 0.013934025214436
10 0.0102143522592409
10 0.013016081768023
10 0.0213006358059474
11 0.0140909520714564
11 -0.00467865079915839
11 0.0124419648459377
11 0.00559506405713785
11 0.0129567083090436
11 0.00267309961093414
11 0.0221866854250682
11 0.0106729704307652
11 0.00120301621622148
11 0.00900896039006627
11 0.123358426505836
11 0.00930432297098568
12 0.00328791449003352
12 0.00872636753916073
12 0.0074908353881857
12 0.00344615067831425
12 0.0100931886279425
12 0.00764360777131804
12 0.0129036735336204
12 0.00911897440553184
12 0.00287737305987484
12 0.0111232081746234
12 0.00685600762819675
12 0.00611199241931862
};
\addlegendentry{[CLS]}
\addplot [only marks, draw=color1,mark size=1.25pt, fill=color1, colormap={mymap}{[1pt]
 rgb(0pt)=(0.01060815,0.01808215,0.10018654);
  rgb(1pt)=(0.01428972,0.02048237,0.10374486);
  rgb(2pt)=(0.01831941,0.0229766,0.10738511);
  rgb(3pt)=(0.02275049,0.02554464,0.11108639);
  rgb(4pt)=(0.02759119,0.02818316,0.11483751);
  rgb(5pt)=(0.03285175,0.03088792,0.11863035);
  rgb(6pt)=(0.03853466,0.03365771,0.12245873);
  rgb(7pt)=(0.04447016,0.03648425,0.12631831);
  rgb(8pt)=(0.05032105,0.03936808,0.13020508);
  rgb(9pt)=(0.05611171,0.04224835,0.13411624);
  rgb(10pt)=(0.0618531,0.04504866,0.13804929);
  rgb(11pt)=(0.06755457,0.04778179,0.14200206);
  rgb(12pt)=(0.0732236,0.05045047,0.14597263);
  rgb(13pt)=(0.0788708,0.05305461,0.14995981);
  rgb(14pt)=(0.08450105,0.05559631,0.15396203);
  rgb(15pt)=(0.09011319,0.05808059,0.15797687);
  rgb(16pt)=(0.09572396,0.06050127,0.16200507);
  rgb(17pt)=(0.10132312,0.06286782,0.16604287);
  rgb(18pt)=(0.10692823,0.06517224,0.17009175);
  rgb(19pt)=(0.1125315,0.06742194,0.17414848);
  rgb(20pt)=(0.11813947,0.06961499,0.17821272);
  rgb(21pt)=(0.12375803,0.07174938,0.18228425);
  rgb(22pt)=(0.12938228,0.07383015,0.18636053);
  rgb(23pt)=(0.13501631,0.07585609,0.19044109);
  rgb(24pt)=(0.14066867,0.0778224,0.19452676);
  rgb(25pt)=(0.14633406,0.07973393,0.1986151);
  rgb(26pt)=(0.15201338,0.08159108,0.20270523);
  rgb(27pt)=(0.15770877,0.08339312,0.20679668);
  rgb(28pt)=(0.16342174,0.0851396,0.21088893);
  rgb(29pt)=(0.16915387,0.08682996,0.21498104);
  rgb(30pt)=(0.17489524,0.08848235,0.2190294);
  rgb(31pt)=(0.18065495,0.09009031,0.22303512);
  rgb(32pt)=(0.18643324,0.09165431,0.22699705);
  rgb(33pt)=(0.19223028,0.09317479,0.23091409);
  rgb(34pt)=(0.19804623,0.09465217,0.23478512);
  rgb(35pt)=(0.20388117,0.09608689,0.23860907);
  rgb(36pt)=(0.20973515,0.09747934,0.24238489);
  rgb(37pt)=(0.21560818,0.09882993,0.24611154);
  rgb(38pt)=(0.22150014,0.10013944,0.2497868);
  rgb(39pt)=(0.22741085,0.10140876,0.25340813);
  rgb(40pt)=(0.23334047,0.10263737,0.25697736);
  rgb(41pt)=(0.23928891,0.10382562,0.2604936);
  rgb(42pt)=(0.24525608,0.10497384,0.26395596);
  rgb(43pt)=(0.25124182,0.10608236,0.26736359);
  rgb(44pt)=(0.25724602,0.10715148,0.27071569);
  rgb(45pt)=(0.26326851,0.1081815,0.27401148);
  rgb(46pt)=(0.26930915,0.1091727,0.2772502);
  rgb(47pt)=(0.27536766,0.11012568,0.28043021);
  rgb(48pt)=(0.28144375,0.11104133,0.2835489);
  rgb(49pt)=(0.2875374,0.11191896,0.28660853);
  rgb(50pt)=(0.29364846,0.11275876,0.2896085);
  rgb(51pt)=(0.29977678,0.11356089,0.29254823);
  rgb(52pt)=(0.30592213,0.11432553,0.29542718);
  rgb(53pt)=(0.31208435,0.11505284,0.29824485);
  rgb(54pt)=(0.31826327,0.1157429,0.30100076);
  rgb(55pt)=(0.32445869,0.11639585,0.30369448);
  rgb(56pt)=(0.33067031,0.11701189,0.30632563);
  rgb(57pt)=(0.33689808,0.11759095,0.3088938);
  rgb(58pt)=(0.34314168,0.11813362,0.31139721);
  rgb(59pt)=(0.34940101,0.11863987,0.3138355);
  rgb(60pt)=(0.355676,0.11910909,0.31620996);
  rgb(61pt)=(0.36196644,0.1195413,0.31852037);
  rgb(62pt)=(0.36827206,0.11993653,0.32076656);
  rgb(63pt)=(0.37459292,0.12029443,0.32294825);
  rgb(64pt)=(0.38092887,0.12061482,0.32506528);
  rgb(65pt)=(0.38727975,0.12089756,0.3271175);
  rgb(66pt)=(0.39364518,0.12114272,0.32910494);
  rgb(67pt)=(0.40002537,0.12134964,0.33102734);
  rgb(68pt)=(0.40642019,0.12151801,0.33288464);
  rgb(69pt)=(0.41282936,0.12164769,0.33467689);
  rgb(70pt)=(0.41925278,0.12173833,0.33640407);
  rgb(71pt)=(0.42569057,0.12178916,0.33806605);
  rgb(72pt)=(0.43214263,0.12179973,0.33966284);
  rgb(73pt)=(0.43860848,0.12177004,0.34119475);
  rgb(74pt)=(0.44508855,0.12169883,0.34266151);
  rgb(75pt)=(0.45158266,0.12158557,0.34406324);
  rgb(76pt)=(0.45809049,0.12142996,0.34540024);
  rgb(77pt)=(0.46461238,0.12123063,0.34667231);
  rgb(78pt)=(0.47114798,0.12098721,0.34787978);
  rgb(79pt)=(0.47769736,0.12069864,0.34902273);
  rgb(80pt)=(0.48426077,0.12036349,0.35010104);
  rgb(81pt)=(0.49083761,0.11998161,0.35111537);
  rgb(82pt)=(0.49742847,0.11955087,0.35206533);
  rgb(83pt)=(0.50403286,0.11907081,0.35295152);
  rgb(84pt)=(0.51065109,0.11853959,0.35377385);
  rgb(85pt)=(0.51728314,0.1179558,0.35453252);
  rgb(86pt)=(0.52392883,0.11731817,0.35522789);
  rgb(87pt)=(0.53058853,0.11662445,0.35585982);
  rgb(88pt)=(0.53726173,0.11587369,0.35642903);
  rgb(89pt)=(0.54394898,0.11506307,0.35693521);
  rgb(90pt)=(0.5506426,0.11420757,0.35737863);
  rgb(91pt)=(0.55734473,0.11330456,0.35775059);
  rgb(92pt)=(0.56405586,0.11235265,0.35804813);
  rgb(93pt)=(0.57077365,0.11135597,0.35827146);
  rgb(94pt)=(0.5774991,0.11031233,0.35841679);
  rgb(95pt)=(0.58422945,0.10922707,0.35848469);
  rgb(96pt)=(0.59096382,0.10810205,0.35847347);
  rgb(97pt)=(0.59770215,0.10693774,0.35838029);
  rgb(98pt)=(0.60444226,0.10573912,0.35820487);
  rgb(99pt)=(0.61118304,0.10450943,0.35794557);
  rgb(100pt)=(0.61792306,0.10325288,0.35760108);
  rgb(101pt)=(0.62466162,0.10197244,0.35716891);
  rgb(102pt)=(0.63139686,0.10067417,0.35664819);
  rgb(103pt)=(0.63812122,0.09938212,0.35603757);
  rgb(104pt)=(0.64483795,0.0980891,0.35533555);
  rgb(105pt)=(0.65154562,0.09680192,0.35454107);
  rgb(106pt)=(0.65824241,0.09552918,0.3536529);
  rgb(107pt)=(0.66492652,0.09428017,0.3526697);
  rgb(108pt)=(0.67159578,0.09306598,0.35159077);
  rgb(109pt)=(0.67824099,0.09192342,0.3504148);
  rgb(110pt)=(0.684863,0.09085633,0.34914061);
  rgb(111pt)=(0.69146268,0.0898675,0.34776864);
  rgb(112pt)=(0.69803757,0.08897226,0.3462986);
  rgb(113pt)=(0.70457834,0.0882129,0.34473046);
  rgb(114pt)=(0.71108138,0.08761223,0.3430635);
  rgb(115pt)=(0.7175507,0.08716212,0.34129974);
  rgb(116pt)=(0.72398193,0.08688725,0.33943958);
  rgb(117pt)=(0.73035829,0.0868623,0.33748452);
  rgb(118pt)=(0.73669146,0.08704683,0.33543669);
  rgb(119pt)=(0.74297501,0.08747196,0.33329799);
  rgb(120pt)=(0.74919318,0.08820542,0.33107204);
  rgb(121pt)=(0.75535825,0.08919792,0.32876184);
  rgb(122pt)=(0.76145589,0.09050716,0.32637117);
  rgb(123pt)=(0.76748424,0.09213602,0.32390525);
  rgb(124pt)=(0.77344838,0.09405684,0.32136808);
  rgb(125pt)=(0.77932641,0.09634794,0.31876642);
  rgb(126pt)=(0.78513609,0.09892473,0.31610488);
  rgb(127pt)=(0.79085854,0.10184672,0.313391);
  rgb(128pt)=(0.7965014,0.10506637,0.31063031);
  rgb(129pt)=(0.80205987,0.10858333,0.30783);
  rgb(130pt)=(0.80752799,0.11239964,0.30499738);
  rgb(131pt)=(0.81291606,0.11645784,0.30213802);
  rgb(132pt)=(0.81820481,0.12080606,0.29926105);
  rgb(133pt)=(0.82341472,0.12535343,0.2963705);
  rgb(134pt)=(0.82852822,0.13014118,0.29347474);
  rgb(135pt)=(0.83355779,0.13511035,0.29057852);
  rgb(136pt)=(0.83850183,0.14025098,0.2876878);
  rgb(137pt)=(0.84335441,0.14556683,0.28480819);
  rgb(138pt)=(0.84813096,0.15099892,0.281943);
  rgb(139pt)=(0.85281737,0.15657772,0.27909826);
  rgb(140pt)=(0.85742602,0.1622583,0.27627462);
  rgb(141pt)=(0.86196552,0.16801239,0.27346473);
  rgb(142pt)=(0.86641628,0.17387796,0.27070818);
  rgb(143pt)=(0.87079129,0.17982114,0.26797378);
  rgb(144pt)=(0.87507281,0.18587368,0.26529697);
  rgb(145pt)=(0.87925878,0.19203259,0.26268136);
  rgb(146pt)=(0.8833417,0.19830556,0.26014181);
  rgb(147pt)=(0.88731387,0.20469941,0.25769539);
  rgb(148pt)=(0.89116859,0.21121788,0.2553592);
  rgb(149pt)=(0.89490337,0.21785614,0.25314362);
  rgb(150pt)=(0.8985026,0.22463251,0.25108745);
  rgb(151pt)=(0.90197527,0.23152063,0.24918223);
  rgb(152pt)=(0.90530097,0.23854541,0.24748098);
  rgb(153pt)=(0.90848638,0.24568473,0.24598324);
  rgb(154pt)=(0.911533,0.25292623,0.24470258);
  rgb(155pt)=(0.9144225,0.26028902,0.24369359);
  rgb(156pt)=(0.91717106,0.26773821,0.24294137);
  rgb(157pt)=(0.91978131,0.27526191,0.24245973);
  rgb(158pt)=(0.92223947,0.28287251,0.24229568);
  rgb(159pt)=(0.92456587,0.29053388,0.24242622);
  rgb(160pt)=(0.92676657,0.29823282,0.24285536);
  rgb(161pt)=(0.92882964,0.30598085,0.24362274);
  rgb(162pt)=(0.93078135,0.31373977,0.24468803);
  rgb(163pt)=(0.93262051,0.3215093,0.24606461);
  rgb(164pt)=(0.93435067,0.32928362,0.24775328);
  rgb(165pt)=(0.93599076,0.33703942,0.24972157);
  rgb(166pt)=(0.93752831,0.34479177,0.25199928);
  rgb(167pt)=(0.93899289,0.35250734,0.25452808);
  rgb(168pt)=(0.94036561,0.36020899,0.25734661);
  rgb(169pt)=(0.94167588,0.36786594,0.2603949);
  rgb(170pt)=(0.94291042,0.37549479,0.26369821);
  rgb(171pt)=(0.94408513,0.3830811,0.26722004);
  rgb(172pt)=(0.94520419,0.39062329,0.27094924);
  rgb(173pt)=(0.94625977,0.39813168,0.27489742);
  rgb(174pt)=(0.94727016,0.4055909,0.27902322);
  rgb(175pt)=(0.94823505,0.41300424,0.28332283);
  rgb(176pt)=(0.94914549,0.42038251,0.28780969);
  rgb(177pt)=(0.95001704,0.42771398,0.29244728);
  rgb(178pt)=(0.95085121,0.43500005,0.29722817);
  rgb(179pt)=(0.95165009,0.44224144,0.30214494);
  rgb(180pt)=(0.9524044,0.44944853,0.3072105);
  rgb(181pt)=(0.95312556,0.45661389,0.31239776);
  rgb(182pt)=(0.95381595,0.46373781,0.31769923);
  rgb(183pt)=(0.95447591,0.47082238,0.32310953);
  rgb(184pt)=(0.95510255,0.47787236,0.32862553);
  rgb(185pt)=(0.95569679,0.48489115,0.33421404);
  rgb(186pt)=(0.95626788,0.49187351,0.33985601);
  rgb(187pt)=(0.95681685,0.49882008,0.34555431);
  rgb(188pt)=(0.9573439,0.50573243,0.35130912);
  rgb(189pt)=(0.95784842,0.51261283,0.35711942);
  rgb(190pt)=(0.95833051,0.51946267,0.36298589);
  rgb(191pt)=(0.95879054,0.52628305,0.36890904);
  rgb(192pt)=(0.95922872,0.53307513,0.3748895);
  rgb(193pt)=(0.95964538,0.53983991,0.38092784);
  rgb(194pt)=(0.96004345,0.54657593,0.3870292);
  rgb(195pt)=(0.96042097,0.55328624,0.39319057);
  rgb(196pt)=(0.96077819,0.55997184,0.39941173);
  rgb(197pt)=(0.9611152,0.5666337,0.40569343);
  rgb(198pt)=(0.96143273,0.57327231,0.41203603);
  rgb(199pt)=(0.96173392,0.57988594,0.41844491);
  rgb(200pt)=(0.96201757,0.58647675,0.42491751);
  rgb(201pt)=(0.96228344,0.59304598,0.43145271);
  rgb(202pt)=(0.96253168,0.5995944,0.43805131);
  rgb(203pt)=(0.96276513,0.60612062,0.44471698);
  rgb(204pt)=(0.96298491,0.6126247,0.45145074);
  rgb(205pt)=(0.96318967,0.61910879,0.45824902);
  rgb(206pt)=(0.96337949,0.6255736,0.46511271);
  rgb(207pt)=(0.96355923,0.63201624,0.47204746);
  rgb(208pt)=(0.96372785,0.63843852,0.47905028);
  rgb(209pt)=(0.96388426,0.64484214,0.4861196);
  rgb(210pt)=(0.96403203,0.65122535,0.4932578);
  rgb(211pt)=(0.96417332,0.65758729,0.50046894);
  rgb(212pt)=(0.9643063,0.66393045,0.5077467);
  rgb(213pt)=(0.96443322,0.67025402,0.51509334);
  rgb(214pt)=(0.96455845,0.67655564,0.52251447);
  rgb(215pt)=(0.96467922,0.68283846,0.53000231);
  rgb(216pt)=(0.96479861,0.68910113,0.53756026);
  rgb(217pt)=(0.96492035,0.69534192,0.5451917);
  rgb(218pt)=(0.96504223,0.7015636,0.5528892);
  rgb(219pt)=(0.96516917,0.70776351,0.5606593);
  rgb(220pt)=(0.96530224,0.71394212,0.56849894);
  rgb(221pt)=(0.96544032,0.72010124,0.57640375);
  rgb(222pt)=(0.96559206,0.72623592,0.58438387);
  rgb(223pt)=(0.96575293,0.73235058,0.59242739);
  rgb(224pt)=(0.96592829,0.73844258,0.60053991);
  rgb(225pt)=(0.96612013,0.74451182,0.60871954);
  rgb(226pt)=(0.96632832,0.75055966,0.61696136);
  rgb(227pt)=(0.96656022,0.75658231,0.62527295);
  rgb(228pt)=(0.96681185,0.76258381,0.63364277);
  rgb(229pt)=(0.96709183,0.76855969,0.64207921);
  rgb(230pt)=(0.96739773,0.77451297,0.65057302);
  rgb(231pt)=(0.96773482,0.78044149,0.65912731);
  rgb(232pt)=(0.96810471,0.78634563,0.66773889);
  rgb(233pt)=(0.96850919,0.79222565,0.6764046);
  rgb(234pt)=(0.96893132,0.79809112,0.68512266);
  rgb(235pt)=(0.96935926,0.80395415,0.69383201);
  rgb(236pt)=(0.9698028,0.80981139,0.70252255);
  rgb(237pt)=(0.97025511,0.81566605,0.71120296);
  rgb(238pt)=(0.97071849,0.82151775,0.71987163);
  rgb(239pt)=(0.97120159,0.82736371,0.72851999);
  rgb(240pt)=(0.97169389,0.83320847,0.73716071);
  rgb(241pt)=(0.97220061,0.83905052,0.74578903);
  rgb(242pt)=(0.97272597,0.84488881,0.75440141);
  rgb(243pt)=(0.97327085,0.85072354,0.76299805);
  rgb(244pt)=(0.97383206,0.85655639,0.77158353);
  rgb(245pt)=(0.97441222,0.86238689,0.78015619);
  rgb(246pt)=(0.97501782,0.86821321,0.78871034);
  rgb(247pt)=(0.97564391,0.87403763,0.79725261);
  rgb(248pt)=(0.97628674,0.87986189,0.8057883);
  rgb(249pt)=(0.97696114,0.88568129,0.81430324);
  rgb(250pt)=(0.97765722,0.89149971,0.82280948);
  rgb(251pt)=(0.97837585,0.89731727,0.83130786);
  rgb(252pt)=(0.97912374,0.90313207,0.83979337);
  rgb(253pt)=(0.979891,0.90894778,0.84827858);
  rgb(254pt)=(0.98067764,0.91476465,0.85676611);
  rgb(255pt)=(0.98137749,0.92061729,0.86536915)
}]
table{%
x                      y
1 0.0109597003400535
1 0.00579576312899923
1 0.0267235735221794
1 0.0208010347517276
1 0.0467622775513663
1 0.0101307624835181
1 0.015042024476416
1 0.0287456860614518
1 0.0175536161166103
1 0.027654841084553
1 0.0244924136849491
1 0.0129420548348799
2 0.0565382135424874
2 0.0330865102700901
2 0.0227830243502845
2 0.0641466964148627
2 0.0318028422541135
2 0.0718362412583372
2 0.0761978893728106
2 0.0493336843802458
2 0.0561529749582231
2 0.0928169518934103
2 0.0386948105555714
2 0.0359654334280668
3 0.0337315866986497
3 0.132710189884115
3 0.217985387363558
3 0.106378645124672
3 0.0755562513023416
3 0.0509983590326202
3 0.0910765491941766
3 0.0755767141306364
3 0.0504146517869421
3 0.0332470304273007
3 0.0521610280674537
3 0.0601035418298195
4 0.105607942094756
4 0.0298693852713991
4 0.0301616440374218
4 0.0405849369298951
4 0.0396197630794612
4 0.0240708792151162
4 0.0401622430941847
4 0.0407909957989266
4 0.0707976977829868
4 0.0674541717202592
4 0.0397728048972703
4 0.0865788271815516
5 0.0279291185378686
5 0.0136811921605422
5 0.0329280456928743
5 0.0443933999424826
5 0.0158548871758655
5 0.0264318574030602
5 0.0284264710673983
5 0.0352605731776525
5 0.0315679890342824
5 0.0320203208654497
5 0.023907413288833
5 0.0213801845735115
6 0.0351673261484567
6 0.0345892100867133
6 0.0292892233713462
6 0.0305162875428976
6 0.0372182070249294
6 0.0424959113842063
6 0.0418828481772814
6 0.0347443542371881
6 0.0347492997025113
6 0.0276733253318874
6 0.0385014672205696
6 0.0403855122446477
7 0.0363344091006942
7 0.0286396985493269
7 0.0333894935246396
7 0.0356703547459398
7 0.0286562658850812
7 0.031161090948616
7 0.031735218791905
7 0.0355943284243728
7 0.0320467635543565
7 0.0418708741739134
7 0.0367534897345324
7 0.0255621669821822
8 0.0272741854502495
8 0.0306077340487759
8 0.0354742327690097
8 0.0382050521637575
8 0.0252486210656095
8 0.0279584832711808
8 0.0415851271858847
8 0.041665655243874
8 0.0314768577186116
8 0.0339652279629701
8 0.0291869349877717
8 0.0383136535335426
9 0.0304090189436774
9 0.0264237837666243
9 0.0433131955208842
9 0.0285578690824672
9 0.0310117713577
9 0.0354673853996664
9 0.0352584798322505
9 0.0269305965334478
9 0.0346621580395822
9 0.0372078290484769
9 0.0291872100978211
9 0.0320969062126013
10 0.03235552081078
10 0.0285700296265236
10 0.024979441062404
10 0.0386110882777119
10 0.0272290506004654
10 0.0257220826774984
10 0.0437471400640023
10 0.0300590326269722
10 0.0342796983801969
10 0.0465360685616897
10 0.0279340397875722
10 0.0361094493304357
11 0.0337544273947793
11 0.0150331366817982
11 0.0363787335081839
11 0.05327114026398
11 0.0397653805647065
11 0.018042264215326
11 0.0509913751126531
11 0.0412752138299558
11 0.0464988060063162
11 0.0499716441864039
11 0.136451194011997
11 0.026097640339026
12 0.015121629990012
12 0.0159998177807866
12 0.0202458671481145
12 0.0195377144785695
12 0.0194439195054026
12 0.0314878737252256
12 0.0257687623857373
12 0.0129372996465148
12 0.0117009676520132
12 0.0251038898593178
12 0.0167577639894977
12 0.00941720214328911
};
\addlegendentry{[SEP]}
\addplot [only marks, draw=color2,mark size=1.25pt, fill=color2, colormap={mymap}{[1pt]
 rgb(0pt)=(0.01060815,0.01808215,0.10018654);
  rgb(1pt)=(0.01428972,0.02048237,0.10374486);
  rgb(2pt)=(0.01831941,0.0229766,0.10738511);
  rgb(3pt)=(0.02275049,0.02554464,0.11108639);
  rgb(4pt)=(0.02759119,0.02818316,0.11483751);
  rgb(5pt)=(0.03285175,0.03088792,0.11863035);
  rgb(6pt)=(0.03853466,0.03365771,0.12245873);
  rgb(7pt)=(0.04447016,0.03648425,0.12631831);
  rgb(8pt)=(0.05032105,0.03936808,0.13020508);
  rgb(9pt)=(0.05611171,0.04224835,0.13411624);
  rgb(10pt)=(0.0618531,0.04504866,0.13804929);
  rgb(11pt)=(0.06755457,0.04778179,0.14200206);
  rgb(12pt)=(0.0732236,0.05045047,0.14597263);
  rgb(13pt)=(0.0788708,0.05305461,0.14995981);
  rgb(14pt)=(0.08450105,0.05559631,0.15396203);
  rgb(15pt)=(0.09011319,0.05808059,0.15797687);
  rgb(16pt)=(0.09572396,0.06050127,0.16200507);
  rgb(17pt)=(0.10132312,0.06286782,0.16604287);
  rgb(18pt)=(0.10692823,0.06517224,0.17009175);
  rgb(19pt)=(0.1125315,0.06742194,0.17414848);
  rgb(20pt)=(0.11813947,0.06961499,0.17821272);
  rgb(21pt)=(0.12375803,0.07174938,0.18228425);
  rgb(22pt)=(0.12938228,0.07383015,0.18636053);
  rgb(23pt)=(0.13501631,0.07585609,0.19044109);
  rgb(24pt)=(0.14066867,0.0778224,0.19452676);
  rgb(25pt)=(0.14633406,0.07973393,0.1986151);
  rgb(26pt)=(0.15201338,0.08159108,0.20270523);
  rgb(27pt)=(0.15770877,0.08339312,0.20679668);
  rgb(28pt)=(0.16342174,0.0851396,0.21088893);
  rgb(29pt)=(0.16915387,0.08682996,0.21498104);
  rgb(30pt)=(0.17489524,0.08848235,0.2190294);
  rgb(31pt)=(0.18065495,0.09009031,0.22303512);
  rgb(32pt)=(0.18643324,0.09165431,0.22699705);
  rgb(33pt)=(0.19223028,0.09317479,0.23091409);
  rgb(34pt)=(0.19804623,0.09465217,0.23478512);
  rgb(35pt)=(0.20388117,0.09608689,0.23860907);
  rgb(36pt)=(0.20973515,0.09747934,0.24238489);
  rgb(37pt)=(0.21560818,0.09882993,0.24611154);
  rgb(38pt)=(0.22150014,0.10013944,0.2497868);
  rgb(39pt)=(0.22741085,0.10140876,0.25340813);
  rgb(40pt)=(0.23334047,0.10263737,0.25697736);
  rgb(41pt)=(0.23928891,0.10382562,0.2604936);
  rgb(42pt)=(0.24525608,0.10497384,0.26395596);
  rgb(43pt)=(0.25124182,0.10608236,0.26736359);
  rgb(44pt)=(0.25724602,0.10715148,0.27071569);
  rgb(45pt)=(0.26326851,0.1081815,0.27401148);
  rgb(46pt)=(0.26930915,0.1091727,0.2772502);
  rgb(47pt)=(0.27536766,0.11012568,0.28043021);
  rgb(48pt)=(0.28144375,0.11104133,0.2835489);
  rgb(49pt)=(0.2875374,0.11191896,0.28660853);
  rgb(50pt)=(0.29364846,0.11275876,0.2896085);
  rgb(51pt)=(0.29977678,0.11356089,0.29254823);
  rgb(52pt)=(0.30592213,0.11432553,0.29542718);
  rgb(53pt)=(0.31208435,0.11505284,0.29824485);
  rgb(54pt)=(0.31826327,0.1157429,0.30100076);
  rgb(55pt)=(0.32445869,0.11639585,0.30369448);
  rgb(56pt)=(0.33067031,0.11701189,0.30632563);
  rgb(57pt)=(0.33689808,0.11759095,0.3088938);
  rgb(58pt)=(0.34314168,0.11813362,0.31139721);
  rgb(59pt)=(0.34940101,0.11863987,0.3138355);
  rgb(60pt)=(0.355676,0.11910909,0.31620996);
  rgb(61pt)=(0.36196644,0.1195413,0.31852037);
  rgb(62pt)=(0.36827206,0.11993653,0.32076656);
  rgb(63pt)=(0.37459292,0.12029443,0.32294825);
  rgb(64pt)=(0.38092887,0.12061482,0.32506528);
  rgb(65pt)=(0.38727975,0.12089756,0.3271175);
  rgb(66pt)=(0.39364518,0.12114272,0.32910494);
  rgb(67pt)=(0.40002537,0.12134964,0.33102734);
  rgb(68pt)=(0.40642019,0.12151801,0.33288464);
  rgb(69pt)=(0.41282936,0.12164769,0.33467689);
  rgb(70pt)=(0.41925278,0.12173833,0.33640407);
  rgb(71pt)=(0.42569057,0.12178916,0.33806605);
  rgb(72pt)=(0.43214263,0.12179973,0.33966284);
  rgb(73pt)=(0.43860848,0.12177004,0.34119475);
  rgb(74pt)=(0.44508855,0.12169883,0.34266151);
  rgb(75pt)=(0.45158266,0.12158557,0.34406324);
  rgb(76pt)=(0.45809049,0.12142996,0.34540024);
  rgb(77pt)=(0.46461238,0.12123063,0.34667231);
  rgb(78pt)=(0.47114798,0.12098721,0.34787978);
  rgb(79pt)=(0.47769736,0.12069864,0.34902273);
  rgb(80pt)=(0.48426077,0.12036349,0.35010104);
  rgb(81pt)=(0.49083761,0.11998161,0.35111537);
  rgb(82pt)=(0.49742847,0.11955087,0.35206533);
  rgb(83pt)=(0.50403286,0.11907081,0.35295152);
  rgb(84pt)=(0.51065109,0.11853959,0.35377385);
  rgb(85pt)=(0.51728314,0.1179558,0.35453252);
  rgb(86pt)=(0.52392883,0.11731817,0.35522789);
  rgb(87pt)=(0.53058853,0.11662445,0.35585982);
  rgb(88pt)=(0.53726173,0.11587369,0.35642903);
  rgb(89pt)=(0.54394898,0.11506307,0.35693521);
  rgb(90pt)=(0.5506426,0.11420757,0.35737863);
  rgb(91pt)=(0.55734473,0.11330456,0.35775059);
  rgb(92pt)=(0.56405586,0.11235265,0.35804813);
  rgb(93pt)=(0.57077365,0.11135597,0.35827146);
  rgb(94pt)=(0.5774991,0.11031233,0.35841679);
  rgb(95pt)=(0.58422945,0.10922707,0.35848469);
  rgb(96pt)=(0.59096382,0.10810205,0.35847347);
  rgb(97pt)=(0.59770215,0.10693774,0.35838029);
  rgb(98pt)=(0.60444226,0.10573912,0.35820487);
  rgb(99pt)=(0.61118304,0.10450943,0.35794557);
  rgb(100pt)=(0.61792306,0.10325288,0.35760108);
  rgb(101pt)=(0.62466162,0.10197244,0.35716891);
  rgb(102pt)=(0.63139686,0.10067417,0.35664819);
  rgb(103pt)=(0.63812122,0.09938212,0.35603757);
  rgb(104pt)=(0.64483795,0.0980891,0.35533555);
  rgb(105pt)=(0.65154562,0.09680192,0.35454107);
  rgb(106pt)=(0.65824241,0.09552918,0.3536529);
  rgb(107pt)=(0.66492652,0.09428017,0.3526697);
  rgb(108pt)=(0.67159578,0.09306598,0.35159077);
  rgb(109pt)=(0.67824099,0.09192342,0.3504148);
  rgb(110pt)=(0.684863,0.09085633,0.34914061);
  rgb(111pt)=(0.69146268,0.0898675,0.34776864);
  rgb(112pt)=(0.69803757,0.08897226,0.3462986);
  rgb(113pt)=(0.70457834,0.0882129,0.34473046);
  rgb(114pt)=(0.71108138,0.08761223,0.3430635);
  rgb(115pt)=(0.7175507,0.08716212,0.34129974);
  rgb(116pt)=(0.72398193,0.08688725,0.33943958);
  rgb(117pt)=(0.73035829,0.0868623,0.33748452);
  rgb(118pt)=(0.73669146,0.08704683,0.33543669);
  rgb(119pt)=(0.74297501,0.08747196,0.33329799);
  rgb(120pt)=(0.74919318,0.08820542,0.33107204);
  rgb(121pt)=(0.75535825,0.08919792,0.32876184);
  rgb(122pt)=(0.76145589,0.09050716,0.32637117);
  rgb(123pt)=(0.76748424,0.09213602,0.32390525);
  rgb(124pt)=(0.77344838,0.09405684,0.32136808);
  rgb(125pt)=(0.77932641,0.09634794,0.31876642);
  rgb(126pt)=(0.78513609,0.09892473,0.31610488);
  rgb(127pt)=(0.79085854,0.10184672,0.313391);
  rgb(128pt)=(0.7965014,0.10506637,0.31063031);
  rgb(129pt)=(0.80205987,0.10858333,0.30783);
  rgb(130pt)=(0.80752799,0.11239964,0.30499738);
  rgb(131pt)=(0.81291606,0.11645784,0.30213802);
  rgb(132pt)=(0.81820481,0.12080606,0.29926105);
  rgb(133pt)=(0.82341472,0.12535343,0.2963705);
  rgb(134pt)=(0.82852822,0.13014118,0.29347474);
  rgb(135pt)=(0.83355779,0.13511035,0.29057852);
  rgb(136pt)=(0.83850183,0.14025098,0.2876878);
  rgb(137pt)=(0.84335441,0.14556683,0.28480819);
  rgb(138pt)=(0.84813096,0.15099892,0.281943);
  rgb(139pt)=(0.85281737,0.15657772,0.27909826);
  rgb(140pt)=(0.85742602,0.1622583,0.27627462);
  rgb(141pt)=(0.86196552,0.16801239,0.27346473);
  rgb(142pt)=(0.86641628,0.17387796,0.27070818);
  rgb(143pt)=(0.87079129,0.17982114,0.26797378);
  rgb(144pt)=(0.87507281,0.18587368,0.26529697);
  rgb(145pt)=(0.87925878,0.19203259,0.26268136);
  rgb(146pt)=(0.8833417,0.19830556,0.26014181);
  rgb(147pt)=(0.88731387,0.20469941,0.25769539);
  rgb(148pt)=(0.89116859,0.21121788,0.2553592);
  rgb(149pt)=(0.89490337,0.21785614,0.25314362);
  rgb(150pt)=(0.8985026,0.22463251,0.25108745);
  rgb(151pt)=(0.90197527,0.23152063,0.24918223);
  rgb(152pt)=(0.90530097,0.23854541,0.24748098);
  rgb(153pt)=(0.90848638,0.24568473,0.24598324);
  rgb(154pt)=(0.911533,0.25292623,0.24470258);
  rgb(155pt)=(0.9144225,0.26028902,0.24369359);
  rgb(156pt)=(0.91717106,0.26773821,0.24294137);
  rgb(157pt)=(0.91978131,0.27526191,0.24245973);
  rgb(158pt)=(0.92223947,0.28287251,0.24229568);
  rgb(159pt)=(0.92456587,0.29053388,0.24242622);
  rgb(160pt)=(0.92676657,0.29823282,0.24285536);
  rgb(161pt)=(0.92882964,0.30598085,0.24362274);
  rgb(162pt)=(0.93078135,0.31373977,0.24468803);
  rgb(163pt)=(0.93262051,0.3215093,0.24606461);
  rgb(164pt)=(0.93435067,0.32928362,0.24775328);
  rgb(165pt)=(0.93599076,0.33703942,0.24972157);
  rgb(166pt)=(0.93752831,0.34479177,0.25199928);
  rgb(167pt)=(0.93899289,0.35250734,0.25452808);
  rgb(168pt)=(0.94036561,0.36020899,0.25734661);
  rgb(169pt)=(0.94167588,0.36786594,0.2603949);
  rgb(170pt)=(0.94291042,0.37549479,0.26369821);
  rgb(171pt)=(0.94408513,0.3830811,0.26722004);
  rgb(172pt)=(0.94520419,0.39062329,0.27094924);
  rgb(173pt)=(0.94625977,0.39813168,0.27489742);
  rgb(174pt)=(0.94727016,0.4055909,0.27902322);
  rgb(175pt)=(0.94823505,0.41300424,0.28332283);
  rgb(176pt)=(0.94914549,0.42038251,0.28780969);
  rgb(177pt)=(0.95001704,0.42771398,0.29244728);
  rgb(178pt)=(0.95085121,0.43500005,0.29722817);
  rgb(179pt)=(0.95165009,0.44224144,0.30214494);
  rgb(180pt)=(0.9524044,0.44944853,0.3072105);
  rgb(181pt)=(0.95312556,0.45661389,0.31239776);
  rgb(182pt)=(0.95381595,0.46373781,0.31769923);
  rgb(183pt)=(0.95447591,0.47082238,0.32310953);
  rgb(184pt)=(0.95510255,0.47787236,0.32862553);
  rgb(185pt)=(0.95569679,0.48489115,0.33421404);
  rgb(186pt)=(0.95626788,0.49187351,0.33985601);
  rgb(187pt)=(0.95681685,0.49882008,0.34555431);
  rgb(188pt)=(0.9573439,0.50573243,0.35130912);
  rgb(189pt)=(0.95784842,0.51261283,0.35711942);
  rgb(190pt)=(0.95833051,0.51946267,0.36298589);
  rgb(191pt)=(0.95879054,0.52628305,0.36890904);
  rgb(192pt)=(0.95922872,0.53307513,0.3748895);
  rgb(193pt)=(0.95964538,0.53983991,0.38092784);
  rgb(194pt)=(0.96004345,0.54657593,0.3870292);
  rgb(195pt)=(0.96042097,0.55328624,0.39319057);
  rgb(196pt)=(0.96077819,0.55997184,0.39941173);
  rgb(197pt)=(0.9611152,0.5666337,0.40569343);
  rgb(198pt)=(0.96143273,0.57327231,0.41203603);
  rgb(199pt)=(0.96173392,0.57988594,0.41844491);
  rgb(200pt)=(0.96201757,0.58647675,0.42491751);
  rgb(201pt)=(0.96228344,0.59304598,0.43145271);
  rgb(202pt)=(0.96253168,0.5995944,0.43805131);
  rgb(203pt)=(0.96276513,0.60612062,0.44471698);
  rgb(204pt)=(0.96298491,0.6126247,0.45145074);
  rgb(205pt)=(0.96318967,0.61910879,0.45824902);
  rgb(206pt)=(0.96337949,0.6255736,0.46511271);
  rgb(207pt)=(0.96355923,0.63201624,0.47204746);
  rgb(208pt)=(0.96372785,0.63843852,0.47905028);
  rgb(209pt)=(0.96388426,0.64484214,0.4861196);
  rgb(210pt)=(0.96403203,0.65122535,0.4932578);
  rgb(211pt)=(0.96417332,0.65758729,0.50046894);
  rgb(212pt)=(0.9643063,0.66393045,0.5077467);
  rgb(213pt)=(0.96443322,0.67025402,0.51509334);
  rgb(214pt)=(0.96455845,0.67655564,0.52251447);
  rgb(215pt)=(0.96467922,0.68283846,0.53000231);
  rgb(216pt)=(0.96479861,0.68910113,0.53756026);
  rgb(217pt)=(0.96492035,0.69534192,0.5451917);
  rgb(218pt)=(0.96504223,0.7015636,0.5528892);
  rgb(219pt)=(0.96516917,0.70776351,0.5606593);
  rgb(220pt)=(0.96530224,0.71394212,0.56849894);
  rgb(221pt)=(0.96544032,0.72010124,0.57640375);
  rgb(222pt)=(0.96559206,0.72623592,0.58438387);
  rgb(223pt)=(0.96575293,0.73235058,0.59242739);
  rgb(224pt)=(0.96592829,0.73844258,0.60053991);
  rgb(225pt)=(0.96612013,0.74451182,0.60871954);
  rgb(226pt)=(0.96632832,0.75055966,0.61696136);
  rgb(227pt)=(0.96656022,0.75658231,0.62527295);
  rgb(228pt)=(0.96681185,0.76258381,0.63364277);
  rgb(229pt)=(0.96709183,0.76855969,0.64207921);
  rgb(230pt)=(0.96739773,0.77451297,0.65057302);
  rgb(231pt)=(0.96773482,0.78044149,0.65912731);
  rgb(232pt)=(0.96810471,0.78634563,0.66773889);
  rgb(233pt)=(0.96850919,0.79222565,0.6764046);
  rgb(234pt)=(0.96893132,0.79809112,0.68512266);
  rgb(235pt)=(0.96935926,0.80395415,0.69383201);
  rgb(236pt)=(0.9698028,0.80981139,0.70252255);
  rgb(237pt)=(0.97025511,0.81566605,0.71120296);
  rgb(238pt)=(0.97071849,0.82151775,0.71987163);
  rgb(239pt)=(0.97120159,0.82736371,0.72851999);
  rgb(240pt)=(0.97169389,0.83320847,0.73716071);
  rgb(241pt)=(0.97220061,0.83905052,0.74578903);
  rgb(242pt)=(0.97272597,0.84488881,0.75440141);
  rgb(243pt)=(0.97327085,0.85072354,0.76299805);
  rgb(244pt)=(0.97383206,0.85655639,0.77158353);
  rgb(245pt)=(0.97441222,0.86238689,0.78015619);
  rgb(246pt)=(0.97501782,0.86821321,0.78871034);
  rgb(247pt)=(0.97564391,0.87403763,0.79725261);
  rgb(248pt)=(0.97628674,0.87986189,0.8057883);
  rgb(249pt)=(0.97696114,0.88568129,0.81430324);
  rgb(250pt)=(0.97765722,0.89149971,0.82280948);
  rgb(251pt)=(0.97837585,0.89731727,0.83130786);
  rgb(252pt)=(0.97912374,0.90313207,0.83979337);
  rgb(253pt)=(0.979891,0.90894778,0.84827858);
  rgb(254pt)=(0.98067764,0.91476465,0.85676611);
  rgb(255pt)=(0.98137749,0.92061729,0.86536915)
}]
table{%
x                      y
1 0.0615945198681702
1 0.0470157215132844
1 0.079246041180504
1 0.0903657357679104
1 0.0526154887039798
1 0.0923832975872036
1 0.0579270170053269
1 0.0749210886743333
1 0.0475468745269669
1 0.0694958448940233
1 0.0696017952659514
1 0.00736943936160203
2 0.108294284692883
2 0.0518210336653495
2 0.0908439801987521
2 0.12779681998558
2 0.0597662446907971
2 0.116452161561392
2 0.0524012807183584
2 0.13859562337899
2 0.0922783122963243
2 0.0725196762480033
2 0.108130116953586
2 0.100999905025512
3 0.0829307431294015
3 0.0745046819792106
3 0.0765907958166007
3 0.0791930735667293
3 0.0535255794470864
3 0.100351070566075
3 0.0981264408150325
3 0.0520566442530147
3 0.0838934842534086
3 0.0831265037182453
3 0.0722144118298308
3 0.0939263277898666
4 0.173171904887567
4 0.0599356361885883
4 0.138727672826751
4 0.0611094043706372
4 0.00519310927090847
4 0.0525238853455329
4 0.112901518169558
4 0.0693754476786355
4 0.0608172238500697
4 0.0413964375703182
4 0.0493134518782322
4 0.0666103232602997
5 -0.0026485145879878
5 0.051641240590904
5 0.0623611585951521
5 0.0516628737865963
5 0.0304389113413801
5 0.0532592630190357
5 0.0397005194033794
5 0.0343728577414366
5 0.0793387577436172
5 0.0655021668456183
5 0.0536170846507792
5 0.0254979264537173
6 0.0391221593596291
6 0.0302571477604251
6 0.0538390770569962
6 0.0535094216820067
6 0.0692036698064926
6 0.02706684571687
6 0.0250028962165954
6 0.0274346944333726
6 0.0383428204843139
6 0.0430009982390283
6 0.051891462035359
6 0.0480157083896929
7 0.04372775837847
7 0.0490772144960798
7 0.0511335945360206
7 0.00860762941985177
7 0.017416617356051
7 0.0215267696835319
7 0.0306838092818919
7 0.0471370601884031
7 0.0593275336844154
7 0.0755554335574164
7 0.0455870759406165
7 0.0657621862602496
8 0.115169506446741
8 0.045079876103427
8 0.0448774977238551
8 -0.00103559448812834
8 0.0883238431069557
8 0.0151862904760152
8 0.0117641913681162
8 0.0336261322551678
8 0.143820551742626
8 0.0511702255856811
8 0.0144666654540233
8 0.0124541350425131
9 0.043356527710874
9 0.0883079816595838
9 0.0384587874244441
9 0.0734552961824789
9 0.0636001524259159
9 0.0246177766978301
9 0.0679577033265391
9 0.0604094333608688
9 0.0764208756987677
9 0.0500322237810296
9 0.0208820347835544
9 0.064780844362963
10 0.0790918437577209
10 0.0476214321196282
10 0.0915956212916938
10 0.121268465193346
10 0.0349721282145248
10 0.0537828743359333
10 0.139902190486972
10 0.0336105431257683
10 0.107701818601332
10 0.0810823823105411
10 0.0602274398021768
10 0.103962799261393
11 0.0601974413267434
11 0.0870655758410429
11 0.159926530146423
11 0.142868609966175
11 0.125720065466382
11 0.0714159208971926
11 0.12256188830187
11 0.131297706539876
11 0.0918624390858074
11 0.145967702493263
11 0.139298836744761
11 0.101084355272347
12 0.0427329309317301
12 0.0714806175499002
12 0.0474253488900034
12 0.0480731060615894
12 0.0538861852033716
12 0.147600928566534
12 0.0780998267321932
12 0.0554104377755975
12 0.0538621565193344
12 0.0603027819191373
12 0.0511105841652514
12 0.0761009122280644
};
\addlegendentry{. or ,}
\addplot [semithick, color0, forget plot]
table {%
1 0.0227736548995818
2 0.106517683876547
3 0.0779425207539609
4 0.0490439203254898
5 0.0222228950895668
6 0.0149267111542276
7 0.0120155348163907
8 0.00575001686950369
9 0.0104030039963339
10 0.0137259502310503
11 0.0182344600028579
12 0.00747327447634338
};
\addplot [semithick, color1, forget plot]
table {%
1 0.0206336456697253
2 0.0524462727232086
3 0.0816616612368571
4 0.0512892742586024
5 0.0278151210766517
6 0.0356010810393863
7 0.0331178462012967
8 0.0334134804501031
9 0.0325438503195999
10 0.0330110534838544
11 0.0456275796762605
12 0.0186268923587067
};
\addplot [semithick, color2, forget plot]
table {%
1 0.062506905362438
2 0.0933249532846272
3 0.0792033130970419
4 0.0742563346080915
5 0.0453953537986357
6 0.0422239084317318
7 0.0429618902319165
8 0.0479086100680827
9 0.0560233031179041
10 0.0795682948750858
11 0.11493892267349
12 0.0655071513785589
};
\end{axis}

\end{tikzpicture}

%% file: sections/context.tex
\input{sections/token.tex}

\section{Attribution Analysis to Identify Context Contribution}\label{sec:attributionPart}
We consider the role of the contextual information in the hidden embeddings, which is accumulated through multiple paths in a multi-layer network. To shed more light on this process,
we introduce \emph{Hidden Token Attribution}, a context quantification method based on gradient attribution~\citep{gradientAttribution} to investigate the hidden tokens' sensitivity with respect to the input tokens. 

Gradient based attribution approximates the neural network function $f(\mX)$ around a given sequence of input word embeddings $\mX\in \R^{d_s\times d}$ by the linear part of the Taylor expansion: 
\begin{equation}
f(\mX + \Delta \mX) \approx f(\mX) + \nabla_{\mX} f(\mX)^T\cdot\Delta \mX
\end{equation}

With this, the network sensitivity is analyzed by looking at how small changes $\Delta \mX$ at the input correlate with changes at the output. Since in the linear approximation this change is given by the gradient $\nabla_{\vx_i}f=\frac{\delta f(\mX)}{\delta \vx_i}$ for a change in the $i$-th input token $\vx_i\in \R^d$ of $\mX$, the attribution of how much input token $\vx_i$ affects the network output $f(\mX)$ can be approximated by the $L_2$ norm of the respective gradient: $\text{attr}(\vx_i) = \left|\left|\nabla_{\vx_i}f\right|\right|_2$. Since we are interested in how much a given hidden embedding $\ve_j^l$ attributes to the input tokens $\vx_i$, $i\in [1, 2, \dots, d_s]$, we define the relative input contribution $c^l_{i,j}$ of input $\vx_i$ to output $f(\mX)=\ve_j^l$ as
\begin{equation}
c_{i,j}^l = \frac{||\nabla_{i,j}^l||_2}{\sum_{k=0}^{d_s}||\nabla_{k,j}^l||_2}\qquad\text{with}\quad \nabla_{i,j}^l =\frac{\delta \ve_j^l}{\delta \vx_i}
\end{equation}
Since we normalize by dividing by the sum of the attribution values to all input tokens, we obtain values between 0 and 1 that represent the \emph{contribution} of each input token $\vx_i$ to the hidden embedding $\ve_j^l$. \emph{Hidden Token Attribution} differs from the standard use of gradient attribution in that, instead of taking the gradients of the output of the model with respect to the inputs in order to explain the model's decision, we calculate the contribution of the inputs to intermediate embeddings in order to track the mixing of information. Further details of this method are discussed in Appendix~\ref{AppGrad}.

\begin{figure}[t]
\centering
\begin{subfigure}[t]{.49\textwidth}
\centering
\input{gradient_notfinetuned_task_indep_figs/contribution_all_task_independent_not_fine_tuned_mrpc.tex}
\caption{}
\label{contr_all}
\end{subfigure}
\hfill
\begin{subfigure}[t]{.49\textwidth}
\centering
\input{gradient_notfinetuned_task_indep_figs/rank_all_task_independent_not_fine_tuned_mrpc.tex}
\caption{}
\label{rank_all}
\end{subfigure}%
\caption{(a) Contribution of the input token to the embedding at the same position. The orange line represents the median value and outliers are not shown. (b) Percentage of tokens $\tilde{P}$ that are \emph{not} the main contributors to their corresponding contextual embedding at each 
layer.}
\end{figure}

\subsection{Token Mixing: Contribution of Input Tokens} 

We use Hidden Token Attribution to extend the results of Section~\ref{IdenTok} showing \emph{how much} of the input token is contained in a given hidden embedding. 
In Figure~\ref{contr_all} we report the contribution $c_{j,j}^l$ of input tokens $\vx_j$ to their corresponding hidden embeddings $\ve_j^l$ at the same position $j$ for each layer $l$. After the first layer the median contribution of the input token is less than a third (30.6\%). The contribution then decreases monotonically with depth; at layer 6 the median is only 14.4\% and after the last layer it is 10.7\%. In Appendix~\ref{AppPOS} we provide detailed results by word type.
Next, we study which input token is the largest contributor to a given hidden embedding $\ve_j^l$. The corresponding input token $\vx_j$ generally has the largest contribution. Figure~\ref{rank_all} shows the percentage $\tilde{P}$ of tokens that are \emph{not} the highest contributor to their hidden embedding at each layer. In the first three layers the original input $\vx_j$ always contributes the most to the embedding $\ve_j^l$. In subsequent layers, $\tilde{P}$ increases monotonically, reaching 18\% in the sixth layer and 30\% in the last two layers. 

These results show that, starting from layer three, self-attention strongly mixes the input information by aggregating context into the hidden embeddings. This is in line with the results from Section~\ref{IdenTok}, where we see a decrease in token identifiability rate after layer three. Nevertheless, $\tilde{P}$ is always higher than the token identifiability error at the same layer, indicating that tokens are mixed in a way that often permits recovering token identity even if the contribution of the original token is outweighed by other tokens. This suggests that there is some \quotes{identity information} that is preserved through the layers. 

The strong mixing of information questions the common assumption that attention distributions can be interpreted as \quotes{how much a word attends to another word}. However, the fact that tokens remain identifiable despite information mixing opens a number of new interesting questions to be addressed by future research. In particular, this seeming contradiction may be solved by investigating the space in which hidden embeddings operate: is there a relation between semantics and geometric distance for hidden embeddings? Are some embedding dimensions more important than others?

\subsection{Contribution of Context to Hidden Tokens}\label{sec:Contxt} 

In this section we study how context is aggregated into hidden embeddings. Figure~\ref{rel_attr} shows the relative contribution of neighbouring tokens at each layer for the relative positions: first, second, third, fourth and fifth together, sixth to 10th together, and the rest. The closest neighbours (1st) contribute significantly more in the first layers than in later layers. Conversely, the most distant neighbours (11th onwards) contribute the most in deeper layers (cf. Appendix~\ref{AppCtxtDet}).
\begin{figure}[t]
\begin{center}
\begin{subfigure}[t]{.49\textwidth}
\input{gradient_notfinetuned_task_indep_figs/relative_attr_per_layer.tex}
\caption{}\label{rel_attr}
\end{subfigure}%
\begin{subfigure}[t]{.49\textwidth}
\input{gradient_notfinetuned_task_indep_figs/context_per_token.tex}
\caption{}\label{tot_attr}
\end{subfigure}%
\end{center}
\caption{(a) Relative contribution per layer of neighbours at different positions. 
(b) Total contribution per neighbour for the first, middle and last layers. 
}\label{self_attention_is_local}
\end{figure}
Despite the progressive increase in long-range dependencies, the context in the hidden embeddings remains mostly local. Figure~\ref{tot_attr} represents the normalized total contribution aggregated over all tokens from each of their neighbours at the first, middle and last layer. This figure shows that the closest neighbours consistently contribute the most to the contextual word embedding regardless of depth. On the other hand, we indeed observe an increase of distant contributions at later layers.

The results of this section suggest that BERT learns local operators from data in an unsupervised manner, in the absence of any such prior in the architecture. This behavior is not obvious, since attention is a highly non-local operator, and in turn indicates the importance of local dependencies in natural language. While contribution is local \emph{on average}, we find that there are exceptions, such as the [CLS] token (cf. Appendix~\ref{AppCLS}). Furthermore, using our Hidden Token Attribution method, one can track how context is aggregated for specific tokens (cf. Appendix~\ref{AppTrack}).

%% file: sections/token.tex
\section{Token Identifiability}\label{IdenTok}
We now study the other fundamental element of transformers; the hidden vector representations of tokens, or contextual word embeddings. 
It is commonly assumed that a contextual word embedding keeps its \quotes{identity}, which is tied to the input word, as it passes through the self-attention layers. 
Specifically, we identify three cases where this assumption is made implicitly without justification.

\begin{itemize}
    \item Visualizations/interpretations linking attention weights to attention between words, when in fact the attention is between embeddings, i.e., mixtures of multiple words~\citep{attentionIsAllYouNeed,BERT,vigTransformerViz,whatBertLooksAt,transfEncAnalysis,heavyliftingpruning,analysisAttnWordSenseDisam,attendToMathTransformer,bertpassageranking,transformerheadstransparency?,universalTransformers,wordAlignment}.
    \item Attention accumulation methods that sum the attention to a specific sequence position over layers and/or attention heads, when the given position might encode a different mixture of inputs in each layer~\citep{whatBertLooksAt,transformerheadstransparency?,commonsenseReasoning,visualizingBertGeometry}.
   \item Using classifiers to probe hidden embeddings for word-specific aspects without factoring in how much the word is still represented~\citep{lin2019opensesame,DissectingContextualEmbeddings}.
\end{itemize}
To investigate this assumption we introduce the concept of \emph{token identifiability}, as the existence of a mapping assigning contextual embeddings to their corresponding input tokens. 
Formally, we state that an embedding $\ve_i^l$ is identifiable if there exists a classification function $c(\cdot)$ such that $c(\ve_i^l) = \vx_i$. For identifiability we only require $c(\ve_i^l)$ to recover $\vx_i$ in a nearest neighbour sense within the same input sentence. 
Therefore, for each layer $l$ we define $c_l(\cdot)=NN(g_l(\cdot))$, where $NN$ is a 1-nearest neighbour lookup, and $g_l : \mathbb{R}^d \rightarrow \mathbb{R}^d$ is a continuous function mapping embeddings to vectors of real numbers.
Since we cannot prove the existence of $g_l$ analytically, we instead use a function approximator $\hat{g}_l(\ve_i^l)=\hat{\vx}_i$, trained on a dataset of $(\ve_i^l,\vx_i)$ pairs. 
We then say that $\ve_i^l$ is identifiable if $\hat{c}_l(\ve_i^l)=NN(\hat{g}_l(\ve_i^l))=\vx_i$. For evaluation we report the \emph{token identifiability rate} defined as the percentage of correctly identified tokens.

\subsection{Setup}\label{subsec:tokenIdentExpSetup}
For the experiments in this and subsequent sections we use the development dataset from the Microsoft Research Paraphrase Corpus (MRPC) dataset~\citep{MRPC}, while in Appendix \ref{AppDataSets} we provide results on two additional datasets. The MRPC development set contains 408 examples with a sequence length $d_s$ between 26 and 92 tokens, with 58 tokens on average.
We pass all 408 sentences (21,723 tokens) through BERT and extract for each token the input embeddings $\vx_i$ and the hidden embeddings $\ve_i^l$ at all layers. We then train $\hat{g}$ on the regression task of predicting input tokens $\vx_i$ from hidden tokens $\ve_i^l$. We experiment with two loss functions and similarity measures for finding the nearest neighbour; cosine distance and $L_2$ distance. We use 10-fold cross validation with 70/15/15 train/validation/test splits per fold and ensure that tokens from the same sentence are not split across sets. The validation set is used for early stopping. See Appendix~\ref{App_tokenClassifier_SetupDetails} for further details.

\subsection{Experimental Results and Discussion}\label{subsec:tokenIdentExperiments}

\begin{figure}[t]
\centering
\begin{subfigure}[t]{.49\textwidth}
\centering
\input{token_classification_experiments/rebuttal_plots/MainTokenIdentityResults/TokenIdentity_mlpLinearBaseline_with_cosAndL2.tex}
\caption{}
\label{fig:tokenclassifierresults_a}
\end{subfigure}
\hfill
\begin{subfigure}[t]{.49\textwidth}
\centering
\input{token_classification_experiments/rebuttal_plots/GeneralizingAcrossLayers/TokenIdentity_GeneralizeAcrossLayers_LINEAR_COSINE_Layers1-6-11-12_10FoldCV.tex}
\caption{}
\label{fig:tokenclassifierresults_b}
\end{subfigure}%
\caption{(a) Identifiability of contextual word embeddings at different layers. Here, $\hat{g}$ is trained and tested on the same layer.
(b) $g_{cos,l}^{lin}$ trained on layer $l$ and tested on all layers.}%\label{fig:tokenclassifierresults}
\end{figure}

In a first experiment, we use a linear perceptron without bias and a non-linear MLP $\hat{g}_l^{MLP}$, where training, validation and test data all come from layer $l$. 
Figure~\ref{fig:tokenclassifierresults_a} shows the test set token identifiability rate of $\hat{c}_l$ for $l=[1,...,12]$. We also report a naive baseline $\hat{g}_l^{naive}(\ve_i^l) = \ve_i^l$, i.e., we directly retrieve the nearest neighbour of $\ve_i^l$ from the input tokens.
The results for $\hat{g}_l^{naive}$ show that, according to both similarity measures, contextual embeddings in BERT stay close to their input embeddings up to layer 4, followed by a linear decrease in token identifiability rate. %If we train
By training a transformation to recover the original embedding, we see that most of the identity information is still present in the contextualized embeddings. Specifically, a linear projection is enough to recover 93\% of the tokens in the last layer based on a cosine distance nearest neighbour lookup.

This experiment shows that although the identifiablity rate decreases with depth, tokens remain mostly identifiable across layers. Furthermore, we find that cosine distance is more effective to recover identity than $L_2$ distance. Therefore, we conjecture that BERT encodes most of the identity information in the angle of the embeddings.
Finally, \cite{lin2019opensesame} show that BERT discards much of the positional information after layer 3. 
However, tokens remain largely identifiable throughout the model, indicating that BERT does not only rely on the positional embeddings to track token identity.
To provide further insights into contextual word embeddings, Appendix~\ref{appendix:neighbouringTokenIdentity} shows results for recovering neighbouring input tokens.

In a second experiment we test how well the $\hat{g}_{cos,l}^{lin}$ trained only on $(\ve_i^l,\vx_i)$ pairs from one layer $l$ generalizes to all layers, see Figure~\ref{fig:tokenclassifierresults_b}. For $l=1$, the token identifiability rate on subsequent layers drops quickly to below 70\% at layers 11 and 12. Interestingly, for $l=12$ a very different pattern can be observed, where the identifiability is 94\% for layer 12 and then almost monotonically increases when testing on earlier layers. Further, for $l=6$ we see both patterns.

This experiment suggests that the nature of token identity changes as tokens pass through the model, and patterns learned on data from later layers transfer well to earlier layers. 
The experiment also shows that layer 12 is behaving differently than the other layers. In particular, generalizing \emph{to} layer 12 from layer 11 seems to be difficult, signified by a sudden drop in token identifiability rate.
We believe this is due to a task dependent parameter adaptation
induced in the last layer by the next-sentence prediction task which only uses the CLS token (cf. Appendix~\ref{appendix:tokenident:hiddentohidden} for additional hints that the last layer(s) behave differently). See Appendix~\ref{App_tokenClassifier_GeneralizingLayersFullResults} for results of $\hat{g}_{l2,l}^{lin}$, $\hat{g}_{l2,l}^{MLP}$ and $\hat{g}_{cos,l}^{MLP}$.

Overall, the results of this section suggest that one can associate most hidden embeddings with their input token, for example for drawing conclusions based on (effective) attention weights. 
However, self-attention has the potential to strongly mix tokens across multiple layers, and hence it is unclear whether token identifiability alone is enough to equate hidden embeddings with their input words, or whether we also need to take into account exactly \emph{how much} of the word is still contained in the hidden embedding. 
In order to address this question, we now study the degree of information mixing among embeddings, and introduce a tool to track the contributions of tokens to embeddings throughout the model.

%% file: token_classification_experiments/rebuttal_plots/MainTokenIdentityResults/TokenIdentity_mlpLinearBaseline_with_cosAndL2.tex
% This file was created by tikzplotlib v0.8.2.
\begin{tikzpicture}

\begin{axis}[
legend cell align={left},
legend columns = 2,
legend style={at={(0,0)},anchor=south west,font=\tiny, draw=white!80.0!black},
tick align=outside,
tick pos=both,
x grid style={lightgray!92.02614379084967!black},
xlabel={Layer},
xmajorgrids,
width = 0.98\linewidth,
height = \tokenClassifierPlotHeight,
xmin=1, xmax=12,
xtick style={color=black},
y grid style={lightgray!92.02614379084967!black},
ylabel={Identifiability Rate},
ymajorgrids,
ymin=0.1, ymax=1.05,
ytick style={color=black}
]
\addplot [thick, blue]
table {%
1 1
2 1
3 1
4 0.999720562512722
5 0.999411571563954
6 0.997551603413987
7 0.995041071100503
8 0.99288084860688
9 0.988352926354295
10 0.977237935066728
11 0.948640573198221
12 0.939551531117988
};
\addlegendentry{$\hat{g}_{cos,l}^{MLP}$}
\addplot [thick, blue, dotted]
table {%
1 1
2 1
3 1
4 0.999954022988506
5 0.998716952546194
6 0.995661791087932
7 0.993190312694488
8 0.990985247942984
9 0.986851925862233
10 0.978540890445112
11 0.953781119749692
12 0.934482775598561
};
\addlegendentry{$\hat{g}_{cos,l}^{lin}$}
\addplot [thick, blue, dashed]
table {%
1 1
2 0.999447590111863
3 0.982414951894306
4 0.966441099295677
5 0.901993279013028
6 0.814896653316761
7 0.737467200662892
8 0.617318049993095
9 0.551857478248861
10 0.411499332504718
11 0.211941260415228
12 0.489665331676104
};
\addlegendentry{$\hat{g}_{cos,l}^{naive}$}
\addplot [thick, black]
table {%
1 1
2 1
3 0.999954022988506
4 0.998297522559177
5 0.988753483600016
6 0.978824743402777
7 0.96535017058847
8 0.952245663400465
9 0.930002052525579
10 0.880599672426699
11 0.82392226130497
12 0.819659406779177
};
\addlegendentry{$\hat{g}_{l2,l}^{MLP}$}
\addplot [thick, black, dotted]
table {%
1 1
2 1
3 0.999125258026798
4 0.988475486146777
5 0.981853030065317
6 0.969192708945533
7 0.945448901786302
8 0.923599170471342
9 0.893778659012765
10 0.817093700077433
11 0.769724237587454
12 0.73293072117683
};
\addlegendentry{$\hat{g}_{l2,l}^{lin}$}
\addplot [thick, black, dashed]
table {%
1 1
2 0.999677760898587
3 0.988721631450536
4 0.969019012106983
5 0.894351608893799
6 0.79763384431248
7 0.699442986696128
8 0.521152695299912
9 0.408599180591999
10 0.276297012383188
11 0.132670441467569
12 0.187589191179855
};
\addlegendentry{$\hat{g}_{l2,l}^{naive}$}
\end{axis}

\end{tikzpicture}

%% file: token_classification_experiments/rebuttal_plots/GeneralizingAcrossLayers/TokenIdentity_GeneralizeAcrossLayers_LINEAR_COSINE_Layers1-6-11-12_10FoldCV.tex
% This file was created by tikzplotlib v0.8.2.
\begin{tikzpicture}

\definecolor{color3}{rgb}{0.83921568627451,0.152941176470588,0.156862745098039}
\definecolor{color0}{rgb}{0.12156862745098,0.466666666666667,0.705882352941177}
\definecolor{color2}{rgb}{0.172549019607843,0.627450980392157,0.172549019607843}
\definecolor{color1}{rgb}{1,0.498039215686275,0.0549019607843137}

\begin{axis}[
legend cell align={left},
legend columns = 2,
legend style={at={(0,0)},anchor=south west,font=\tiny, draw=white!80.0!black},
tick align=outside,
tick pos=both,
x grid style={lightgray!92.02614379084967!black},
xlabel={Layer},
xmajorgrids,
width = 0.98\linewidth,
height = \tokenClassifierPlotHeight,
xmin=1, xmax=12,
xtick style={color=black},
y grid style={lightgray!92.02614379084967!black},
ylabel={Identifiability Rate},
ymajorgrids,
ymin=0.65, ymax=1.05,
ytick style={color=black}
]
\addplot [thick, color0]
table {%
1 1
2 1
3 0.981943658370596
4 0.974384512625004
5 0.967806366904788
6 0.95981816347002
7 0.954210191049799
8 0.933868968553002
9 0.89102516263228
10 0.812343264286084
11 0.675274297049946
12 0.685153644930376
};
\addlegendentry{l=1}
\addplot [thick, color1]
table {%
1 0.997509973478275
2 0.999172382880112
3 0.995202567379097
4 0.994312859244359
5 0.993978270232952
6 0.995661791087932
7 0.989965625665474
8 0.981032771467164
9 0.934871552377645
10 0.876374432850787
11 0.821774222675968
12 0.773986982968302
};
\addlegendentry{l=6}
\addplot [thick, color2]
table {%
1 0.991954410179278
2 0.993359342682405
3 0.979693386659757
4 0.976592349902883
5 0.976372578990465
6 0.973976793098691
7 0.971681246854108
8 0.967950966808659
9 0.957353785067388
10 0.924814518843242
11 0.953920470832959
12 0.837012492127621
};
\addlegendentry{l=11}
\addplot [thick, color3]
table {%
1 0.975025294522085
2 0.984020283882439
3 0.977045565463492
4 0.976195811608137
5 0.975158005152879
6 0.973111934434619
7 0.971321194413633
8 0.966590231835631
9 0.952739332037229
10 0.922344687457931
11 0.916836991108777
12 0.935090715445356
};
\addlegendentry{l=12}
\end{axis}

\end{tikzpicture}

%% file: gradient_notfinetuned_task_indep_figs/contribution_all_task_independent_not_fine_tuned_mrpc.tex
% This file was created by tikzplotlib v0.8.2.
\begin{tikzpicture}

\definecolor{color0}{rgb}{1,0.498039215686275,0.0549019607843137}

\begin{axis}[
tick align=outside,
tick pos=both,
x grid style={white!69.01960784313725!black},
width=0.97\linewidth,
xlabel={Layer},
xmin=0.5, xmax=12.5,
height=\contributionPlotsHeightFirst,
xtick style={color=black},
xtick={1,2,3,4,5,6,7,8,9,10,11,12},
xticklabels={1,2,3,4,5,6,7,8,9,10,11,12},
y grid style={white!69.01960784313725!black},
ylabel={Contribution [\%]},
ytick={0,0.1,0.2,0.3},
yticklabels={0,10, 20, 30},
ymin=-0.01, ymax=0.4,
ytick style={color=black},
xmajorgrids,
ymajorgrids
]
\addplot [black]
table {%
0.75 0.288386180996895
1.25 0.288386180996895
1.25 0.323633223772049
0.75 0.323633223772049
0.75 0.288386180996895
};
\addplot [black]
table {%
1 0.288386180996895
1 0.235526546835899
};
\addplot [black]
table {%
1 0.323633223772049
1 0.376225084066391
};
\addplot [black]
table {%
0.875 0.235526546835899
1.125 0.235526546835899
};
\addplot [black]
table {%
0.875 0.376225084066391
1.125 0.376225084066391
};
\addplot [black]
table {%
1.75 0.215413510799408
2.25 0.215413510799408
2.25 0.245689995586872
1.75 0.245689995586872
1.75 0.215413510799408
};
\addplot [black]
table {%
2 0.215413510799408
2 0.170352607965469
};
\addplot [black]
table {%
2 0.245689995586872
2 0.291104108095169
};
\addplot [black]
table {%
1.875 0.170352607965469
2.125 0.170352607965469
};
\addplot [black]
table {%
1.875 0.291104108095169
2.125 0.291104108095169
};
\addplot [black]
table {%
2.75 0.197865888476372
3.25 0.197865888476372
3.25 0.231675706803799
2.75 0.231675706803799
2.75 0.197865888476372
};
\addplot [black]
table {%
3 0.197865888476372
3 0.147297322750092
};
\addplot [black]
table {%
3 0.231675706803799
3 0.282358914613724
};
\addplot [black]
table {%
2.875 0.147297322750092
3.125 0.147297322750092
};
\addplot [black]
table {%
2.875 0.282358914613724
3.125 0.282358914613724
};
\addplot [black]
table {%
3.75 0.165827795863152
4.25 0.165827795863152
4.25 0.213764615356922
3.75 0.213764615356922
3.75 0.165827795863152
};
\addplot [black]
table {%
4 0.165827795863152
4 0.101034246385098
};
\addplot [black]
table {%
4 0.213764615356922
4 0.285463392734528
};
\addplot [black]
table {%
3.875 0.101034246385098
4.125 0.101034246385098
};
\addplot [black]
table {%
3.875 0.285463392734528
4.125 0.285463392734528
};
\addplot [black]
table {%
4.75 0.127333022654057
5.25 0.127333022654057
5.25 0.19246032088995
4.75 0.19246032088995
4.75 0.127333022654057
};
\addplot [black]
table {%
5 0.127333022654057
5 0.0443256124854088
};
\addplot [black]
table {%
5 0.19246032088995
5 0.289989143610001
};
\addplot [black]
table {%
4.875 0.0443256124854088
5.125 0.0443256124854088
};
\addplot [black]
table {%
4.875 0.289989143610001
5.125 0.289989143610001
};
\addplot [black]
table {%
5.75 0.10745033621788
6.25 0.10745033621788
6.25 0.188649028539658
5.75 0.188649028539658
5.75 0.10745033621788
};
\addplot [black]
table {%
6 0.10745033621788
6 0.0423746518790722
};
\addplot [black]
table {%
6 0.188649028539658
6 0.308105558156967
};
\addplot [black]
table {%
5.875 0.0423746518790722
6.125 0.0423746518790722
};
\addplot [black]
table {%
5.875 0.308105558156967
6.125 0.308105558156967
};
\addplot [black]
table {%
6.75 0.0980302430689335
7.25 0.0980302430689335
7.25 0.175819225609303
6.75 0.175819225609303
6.75 0.0980302430689335
};
\addplot [black]
table {%
7 0.0980302430689335
7 0.0185092762112617
};
\addplot [black]
table {%
7 0.175819225609303
7 0.292465448379517
};
\addplot [black]
table {%
6.875 0.0185092762112617
7.125 0.0185092762112617
};
\addplot [black]
table {%
6.875 0.292465448379517
7.125 0.292465448379517
};
\addplot [black]
table {%
7.75 0.0921693108975887
8.25 0.0921693108975887
8.25 0.16691966354847
7.75 0.16691966354847
7.75 0.0921693108975887
};
\addplot [black]
table {%
8 0.0921693108975887
8 0.0278278142213821
};
\addplot [black]
table {%
8 0.16691966354847
8 0.278807669878006
};
\addplot [black]
table {%
7.875 0.0278278142213821
8.125 0.0278278142213821
};
\addplot [black]
table {%
7.875 0.278807669878006
8.125 0.278807669878006
};
\addplot [black]
table {%
8.75 0.0832627937197685
9.25 0.0832627937197685
9.25 0.155081644654274
8.75 0.155081644654274
8.75 0.0832627937197685
};
\addplot [black]
table {%
9 0.0832627937197685
9 0.0146198440343142
};
\addplot [black]
table {%
9 0.155081644654274
9 0.262605488300323
};
\addplot [black]
table {%
8.875 0.0146198440343142
9.125 0.0146198440343142
};
\addplot [black]
table {%
8.875 0.262605488300323
9.125 0.262605488300323
};
\addplot [black]
table {%
9.75 0.068084005266428
10.25 0.068084005266428
10.25 0.141534626483917
9.75 0.141534626483917
9.75 0.068084005266428
};
\addplot [black]
table {%
10 0.068084005266428
10 0.0151265282183886
};
\addplot [black]
table {%
10 0.141534626483917
10 0.251226812601089
};
\addplot [black]
table {%
9.875 0.0151265282183886
10.125 0.0151265282183886
};
\addplot [black]
table {%
9.875 0.251226812601089
10.125 0.251226812601089
};
\addplot [black]
table {%
10.75 0.0642365738749504
11.25 0.0642365738749504
11.25 0.138205736875534
10.75 0.138205736875534
10.75 0.0642365738749504
};
\addplot [black]
table {%
11 0.0642365738749504
11 0.00840953178703785
};
\addplot [black]
table {%
11 0.138205736875534
11 0.24856986105442
};
\addplot [black]
table {%
10.875 0.00840953178703785
11.125 0.00840953178703785
};
\addplot [black]
table {%
10.875 0.24856986105442
11.125 0.24856986105442
};
\addplot [black]
table {%
11.75 0.0661351755261421
12.25 0.0661351755261421
12.25 0.139503479003906
11.75 0.139503479003906
11.75 0.0661351755261421
};
\addplot [black]
table {%
12 0.0661351755261421
12 0.00533802295103669
};
\addplot [black]
table {%
12 0.139503479003906
12 0.24939951300621
};
\addplot [black]
table {%
11.875 0.00533802295103669
12.125 0.00533802295103669
};
\addplot [black]
table {%
11.875 0.24939951300621
12.125 0.24939951300621
};
\addplot [color0]
table {%
0.75 0.305870771408081
1.25 0.305870771408081
};
\addplot [color0]
table {%
1.75 0.229077324271202
2.25 0.229077324271202
};
\addplot [color0]
table {%
2.75 0.212792798876762
3.25 0.212792798876762
};
\addplot [color0]
table {%
3.75 0.187422603368759
4.25 0.187422603368759
};
\addplot [color0]
table {%
4.75 0.157094225287437
5.25 0.157094225287437
};
\addplot [color0]
table {%
5.75 0.144123941659927
6.25 0.144123941659927
};
\addplot [color0]
table {%
6.75 0.133633121848106
7.25 0.133633121848106
};
\addplot [color0]
table {%
7.75 0.126510694622993
8.25 0.126510694622993
};
\addplot [color0]
table {%
8.75 0.116721294820309
9.25 0.116721294820309
};
\addplot [color0]
table {%
9.75 0.102505199611187
10.25 0.102505199611187
};
\addplot [color0]
table {%
10.75 0.0980665758252144
11.25 0.0980665758252144
};
\addplot [color0]
table {%
11.75 0.0988656580448151
12.25 0.0988656580448151
};
\end{axis}

\end{tikzpicture}

%% file: gradient_notfinetuned_task_indep_figs/rank_all_task_independent_not_fine_tuned_mrpc.tex
% This file was created by tikzplotlib v0.8.2.
\begin{tikzpicture}

\definecolor{color0}{rgb}{0.12156862745098,0.466666666666667,0.705882352941177}

\begin{axis}[
tick align=outside,
tick pos=both,
x grid style={white!69.01960784313725!black},
xlabel={Layer},
height=\contributionPlotsHeightFirst,
width=0.97\linewidth,
xmajorgrids,
ymajorgrids,
xmin=0, xmax=11.5,
xtick style={color=black},
xtick={0,1,2,3,4,5,6,7,8,9,10,11},
xticklabels={1,2,3,4,5,6,7,8,9,10,11,12},
y grid style={white!69.01960784313725!black},
ylabel={$\tilde{P}$ [\%]},
ymin=-0.01,
ymax=0.34,
ytick style={color=black},
ytick={0,0.1,0.2,0.3},
yticklabels={0,10, 20, 30}
]
\addplot [line width=1.5pt, color0]
table {%
0 0
1 0
2 0.000460341573447498
3 0.0334207982322884
4 0.0752658472586659
5 0.136353174055149
6 0.14993325047185
7 0.180499930948764
8 0.212723841090089
9 0.298761681167426
10 0.319016710399116
11 0.316024490171707
};
\end{axis}

\end{tikzpicture}

%% file: gradient_notfinetuned_task_indep_figs/relative_attr_per_layer.tex
% This file was created by tikzplotlib v0.8.2.
\begin{tikzpicture}

\definecolor{color0}{rgb}{0.12156862745098,0.466666666666667,0.705882352941177}
\definecolor{color1}{rgb}{1,0.498039215686275,0.0549019607843137}
\definecolor{color2}{rgb}{0.172549019607843,0.627450980392157,0.172549019607843}
\definecolor{color3}{rgb}{0.83921568627451,0.152941176470588,0.156862745098039}
\definecolor{color4}{rgb}{0.580392156862745,0.403921568627451,0.741176470588235}
\definecolor{color5}{rgb}{0.549019607843137,0.337254901960784,0.294117647058824}

\begin{axis}[
legend cell align={left},
legend columns = 2,
legend style={nodes={scale=0.6, transform shape},at={(0.5,0.01)}, anchor=south, draw=white!80.0!black},
width=0.97\linewidth,
height=\localKernelPlotsHeight,
tick align=outside,
tick pos=both,
ytick={0.06,0.08,0.1},
yticklabels={6,8,10},
x grid style={white!69.01960784313725!black},
xlabel={Layer},
xmin=-0.55, xmax=11.55,
xtick style={color=black},
xtick={0,1,2,3,4,5,6,7,8,9,10,11},
xticklabels={1,2,3,4,5,6,7,8,9,10,11,12},
y grid style={white!69.01960784313725!black},
ylabel={Relative attribution per layer},
ylabel={Rel. Contribution (\%)},
ymin=0.0492504438753602, ymax=0.104405610403338,
ytick style={color=black}
]
\addplot [very thick, color0]
table {%
0 0.100391137138722
1 0.0930239341999645
2 0.101898557379339
3 0.0958146104552492
4 0.0872389935020662
5 0.0843665530839123
6 0.0818698766798712
7 0.0780590191691875
8 0.07374039982031
9 0.0693933481019662
10 0.066796888327092
11 0.0674066821423199
};
\addlegendentry{1st}
\addplot [thin, color1]
table {%
0 0.094683176188989
1 0.082621417094353
2 0.0823335925610944
3 0.0873090274247549
4 0.0830373205256345
5 0.0851649146430911
6 0.0851552057899173
7 0.0840080934726321
8 0.082223723368787
9 0.079393136665803
10 0.0770193567069837
11 0.0770510355579601
};
\addlegendentry{2nd}
\addplot [thin, color2]
table {%
0 0.0826743611776792
1 0.0778941853021008
2 0.0804940692463341
3 0.086601261573432
4 0.0837691056961211
5 0.0867327960951666
6 0.0869974558278597
7 0.0862840180066007
8 0.0851302061263758
9 0.08295629737363
10 0.0804825784179402
11 0.0799836651567599
};
\addlegendentry{3rd}
\addplot [thin, color3]
table {%
0 0.0770747668691744
1 0.0759554192583215
2 0.0773628486536306
3 0.0804987139466133
4 0.0815562168796155
5 0.085194377762732
6 0.0863205477148657
7 0.0869098303652524
8 0.0886169521808653
9 0.0883112867488208
10 0.0861213893960961
11 0.0860776502240124
};
\addlegendentry{4th and 5th}
\addplot [thin, color4]
table {%
0 0.0766612664221415
1 0.0790444366399607
2 0.0782638999221383
3 0.0779261951816118
4 0.0801782196700621
5 0.0830033578154011
6 0.0845605015738016
7 0.0852477881534244
8 0.0883839647035467
9 0.0891883943621883
10 0.0885788638144604
11 0.088963111741263
};
\addlegendentry{6th to 10th}
\addplot [very thick, color5]
table {%
0 0.0517574968993592
1 0.0705369437717982
2 0.0708892775170494
3 0.0755238618375964
4 0.0841538484028635
5 0.084829212234089
6 0.0868095727888424
7 0.0895000509482599
8 0.091858975225766
9 0.0961599263696056
10 0.099074340967948
11 0.0989064930368223
};
\addlegendentry{11th onwards}
\end{axis}

\end{tikzpicture}

%% file: sections/related.tex
\section{Related Work}

Input-output mappings play a key role in NLP. For example, in machine translation, they were introduced in the form of explicit \emph{alignments} between source and target words~\citep{brown-etal-1993-mathematics}. Neural translation architectures re-introduced this concept in the form of \emph{attention}~\citep{originalattentionpaper}.
The development of multi-head self-attention~\citep{attentionIsAllYouNeed} has led to many impressive results in NLP. As a consequence, much work has been devoted to better understand what these models learn, with a particular focus on using attention to explain model decisions. 

\cite{attentionIsNotExplanation} show that attention distributions of LSTM based encoder-decoder models are not unique, and that adversarial attention distributions that do not change the model's decision can be constructed. They further show that attention distributions only correlate weakly to moderately with dot-product based gradient attribution. \cite{attentionisNotNotexplanation} also find that adversarial attention distributions can be easily found, but that these alternative distributions perform worse on a simple diagnostic task. \cite{isattentioninterprable} find that zero-ing out attention weights based on gradient attribution changes the output of a multi-class prediction task more quickly than zero-ing out based on attention weights, thus showing that attention is not the best predictor of learned feature importance. 
\cite{pruthi2019learning} demonstrate that self-attention models can be manipulated to produce different attention masks with very little cost in accuracy.
These papers differ in their approaches, but they all provide empirical evidence showing that attention distributions are not unique with respect to downstream parts of the model (e.g., output) and hence should be interpreted with care.
Here, we support these empirical findings by presenting a theoretical proof of the identifiability of attention weights. Further, while these works focus on RNN-based language models with a single layer of attention, we instead consider multi-head multi-layer self-attention models. Our token classification and token mixing experiments show that non-identifiable tokens increase with depth, further reinforcing the point that the factors that contribute to the mixing of information are complex and deserve further study.

\cite{heavyliftingpruning} and \cite{sixteenheadsbetterthanone} find that only a small number of heads in BERT have a relevant effect on the output. These results are akin to ours about the non-identifiability of attention weights, showing that a significant part of  attention weights do not affect downstream components. One line of work investigates the internal representations of transformers by attaching probing classifiers to different parts of the model. \cite{bertRediscoversNlpPipeline} find that BERT has learned to perform steps from the classical NLP pipeline. Similarly, \cite{whatdoesbertlearnaboutlanguagestructure} show that lower layers of BERT learn syntactic features, while higher layers learn semantic features. They also argue that long-range features are learned in later layers, which agrees with our attribution-based experiments.

%% file: sections/conclusion.tex
\section{Conclusion}

We used the notion of identifiability to gain a better understanding of transformers from different yet complementary angles. We started by proving that attention weights are non-identifiable when the sequence length is longer than the attention head dimension. Thus, infinitely many attention distributions can lead to the same internal representation and model output. As an alternative, we propose \emph{effective attention}, a method that improves the interpretability of attention weights by projecting out the null space. 
Second, we show that tokens remain largely identifiable through a learned linear transformation followed by a nearest neighbor lookup based on cosine similarity. However, input tokens gradually become less identifiable in later layers.
Finally, we present \emph{Hidden Token Attribution}, a gradient-based method to quantify information mixing. This method is general and can be used to investigate contextual embeddings in self-attention based models. In this work, we use it to demonstrate that input tokens mix heavily inside transformers. This result means that attention-based interpretations, which suggest that a word at some layer is attending to another word can be improved by accounting for how the tokens are mixed inside the model. We further show that context is progressively aggregated into the hidden embeddings while some identity information is preserved. Moreover, we show that context aggregation is mostly local and that distant dependencies become relevant only in the last layers, which highlights the importance of local information for natural language understanding. Our results suggest that some of the conclusions in prior work~\citep{attentionIsAllYouNeed,vigTransformerViz,extractSyntTreesTransformer,whatBertLooksAt,transfEncAnalysis,heavyliftingpruning,analysisAttnWordSenseDisam,attendToMathTransformer,bertpassageranking,transformerheadstransparency?,lin2019opensesame,universalTransformers,wordAlignment,commonsenseReasoning,visualizingBertGeometry} may be worth re-examining from this perspective.

There are still many open questions for future research. 
For one, by constraining \emph{effective attention} to the probability simplex, one could better compare it to standard attention, although in this case non-influencing parts would be included in the weights. More research is needed to better understand the differences between these formulations.
Further, we find that tokens mix \emph{and} remain largely identifiable. 
While these two conclusions are not necessarily at odds - a token can both gather context information and still retain the essence of the original input word - we believe that the relationship between mixing and identifiability warrants further investigation.
Moreover, it is becoming increasingly difficult to compare all the new transformer variants, and it is hence important to gain a deeper understanding of this class of models. The concepts introduced in this paper could help in identifying fundamental differences and commonalities between variants of self-attention models.

\subsubsection*{Acknowledgements}
 We would like to thank the reviewers and area chairs for their thorough technical comments and valuable suggestions. We would also like to thank Jacob Devlin for feedback on an early draft. This research was supported with Cloud TPUs from Google's TensorFlow Research Cloud (TFRC).

%% file: sections/appendix.tex
\section{Identifiability of Self-Attention}

\subsection{Background on attention identifiability}\label{app:structural_identif_background}
Often, the identifiability issue arises for a model with a large number of unknown parameters and limited observations. Taking a simple linear model $y=x_1\beta_1  +  x_2 \beta_2$ as an example, when there is only one observation $(y, x_1, x_2)$, model parameters $\beta_1$ and $\beta_2$ cannot be uniquely determined. Moreover, in the matrix form $Y=X\beta$, by definition the parameter $\beta$ is identifiable only if $Y=X\beta_1$ and $Y=X\beta_2$ imply $\beta_1=\beta_2$. So if the null space contains only the zero solution $\{\beta|X\beta=0\}=\{{0}\}$, i.e., $X\beta_1-X\beta_2=X(\beta_1-\beta_2)=0 \implies \beta_1-\beta_2={0}$, the model is identifiable. Therefore, the identifiability of parameters in a linear model is linked to the dimension of the null space, which in the end is determined by the rank of $X$.

\subsection{Additional Results of the Effective Attention vs. Raw Attention Results}
\label{app:additional_effective_attention_vs_raw_attention}

\begin{figure}[!htb]
\begin{center}
\begin{subfigure}[t]{.5\textwidth}
\input{identifiability/avg_attn2.tex}
\caption{}
\end{subfigure}% 
\begin{subfigure}[t]{.5\textwidth}
\input{identifiability/avg_attn2_org.tex}
\caption{}
\end{subfigure}%
\end{center}
\caption{Effective attention (a) vs. raw attention (b). (a) Each point represents the average effective attention from a token type to a token type. Solid lines are the average effective attention of corresponding points in each layer. (b) is the corresponding figures using raw attention weights.}\label{fig:effective_attn_additional}
\end{figure}
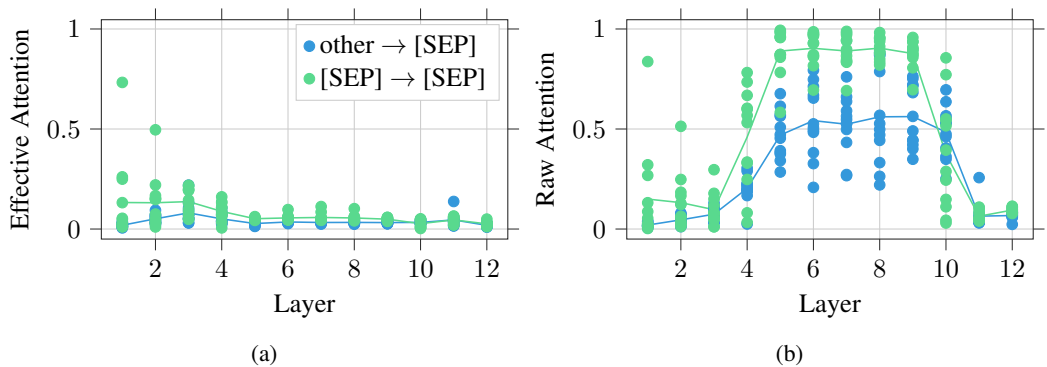

In Figure~\ref{fig:effective_attn_additional} we provide a recreation of the figure regarding the attention of tokens towards the [SEP] token found in~\citep[Figure~2]{whatBertLooksAt} with average attention as well as average effective attention. Again, we see that most of the raw attention lies effectively in the null space, confirming the pattern of Figure~\ref{fig:effective_attn}. The figures are produced using the code from~\cite{whatBertLooksAt}.

\subsection{A closer look at effective attention weights}
\label{app:additional_effective_attention_depth}
Here we discuss an example of how effective attention might lead to interpretive conclusions that differ from raw attention. Figure~\ref{fig:att-weight-3} plots the attention weights (raw, effective, null) from one of the attention heads in BERT's layer 4, for the following passage:

"[CLS] research into military brats has consistently shown them to
be better behaved than their civilian counterparts. [SEP] hypotheses as to why brats are better behaved: firstly, military parents have a lower threshold for misbehavior in their children; secondly, the mobility of teenagers might make them less likely to attract attention to themselves, as many want to fit in and are less secure with their surroundings; and thirdly, normative constraints are greater, with brats knowing that their behavior is under scrutiny and can affect the military member's career. teenage years are typically a period when people establish
independence by taking some risks away from their [SEP]".

For readability, on the y-axis, we consider just the sentence "the mobility of teenagers might make them less likely to attract attention to themselves, as many want to fit in and are less secure with their surroundings".

The following seems worth noticing:
\begin{itemize}
    \item Raw attention weights are by and large concentrated either on the structural components, [CLS] and [SEP], or on the semi-monotonic, near diagonal alignments.
    \item Effective attention weights are more uniform, in general. They are still concentrated near the diagonal elements, although less so than in raw attention. However, the attention on [CLS] and [SEP] has vanished. The collapse of the [CLS] and [SEP] weights brings to the surface other interesting things. As an example, we point out the highest weight on the attention matrix, that is not on the diagonal. This involves (highlighted by means of the yellow lines) the main verb of the selected sentence, "make", whose object is "them" (teenagers), and the pronoun "them" (the direct object of the first sentence, "military brats", 48 positions away). The two are co-referential, as both refer to the main subject of the passage, military brats.
    \item Null attention weights are also more uniform than raw attention ones. Interestingly, they seem to carry all the mass of the [CLS] and [SEP] tokens. There is a visible degree of redundancy between the null attention weights and the effective ones, but also clear complementary elements. 
\end{itemize}
One should not extrapolate too much from a single observation. Further research is needed on this topic. However, this example is a proof of concept that raw and effective attention can diverge qualitatively, in significant ways. It agrees with the hypothesis that the weights on the structural components may act as sinks, as observed in~\citep{whatBertLooksAt}, but also tells us how this happens. Furthermore, it indicates that attention in the null space can obfuscate other valuable interactions that may be recoverable by inspecting effective attention.

\clearpage

\begin{landscape}
\begin{figure*}[]
\centering
  \includegraphics[width=1.13\textwidth]{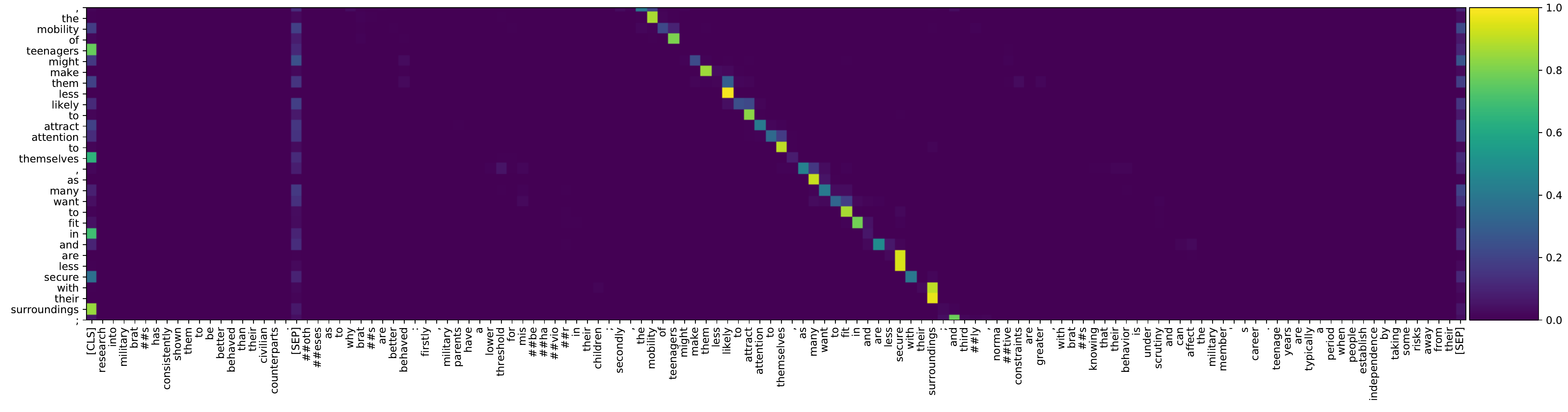}
\end{figure*}
\begin{figure*}[]
\centering
  \includegraphics[width=1.13\textwidth]{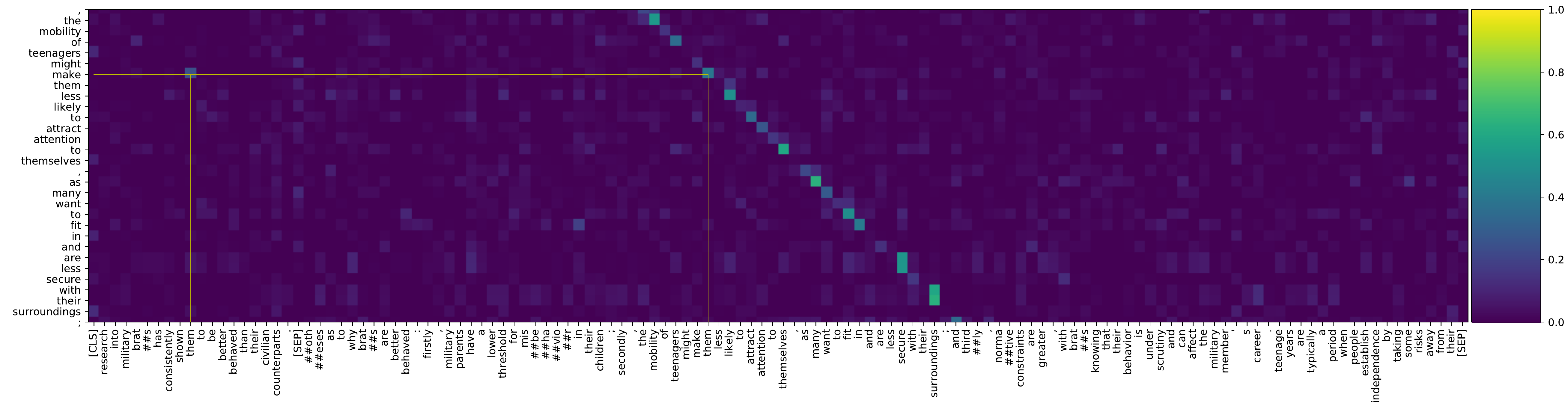}
\end{figure*}
\begin{figure*}[]
\centering
  \includegraphics[width=1.13\textwidth]{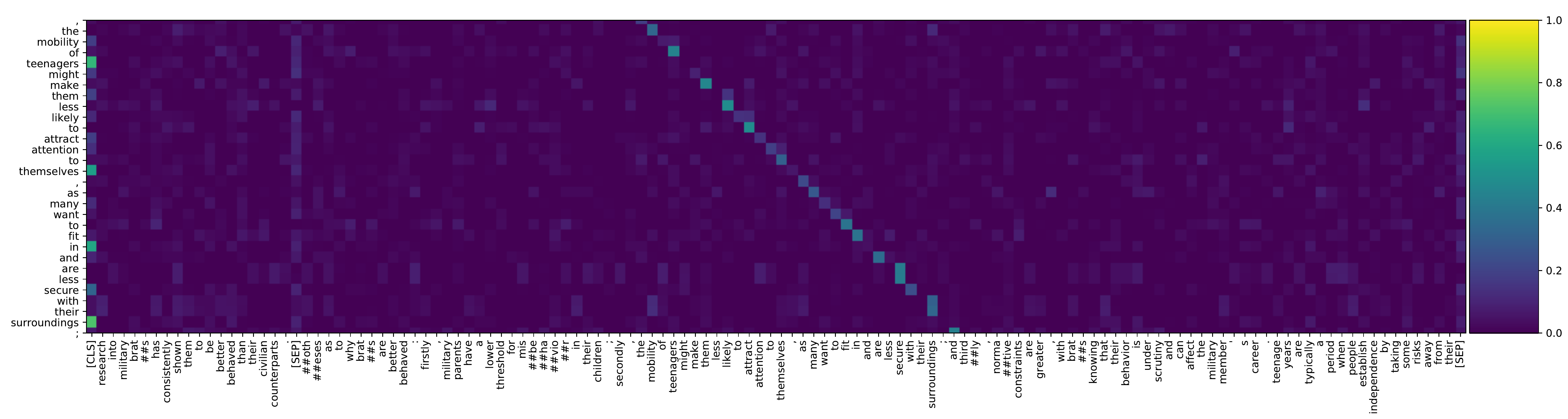}
\caption{Raw attention weights (top), Effective attention weights (middle) and Null attention weights (bottom).}
\label{fig:att-weight-3}
\end{figure*}

\end{landscape}

\clearpage

\section{Token Identifiability Experiments}\label{App_TokenClassifier}
\subsection{Experimental Setup and Training Details}\label{App_tokenClassifier_SetupDetails}

The linear perceptron and MLP are both trained by either minimizing the L2 or cosine distance loss using the ADAM optimizer~\citep{adam} with a learning rate of $\alpha=0.0001$, $\beta_1=0.9$ and $\beta_2=0.999$. We use a batch size of 256. We monitor performance on the validation set and stop training if there is no improvement for 20 epochs.  The input and output dimension of the models is $d=768$; the dimension of the contextual word embeddings.
For both models we performed a learning rate search over the values $\alpha \in[0.003,0.001, 0.0003,0.0001,0.00003,0.00001,0.000003]$. The weights are initialized with the Glorot Uniform initializer~\citep{glorotuniform}. 
The MLP has one hidden layer with 1000 neurons and uses the gelu activation function~\citep{geluactivation}, following the feed-forward layers in BERT and GPT. We chose a hidden layer size of 1000 in order to avoid a bottleneck. We experimented with using a larger hidden layer of size 3072 and adding dropout to more closely match the feed-forward layers in BERT. This only resulted in increased training times and we hence deferred from further architecture search. 

We split the data by sentences into train/validation/test according to a 70/15/15 split. This way of splitting the data ensures that the models have never seen the test sentences (i.e., contexts) during training.
In order to get a more robust estimate of performance we perform the experiments in Figure~\ref{fig:tokenclassifierresults_a} using 10-fold cross validation. The variance, due to the random assignment of sentences to train/validation/test sets, is small, and hence not shown.

\subsection{Generalization Error}\label{App_tokenClassifier_GenError}

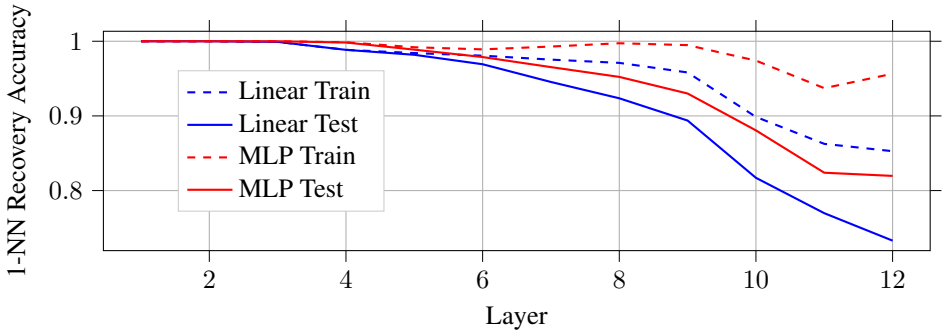
\begin{figure}[h]
\begin{center}
\input{token_classification_experiments/PredictTokens_AllLayers_NearestNeighbour_TestVsTrain.tex}
\end{center}
\caption{Train and test token identifiability rates for the linear perceptron and MLP.}
\label{fig:token_prediction_nn_train_vs_test}
\end{figure}

Figure~\ref{fig:token_prediction_nn_train_vs_test} shows the token identifiability rate for train and test set for both models, linear and MLP, when using L2 distance. Both models are overfitting to the same degree. The fact that the linear model has about the same generalization error as the MLP suggests that more training data would not significantly increase performance on the test set.
Further, we trained the MLP on layer 11 using 50\%, 80\%, 90\% and 100\% of the training data set. The MLP achieved the following token identifiability rate on the test set: 0.74, 0.8, 0.81, 0.82. This indicates that the MLP would not profit much from more data.

We do not report the generalization error for the models trained to minimize cosine distance, as the linear and non-linear perceptrons perform almost equally. 

\clearpage

\subsection{Additional results for Figure~\ref{fig:tokenclassifierresults_b}}\label{App_tokenClassifier_GeneralizingLayersFullResults}

\begin{figure}[h]
\begin{center}
\input{token_classification_experiments/rebuttal_plots/GeneralizingAcrossLayers/TokenIdentity_GeneralizeAcrossLayers_LINEAR_L2_Layers1-6-11-12_10FoldCV.tex}
\end{center}
\caption{Linear Perceptron trained to minimize L2 distance generalizing to all layers.}
\label{fig:linear_l2_generalizing_to_all_layers}
\end{figure}
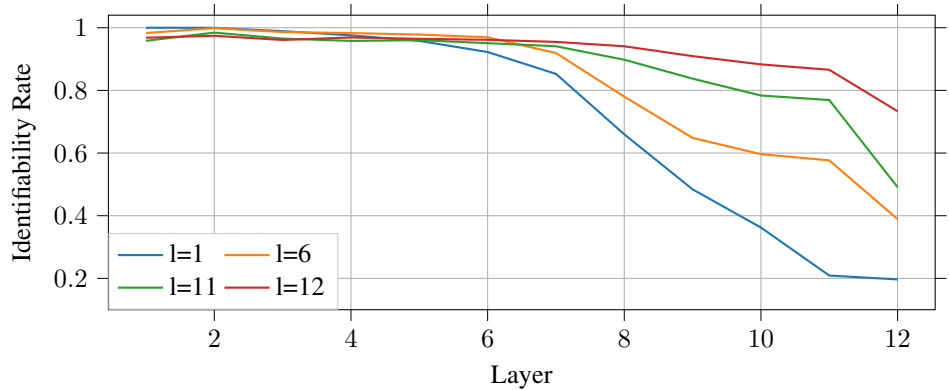

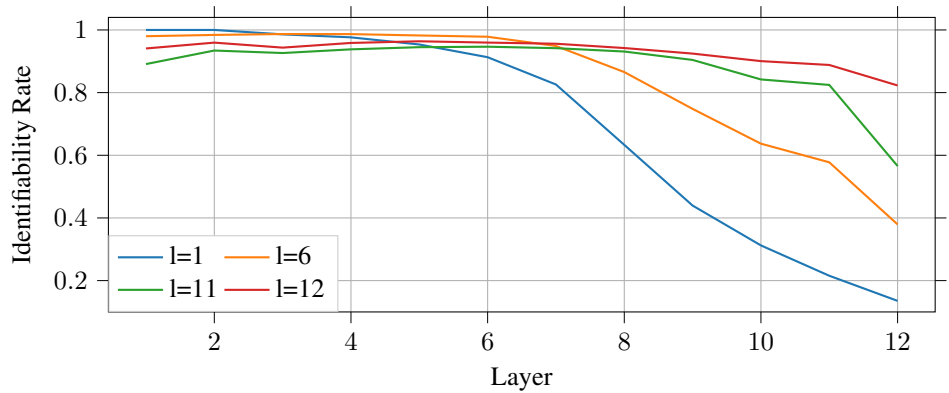
\begin{figure}[h]
\begin{center}
\input{token_classification_experiments/rebuttal_plots/GeneralizingAcrossLayers/TokenIdentity_GeneralizeAcrossLayers_MLP_L2_Layers1-6-11-12_10FoldCV.tex}
\end{center}
\caption{MLP trained to minimize L2 distance generalizing to all layers.}
\label{fig:mlp_l2_generalizing_to_all_layers}
\end{figure}

\begin{figure}[h]
\begin{center}
\input{token_classification_experiments/rebuttal_plots/GeneralizingAcrossLayers/TokenIdentity_GeneralizeAcrossLayers_MLP_COSINE_Layers1-6-11-12_10FoldCV.tex}
\end{center}
\caption{MLP trained to minimize cosine distance generalizing to all layers.}
\label{fig:mlp_cosine_generalizing_to_all_layers}
\end{figure}
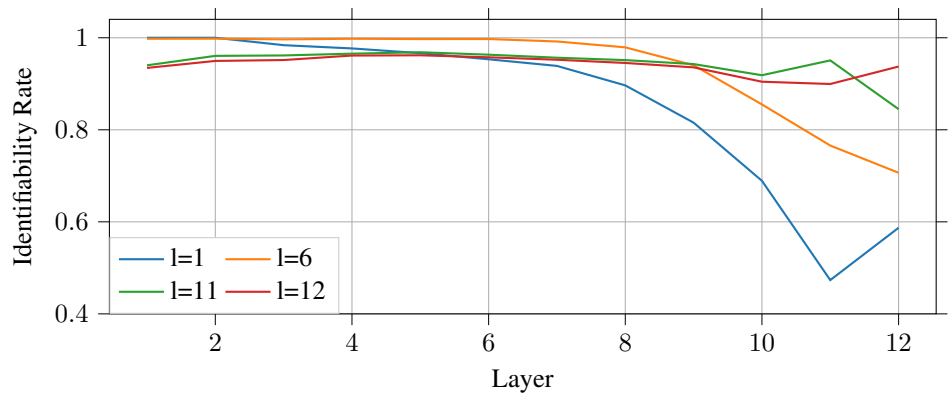

Figure~\ref{fig:tokenclassifierresults_b} in the main text only shows results of the linear perceptron trained to minimize cosine distance on layers $l=[1,6,11,12]$ and tested on all other layers.
Figures~\ref{fig:linear_l2_generalizing_to_all_layers}, \ref{fig:mlp_l2_generalizing_to_all_layers} and \ref{fig:mlp_cosine_generalizing_to_all_layers} show the corresponding results for the linear perceptron trained to minimize cosine distance, and for the MLP trained to minimize L2 and cosine distance respectively. 
Overall, all figures show the same qualitative trends as presented in Section~\ref{IdenTok} of the main text: Generalizing to later layers works considerably worse than the other way around. The linear perceptrons seem to generalize better across layers, likely due to the MLPs overfitting more to the particular layers they are trained on.

\clearpage

\subsection{Token Identity - From Hidden Tokens to Hidden Tokens}\label{appendix:tokenident:hiddentohidden}

Figure~\ref{fig:appendix:hiddentohidden_nonFinetuned_mrpc} shows results for identifying tokens across single layers of BERT, i.e., the input to $g$ is $(\ve_i^1,\vx_i)$ in the first layer, and subsequently $(\ve_i^l,\ve_i^{l-1})$, where $l=[1,...,12]$. This experiment gives further insight into what kinds of transformations are applied by each transformer layer separately, as opposed to the cumulative transformations shown in Section~\ref{IdenTok} of the main text. 
Interestingly, even the naive baselines perform well across single layers. This shows that BERT only applies small changes to the contextual word embeddings, whereas overall the angle (as indicated by the naive baseline using cosine distance) is affected less than the magnitude of the word embeddings (indicated by the naive baseline using L2 distance).

Figure~\ref{fig:appendix:hiddentohidden_nonFinetuned_mrpc} shows that tokens are on average more difficult to identify across later layers. In the main text we hypothesize that the qualitative change seen in later layers could be due to a task-specific parameter adaptation during the second (next sentence prediction) pre-training phase. A possible reason is that during this pre-training-phase, BERT only needs the [CLS] token in the last layer, which is qualitatively very different form the first (masked language modeling) pre-training phase, where potentially all the tokens are needed in the last layer. 

To further verify this hypothesis we experimented with BERT fine-tuned on two datasets, MRPC and CoLA~\citep{warstadt2018neural}. 
During the fine-tuning phase, similar to the next sentence prediction pre-training phase, only the [CLS] token is needed at the last layer. If task-dependent parameter adaptation indeed has a different influence on the last layer(s) than on earlier layers, then we should be able to see a difference between the finetuned and non-finetuned cases. Figures~\ref{fig:appendix:hiddentohidden_baselines_finetuned_mrpc} and \ref{fig:appendix:hiddentohidden_baselines_finetuned_cola} compare the naive baselines across single layers for BERT finetuned on MRPC and
CoLA, respectively. Indeed, one can see a remarkable decrease in identifiability across the last layer for L2-based nearest neighbour lookup, further indicating that the last layers are indeed more strongly affected by different fine-tuning objectives. Nearest neighbour lookup based on cosine distance is affected much less, indicating that in terms of token identifiability, the last layers are only slightly affected by fine-tuning.

\begin{figure}[h]
\begin{center}
\input{token_classification_experiments/rebuttal_plots/HiddenToHidden/TokenIdentity_HiddenToHidden_withBaselines_10FoldCV.tex}
\end{center}
\caption{Token identifiability across single layers. These results are for non fine-tuned BERT on MRPC.}
\label{fig:appendix:hiddentohidden_nonFinetuned_mrpc}
\end{figure}
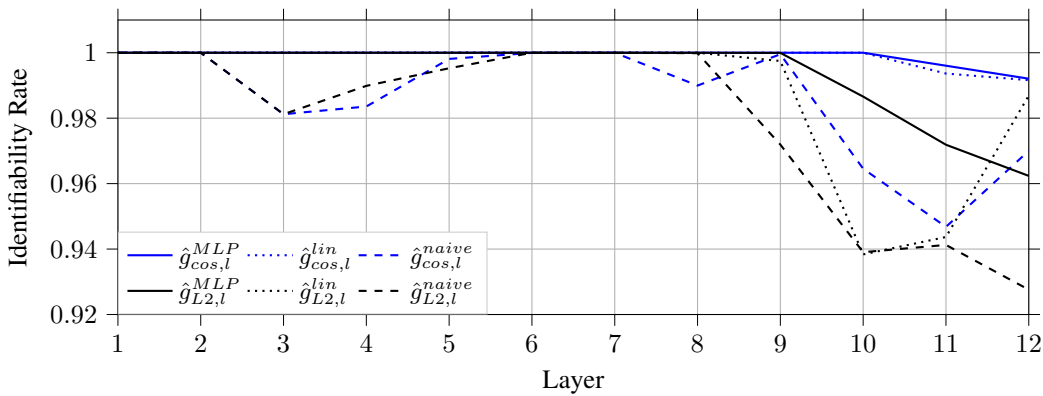

\begin{figure}[h]
\begin{center}
\input{token_classification_experiments/rebuttal_plots/HiddenToHidden/TokenIdentity_HiddenToHidden_mrpcFinetuned_baselines.tex}
\end{center}
\caption{Token identifiability across single layers, comparing non fine-tuned (dashed) BERT against BERT fine-tuned on MRPC (solid).}
\label{fig:appendix:hiddentohidden_baselines_finetuned_mrpc}
\end{figure}
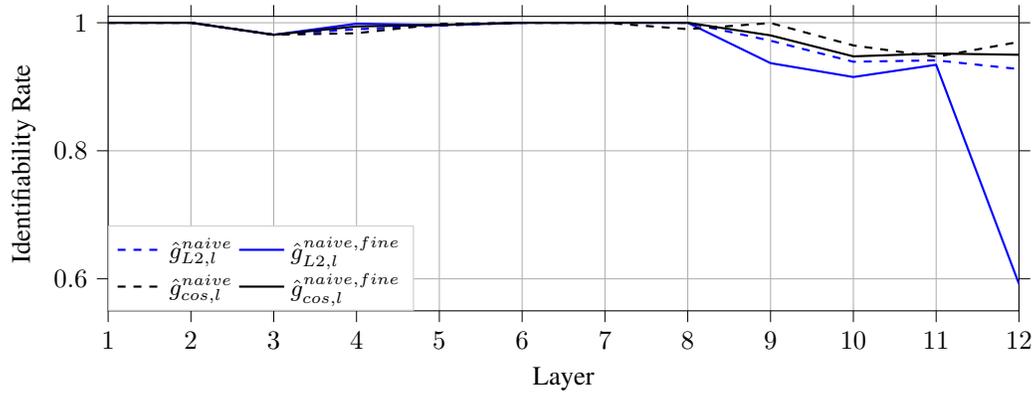

\begin{figure}[h]
\begin{center}
\input{token_classification_experiments/rebuttal_plots/HiddenToHidden/TokenIdentity_HiddenToHidden_colaFinetuned_baselines.tex}
\end{center}
\caption{Token identifiability across single layers, comparing non fine-tuned BERT (dashed) against BERT fine-tuned on CoLA (solid).}
\label{fig:appendix:hiddentohidden_baselines_finetuned_cola}
\end{figure}
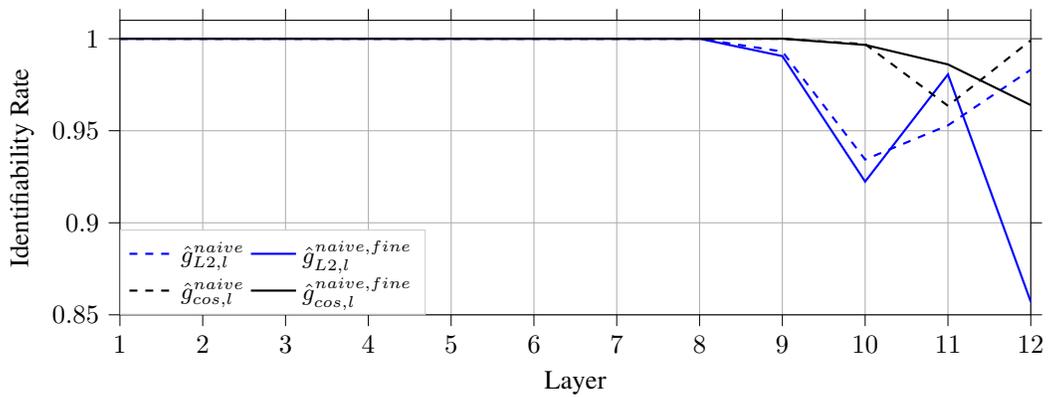

\clearpage

\subsection{Token Identity - Recover Neighbouring Input Tokens}\label{appendix:neighbouringTokenIdentity}

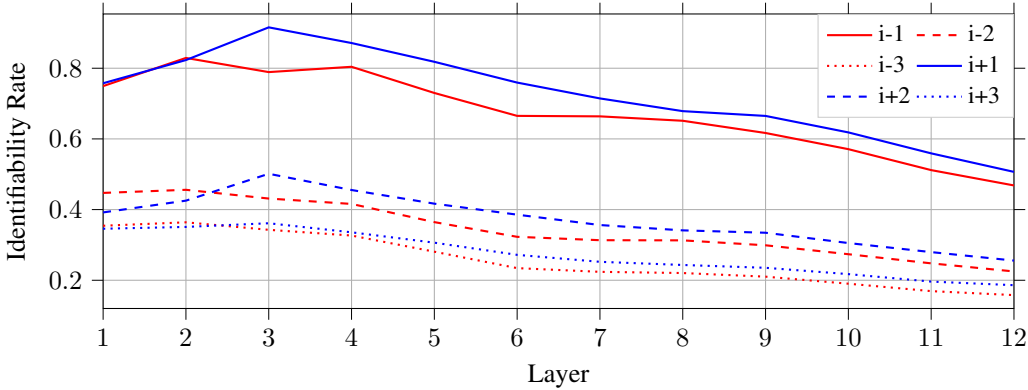
\begin{figure}[h]
\begin{center}
\input{token_classification_experiments/rebuttal_plots/NeighbouringTokens/TokenIdentity_NeighbouringTokens_linear_cosine_mrpc.tex}
\end{center}
\caption{Recovering neighbouring input tokens using $\hat{g}_{cos,l}^{lin}$.}
\label{fig:appendix:neighbouringTokenIdentity_linear_cos}
\end{figure}

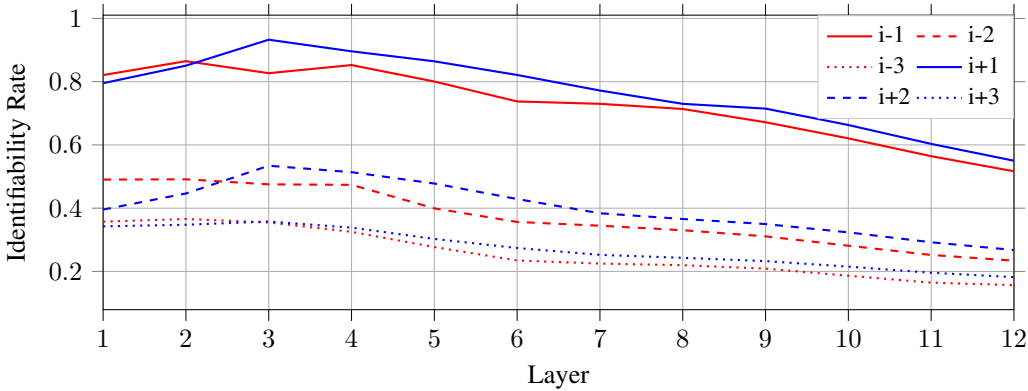
\begin{figure}[h]
\begin{center}
\input{token_classification_experiments/rebuttal_plots/NeighbouringTokens/TokenIdentity_NeighbouringTokens_mlp_cosine_mrpc.tex}
\end{center}
\caption{Recovering neighbouring input tokens using $\hat{g}_{cos,l}^{mlp}$.}
\label{fig:appendix:neighbouringTokenIdentity_mlp_cos}
\end{figure}

In Section~\ref{IdenTok} of the main text we show that tokens at position $i$ remain largely identifiable throughout the layers of BERT. In this section we show results of a related experiment, where we test how much information about tokens at neighbouring positions is contained in a contextual word embedding. More formally, the input to $g$ is $(\ve_i^l,\vx_{i\pm k})$, where $k\in\{1,2,3\}$. 
Thus, we try to recover input token $\vx_{i+k}$ from hidden token $\ve_i^l$. 
Figures~\ref{fig:appendix:neighbouringTokenIdentity_linear_cos}, \ref{fig:appendix:neighbouringTokenIdentity_mlp_cos}, \ref{fig:appendix:neighbouringTokenIdentity_mlp_l2} and \ref{fig:appendix:neighbouringTokenIdentity_linear_l2} show the results of for $\hat{g}_{cos,l}^{lin}$, $\hat{g}_{cos,l}^{MLP}$, $\hat{g}_{L2,l}^{MLP}$ and $\hat{g}_{L2,l}^{lin}$, respectively. In the figures, blue corresponds to \quotes{previous} tokens and red to \quotes{next} tokens.

From the figures we can see that tokens do contain information about neighbouring tokens that lets us recover the neighbouring tokens based on a transformation and subsequent nearest neighbour lookup. The identifiability rate drops both with increasing $k$, but also with increasing depth. 
Interestingly, recovering left (blue) and right (red) neighbours shows different behaviour, indicating that BERT is treating left and right context differently, despite having been pre-trained using a bi-directional language modeling objective. 

Overall, neighbouring tokens can be recovered to a much lower degree than same-position tokens (cf.~Section\ref{IdenTok}).

\begin{figure}[h]
\begin{center}
\input{token_classification_experiments/rebuttal_plots/NeighbouringTokens/TokenIdentity_NeighbouringTokens_mlp_l2_mrpc.tex}
\end{center}
\caption{Recovering neighbouring input tokens using $\hat{g}_{L2,l}^{mlp}$.}
\label{fig:appendix:neighbouringTokenIdentity_mlp_l2}
\end{figure}
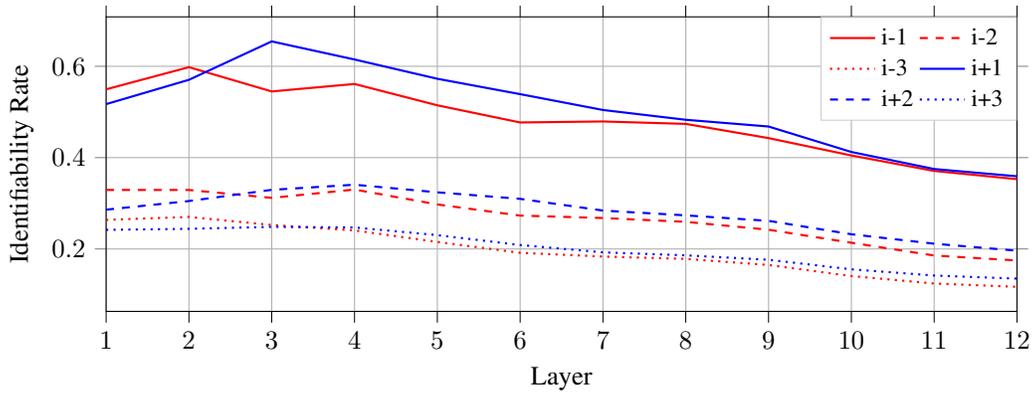

\begin{figure}[h]
\begin{center}
\input{token_classification_experiments/rebuttal_plots/NeighbouringTokens/TokenIdentity_NeighbouringTokens_linear_l2_mrpc.tex}
\end{center}
\caption{Recovering neighbouring input tokens using $\hat{g}_{l2,l}^{lin}$.}
\label{fig:appendix:neighbouringTokenIdentity_linear_l2}
\end{figure}
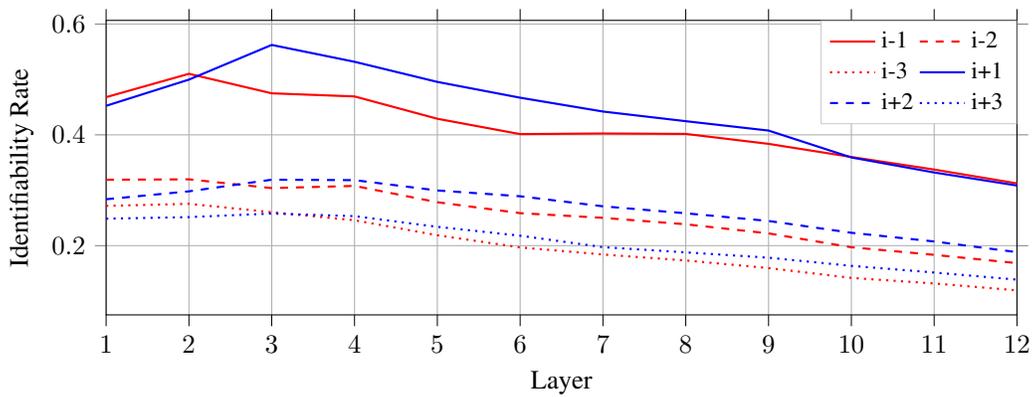

\clearpage
\section{Context Contribution Analysis} 

\subsection{Hidden Token Attribution: Details}\label{AppGrad}

The attribution method proposed in Section~\ref{subsec:gradientAttr} to calculate the contribution of input tokens to a given embedding does not look at the output of the model but at the intermediate hidden representations and therefore is task independent. Since the contribution values do not depend on the task that is evaluated, we can compare these values directly to attention distributions, which are also task-independent. In this way, we can compare to other works in the literature~\citep{vigTransformerViz,whatBertLooksAt,commonsenseReasoning,visualizingBertGeometry,lin2019opensesame} by using the publicly available pretrained BERT model in our analyses without fine-tuning it to a specific task. 

Furthermore, since we are not interested in analysing how the input affects the output of the model but in quantifying the absolute contribution of the input tokens to the hidden embeddings, we use the $L_2$ norm of the gradients. If we were analyzing whether the input contributed positively or negatively to a given decision, the dot-product of the input token embedding with the gradient would be the natural attribution choice~\citep{attributionMethodEval}.

\subsection{Context Identifiability: Details}\label{AppCtxtDet} %Analyze the "contextual" in contextual word embedding

To calculate the relative contribution values shown in Figure~\ref{rel_attr} we firstly calculate the mean of the left and right neighbours for each of the groups of neighbours, i.e., first, second, third, fourth and fifth, sixth to 10th and, from 11th onwards. Then we aggregate the values averaging over all the tokens in the MRPC evaluation set. Finally, we normalize for each group so that the sum of the contribution values of each group is one. In this way, we can observe in which layer the contribution of a given group of neighbours is the largest.  

Our results on context identifiability from Section~\ref{sec:Contxt} complement some of the studies in previous literature. In~\citep{whatdoesbertlearnaboutlanguagestructure} the authors observe that transformers learn local syntactic tasks in the first layers and long range semantic tasks in the last layers. We explain this behavior from the point of view of context aggregation by showing that distant context acquires more importance in the last layers (semantic tasks) while the first layers aggregate local context (syntactic tasks). Furthermore, the results showing that the context aggregation is mainly local, specially in the first layers, provide an explanation for the increase in performance observed in \citep{localnesforselfattention}. In that work, the authors enforce a locality constraint in the first layers of transformers, which pushes the model towards the local operators that it naturally tends to learn, as we show in Figure~\ref{tot_attr}, improving in this way the overall performance.

\subsection{Context Contribution to CLS token}\label{AppCLS}

In this section we use Hidden Token Attribution to look at the contribution of the context to the [CLS] token, which is added to the beginning of the input sequence by the BERT pre-processing pipeline. This is an especially interesting token to look at because the decision of BERT for a classification task is based on the output in the [CLS] token. Furthermore, like the [SEP] token, it does not correspond to a natural language word and its position in the input sequence does not have any meaning. Therefore, the conclusion that context is on average predominantly local (cf. Section~\ref{sec:Contxt}), is likely to not hold for [CLS]. 

\begin{figure}[h]
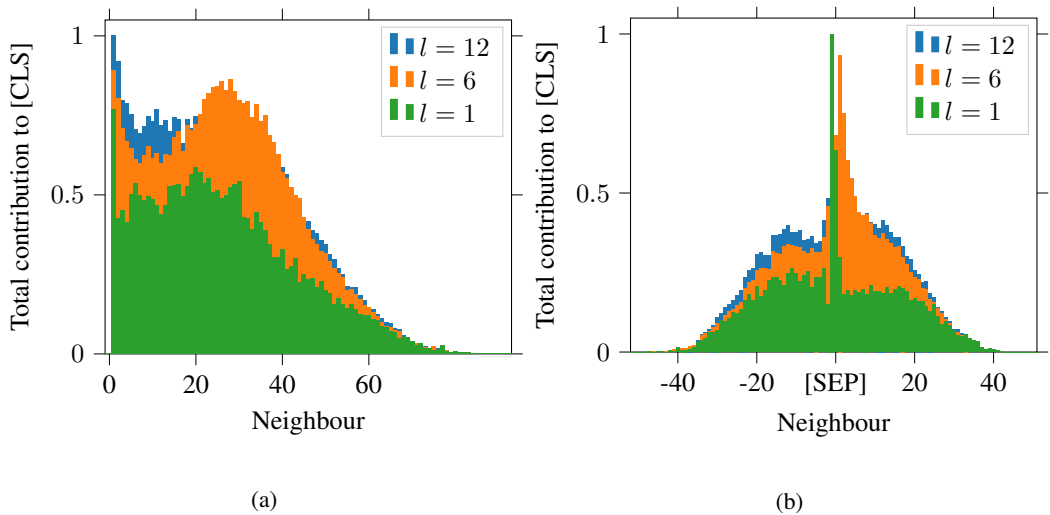

\begin{center}
\begin{subfigure}{.5\textwidth}
\input{CLS/CLS.tex}
\caption{}\label{CLS_cent}
\end{subfigure}%
\begin{subfigure}{.5\textwidth}
\input{CLS/CLS_SEP_centered.tex}
\caption{}\label{CLS_SEP}
\end{subfigure}%
\end{center}
\caption{Normalized total contribution to the [CLS] token (a) centered around [CLS] at position 0 (b) centered around [SEP].}\label{CLS}
\end{figure}

The second and final pre-training task that BERT is trained on is next sentence prediction. During this task, BERT receives two sentences separated by a [SEP] token as input, and then has to predict whether the second sentence follows the first sentence or not. 
Therefore, it is expected that the context aggregated into the [CLS] token comes mainly from the tokens around the first [SEP] token, which marks the border between the first and second sentence in the input sequence. In Figure~\ref{CLS} we show the contribution to the [CLS] token from all of its neighbours averaged over all the examples in the MRPC evaluation set for the first, middle and last layers. In Figure~\ref{CLS_cent}, the [CLS] token is placed at position 0 and we see how the context contribution comes mainly from the tokens around position 30, which is roughly the middle of the input examples. In Figure \ref{CLS_SEP} we center the contribution around the first [SEP] token and indeed, it becomes clear that the [CLS] token is aggregating most of its context from the tokens around [SEP], i.e., from the junction between both sentences. In particular, the two tokens with the highest contribution are the tokens directly before and after [SEP]. Also, it seems that the second sentence contributes more to [CLS] than the first one.

These results give an insight on what information BERT uses to solve next sentence prediction and serves as an illustrative example of how Hidden Token Attribution can be used to analyze transformers.

\subsection{Tracking Context Contribution}\label{AppTrack}

Here we show examples of how Hidden Token Attribution can track how context is aggregated for a given word at each layer. For reasons of space we show only few words of a randomly picked sentence of the MRPC evaluation set, which is tokenized as follows: 

\texttt{[CLS] he said the foods \#\#er \#\#vic \#\#e pie business doesn ' t fit the company ' s long - term growth strategy . [SEP] " the foods \#\#er \#\#vic \#\#e pie business does not fit our long - term growth strategy . [SEP]}

\begin{figure}[h]
\begin{center}
\includegraphics[width=1\linewidth]{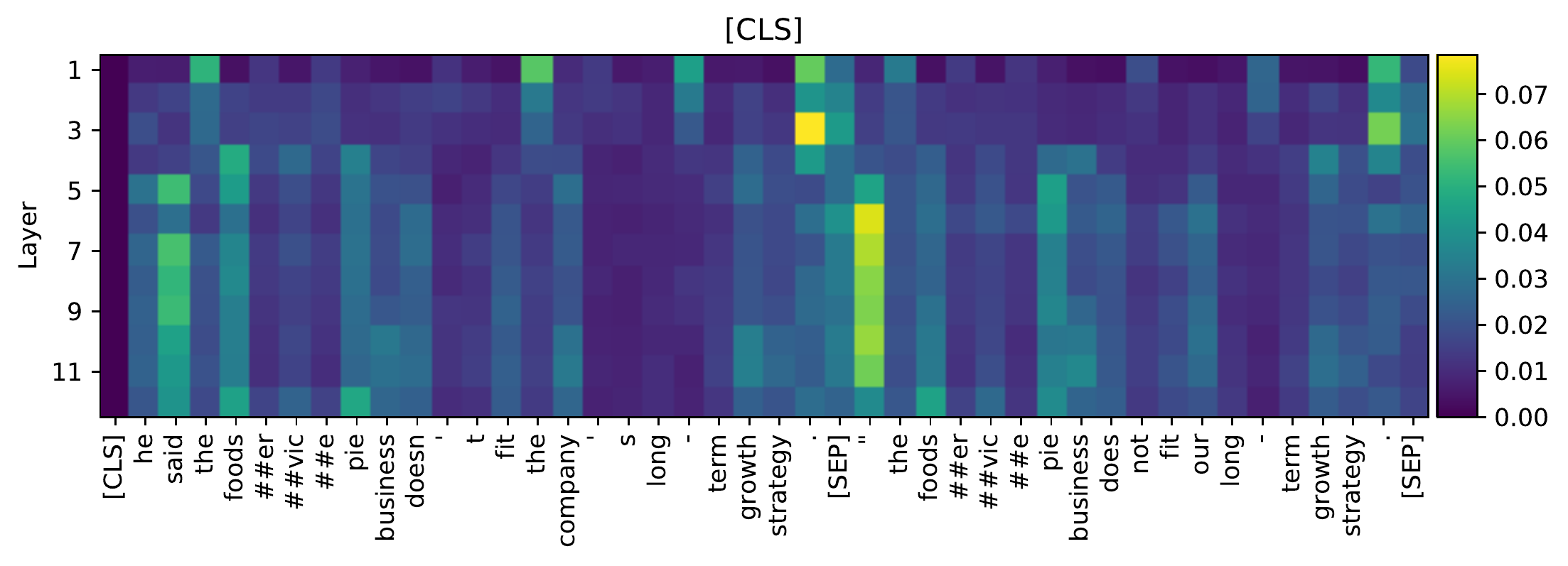}
\caption{ [CLS]: Aggregates context from all tokens but more strongly from those around the first [SEP] token. We hypothesize that this is due to the Next Sentence Prediction pre-training.}
\end{center}
\end{figure}

\begin{figure}[h]
\begin{center}
\includegraphics[width=1\linewidth]{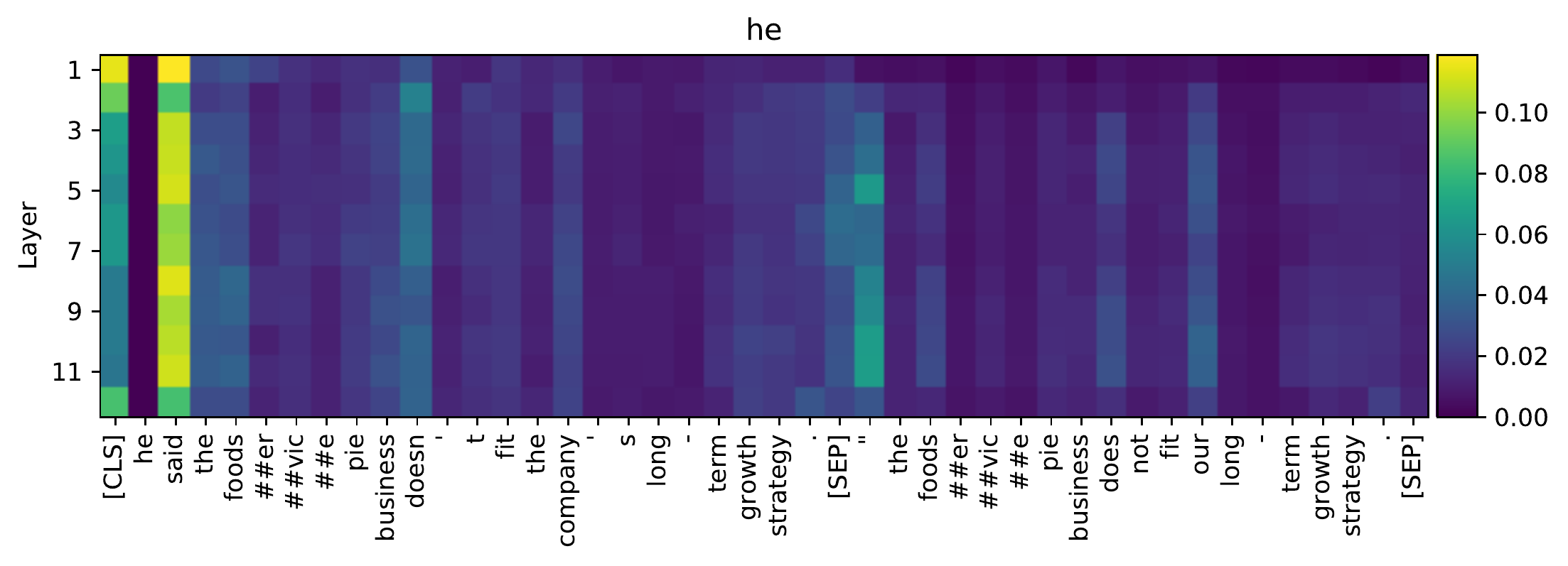}
\caption{he: Aggregates most context from the main verb of the sentence, "said".}
\end{center}
\end{figure}

\begin{figure}[h]
\begin{center}
\includegraphics[width=1\linewidth]{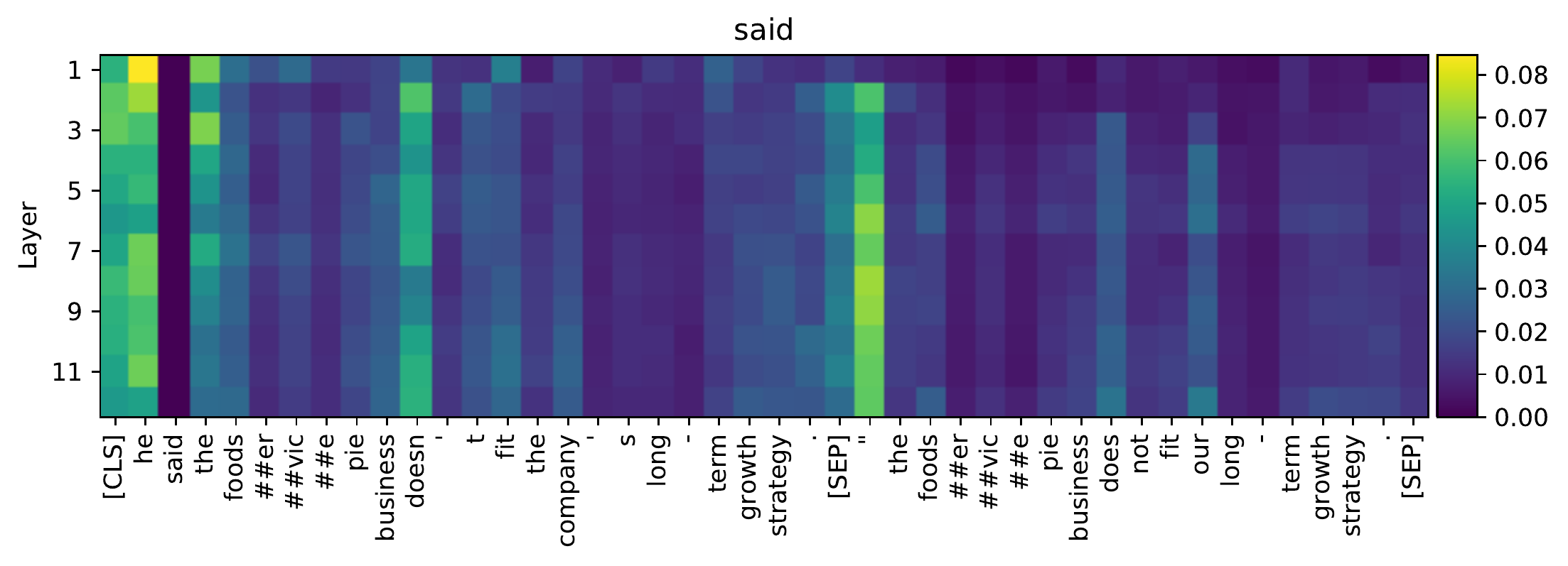}
\caption{said: Aggregates context mainly from its neighborhood, the main verb of the subordinate sentence and the border between the two input sentences.}
\end{center}
\end{figure}

\begin{figure}[h]
\begin{center}
\includegraphics[width=1\linewidth]{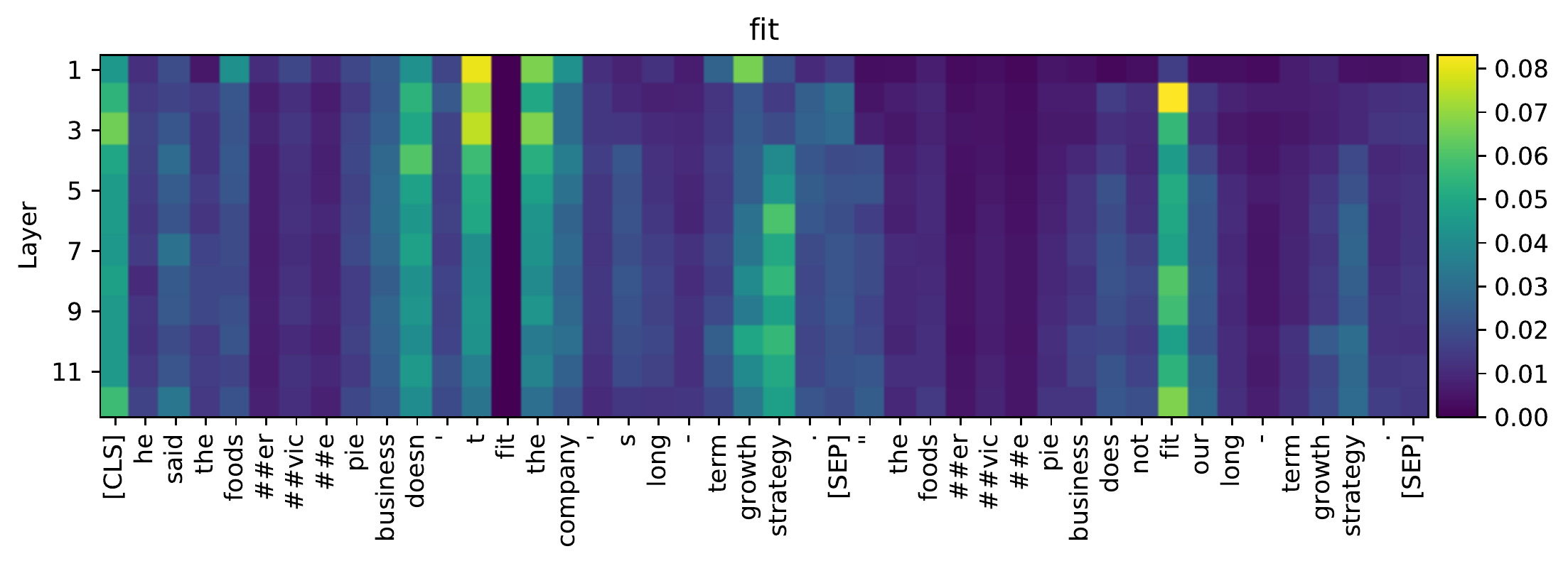}
\caption{fit: In the first layers it aggregates most context from its neighborhood and towards the last layers it gets the context from its direct object (strategy) and from the token with the same meaning in the second sentence.}
\end{center}
\end{figure}

\begin{figure}[h]
\begin{center}
\includegraphics[width=1\linewidth]{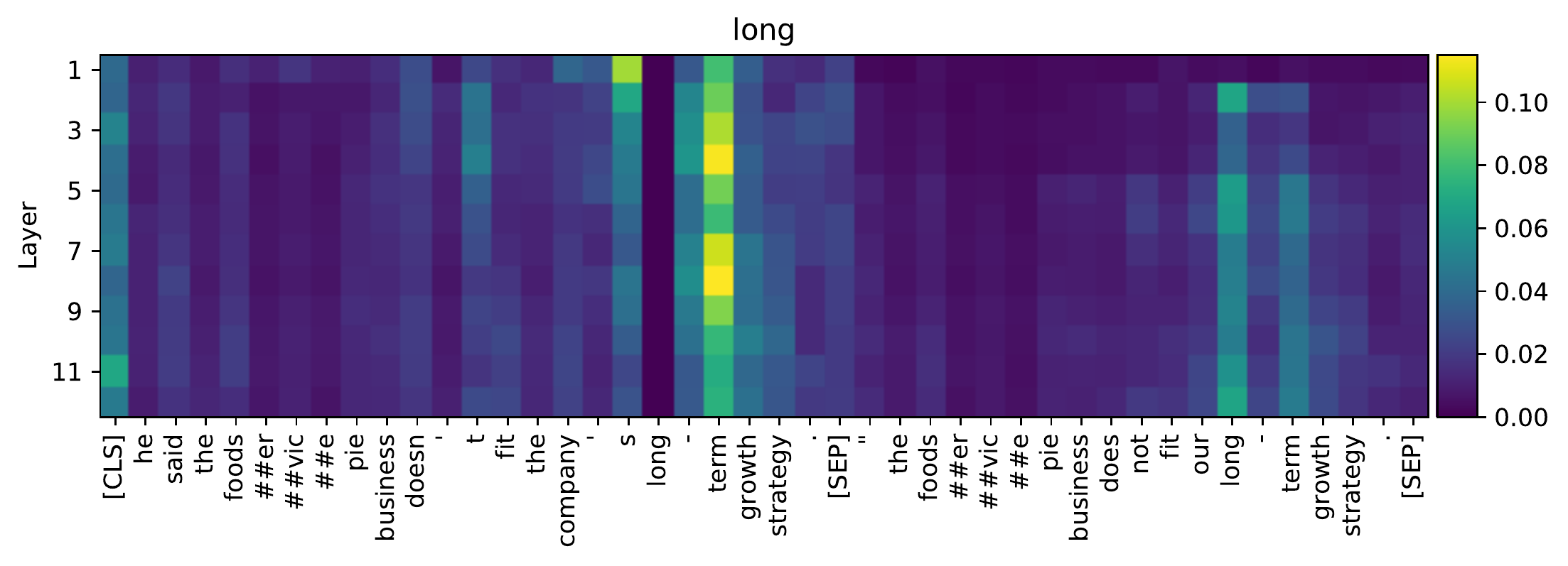}
\caption{long: It is part of a composed adjective (long-term) and aggregates most of its context from the other part of the adjective (term) as well as from the same tokens in the second sentence. Interestingly, it mostly ignores the hyphen.}
\end{center}
\end{figure}

\begin{figure}[h]
\begin{center}
\includegraphics[width=1\linewidth]{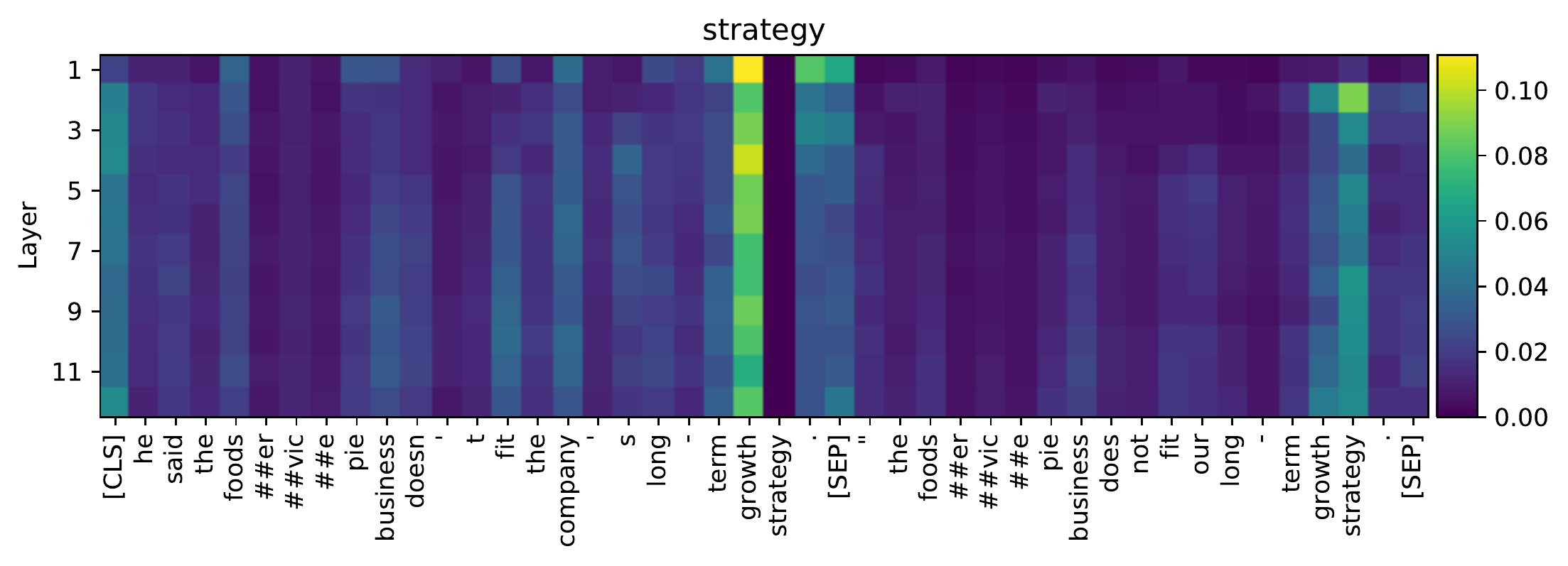}
\caption{strategy: Aggregates context from the word growth, which is the first one of the noun phrase "growth strategy".}
\end{center}
\end{figure}

\begin{figure}[h]
\begin{center}
\includegraphics[width=1\linewidth]{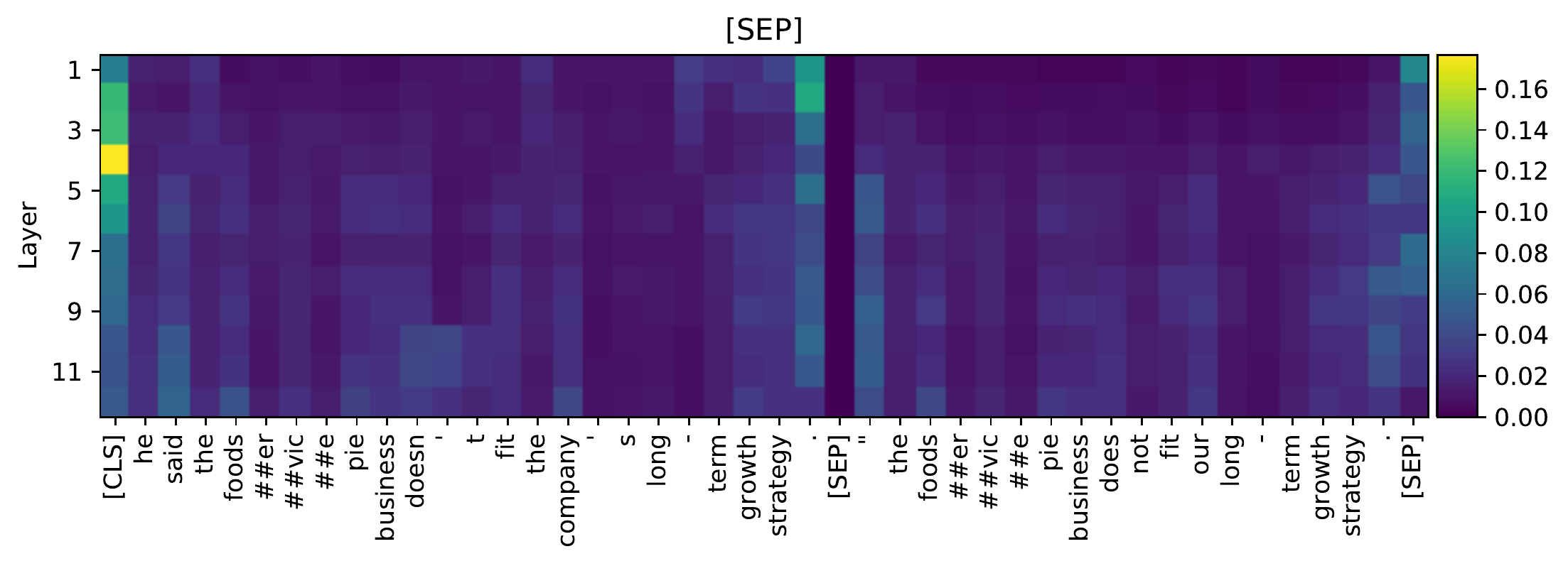}
\caption{[SEP]: This token that has no semantic meaning aggregates context mostly from [CLS] and its own neighborhood.}
\end{center}
\end{figure}

\clearpage
\subsection{Token Contributions by POS Tag}\label{AppPOS}
Here we show the contribution of input tokens to hidden representations in all layers split by part-of-speech (POS) tag~\citep{POStagger}. The POS tags are ordered according to the contribution in layer 12.

\begin{figure}[h]
\begin{center}
\includegraphics[width=1\linewidth]{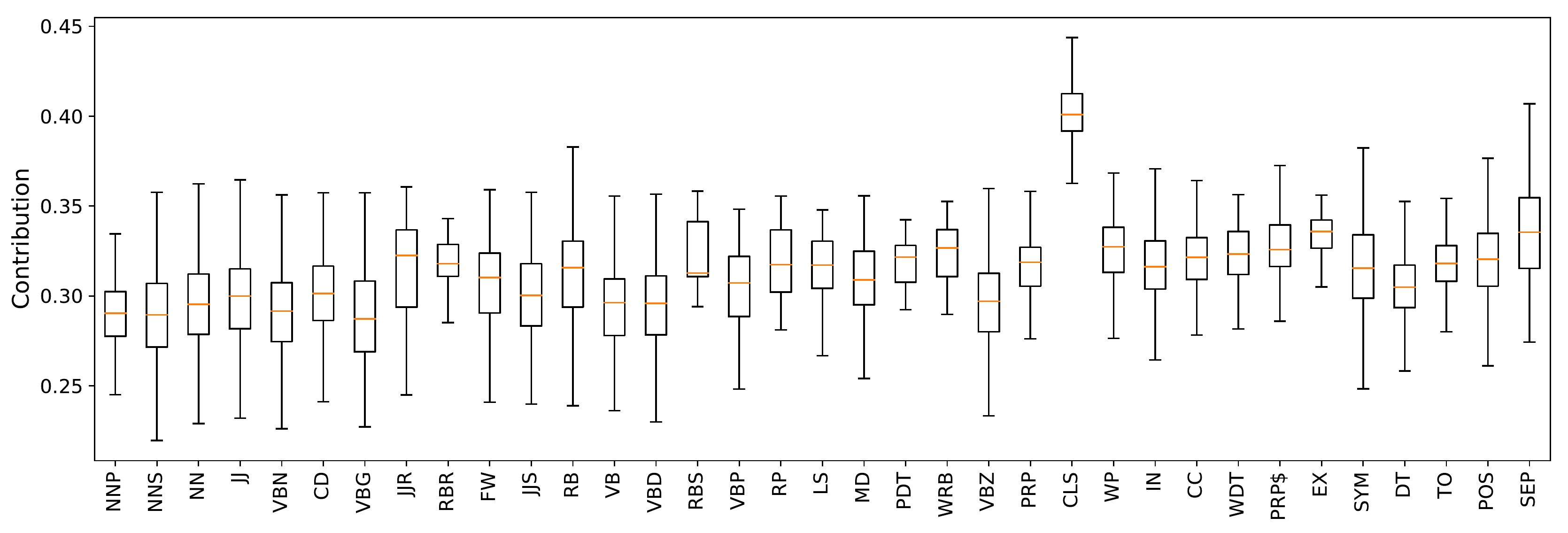}
\caption{Layer 1: Most token types are equally mixed and have already less than 35\% median contribution from their corresponding input. The only exception are the [CLS] tokens, which remain with over 40\% median original contribution.}
\end{center}
\end{figure}

\begin{figure}[h]
\begin{center}
\includegraphics[width=1\linewidth]{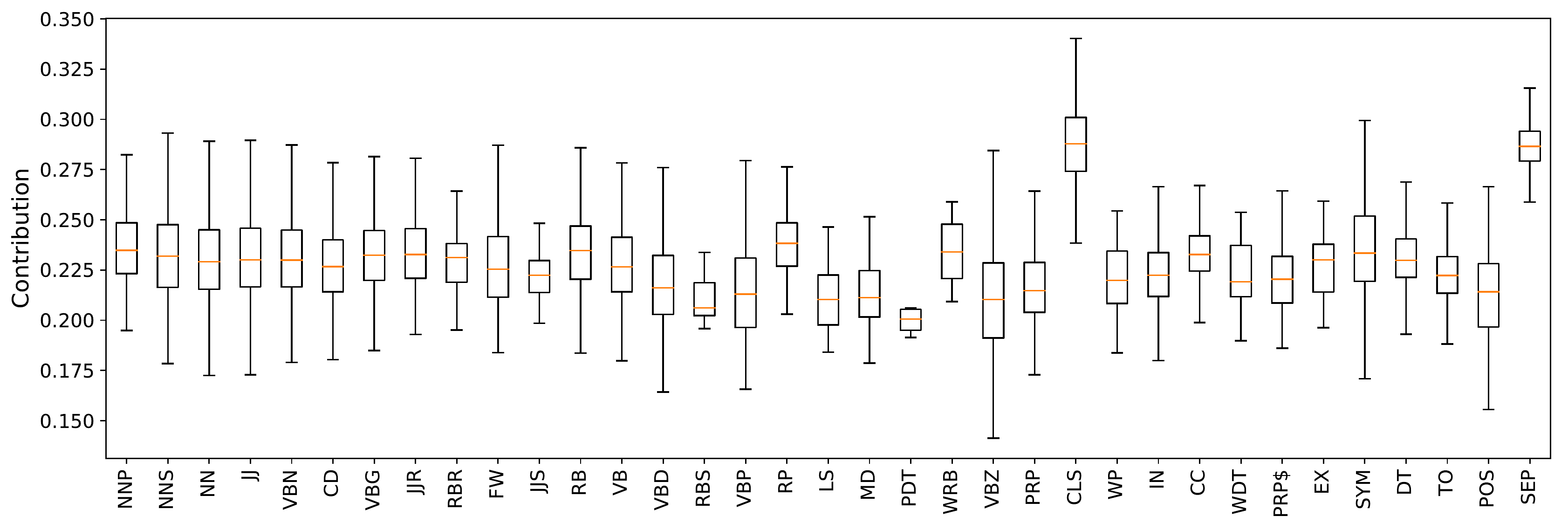}
\caption{Layer 2: Similar to the previous layer with less contribution over all and [SEP] behaving similarly to [CLS].}
\end{center}
\end{figure}

\begin{figure}[h]
\begin{center}
\includegraphics[width=1\linewidth]{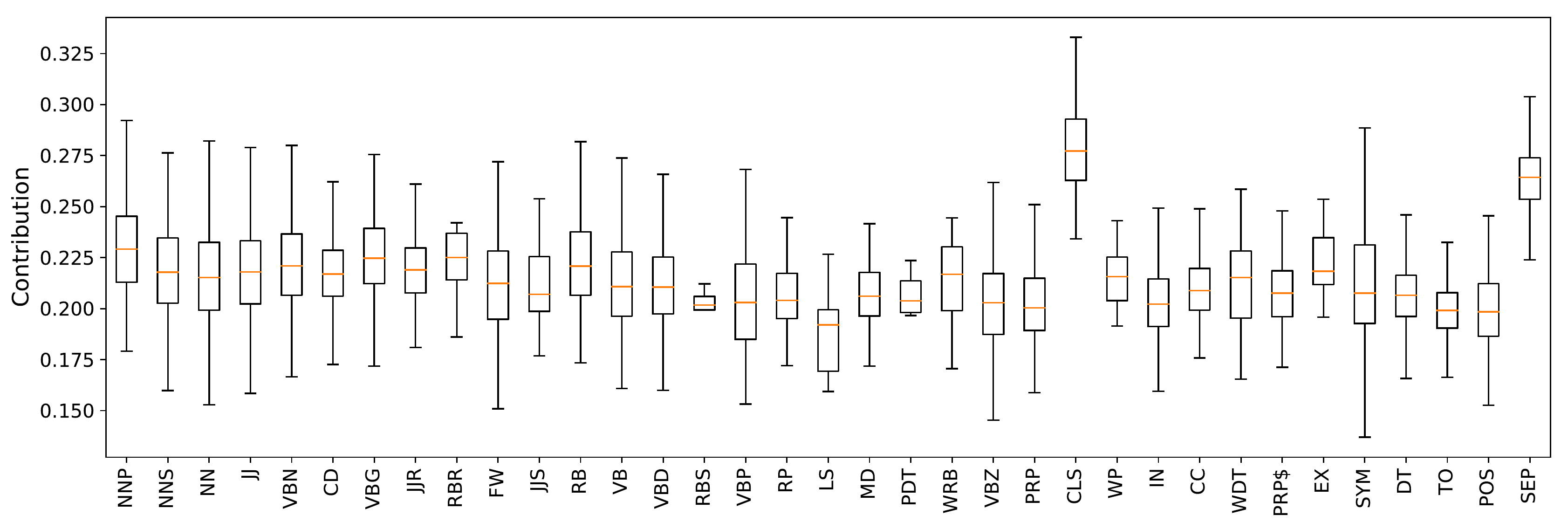}
\caption{Layer 3: Similar to layer 2 with decreasing contribution overall.}
\end{center}
\end{figure}

\begin{figure}[h]
\begin{center}
\includegraphics[width=1\linewidth]{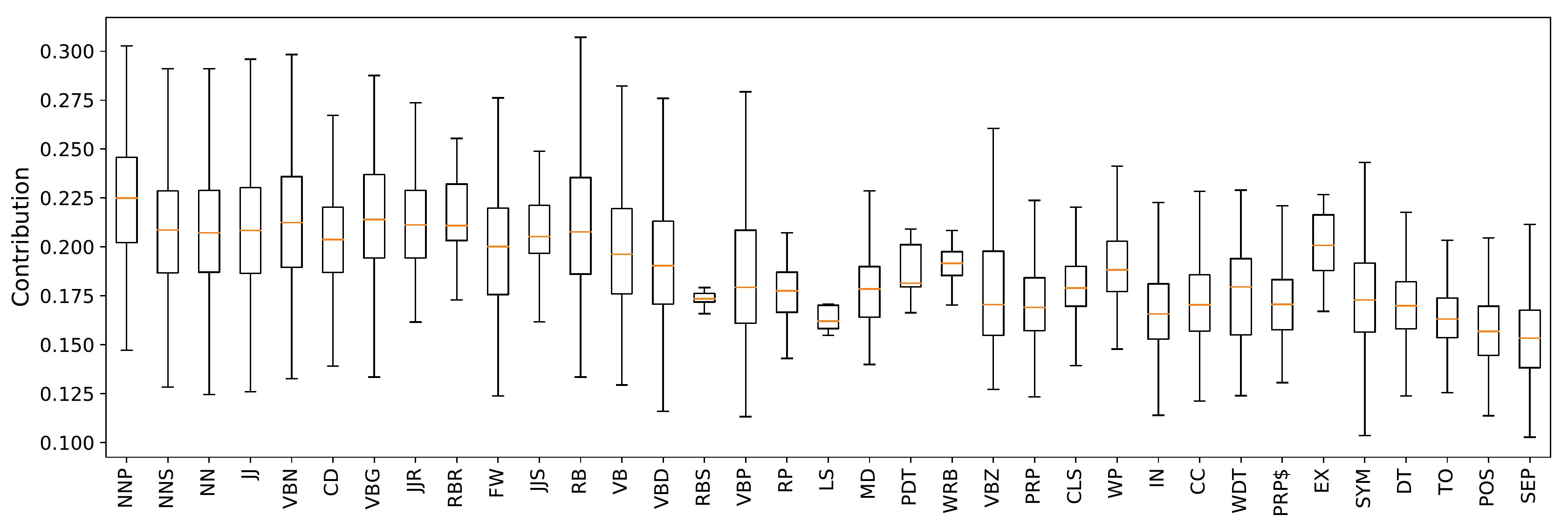}
\caption{Layer 4: The original input contribution to [CLS] and [SEP] falls significantly. The trend that the word types will follow until the last layer is already clear: Most nouns (NNP, NNS, NN), verbs (VBN, VB, VBD, VBP), adjectives (JJ, JJS) and adverbs (RBR, RBS) keep more contribution from their corresponding input embeddings than words with \quotes{less} semantic meaning like Wh-pronouns and determiners (WP, WDT), prepositons (IN), coordinating conjunctions (CC), symbols (SYM), possessives (PRP\$, POS) or determiners (DT). }
\end{center}
\end{figure}

\begin{figure}[h]
\begin{center}
\includegraphics[width=1\linewidth]{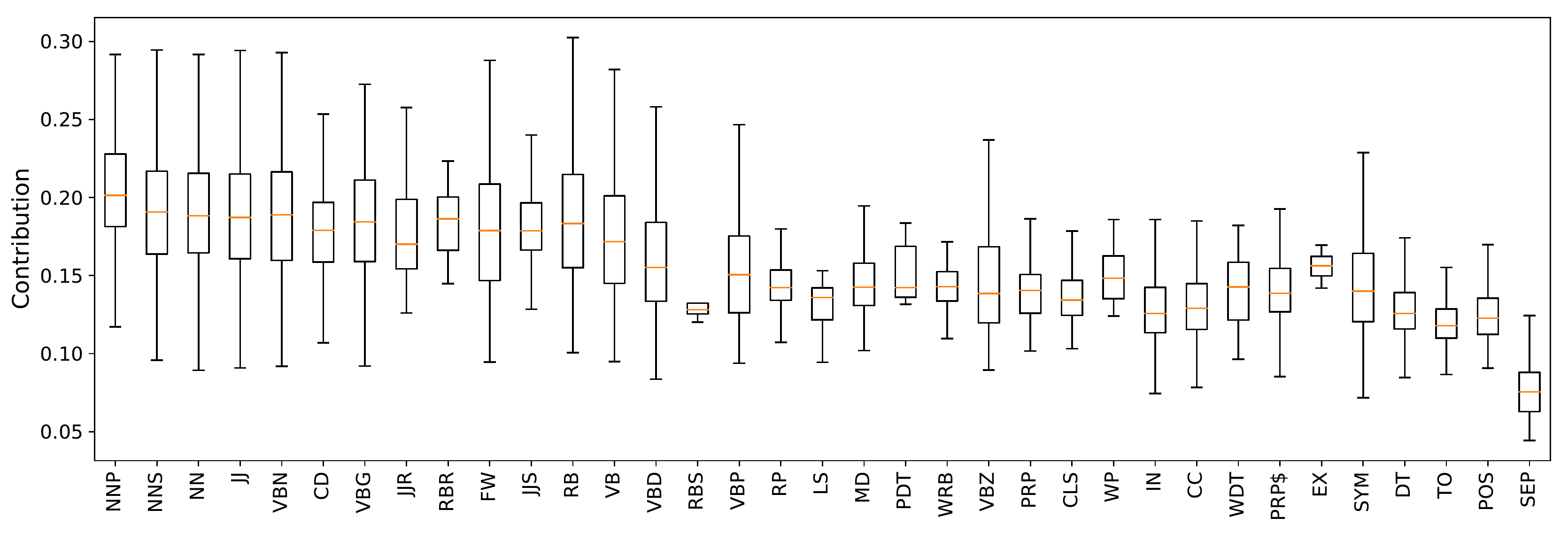}
\caption{Layer 5: The trend started in the previous layer continues, with a reduction of internal variability within those word types with less original contribution.}
\end{center}
\end{figure}

\begin{figure}[h]
\begin{center}
\includegraphics[width=1\linewidth]{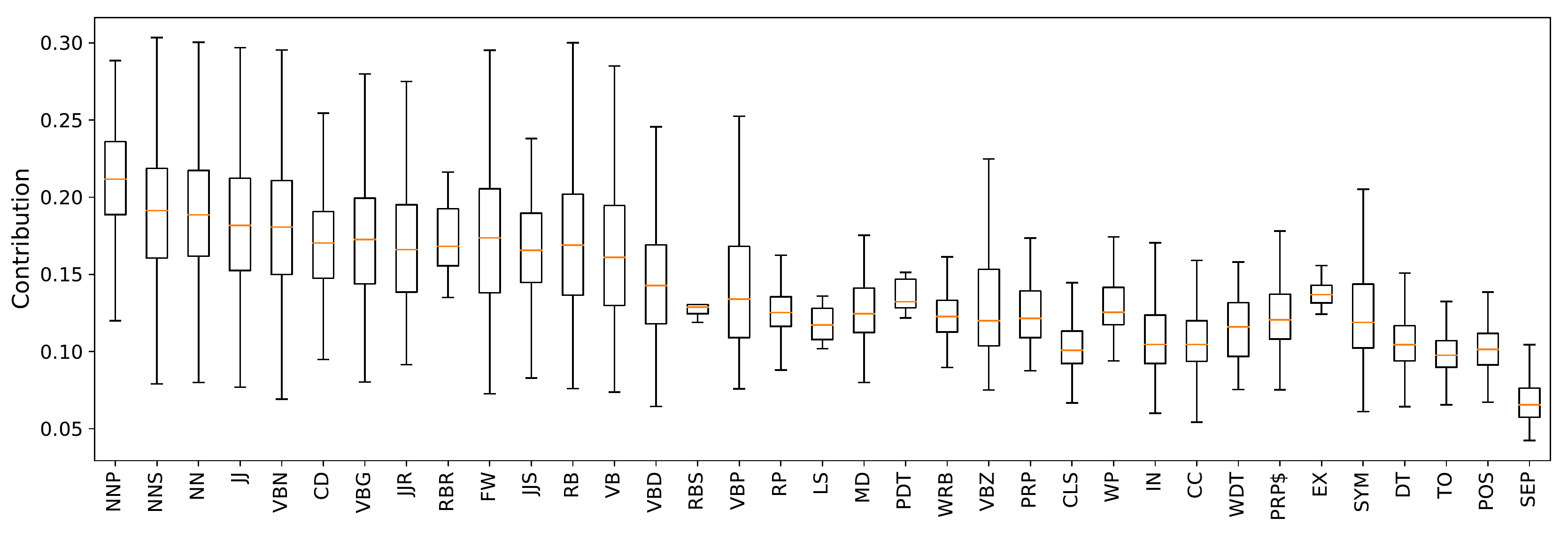}
\caption{Layer 6: Similar behavior as in the previous layer with minor evolution.}
\end{center}
\end{figure}

\begin{figure}[h]
\begin{center}
\includegraphics[width=1\linewidth]{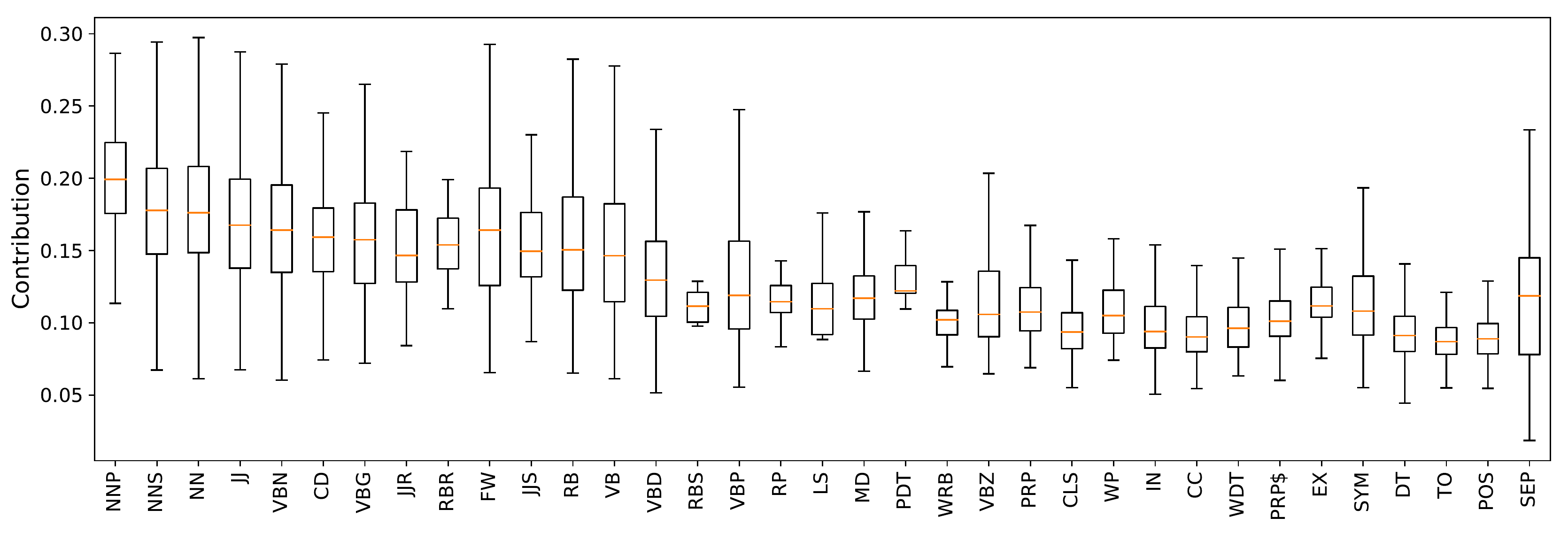}
\caption{Layer 7: Minor changes with respect to Layer 6.}
\end{center}
\end{figure}

\begin{figure}[h]
\begin{center}
\includegraphics[width=1\linewidth]{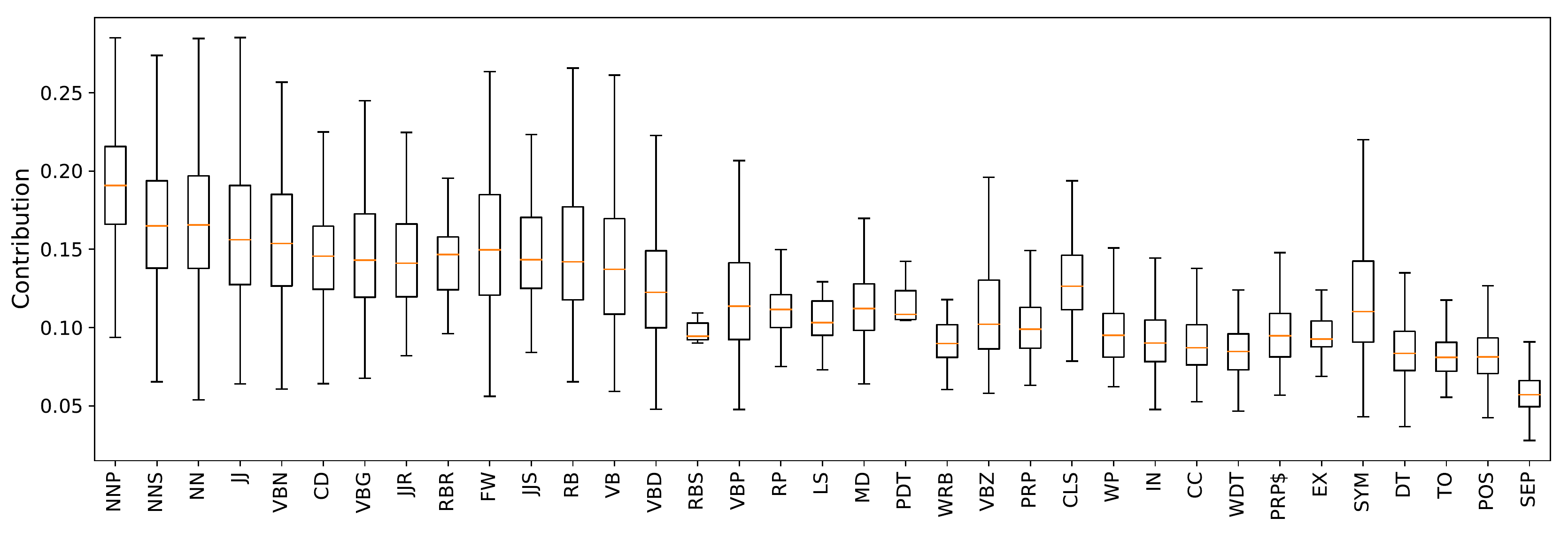}
\caption{Layer 8: At this point there is clearly a different behavior between the tokens with most contribution which present more intra-class variability, and those with less contribution, which are more uniform.}
\end{center}
\end{figure}

\begin{figure}[h]
\begin{center}
\includegraphics[width=1\linewidth]{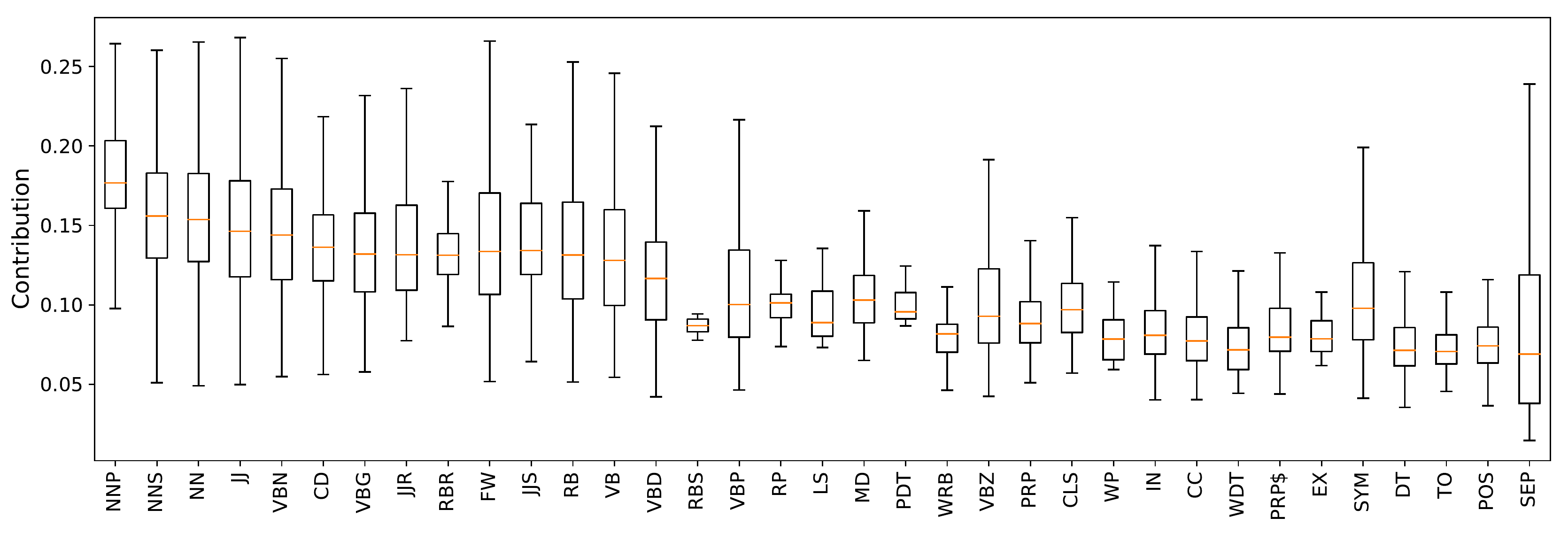}
\caption{Layer 9: SEP changes increasing the contribution, while the rest stays similar.}
\end{center}
\end{figure}

\begin{figure}[h]
\begin{center}
\includegraphics[width=1\linewidth]{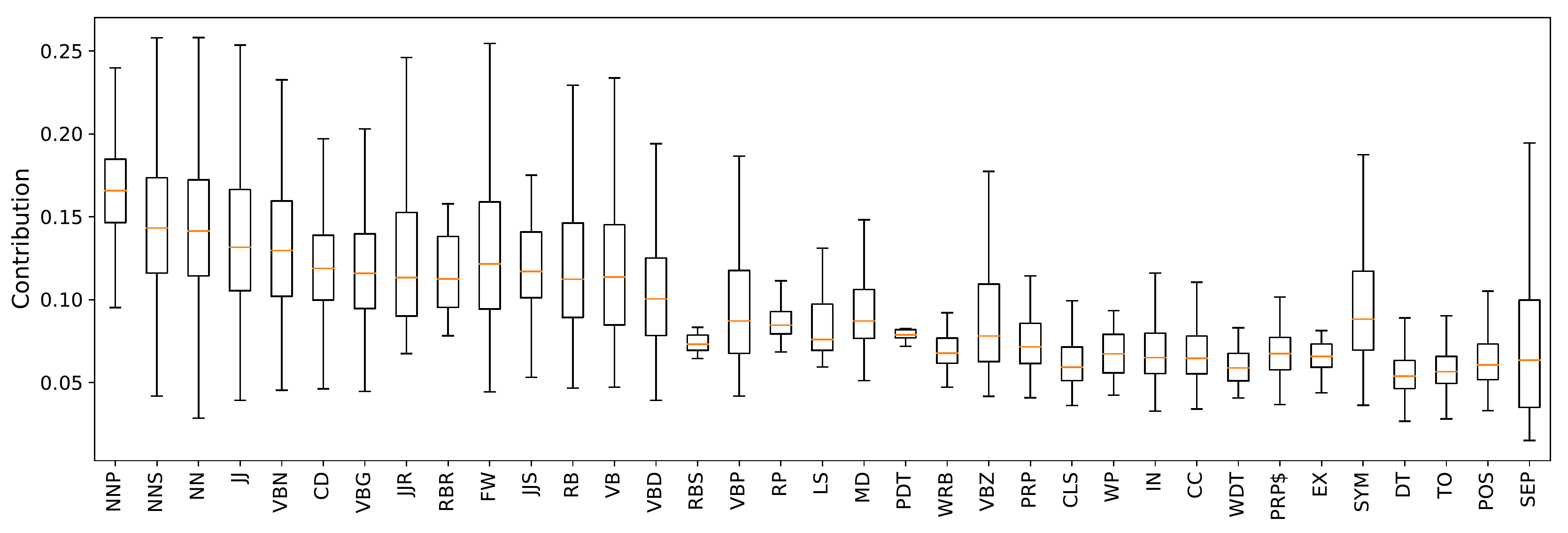}
\caption{Layer 10: The contribution evolves with the same pattern as in previous layers.}
\end{center}
\end{figure}

\begin{figure}[h]
\begin{center}
\includegraphics[width=1\linewidth]{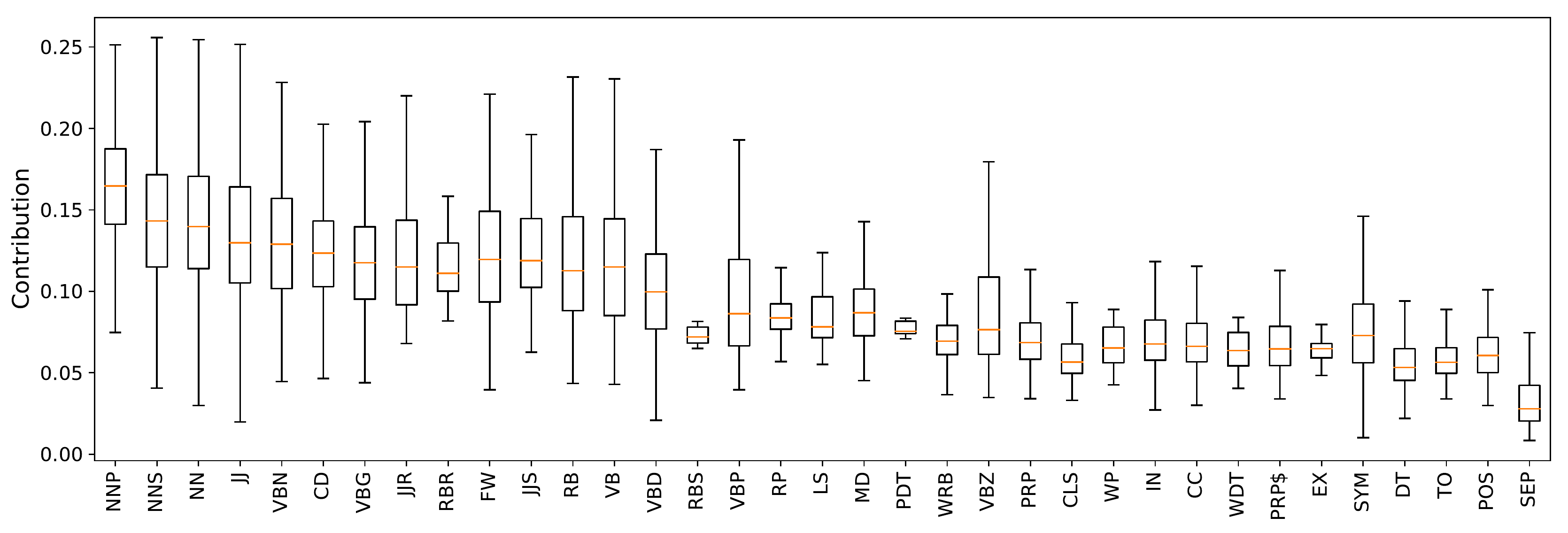}
\caption{Layer 11:The contribution evolves with the same pattern as in previous layers.}
\end{center}
\end{figure}

\begin{figure}[h]
\begin{center}
\includegraphics[width=1\linewidth]{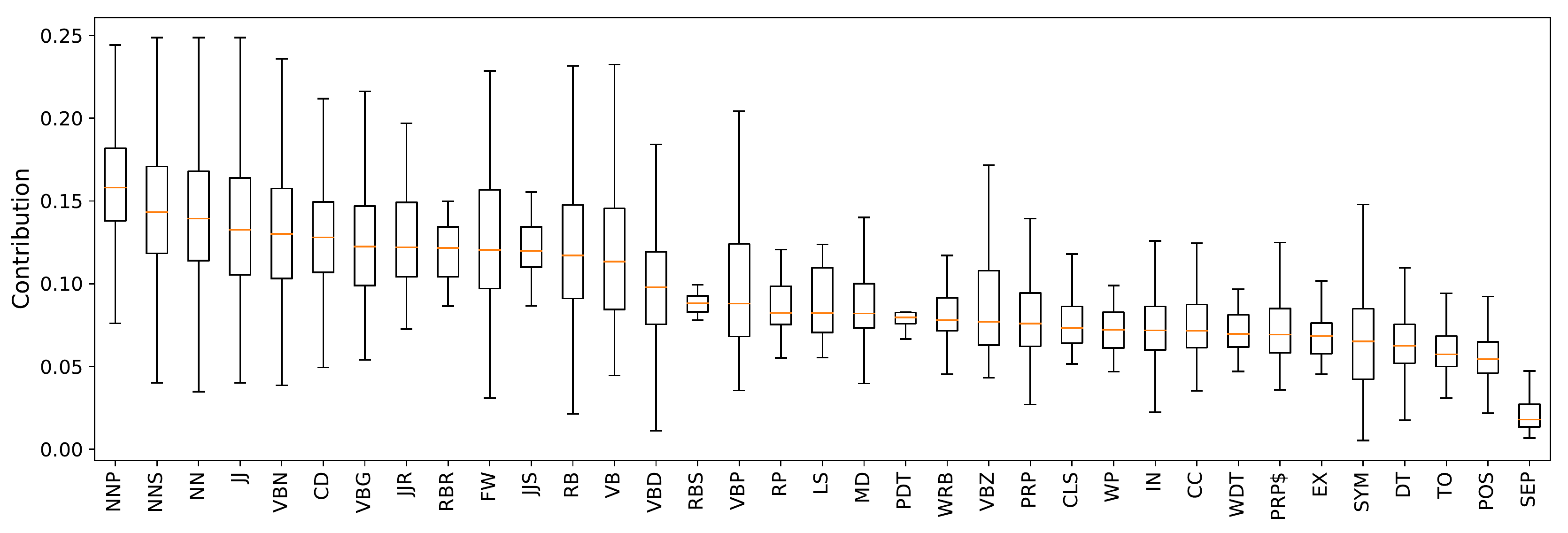}
\caption{Layer 12: Finally, nouns, verbs, adjectives, adverbs, receive more contribution from their corresponding input than determiners, prepositions, pronouns, "to" words and symbols.}
\end{center}
\end{figure}

\clearpage
\section{Generalization to other Datasets}\label{AppDataSets}
In this appendix we reproduce several experiments from the main text using the development sets of two additional datasets from the GLUE benchmark: The Corpus of Linguistic Acceptability (CoLA)~\citep{warstadt2018neural}, and the matched Multi-Genre Natural Language Inference corpus (MNLI-matched)~\citep{N18-1101}. CoLA is a dataset about grammatical acceptability of sentences and MNLI consists of pairs of sentences where the second sentence entails, contradicts or is neutral about the first one. These datasets differ significantly from MRPC. The development set of CoLa has 1043 examples with sequence length $d_s$ between 5 and 35 tokens, and 11 tokens on average. The development set of MNLI-m consits of 9815 examples although we restrict the experiments to the first 4000 examples without loss of generality. These contain a total of 155964 tokens, with a sequence length comprised between 6 and 128 tokens and an average of 39 tokens per example.

The results presented in this appendix are qualitatively similar to those presented in the main text, which shows that our empirical conclusions about BERT are general across data domains.

\subsection{Token Identifiability}

Here we reproduce the main token identifiability results of Section~\ref{IdenTok} on two additional datsets: CoLA and MNLI. Qualitatively, the results are in line with those for MRPC. Note that a random classifier would achieve an accuracy of $1/\bar{d_s}$, where $\bar{d_s}$ denotes the average sentence length. Thus, the random guessing baselines for MRPC, CoLA and MNLI are $1.7\%$, $9\%$ and $2.6\%$ respectively. 

\subsubsection{CoLA}

\begin{figure}[h]
\begin{center}
\input{token_classification_experiments/rebuttal_plots/TokenIdentityColaMNLI/TokenIdentity_COLA.tex}
\end{center}
\caption{Identifiability of contextual word embeddings at different layers on CoLA.}
\label{fig:appendix_TokenIdentity_cola}
\end{figure}
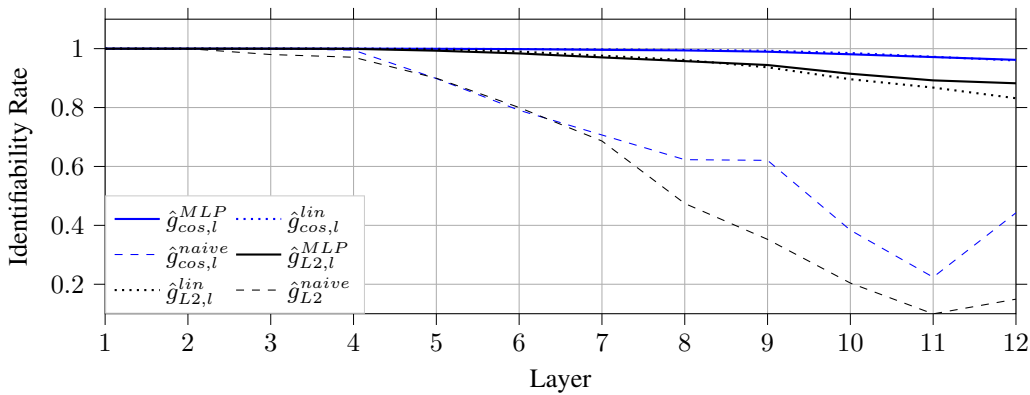

Figure~\ref{fig:appendix_TokenIdentity_cola} shows the token identifiability results for CoLA. 

\clearpage 

\subsubsection{MNLI}

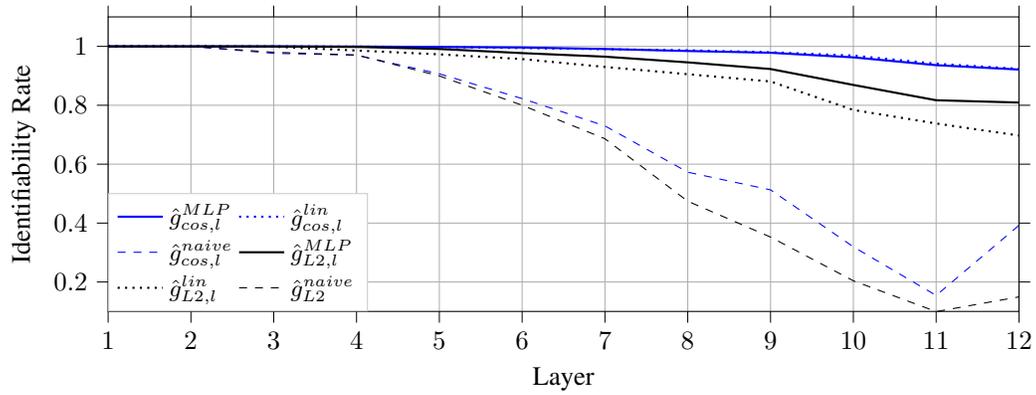
\begin{figure}[h]
\begin{center}
\input{token_classification_experiments/rebuttal_plots/TokenIdentityColaMNLI/TokenIdentity_MNLI_first500.tex}
\end{center}
\caption{Identifiability of contextual word embeddings at different layers on a the first 500 sentences of MNLI-matched (19,839 tokens).}
\label{fig:appendix_TokenIdentity_mnli}
\end{figure}

Figure~\ref{fig:appendix_TokenIdentity_mnli} shows the token identifiability results for the first 500 sentences (19,839 tokens) of MNLI-matched. 

\clearpage

\subsection{Attribution Analysis}

\subsubsection{CoLA Experiments}

Figure~\ref{cola_1} shows the token mixing analysis for the CoLA dataset. The behavior is very similar to MRPC with the only difference that both, the contribution of the original token and the percentage of tokens that are not maximum contributors to their embeddings are slightly larger across layers. However, this increase is explained by CoLA consisting of much shorter sequences on average than MRPC.

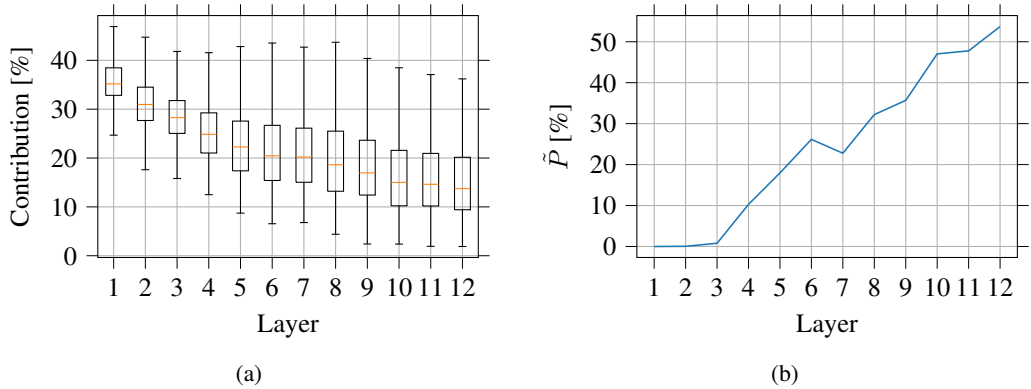
\begin{figure}[h]
\centering
\begin{subfigure}[h]{.49\textwidth}
\centering
\input{gradient_notfinetuned_task_indep_figs/cola/contribution_all_task_independent_not_fine_tuned_cola.tex}
\caption{}
\label{contr_all_cola}
\end{subfigure}
\hfill
\begin{subfigure}[h]{.49\textwidth}
\centering
\input{gradient_notfinetuned_task_indep_figs/cola/rank_percent_task_independent_not_fine_tuned_cola.tex}
\caption{}
\label{rank_all_cola}
\end{subfigure}%
\caption{(a) Contribution of the input token to the embedding at the same position. (b) Percentage of tokens $\tilde{P}$ that are \emph{not} the main contributors to their corresponding contextual embedding at each layer.}
\end{figure}\label{cola_1}

Figure \ref{cola_2} presents the context aggregation for the CoLA development set. We observe the same general trend as for MRPC, with the context being aggregated mostly locally and long range dependencies increasing in the later layers. The fact that examples in CoLA have an average sequence length of 11 tokens explains the smaller relative contribution of tokens beyond the 10th neighbour.

\begin{figure}[h]
\begin{center}
\begin{subfigure}[h]{.49\textwidth}
\input{gradient_notfinetuned_task_indep_figs/cola/relative_attr_per_layer.tex}
\caption{}\label{rel_attr_cola}
\end{subfigure}%
\begin{subfigure}[h]{.49\textwidth}
\input{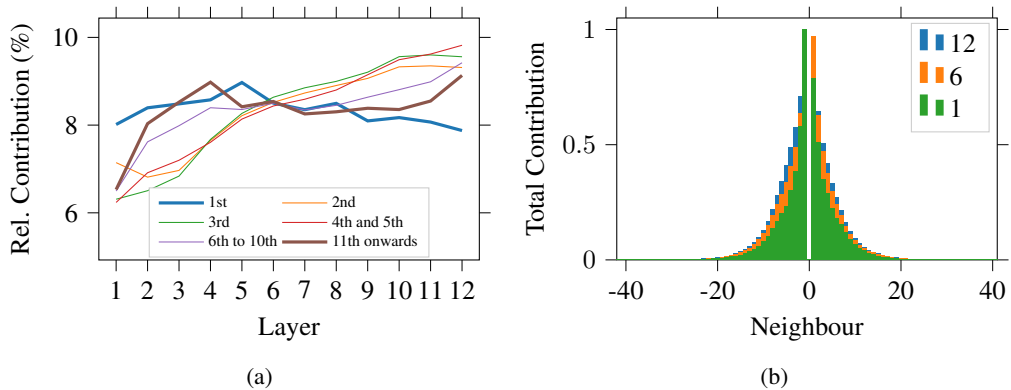}
\caption{}\label{tot_attr_cola}
\end{subfigure}%
\end{center}
\caption{(a) Relative contribution per layer of neighbours at different positions. 
(b) Total contribution per neighbour for the first, middle and last layers. 
}\label{self_attention_is_local_cola}
\end{figure}\label{cola_2}

\clearpage
%%%%%%%%%%%%%%%
\subsubsection{MNLI Experiments}

As shown by Figures \ref{mnli_1} and \ref{mnli_2}, the results with the MNLI matched dataset are very similar to the ones presented in the main text. No meaningful discrepancy exists in this case.

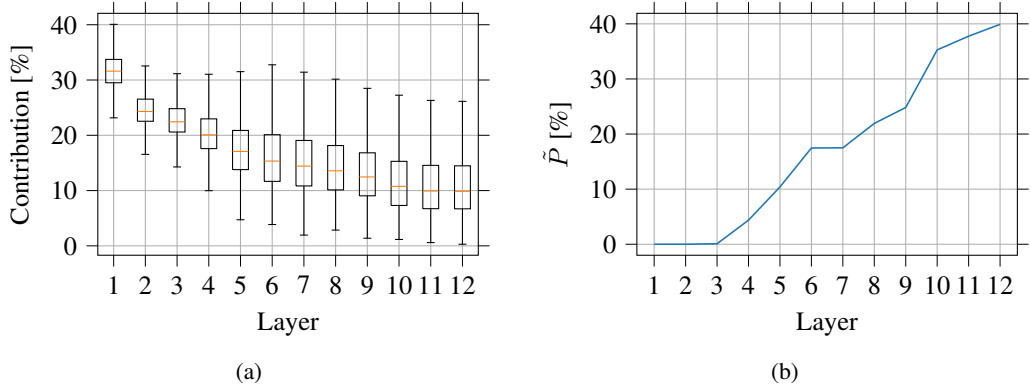
\begin{figure}[h]
\centering
\begin{subfigure}[h]{.49\textwidth}
\centering
\input{gradient_notfinetuned_task_indep_figs/mnlim/contribution_all_task_independent_not_fine_tuned_mnlim.tex}
\caption{}
\label{contr_all_mnlim}
\end{subfigure}
\hfill
\begin{subfigure}[h]{.49\textwidth}
\centering
\input{gradient_notfinetuned_task_indep_figs/mnlim/rank_percent_task_independent_not_fine_tuned_mnlim.tex}
\caption{}
\label{rank_all_mnlim}
\end{subfigure}%
\caption{(a) Contribution of the input token to the embedding at the same position. (b) Percentage of tokens $\tilde{P}$ that are \emph{not} the main contributors to their corresponding contextual embedding at each 
layer.}
\end{figure}\label{mnli_1}

\begin{figure}[h]
\begin{center}
\begin{subfigure}[h]{.49\textwidth}
\input{gradient_notfinetuned_task_indep_figs/mnlim/relative_attr_per_layer.tex}
\caption{}\label{rel_attr_mnlim}
\end{subfigure}%
\begin{subfigure}[h]{.49\textwidth}
\input{gradient_notfinetuned_task_indep_figs/mnlim/context_per_token.tex}
\caption{}\label{tot_attr_mnlim}
\end{subfigure}%
\end{center}
\caption{(a) Relative contribution per layer of neighbours at different positions. 
(b) Total contribution per neighbour for the first, middle and last layers. 
}\label{self_attention_is_local_mnlim}
\end{figure}\label{mnli_2}

%% file: identifiability/avg_attn2.tex
% This file was created by tikzplotlib v0.8.2.
\begin{tikzpicture}

\definecolor{color0}{rgb}{0.203921568627451,0.596078431372549,0.858823529411765}
\definecolor{color1}{rgb}{0.349019607843137,0.850980392156863,0.556862745098039}

\begin{axis}[
axis line style={white!15.0!black},
legend cell align={left},
legend style={draw=white!80.0!black},
tick align=outside,
tick pos=both,
x grid style={white!80.0!black},
xlabel={Layer},
xmajorgrids,
width=\linewidth,
height=\effectiveAttnPlotHeight,
xmin=0.367277355720007, xmax=12.63272264428,
xtick style={color=white!15.0!black},
y grid style={white!80.0!black},
ylabel={Effective Attention},
ymajorgrids,
ymin=-0.050135019036875, ymax=1.0463631746271,
ytick style={color=white!15.0!black}
]
\addplot [only marks, draw=color0, fill=color0, colormap={mymap}{[1pt]
 rgb(0pt)=(0.01060815,0.01808215,0.10018654);
  rgb(1pt)=(0.01428972,0.02048237,0.10374486);
  rgb(2pt)=(0.01831941,0.0229766,0.10738511);
  rgb(3pt)=(0.02275049,0.02554464,0.11108639);
  rgb(4pt)=(0.02759119,0.02818316,0.11483751);
  rgb(5pt)=(0.03285175,0.03088792,0.11863035);
  rgb(6pt)=(0.03853466,0.03365771,0.12245873);
  rgb(7pt)=(0.04447016,0.03648425,0.12631831);
  rgb(8pt)=(0.05032105,0.03936808,0.13020508);
  rgb(9pt)=(0.05611171,0.04224835,0.13411624);
  rgb(10pt)=(0.0618531,0.04504866,0.13804929);
  rgb(11pt)=(0.06755457,0.04778179,0.14200206);
  rgb(12pt)=(0.0732236,0.05045047,0.14597263);
  rgb(13pt)=(0.0788708,0.05305461,0.14995981);
  rgb(14pt)=(0.08450105,0.05559631,0.15396203);
  rgb(15pt)=(0.09011319,0.05808059,0.15797687);
  rgb(16pt)=(0.09572396,0.06050127,0.16200507);
  rgb(17pt)=(0.10132312,0.06286782,0.16604287);
  rgb(18pt)=(0.10692823,0.06517224,0.17009175);
  rgb(19pt)=(0.1125315,0.06742194,0.17414848);
  rgb(20pt)=(0.11813947,0.06961499,0.17821272);
  rgb(21pt)=(0.12375803,0.07174938,0.18228425);
  rgb(22pt)=(0.12938228,0.07383015,0.18636053);
  rgb(23pt)=(0.13501631,0.07585609,0.19044109);
  rgb(24pt)=(0.14066867,0.0778224,0.19452676);
  rgb(25pt)=(0.14633406,0.07973393,0.1986151);
  rgb(26pt)=(0.15201338,0.08159108,0.20270523);
  rgb(27pt)=(0.15770877,0.08339312,0.20679668);
  rgb(28pt)=(0.16342174,0.0851396,0.21088893);
  rgb(29pt)=(0.16915387,0.08682996,0.21498104);
  rgb(30pt)=(0.17489524,0.08848235,0.2190294);
  rgb(31pt)=(0.18065495,0.09009031,0.22303512);
  rgb(32pt)=(0.18643324,0.09165431,0.22699705);
  rgb(33pt)=(0.19223028,0.09317479,0.23091409);
  rgb(34pt)=(0.19804623,0.09465217,0.23478512);
  rgb(35pt)=(0.20388117,0.09608689,0.23860907);
  rgb(36pt)=(0.20973515,0.09747934,0.24238489);
  rgb(37pt)=(0.21560818,0.09882993,0.24611154);
  rgb(38pt)=(0.22150014,0.10013944,0.2497868);
  rgb(39pt)=(0.22741085,0.10140876,0.25340813);
  rgb(40pt)=(0.23334047,0.10263737,0.25697736);
  rgb(41pt)=(0.23928891,0.10382562,0.2604936);
  rgb(42pt)=(0.24525608,0.10497384,0.26395596);
  rgb(43pt)=(0.25124182,0.10608236,0.26736359);
  rgb(44pt)=(0.25724602,0.10715148,0.27071569);
  rgb(45pt)=(0.26326851,0.1081815,0.27401148);
  rgb(46pt)=(0.26930915,0.1091727,0.2772502);
  rgb(47pt)=(0.27536766,0.11012568,0.28043021);
  rgb(48pt)=(0.28144375,0.11104133,0.2835489);
  rgb(49pt)=(0.2875374,0.11191896,0.28660853);
  rgb(50pt)=(0.29364846,0.11275876,0.2896085);
  rgb(51pt)=(0.29977678,0.11356089,0.29254823);
  rgb(52pt)=(0.30592213,0.11432553,0.29542718);
  rgb(53pt)=(0.31208435,0.11505284,0.29824485);
  rgb(54pt)=(0.31826327,0.1157429,0.30100076);
  rgb(55pt)=(0.32445869,0.11639585,0.30369448);
  rgb(56pt)=(0.33067031,0.11701189,0.30632563);
  rgb(57pt)=(0.33689808,0.11759095,0.3088938);
  rgb(58pt)=(0.34314168,0.11813362,0.31139721);
  rgb(59pt)=(0.34940101,0.11863987,0.3138355);
  rgb(60pt)=(0.355676,0.11910909,0.31620996);
  rgb(61pt)=(0.36196644,0.1195413,0.31852037);
  rgb(62pt)=(0.36827206,0.11993653,0.32076656);
  rgb(63pt)=(0.37459292,0.12029443,0.32294825);
  rgb(64pt)=(0.38092887,0.12061482,0.32506528);
  rgb(65pt)=(0.38727975,0.12089756,0.3271175);
  rgb(66pt)=(0.39364518,0.12114272,0.32910494);
  rgb(67pt)=(0.40002537,0.12134964,0.33102734);
  rgb(68pt)=(0.40642019,0.12151801,0.33288464);
  rgb(69pt)=(0.41282936,0.12164769,0.33467689);
  rgb(70pt)=(0.41925278,0.12173833,0.33640407);
  rgb(71pt)=(0.42569057,0.12178916,0.33806605);
  rgb(72pt)=(0.43214263,0.12179973,0.33966284);
  rgb(73pt)=(0.43860848,0.12177004,0.34119475);
  rgb(74pt)=(0.44508855,0.12169883,0.34266151);
  rgb(75pt)=(0.45158266,0.12158557,0.34406324);
  rgb(76pt)=(0.45809049,0.12142996,0.34540024);
  rgb(77pt)=(0.46461238,0.12123063,0.34667231);
  rgb(78pt)=(0.47114798,0.12098721,0.34787978);
  rgb(79pt)=(0.47769736,0.12069864,0.34902273);
  rgb(80pt)=(0.48426077,0.12036349,0.35010104);
  rgb(81pt)=(0.49083761,0.11998161,0.35111537);
  rgb(82pt)=(0.49742847,0.11955087,0.35206533);
  rgb(83pt)=(0.50403286,0.11907081,0.35295152);
  rgb(84pt)=(0.51065109,0.11853959,0.35377385);
  rgb(85pt)=(0.51728314,0.1179558,0.35453252);
  rgb(86pt)=(0.52392883,0.11731817,0.35522789);
  rgb(87pt)=(0.53058853,0.11662445,0.35585982);
  rgb(88pt)=(0.53726173,0.11587369,0.35642903);
  rgb(89pt)=(0.54394898,0.11506307,0.35693521);
  rgb(90pt)=(0.5506426,0.11420757,0.35737863);
  rgb(91pt)=(0.55734473,0.11330456,0.35775059);
  rgb(92pt)=(0.56405586,0.11235265,0.35804813);
  rgb(93pt)=(0.57077365,0.11135597,0.35827146);
  rgb(94pt)=(0.5774991,0.11031233,0.35841679);
  rgb(95pt)=(0.58422945,0.10922707,0.35848469);
  rgb(96pt)=(0.59096382,0.10810205,0.35847347);
  rgb(97pt)=(0.59770215,0.10693774,0.35838029);
  rgb(98pt)=(0.60444226,0.10573912,0.35820487);
  rgb(99pt)=(0.61118304,0.10450943,0.35794557);
  rgb(100pt)=(0.61792306,0.10325288,0.35760108);
  rgb(101pt)=(0.62466162,0.10197244,0.35716891);
  rgb(102pt)=(0.63139686,0.10067417,0.35664819);
  rgb(103pt)=(0.63812122,0.09938212,0.35603757);
  rgb(104pt)=(0.64483795,0.0980891,0.35533555);
  rgb(105pt)=(0.65154562,0.09680192,0.35454107);
  rgb(106pt)=(0.65824241,0.09552918,0.3536529);
  rgb(107pt)=(0.66492652,0.09428017,0.3526697);
  rgb(108pt)=(0.67159578,0.09306598,0.35159077);
  rgb(109pt)=(0.67824099,0.09192342,0.3504148);
  rgb(110pt)=(0.684863,0.09085633,0.34914061);
  rgb(111pt)=(0.69146268,0.0898675,0.34776864);
  rgb(112pt)=(0.69803757,0.08897226,0.3462986);
  rgb(113pt)=(0.70457834,0.0882129,0.34473046);
  rgb(114pt)=(0.71108138,0.08761223,0.3430635);
  rgb(115pt)=(0.7175507,0.08716212,0.34129974);
  rgb(116pt)=(0.72398193,0.08688725,0.33943958);
  rgb(117pt)=(0.73035829,0.0868623,0.33748452);
  rgb(118pt)=(0.73669146,0.08704683,0.33543669);
  rgb(119pt)=(0.74297501,0.08747196,0.33329799);
  rgb(120pt)=(0.74919318,0.08820542,0.33107204);
  rgb(121pt)=(0.75535825,0.08919792,0.32876184);
  rgb(122pt)=(0.76145589,0.09050716,0.32637117);
  rgb(123pt)=(0.76748424,0.09213602,0.32390525);
  rgb(124pt)=(0.77344838,0.09405684,0.32136808);
  rgb(125pt)=(0.77932641,0.09634794,0.31876642);
  rgb(126pt)=(0.78513609,0.09892473,0.31610488);
  rgb(127pt)=(0.79085854,0.10184672,0.313391);
  rgb(128pt)=(0.7965014,0.10506637,0.31063031);
  rgb(129pt)=(0.80205987,0.10858333,0.30783);
  rgb(130pt)=(0.80752799,0.11239964,0.30499738);
  rgb(131pt)=(0.81291606,0.11645784,0.30213802);
  rgb(132pt)=(0.81820481,0.12080606,0.29926105);
  rgb(133pt)=(0.82341472,0.12535343,0.2963705);
  rgb(134pt)=(0.82852822,0.13014118,0.29347474);
  rgb(135pt)=(0.83355779,0.13511035,0.29057852);
  rgb(136pt)=(0.83850183,0.14025098,0.2876878);
  rgb(137pt)=(0.84335441,0.14556683,0.28480819);
  rgb(138pt)=(0.84813096,0.15099892,0.281943);
  rgb(139pt)=(0.85281737,0.15657772,0.27909826);
  rgb(140pt)=(0.85742602,0.1622583,0.27627462);
  rgb(141pt)=(0.86196552,0.16801239,0.27346473);
  rgb(142pt)=(0.86641628,0.17387796,0.27070818);
  rgb(143pt)=(0.87079129,0.17982114,0.26797378);
  rgb(144pt)=(0.87507281,0.18587368,0.26529697);
  rgb(145pt)=(0.87925878,0.19203259,0.26268136);
  rgb(146pt)=(0.8833417,0.19830556,0.26014181);
  rgb(147pt)=(0.88731387,0.20469941,0.25769539);
  rgb(148pt)=(0.89116859,0.21121788,0.2553592);
  rgb(149pt)=(0.89490337,0.21785614,0.25314362);
  rgb(150pt)=(0.8985026,0.22463251,0.25108745);
  rgb(151pt)=(0.90197527,0.23152063,0.24918223);
  rgb(152pt)=(0.90530097,0.23854541,0.24748098);
  rgb(153pt)=(0.90848638,0.24568473,0.24598324);
  rgb(154pt)=(0.911533,0.25292623,0.24470258);
  rgb(155pt)=(0.9144225,0.26028902,0.24369359);
  rgb(156pt)=(0.91717106,0.26773821,0.24294137);
  rgb(157pt)=(0.91978131,0.27526191,0.24245973);
  rgb(158pt)=(0.92223947,0.28287251,0.24229568);
  rgb(159pt)=(0.92456587,0.29053388,0.24242622);
  rgb(160pt)=(0.92676657,0.29823282,0.24285536);
  rgb(161pt)=(0.92882964,0.30598085,0.24362274);
  rgb(162pt)=(0.93078135,0.31373977,0.24468803);
  rgb(163pt)=(0.93262051,0.3215093,0.24606461);
  rgb(164pt)=(0.93435067,0.32928362,0.24775328);
  rgb(165pt)=(0.93599076,0.33703942,0.24972157);
  rgb(166pt)=(0.93752831,0.34479177,0.25199928);
  rgb(167pt)=(0.93899289,0.35250734,0.25452808);
  rgb(168pt)=(0.94036561,0.36020899,0.25734661);
  rgb(169pt)=(0.94167588,0.36786594,0.2603949);
  rgb(170pt)=(0.94291042,0.37549479,0.26369821);
  rgb(171pt)=(0.94408513,0.3830811,0.26722004);
  rgb(172pt)=(0.94520419,0.39062329,0.27094924);
  rgb(173pt)=(0.94625977,0.39813168,0.27489742);
  rgb(174pt)=(0.94727016,0.4055909,0.27902322);
  rgb(175pt)=(0.94823505,0.41300424,0.28332283);
  rgb(176pt)=(0.94914549,0.42038251,0.28780969);
  rgb(177pt)=(0.95001704,0.42771398,0.29244728);
  rgb(178pt)=(0.95085121,0.43500005,0.29722817);
  rgb(179pt)=(0.95165009,0.44224144,0.30214494);
  rgb(180pt)=(0.9524044,0.44944853,0.3072105);
  rgb(181pt)=(0.95312556,0.45661389,0.31239776);
  rgb(182pt)=(0.95381595,0.46373781,0.31769923);
  rgb(183pt)=(0.95447591,0.47082238,0.32310953);
  rgb(184pt)=(0.95510255,0.47787236,0.32862553);
  rgb(185pt)=(0.95569679,0.48489115,0.33421404);
  rgb(186pt)=(0.95626788,0.49187351,0.33985601);
  rgb(187pt)=(0.95681685,0.49882008,0.34555431);
  rgb(188pt)=(0.9573439,0.50573243,0.35130912);
  rgb(189pt)=(0.95784842,0.51261283,0.35711942);
  rgb(190pt)=(0.95833051,0.51946267,0.36298589);
  rgb(191pt)=(0.95879054,0.52628305,0.36890904);
  rgb(192pt)=(0.95922872,0.53307513,0.3748895);
  rgb(193pt)=(0.95964538,0.53983991,0.38092784);
  rgb(194pt)=(0.96004345,0.54657593,0.3870292);
  rgb(195pt)=(0.96042097,0.55328624,0.39319057);
  rgb(196pt)=(0.96077819,0.55997184,0.39941173);
  rgb(197pt)=(0.9611152,0.5666337,0.40569343);
  rgb(198pt)=(0.96143273,0.57327231,0.41203603);
  rgb(199pt)=(0.96173392,0.57988594,0.41844491);
  rgb(200pt)=(0.96201757,0.58647675,0.42491751);
  rgb(201pt)=(0.96228344,0.59304598,0.43145271);
  rgb(202pt)=(0.96253168,0.5995944,0.43805131);
  rgb(203pt)=(0.96276513,0.60612062,0.44471698);
  rgb(204pt)=(0.96298491,0.6126247,0.45145074);
  rgb(205pt)=(0.96318967,0.61910879,0.45824902);
  rgb(206pt)=(0.96337949,0.6255736,0.46511271);
  rgb(207pt)=(0.96355923,0.63201624,0.47204746);
  rgb(208pt)=(0.96372785,0.63843852,0.47905028);
  rgb(209pt)=(0.96388426,0.64484214,0.4861196);
  rgb(210pt)=(0.96403203,0.65122535,0.4932578);
  rgb(211pt)=(0.96417332,0.65758729,0.50046894);
  rgb(212pt)=(0.9643063,0.66393045,0.5077467);
  rgb(213pt)=(0.96443322,0.67025402,0.51509334);
  rgb(214pt)=(0.96455845,0.67655564,0.52251447);
  rgb(215pt)=(0.96467922,0.68283846,0.53000231);
  rgb(216pt)=(0.96479861,0.68910113,0.53756026);
  rgb(217pt)=(0.96492035,0.69534192,0.5451917);
  rgb(218pt)=(0.96504223,0.7015636,0.5528892);
  rgb(219pt)=(0.96516917,0.70776351,0.5606593);
  rgb(220pt)=(0.96530224,0.71394212,0.56849894);
  rgb(221pt)=(0.96544032,0.72010124,0.57640375);
  rgb(222pt)=(0.96559206,0.72623592,0.58438387);
  rgb(223pt)=(0.96575293,0.73235058,0.59242739);
  rgb(224pt)=(0.96592829,0.73844258,0.60053991);
  rgb(225pt)=(0.96612013,0.74451182,0.60871954);
  rgb(226pt)=(0.96632832,0.75055966,0.61696136);
  rgb(227pt)=(0.96656022,0.75658231,0.62527295);
  rgb(228pt)=(0.96681185,0.76258381,0.63364277);
  rgb(229pt)=(0.96709183,0.76855969,0.64207921);
  rgb(230pt)=(0.96739773,0.77451297,0.65057302);
  rgb(231pt)=(0.96773482,0.78044149,0.65912731);
  rgb(232pt)=(0.96810471,0.78634563,0.66773889);
  rgb(233pt)=(0.96850919,0.79222565,0.6764046);
  rgb(234pt)=(0.96893132,0.79809112,0.68512266);
  rgb(235pt)=(0.96935926,0.80395415,0.69383201);
  rgb(236pt)=(0.9698028,0.80981139,0.70252255);
  rgb(237pt)=(0.97025511,0.81566605,0.71120296);
  rgb(238pt)=(0.97071849,0.82151775,0.71987163);
  rgb(239pt)=(0.97120159,0.82736371,0.72851999);
  rgb(240pt)=(0.97169389,0.83320847,0.73716071);
  rgb(241pt)=(0.97220061,0.83905052,0.74578903);
  rgb(242pt)=(0.97272597,0.84488881,0.75440141);
  rgb(243pt)=(0.97327085,0.85072354,0.76299805);
  rgb(244pt)=(0.97383206,0.85655639,0.77158353);
  rgb(245pt)=(0.97441222,0.86238689,0.78015619);
  rgb(246pt)=(0.97501782,0.86821321,0.78871034);
  rgb(247pt)=(0.97564391,0.87403763,0.79725261);
  rgb(248pt)=(0.97628674,0.87986189,0.8057883);
  rgb(249pt)=(0.97696114,0.88568129,0.81430324);
  rgb(250pt)=(0.97765722,0.89149971,0.82280948);
  rgb(251pt)=(0.97837585,0.89731727,0.83130786);
  rgb(252pt)=(0.97912374,0.90313207,0.83979337);
  rgb(253pt)=(0.979891,0.90894778,0.84827858);
  rgb(254pt)=(0.98067764,0.91476465,0.85676611);
  rgb(255pt)=(0.98137749,0.92061729,0.86536915)
}]
table{%
x                      y
1 0.0107924670717722
1 0.00570973556133125
1 0.025054661923022
1 0.0169914512097203
1 0.0358714987191087
1 0.0100900811810611
1 0.0145091704961327
1 0.0290047234112948
1 0.0173454770694382
1 0.0276983030896259
1 0.0209214103229378
1 0.0122964695173727
2 0.0572675723326334
2 0.030113214331045
2 0.0219574579449879
2 0.0643684289336052
2 0.0298948521792757
2 0.0706342485583783
2 0.0695335924655285
2 0.0476823554296123
2 0.0563236924380561
2 0.0937945713782699
2 0.0366580330182029
2 0.0360831208047895
3 0.0311279528604667
3 0.131431387965375
3 0.219396015527826
3 0.105767506276023
3 0.0732854518654737
3 0.0510305649022124
3 0.0917318905404006
3 0.0736439517836188
3 0.0496641730846984
3 0.0307202762887611
3 0.0518521264862243
3 0.0597277644376071
4 0.1047304859349
4 0.0291649371844152
4 0.0296685364014349
4 0.0394971456128234
4 0.0391219904208944
4 0.0239111412826147
4 0.0395742059328457
4 0.0397338129161581
4 0.0697529753030235
4 0.0684336539486116
4 0.0389943322085157
4 0.08587422860472
5 0.0276211490359567
5 0.0131568881535926
5 0.0327676566305423
5 0.0441126923340172
5 0.0152071835021378
5 0.0259951795648523
5 0.0280378813323336
5 0.0349968123858507
5 0.0313069267348076
5 0.0316437250679243
5 0.0234282537999082
5 0.0209734244307462
6 0.0348216569329635
6 0.0343887762954979
6 0.0288318352851097
6 0.0302441403544592
6 0.0367943826479684
6 0.0423937415776229
6 0.041826764702245
6 0.0345686915613855
6 0.0345232609741211
6 0.0271423949622925
6 0.0375697292464972
6 0.0402921092991572
7 0.0361080417694926
7 0.0283713409726196
7 0.0330952317021923
7 0.0355471033482394
7 0.0283314169408913
7 0.0308673564050231
7 0.031374300964449
7 0.0352188445139007
7 0.0318397734074343
7 0.0407525010668167
7 0.0365769404246445
7 0.0246186127166417
8 0.0267261344742693
8 0.0303644712648442
8 0.0350710715878241
8 0.0381642616942803
8 0.0240326317296337
8 0.0276082381741524
8 0.0415381135932845
8 0.041532228152351
8 0.0311143287065869
8 0.03368326339553
8 0.0288609231595324
8 0.0380942674800514
9 0.0301692031968553
9 0.0262370540050696
9 0.0430749298593022
9 0.0281867816874965
9 0.0307340959221589
9 0.0352151062703031
9 0.034969295054526
9 0.0267033728745507
9 0.0344126541858866
9 0.0368798088038268
9 0.0288999956058132
9 0.0318577580802114
10 0.0321420555489659
10 0.0284927473077839
10 0.0251112903383719
10 0.0387363994833254
10 0.0272583768056397
10 0.0260125668089091
10 0.0437432879171499
10 0.0297261295987629
10 0.0347121074283482
10 0.0465079896955402
10 0.0281957383247044
10 0.0366382441111712
11 0.0335902767260268
11 0.0149500939049453
11 0.036021847285607
11 0.0535121261942015
11 0.0396086691081471
11 0.0179927095563089
11 0.0508457343629001
11 0.041310175293269
11 0.0462036840920188
11 0.0497939716551468
11 0.13770555528707
11 0.0259923603681544
12 0.0151232503775109
12 0.0159587593375641
12 0.0200572105351547
12 0.0193264637342359
12 0.0192916999133194
12 0.0312008655203172
12 0.0257706823517697
12 0.0128600504515612
12 0.0114329033203318
12 0.0250598557786066
12 0.0166894110588786
12 0.00938069453033942
};
\addlegendentry{other $\rightarrow$ [SEP]}
\addplot [only marks, draw=color1, fill=color1, colormap={mymap}{[1pt]
 rgb(0pt)=(0.01060815,0.01808215,0.10018654);
  rgb(1pt)=(0.01428972,0.02048237,0.10374486);
  rgb(2pt)=(0.01831941,0.0229766,0.10738511);
  rgb(3pt)=(0.02275049,0.02554464,0.11108639);
  rgb(4pt)=(0.02759119,0.02818316,0.11483751);
  rgb(5pt)=(0.03285175,0.03088792,0.11863035);
  rgb(6pt)=(0.03853466,0.03365771,0.12245873);
  rgb(7pt)=(0.04447016,0.03648425,0.12631831);
  rgb(8pt)=(0.05032105,0.03936808,0.13020508);
  rgb(9pt)=(0.05611171,0.04224835,0.13411624);
  rgb(10pt)=(0.0618531,0.04504866,0.13804929);
  rgb(11pt)=(0.06755457,0.04778179,0.14200206);
  rgb(12pt)=(0.0732236,0.05045047,0.14597263);
  rgb(13pt)=(0.0788708,0.05305461,0.14995981);
  rgb(14pt)=(0.08450105,0.05559631,0.15396203);
  rgb(15pt)=(0.09011319,0.05808059,0.15797687);
  rgb(16pt)=(0.09572396,0.06050127,0.16200507);
  rgb(17pt)=(0.10132312,0.06286782,0.16604287);
  rgb(18pt)=(0.10692823,0.06517224,0.17009175);
  rgb(19pt)=(0.1125315,0.06742194,0.17414848);
  rgb(20pt)=(0.11813947,0.06961499,0.17821272);
  rgb(21pt)=(0.12375803,0.07174938,0.18228425);
  rgb(22pt)=(0.12938228,0.07383015,0.18636053);
  rgb(23pt)=(0.13501631,0.07585609,0.19044109);
  rgb(24pt)=(0.14066867,0.0778224,0.19452676);
  rgb(25pt)=(0.14633406,0.07973393,0.1986151);
  rgb(26pt)=(0.15201338,0.08159108,0.20270523);
  rgb(27pt)=(0.15770877,0.08339312,0.20679668);
  rgb(28pt)=(0.16342174,0.0851396,0.21088893);
  rgb(29pt)=(0.16915387,0.08682996,0.21498104);
  rgb(30pt)=(0.17489524,0.08848235,0.2190294);
  rgb(31pt)=(0.18065495,0.09009031,0.22303512);
  rgb(32pt)=(0.18643324,0.09165431,0.22699705);
  rgb(33pt)=(0.19223028,0.09317479,0.23091409);
  rgb(34pt)=(0.19804623,0.09465217,0.23478512);
  rgb(35pt)=(0.20388117,0.09608689,0.23860907);
  rgb(36pt)=(0.20973515,0.09747934,0.24238489);
  rgb(37pt)=(0.21560818,0.09882993,0.24611154);
  rgb(38pt)=(0.22150014,0.10013944,0.2497868);
  rgb(39pt)=(0.22741085,0.10140876,0.25340813);
  rgb(40pt)=(0.23334047,0.10263737,0.25697736);
  rgb(41pt)=(0.23928891,0.10382562,0.2604936);
  rgb(42pt)=(0.24525608,0.10497384,0.26395596);
  rgb(43pt)=(0.25124182,0.10608236,0.26736359);
  rgb(44pt)=(0.25724602,0.10715148,0.27071569);
  rgb(45pt)=(0.26326851,0.1081815,0.27401148);
  rgb(46pt)=(0.26930915,0.1091727,0.2772502);
  rgb(47pt)=(0.27536766,0.11012568,0.28043021);
  rgb(48pt)=(0.28144375,0.11104133,0.2835489);
  rgb(49pt)=(0.2875374,0.11191896,0.28660853);
  rgb(50pt)=(0.29364846,0.11275876,0.2896085);
  rgb(51pt)=(0.29977678,0.11356089,0.29254823);
  rgb(52pt)=(0.30592213,0.11432553,0.29542718);
  rgb(53pt)=(0.31208435,0.11505284,0.29824485);
  rgb(54pt)=(0.31826327,0.1157429,0.30100076);
  rgb(55pt)=(0.32445869,0.11639585,0.30369448);
  rgb(56pt)=(0.33067031,0.11701189,0.30632563);
  rgb(57pt)=(0.33689808,0.11759095,0.3088938);
  rgb(58pt)=(0.34314168,0.11813362,0.31139721);
  rgb(59pt)=(0.34940101,0.11863987,0.3138355);
  rgb(60pt)=(0.355676,0.11910909,0.31620996);
  rgb(61pt)=(0.36196644,0.1195413,0.31852037);
  rgb(62pt)=(0.36827206,0.11993653,0.32076656);
  rgb(63pt)=(0.37459292,0.12029443,0.32294825);
  rgb(64pt)=(0.38092887,0.12061482,0.32506528);
  rgb(65pt)=(0.38727975,0.12089756,0.3271175);
  rgb(66pt)=(0.39364518,0.12114272,0.32910494);
  rgb(67pt)=(0.40002537,0.12134964,0.33102734);
  rgb(68pt)=(0.40642019,0.12151801,0.33288464);
  rgb(69pt)=(0.41282936,0.12164769,0.33467689);
  rgb(70pt)=(0.41925278,0.12173833,0.33640407);
  rgb(71pt)=(0.42569057,0.12178916,0.33806605);
  rgb(72pt)=(0.43214263,0.12179973,0.33966284);
  rgb(73pt)=(0.43860848,0.12177004,0.34119475);
  rgb(74pt)=(0.44508855,0.12169883,0.34266151);
  rgb(75pt)=(0.45158266,0.12158557,0.34406324);
  rgb(76pt)=(0.45809049,0.12142996,0.34540024);
  rgb(77pt)=(0.46461238,0.12123063,0.34667231);
  rgb(78pt)=(0.47114798,0.12098721,0.34787978);
  rgb(79pt)=(0.47769736,0.12069864,0.34902273);
  rgb(80pt)=(0.48426077,0.12036349,0.35010104);
  rgb(81pt)=(0.49083761,0.11998161,0.35111537);
  rgb(82pt)=(0.49742847,0.11955087,0.35206533);
  rgb(83pt)=(0.50403286,0.11907081,0.35295152);
  rgb(84pt)=(0.51065109,0.11853959,0.35377385);
  rgb(85pt)=(0.51728314,0.1179558,0.35453252);
  rgb(86pt)=(0.52392883,0.11731817,0.35522789);
  rgb(87pt)=(0.53058853,0.11662445,0.35585982);
  rgb(88pt)=(0.53726173,0.11587369,0.35642903);
  rgb(89pt)=(0.54394898,0.11506307,0.35693521);
  rgb(90pt)=(0.5506426,0.11420757,0.35737863);
  rgb(91pt)=(0.55734473,0.11330456,0.35775059);
  rgb(92pt)=(0.56405586,0.11235265,0.35804813);
  rgb(93pt)=(0.57077365,0.11135597,0.35827146);
  rgb(94pt)=(0.5774991,0.11031233,0.35841679);
  rgb(95pt)=(0.58422945,0.10922707,0.35848469);
  rgb(96pt)=(0.59096382,0.10810205,0.35847347);
  rgb(97pt)=(0.59770215,0.10693774,0.35838029);
  rgb(98pt)=(0.60444226,0.10573912,0.35820487);
  rgb(99pt)=(0.61118304,0.10450943,0.35794557);
  rgb(100pt)=(0.61792306,0.10325288,0.35760108);
  rgb(101pt)=(0.62466162,0.10197244,0.35716891);
  rgb(102pt)=(0.63139686,0.10067417,0.35664819);
  rgb(103pt)=(0.63812122,0.09938212,0.35603757);
  rgb(104pt)=(0.64483795,0.0980891,0.35533555);
  rgb(105pt)=(0.65154562,0.09680192,0.35454107);
  rgb(106pt)=(0.65824241,0.09552918,0.3536529);
  rgb(107pt)=(0.66492652,0.09428017,0.3526697);
  rgb(108pt)=(0.67159578,0.09306598,0.35159077);
  rgb(109pt)=(0.67824099,0.09192342,0.3504148);
  rgb(110pt)=(0.684863,0.09085633,0.34914061);
  rgb(111pt)=(0.69146268,0.0898675,0.34776864);
  rgb(112pt)=(0.69803757,0.08897226,0.3462986);
  rgb(113pt)=(0.70457834,0.0882129,0.34473046);
  rgb(114pt)=(0.71108138,0.08761223,0.3430635);
  rgb(115pt)=(0.7175507,0.08716212,0.34129974);
  rgb(116pt)=(0.72398193,0.08688725,0.33943958);
  rgb(117pt)=(0.73035829,0.0868623,0.33748452);
  rgb(118pt)=(0.73669146,0.08704683,0.33543669);
  rgb(119pt)=(0.74297501,0.08747196,0.33329799);
  rgb(120pt)=(0.74919318,0.08820542,0.33107204);
  rgb(121pt)=(0.75535825,0.08919792,0.32876184);
  rgb(122pt)=(0.76145589,0.09050716,0.32637117);
  rgb(123pt)=(0.76748424,0.09213602,0.32390525);
  rgb(124pt)=(0.77344838,0.09405684,0.32136808);
  rgb(125pt)=(0.77932641,0.09634794,0.31876642);
  rgb(126pt)=(0.78513609,0.09892473,0.31610488);
  rgb(127pt)=(0.79085854,0.10184672,0.313391);
  rgb(128pt)=(0.7965014,0.10506637,0.31063031);
  rgb(129pt)=(0.80205987,0.10858333,0.30783);
  rgb(130pt)=(0.80752799,0.11239964,0.30499738);
  rgb(131pt)=(0.81291606,0.11645784,0.30213802);
  rgb(132pt)=(0.81820481,0.12080606,0.29926105);
  rgb(133pt)=(0.82341472,0.12535343,0.2963705);
  rgb(134pt)=(0.82852822,0.13014118,0.29347474);
  rgb(135pt)=(0.83355779,0.13511035,0.29057852);
  rgb(136pt)=(0.83850183,0.14025098,0.2876878);
  rgb(137pt)=(0.84335441,0.14556683,0.28480819);
  rgb(138pt)=(0.84813096,0.15099892,0.281943);
  rgb(139pt)=(0.85281737,0.15657772,0.27909826);
  rgb(140pt)=(0.85742602,0.1622583,0.27627462);
  rgb(141pt)=(0.86196552,0.16801239,0.27346473);
  rgb(142pt)=(0.86641628,0.17387796,0.27070818);
  rgb(143pt)=(0.87079129,0.17982114,0.26797378);
  rgb(144pt)=(0.87507281,0.18587368,0.26529697);
  rgb(145pt)=(0.87925878,0.19203259,0.26268136);
  rgb(146pt)=(0.8833417,0.19830556,0.26014181);
  rgb(147pt)=(0.88731387,0.20469941,0.25769539);
  rgb(148pt)=(0.89116859,0.21121788,0.2553592);
  rgb(149pt)=(0.89490337,0.21785614,0.25314362);
  rgb(150pt)=(0.8985026,0.22463251,0.25108745);
  rgb(151pt)=(0.90197527,0.23152063,0.24918223);
  rgb(152pt)=(0.90530097,0.23854541,0.24748098);
  rgb(153pt)=(0.90848638,0.24568473,0.24598324);
  rgb(154pt)=(0.911533,0.25292623,0.24470258);
  rgb(155pt)=(0.9144225,0.26028902,0.24369359);
  rgb(156pt)=(0.91717106,0.26773821,0.24294137);
  rgb(157pt)=(0.91978131,0.27526191,0.24245973);
  rgb(158pt)=(0.92223947,0.28287251,0.24229568);
  rgb(159pt)=(0.92456587,0.29053388,0.24242622);
  rgb(160pt)=(0.92676657,0.29823282,0.24285536);
  rgb(161pt)=(0.92882964,0.30598085,0.24362274);
  rgb(162pt)=(0.93078135,0.31373977,0.24468803);
  rgb(163pt)=(0.93262051,0.3215093,0.24606461);
  rgb(164pt)=(0.93435067,0.32928362,0.24775328);
  rgb(165pt)=(0.93599076,0.33703942,0.24972157);
  rgb(166pt)=(0.93752831,0.34479177,0.25199928);
  rgb(167pt)=(0.93899289,0.35250734,0.25452808);
  rgb(168pt)=(0.94036561,0.36020899,0.25734661);
  rgb(169pt)=(0.94167588,0.36786594,0.2603949);
  rgb(170pt)=(0.94291042,0.37549479,0.26369821);
  rgb(171pt)=(0.94408513,0.3830811,0.26722004);
  rgb(172pt)=(0.94520419,0.39062329,0.27094924);
  rgb(173pt)=(0.94625977,0.39813168,0.27489742);
  rgb(174pt)=(0.94727016,0.4055909,0.27902322);
  rgb(175pt)=(0.94823505,0.41300424,0.28332283);
  rgb(176pt)=(0.94914549,0.42038251,0.28780969);
  rgb(177pt)=(0.95001704,0.42771398,0.29244728);
  rgb(178pt)=(0.95085121,0.43500005,0.29722817);
  rgb(179pt)=(0.95165009,0.44224144,0.30214494);
  rgb(180pt)=(0.9524044,0.44944853,0.3072105);
  rgb(181pt)=(0.95312556,0.45661389,0.31239776);
  rgb(182pt)=(0.95381595,0.46373781,0.31769923);
  rgb(183pt)=(0.95447591,0.47082238,0.32310953);
  rgb(184pt)=(0.95510255,0.47787236,0.32862553);
  rgb(185pt)=(0.95569679,0.48489115,0.33421404);
  rgb(186pt)=(0.95626788,0.49187351,0.33985601);
  rgb(187pt)=(0.95681685,0.49882008,0.34555431);
  rgb(188pt)=(0.9573439,0.50573243,0.35130912);
  rgb(189pt)=(0.95784842,0.51261283,0.35711942);
  rgb(190pt)=(0.95833051,0.51946267,0.36298589);
  rgb(191pt)=(0.95879054,0.52628305,0.36890904);
  rgb(192pt)=(0.95922872,0.53307513,0.3748895);
  rgb(193pt)=(0.95964538,0.53983991,0.38092784);
  rgb(194pt)=(0.96004345,0.54657593,0.3870292);
  rgb(195pt)=(0.96042097,0.55328624,0.39319057);
  rgb(196pt)=(0.96077819,0.55997184,0.39941173);
  rgb(197pt)=(0.9611152,0.5666337,0.40569343);
  rgb(198pt)=(0.96143273,0.57327231,0.41203603);
  rgb(199pt)=(0.96173392,0.57988594,0.41844491);
  rgb(200pt)=(0.96201757,0.58647675,0.42491751);
  rgb(201pt)=(0.96228344,0.59304598,0.43145271);
  rgb(202pt)=(0.96253168,0.5995944,0.43805131);
  rgb(203pt)=(0.96276513,0.60612062,0.44471698);
  rgb(204pt)=(0.96298491,0.6126247,0.45145074);
  rgb(205pt)=(0.96318967,0.61910879,0.45824902);
  rgb(206pt)=(0.96337949,0.6255736,0.46511271);
  rgb(207pt)=(0.96355923,0.63201624,0.47204746);
  rgb(208pt)=(0.96372785,0.63843852,0.47905028);
  rgb(209pt)=(0.96388426,0.64484214,0.4861196);
  rgb(210pt)=(0.96403203,0.65122535,0.4932578);
  rgb(211pt)=(0.96417332,0.65758729,0.50046894);
  rgb(212pt)=(0.9643063,0.66393045,0.5077467);
  rgb(213pt)=(0.96443322,0.67025402,0.51509334);
  rgb(214pt)=(0.96455845,0.67655564,0.52251447);
  rgb(215pt)=(0.96467922,0.68283846,0.53000231);
  rgb(216pt)=(0.96479861,0.68910113,0.53756026);
  rgb(217pt)=(0.96492035,0.69534192,0.5451917);
  rgb(218pt)=(0.96504223,0.7015636,0.5528892);
  rgb(219pt)=(0.96516917,0.70776351,0.5606593);
  rgb(220pt)=(0.96530224,0.71394212,0.56849894);
  rgb(221pt)=(0.96544032,0.72010124,0.57640375);
  rgb(222pt)=(0.96559206,0.72623592,0.58438387);
  rgb(223pt)=(0.96575293,0.73235058,0.59242739);
  rgb(224pt)=(0.96592829,0.73844258,0.60053991);
  rgb(225pt)=(0.96612013,0.74451182,0.60871954);
  rgb(226pt)=(0.96632832,0.75055966,0.61696136);
  rgb(227pt)=(0.96656022,0.75658231,0.62527295);
  rgb(228pt)=(0.96681185,0.76258381,0.63364277);
  rgb(229pt)=(0.96709183,0.76855969,0.64207921);
  rgb(230pt)=(0.96739773,0.77451297,0.65057302);
  rgb(231pt)=(0.96773482,0.78044149,0.65912731);
  rgb(232pt)=(0.96810471,0.78634563,0.66773889);
  rgb(233pt)=(0.96850919,0.79222565,0.6764046);
  rgb(234pt)=(0.96893132,0.79809112,0.68512266);
  rgb(235pt)=(0.96935926,0.80395415,0.69383201);
  rgb(236pt)=(0.9698028,0.80981139,0.70252255);
  rgb(237pt)=(0.97025511,0.81566605,0.71120296);
  rgb(238pt)=(0.97071849,0.82151775,0.71987163);
  rgb(239pt)=(0.97120159,0.82736371,0.72851999);
  rgb(240pt)=(0.97169389,0.83320847,0.73716071);
  rgb(241pt)=(0.97220061,0.83905052,0.74578903);
  rgb(242pt)=(0.97272597,0.84488881,0.75440141);
  rgb(243pt)=(0.97327085,0.85072354,0.76299805);
  rgb(244pt)=(0.97383206,0.85655639,0.77158353);
  rgb(245pt)=(0.97441222,0.86238689,0.78015619);
  rgb(246pt)=(0.97501782,0.86821321,0.78871034);
  rgb(247pt)=(0.97564391,0.87403763,0.79725261);
  rgb(248pt)=(0.97628674,0.87986189,0.8057883);
  rgb(249pt)=(0.97696114,0.88568129,0.81430324);
  rgb(250pt)=(0.97765722,0.89149971,0.82280948);
  rgb(251pt)=(0.97837585,0.89731727,0.83130786);
  rgb(252pt)=(0.97912374,0.90313207,0.83979337);
  rgb(253pt)=(0.979891,0.90894778,0.84827858);
  rgb(254pt)=(0.98067764,0.91476465,0.85676611);
  rgb(255pt)=(0.98137749,0.92061729,0.86536915)
}]
table{%
x                      y
1 0.0214953962417747
1 0.0112154998920826
1 0.1318650042691
1 0.260804797898187
1 0.732881343983595
1 0.0126936845383038
1 0.0486118252342602
1 0.0124263330213404
1 0.0306663760884549
1 0.0249167347649657
1 0.249465625491663
1 0.0536139298378364
2 0.0105886097632954
2 0.22040415442993
2 0.074793707883967
2 0.0501775477340773
2 0.152006216968899
2 0.147561781355762
2 0.496048594531573
2 0.153367408270154
2 0.0453977737287415
2 0.0312269243472509
2 0.167011795409779
2 0.0285511286945386
3 0.19776051850418
3 0.213274710764733
3 0.129115813014646
3 0.144880392589614
3 0.218616615825017
3 0.0489693892483163
3 0.0497900443820512
3 0.197340741992746
3 0.0976948100283004
3 0.192432541155292
3 0.0716218276849113
3 0.0837775175392058
4 0.160887680165732
4 0.0742496147513814
4 0.0612274251045986
4 0.109115789905416
4 0.0709794405691706
4 0.034134368962705
4 0.0772085842585394
4 0.107393517413349
4 0.136615214020675
4 0.00574679133405293
4 0.0888165842888186
4 0.130968537521943
5 0.047331197158318
5 0.046712344598364
5 0.043032556619789
5 0.0620779792758009
5 0.0566602186207151
5 0.0539425612101519
5 0.0529076243764716
5 0.0518775030611705
5 0.0480149139011994
5 0.0557458561095466
5 0.0540944610910952
5 0.0470060735677239
6 0.0569444867245281
6 0.047216538933284
6 0.0581046728042449
6 0.047661560414517
6 0.0639191427734687
6 0.0489326091989619
6 0.0454161071045805
6 0.0458111028127595
6 0.0489897395910966
6 0.061121938616364
6 0.097200959587128
6 0.0462698978105524
7 0.0505955509664003
7 0.0455462258818912
7 0.0519279883388193
7 0.0434351928010689
7 0.049121749369049
7 0.0496663671949665
7 0.0544730419216296
7 0.0592498147841158
7 0.0450871428104531
7 0.112328379921007
7 0.0478760962574715
7 0.0850060857112348
8 0.0618013969370007
8 0.0459332894364736
8 0.0608733871836926
8 0.0407748517408171
8 0.101855949232084
8 0.0500239243839689
8 0.0445469835197019
8 0.0500715620098249
8 0.0543161854761674
8 0.0517289957117042
8 0.0497256801668446
8 0.0521349749034867
9 0.0455174109934711
9 0.0381877587445659
9 0.0583239322005549
9 0.0519363749656129
9 0.0485053237967922
9 0.0513609705495553
9 0.0534771208288975
9 0.041245687043967
9 0.0503809008224058
9 0.0578731044614312
9 0.0472817230943145
9 0.0471632385531613
10 0.0458038323050724
10 0.0334388157071302
10 0.0166729366764259
10 0.0307164823240577
10 0.0253814996744803
10 0.0074215823986217
10 0.043989825315704
10 0.0510319234041574
10 0.00703792834666538
10 0.0483050371291146
10 0.0114470319482379
10 0.00279537814409716
11 0.0440959195261859
11 0.0202648316235281
11 0.0588625655305336
11 0.0380890266600281
11 0.0496382023279547
11 0.0211642077334076
11 0.0601667423470938
11 0.039072641641225
11 0.0650914866070501
11 0.0611650136556086
11 0.0574264336823797
11 0.0327302785039314
12 0.015019545577579
12 0.0185864997038005
12 0.0321312337645828
12 0.0328465113715861
12 0.0290337538066407
12 0.0495693906344557
12 0.0256478045256876
12 0.0178039989285891
12 0.0285890205479459
12 0.027878036944122
12 0.0210639986185015
12 0.0117171817591182
};
\addlegendentry{[SEP] $\rightarrow$ [SEP]}
\addplot [semithick, color0, forget plot]
table {%
1 0.0188571207977348
2 0.051192594984532
3 0.0807815885015572
4 0.0507047871459131
5 0.0274373144143891
6 0.0352831236532767
7 0.0327251220193621
8 0.0330658277843617
9 0.0322783379621667
10 0.0331064111140561
11 0.0456272669861497
12 0.0185126539091325
};
\addplot [semithick, color1, forget plot]
table {%
1 0.13255471260513
2 0.131427970259831
3 0.137106243560751
4 0.0881119623580318
5 0.0516169407991955
6 0.0556323963642905
7 0.0578594696631756
8 0.0553155983918139
9 0.0492711288378941
10 0.027003522781147
11 0.0456472791532439
12 0.0258239146818841
};
\end{axis}

\end{tikzpicture}

%% file: identifiability/avg_attn2_org.tex
% This file was created by tikzplotlib v0.8.2.
\begin{tikzpicture}

\definecolor{color0}{rgb}{0.203921568627451,0.596078431372549,0.858823529411765}
\definecolor{color1}{rgb}{0.349019607843137,0.850980392156863,0.556862745098039}

\begin{axis}[
axis line style={white!15.0!black},
legend cell align={left},
legend style={at={(0.5,0.09)}, anchor=south, draw=white!80.0!black},
tick align=outside,
tick pos=both,
width=\linewidth,
height=\effectiveAttnPlotHeight,
x grid style={white!80.0!black},
xlabel={Layer},
xmajorgrids,
xmin=0.367277355720007, xmax=12.63272264428,
xtick style={color=white!15.0!black},
y grid style={white!80.0!black},
ylabel={Raw Attention},
ymajorgrids,
ymin=-0.050135019036875, ymax=1.0463631746271,
ytick style={color=white!15.0!black}
]
\addplot [only marks, draw=color0, fill=color0, colormap={mymap}{[1pt]
 rgb(0pt)=(0.01060815,0.01808215,0.10018654);
  rgb(1pt)=(0.01428972,0.02048237,0.10374486);
  rgb(2pt)=(0.01831941,0.0229766,0.10738511);
  rgb(3pt)=(0.02275049,0.02554464,0.11108639);
  rgb(4pt)=(0.02759119,0.02818316,0.11483751);
  rgb(5pt)=(0.03285175,0.03088792,0.11863035);
  rgb(6pt)=(0.03853466,0.03365771,0.12245873);
  rgb(7pt)=(0.04447016,0.03648425,0.12631831);
  rgb(8pt)=(0.05032105,0.03936808,0.13020508);
  rgb(9pt)=(0.05611171,0.04224835,0.13411624);
  rgb(10pt)=(0.0618531,0.04504866,0.13804929);
  rgb(11pt)=(0.06755457,0.04778179,0.14200206);
  rgb(12pt)=(0.0732236,0.05045047,0.14597263);
  rgb(13pt)=(0.0788708,0.05305461,0.14995981);
  rgb(14pt)=(0.08450105,0.05559631,0.15396203);
  rgb(15pt)=(0.09011319,0.05808059,0.15797687);
  rgb(16pt)=(0.09572396,0.06050127,0.16200507);
  rgb(17pt)=(0.10132312,0.06286782,0.16604287);
  rgb(18pt)=(0.10692823,0.06517224,0.17009175);
  rgb(19pt)=(0.1125315,0.06742194,0.17414848);
  rgb(20pt)=(0.11813947,0.06961499,0.17821272);
  rgb(21pt)=(0.12375803,0.07174938,0.18228425);
  rgb(22pt)=(0.12938228,0.07383015,0.18636053);
  rgb(23pt)=(0.13501631,0.07585609,0.19044109);
  rgb(24pt)=(0.14066867,0.0778224,0.19452676);
  rgb(25pt)=(0.14633406,0.07973393,0.1986151);
  rgb(26pt)=(0.15201338,0.08159108,0.20270523);
  rgb(27pt)=(0.15770877,0.08339312,0.20679668);
  rgb(28pt)=(0.16342174,0.0851396,0.21088893);
  rgb(29pt)=(0.16915387,0.08682996,0.21498104);
  rgb(30pt)=(0.17489524,0.08848235,0.2190294);
  rgb(31pt)=(0.18065495,0.09009031,0.22303512);
  rgb(32pt)=(0.18643324,0.09165431,0.22699705);
  rgb(33pt)=(0.19223028,0.09317479,0.23091409);
  rgb(34pt)=(0.19804623,0.09465217,0.23478512);
  rgb(35pt)=(0.20388117,0.09608689,0.23860907);
  rgb(36pt)=(0.20973515,0.09747934,0.24238489);
  rgb(37pt)=(0.21560818,0.09882993,0.24611154);
  rgb(38pt)=(0.22150014,0.10013944,0.2497868);
  rgb(39pt)=(0.22741085,0.10140876,0.25340813);
  rgb(40pt)=(0.23334047,0.10263737,0.25697736);
  rgb(41pt)=(0.23928891,0.10382562,0.2604936);
  rgb(42pt)=(0.24525608,0.10497384,0.26395596);
  rgb(43pt)=(0.25124182,0.10608236,0.26736359);
  rgb(44pt)=(0.25724602,0.10715148,0.27071569);
  rgb(45pt)=(0.26326851,0.1081815,0.27401148);
  rgb(46pt)=(0.26930915,0.1091727,0.2772502);
  rgb(47pt)=(0.27536766,0.11012568,0.28043021);
  rgb(48pt)=(0.28144375,0.11104133,0.2835489);
  rgb(49pt)=(0.2875374,0.11191896,0.28660853);
  rgb(50pt)=(0.29364846,0.11275876,0.2896085);
  rgb(51pt)=(0.29977678,0.11356089,0.29254823);
  rgb(52pt)=(0.30592213,0.11432553,0.29542718);
  rgb(53pt)=(0.31208435,0.11505284,0.29824485);
  rgb(54pt)=(0.31826327,0.1157429,0.30100076);
  rgb(55pt)=(0.32445869,0.11639585,0.30369448);
  rgb(56pt)=(0.33067031,0.11701189,0.30632563);
  rgb(57pt)=(0.33689808,0.11759095,0.3088938);
  rgb(58pt)=(0.34314168,0.11813362,0.31139721);
  rgb(59pt)=(0.34940101,0.11863987,0.3138355);
  rgb(60pt)=(0.355676,0.11910909,0.31620996);
  rgb(61pt)=(0.36196644,0.1195413,0.31852037);
  rgb(62pt)=(0.36827206,0.11993653,0.32076656);
  rgb(63pt)=(0.37459292,0.12029443,0.32294825);
  rgb(64pt)=(0.38092887,0.12061482,0.32506528);
  rgb(65pt)=(0.38727975,0.12089756,0.3271175);
  rgb(66pt)=(0.39364518,0.12114272,0.32910494);
  rgb(67pt)=(0.40002537,0.12134964,0.33102734);
  rgb(68pt)=(0.40642019,0.12151801,0.33288464);
  rgb(69pt)=(0.41282936,0.12164769,0.33467689);
  rgb(70pt)=(0.41925278,0.12173833,0.33640407);
  rgb(71pt)=(0.42569057,0.12178916,0.33806605);
  rgb(72pt)=(0.43214263,0.12179973,0.33966284);
  rgb(73pt)=(0.43860848,0.12177004,0.34119475);
  rgb(74pt)=(0.44508855,0.12169883,0.34266151);
  rgb(75pt)=(0.45158266,0.12158557,0.34406324);
  rgb(76pt)=(0.45809049,0.12142996,0.34540024);
  rgb(77pt)=(0.46461238,0.12123063,0.34667231);
  rgb(78pt)=(0.47114798,0.12098721,0.34787978);
  rgb(79pt)=(0.47769736,0.12069864,0.34902273);
  rgb(80pt)=(0.48426077,0.12036349,0.35010104);
  rgb(81pt)=(0.49083761,0.11998161,0.35111537);
  rgb(82pt)=(0.49742847,0.11955087,0.35206533);
  rgb(83pt)=(0.50403286,0.11907081,0.35295152);
  rgb(84pt)=(0.51065109,0.11853959,0.35377385);
  rgb(85pt)=(0.51728314,0.1179558,0.35453252);
  rgb(86pt)=(0.52392883,0.11731817,0.35522789);
  rgb(87pt)=(0.53058853,0.11662445,0.35585982);
  rgb(88pt)=(0.53726173,0.11587369,0.35642903);
  rgb(89pt)=(0.54394898,0.11506307,0.35693521);
  rgb(90pt)=(0.5506426,0.11420757,0.35737863);
  rgb(91pt)=(0.55734473,0.11330456,0.35775059);
  rgb(92pt)=(0.56405586,0.11235265,0.35804813);
  rgb(93pt)=(0.57077365,0.11135597,0.35827146);
  rgb(94pt)=(0.5774991,0.11031233,0.35841679);
  rgb(95pt)=(0.58422945,0.10922707,0.35848469);
  rgb(96pt)=(0.59096382,0.10810205,0.35847347);
  rgb(97pt)=(0.59770215,0.10693774,0.35838029);
  rgb(98pt)=(0.60444226,0.10573912,0.35820487);
  rgb(99pt)=(0.61118304,0.10450943,0.35794557);
  rgb(100pt)=(0.61792306,0.10325288,0.35760108);
  rgb(101pt)=(0.62466162,0.10197244,0.35716891);
  rgb(102pt)=(0.63139686,0.10067417,0.35664819);
  rgb(103pt)=(0.63812122,0.09938212,0.35603757);
  rgb(104pt)=(0.64483795,0.0980891,0.35533555);
  rgb(105pt)=(0.65154562,0.09680192,0.35454107);
  rgb(106pt)=(0.65824241,0.09552918,0.3536529);
  rgb(107pt)=(0.66492652,0.09428017,0.3526697);
  rgb(108pt)=(0.67159578,0.09306598,0.35159077);
  rgb(109pt)=(0.67824099,0.09192342,0.3504148);
  rgb(110pt)=(0.684863,0.09085633,0.34914061);
  rgb(111pt)=(0.69146268,0.0898675,0.34776864);
  rgb(112pt)=(0.69803757,0.08897226,0.3462986);
  rgb(113pt)=(0.70457834,0.0882129,0.34473046);
  rgb(114pt)=(0.71108138,0.08761223,0.3430635);
  rgb(115pt)=(0.7175507,0.08716212,0.34129974);
  rgb(116pt)=(0.72398193,0.08688725,0.33943958);
  rgb(117pt)=(0.73035829,0.0868623,0.33748452);
  rgb(118pt)=(0.73669146,0.08704683,0.33543669);
  rgb(119pt)=(0.74297501,0.08747196,0.33329799);
  rgb(120pt)=(0.74919318,0.08820542,0.33107204);
  rgb(121pt)=(0.75535825,0.08919792,0.32876184);
  rgb(122pt)=(0.76145589,0.09050716,0.32637117);
  rgb(123pt)=(0.76748424,0.09213602,0.32390525);
  rgb(124pt)=(0.77344838,0.09405684,0.32136808);
  rgb(125pt)=(0.77932641,0.09634794,0.31876642);
  rgb(126pt)=(0.78513609,0.09892473,0.31610488);
  rgb(127pt)=(0.79085854,0.10184672,0.313391);
  rgb(128pt)=(0.7965014,0.10506637,0.31063031);
  rgb(129pt)=(0.80205987,0.10858333,0.30783);
  rgb(130pt)=(0.80752799,0.11239964,0.30499738);
  rgb(131pt)=(0.81291606,0.11645784,0.30213802);
  rgb(132pt)=(0.81820481,0.12080606,0.29926105);
  rgb(133pt)=(0.82341472,0.12535343,0.2963705);
  rgb(134pt)=(0.82852822,0.13014118,0.29347474);
  rgb(135pt)=(0.83355779,0.13511035,0.29057852);
  rgb(136pt)=(0.83850183,0.14025098,0.2876878);
  rgb(137pt)=(0.84335441,0.14556683,0.28480819);
  rgb(138pt)=(0.84813096,0.15099892,0.281943);
  rgb(139pt)=(0.85281737,0.15657772,0.27909826);
  rgb(140pt)=(0.85742602,0.1622583,0.27627462);
  rgb(141pt)=(0.86196552,0.16801239,0.27346473);
  rgb(142pt)=(0.86641628,0.17387796,0.27070818);
  rgb(143pt)=(0.87079129,0.17982114,0.26797378);
  rgb(144pt)=(0.87507281,0.18587368,0.26529697);
  rgb(145pt)=(0.87925878,0.19203259,0.26268136);
  rgb(146pt)=(0.8833417,0.19830556,0.26014181);
  rgb(147pt)=(0.88731387,0.20469941,0.25769539);
  rgb(148pt)=(0.89116859,0.21121788,0.2553592);
  rgb(149pt)=(0.89490337,0.21785614,0.25314362);
  rgb(150pt)=(0.8985026,0.22463251,0.25108745);
  rgb(151pt)=(0.90197527,0.23152063,0.24918223);
  rgb(152pt)=(0.90530097,0.23854541,0.24748098);
  rgb(153pt)=(0.90848638,0.24568473,0.24598324);
  rgb(154pt)=(0.911533,0.25292623,0.24470258);
  rgb(155pt)=(0.9144225,0.26028902,0.24369359);
  rgb(156pt)=(0.91717106,0.26773821,0.24294137);
  rgb(157pt)=(0.91978131,0.27526191,0.24245973);
  rgb(158pt)=(0.92223947,0.28287251,0.24229568);
  rgb(159pt)=(0.92456587,0.29053388,0.24242622);
  rgb(160pt)=(0.92676657,0.29823282,0.24285536);
  rgb(161pt)=(0.92882964,0.30598085,0.24362274);
  rgb(162pt)=(0.93078135,0.31373977,0.24468803);
  rgb(163pt)=(0.93262051,0.3215093,0.24606461);
  rgb(164pt)=(0.93435067,0.32928362,0.24775328);
  rgb(165pt)=(0.93599076,0.33703942,0.24972157);
  rgb(166pt)=(0.93752831,0.34479177,0.25199928);
  rgb(167pt)=(0.93899289,0.35250734,0.25452808);
  rgb(168pt)=(0.94036561,0.36020899,0.25734661);
  rgb(169pt)=(0.94167588,0.36786594,0.2603949);
  rgb(170pt)=(0.94291042,0.37549479,0.26369821);
  rgb(171pt)=(0.94408513,0.3830811,0.26722004);
  rgb(172pt)=(0.94520419,0.39062329,0.27094924);
  rgb(173pt)=(0.94625977,0.39813168,0.27489742);
  rgb(174pt)=(0.94727016,0.4055909,0.27902322);
  rgb(175pt)=(0.94823505,0.41300424,0.28332283);
  rgb(176pt)=(0.94914549,0.42038251,0.28780969);
  rgb(177pt)=(0.95001704,0.42771398,0.29244728);
  rgb(178pt)=(0.95085121,0.43500005,0.29722817);
  rgb(179pt)=(0.95165009,0.44224144,0.30214494);
  rgb(180pt)=(0.9524044,0.44944853,0.3072105);
  rgb(181pt)=(0.95312556,0.45661389,0.31239776);
  rgb(182pt)=(0.95381595,0.46373781,0.31769923);
  rgb(183pt)=(0.95447591,0.47082238,0.32310953);
  rgb(184pt)=(0.95510255,0.47787236,0.32862553);
  rgb(185pt)=(0.95569679,0.48489115,0.33421404);
  rgb(186pt)=(0.95626788,0.49187351,0.33985601);
  rgb(187pt)=(0.95681685,0.49882008,0.34555431);
  rgb(188pt)=(0.9573439,0.50573243,0.35130912);
  rgb(189pt)=(0.95784842,0.51261283,0.35711942);
  rgb(190pt)=(0.95833051,0.51946267,0.36298589);
  rgb(191pt)=(0.95879054,0.52628305,0.36890904);
  rgb(192pt)=(0.95922872,0.53307513,0.3748895);
  rgb(193pt)=(0.95964538,0.53983991,0.38092784);
  rgb(194pt)=(0.96004345,0.54657593,0.3870292);
  rgb(195pt)=(0.96042097,0.55328624,0.39319057);
  rgb(196pt)=(0.96077819,0.55997184,0.39941173);
  rgb(197pt)=(0.9611152,0.5666337,0.40569343);
  rgb(198pt)=(0.96143273,0.57327231,0.41203603);
  rgb(199pt)=(0.96173392,0.57988594,0.41844491);
  rgb(200pt)=(0.96201757,0.58647675,0.42491751);
  rgb(201pt)=(0.96228344,0.59304598,0.43145271);
  rgb(202pt)=(0.96253168,0.5995944,0.43805131);
  rgb(203pt)=(0.96276513,0.60612062,0.44471698);
  rgb(204pt)=(0.96298491,0.6126247,0.45145074);
  rgb(205pt)=(0.96318967,0.61910879,0.45824902);
  rgb(206pt)=(0.96337949,0.6255736,0.46511271);
  rgb(207pt)=(0.96355923,0.63201624,0.47204746);
  rgb(208pt)=(0.96372785,0.63843852,0.47905028);
  rgb(209pt)=(0.96388426,0.64484214,0.4861196);
  rgb(210pt)=(0.96403203,0.65122535,0.4932578);
  rgb(211pt)=(0.96417332,0.65758729,0.50046894);
  rgb(212pt)=(0.9643063,0.66393045,0.5077467);
  rgb(213pt)=(0.96443322,0.67025402,0.51509334);
  rgb(214pt)=(0.96455845,0.67655564,0.52251447);
  rgb(215pt)=(0.96467922,0.68283846,0.53000231);
  rgb(216pt)=(0.96479861,0.68910113,0.53756026);
  rgb(217pt)=(0.96492035,0.69534192,0.5451917);
  rgb(218pt)=(0.96504223,0.7015636,0.5528892);
  rgb(219pt)=(0.96516917,0.70776351,0.5606593);
  rgb(220pt)=(0.96530224,0.71394212,0.56849894);
  rgb(221pt)=(0.96544032,0.72010124,0.57640375);
  rgb(222pt)=(0.96559206,0.72623592,0.58438387);
  rgb(223pt)=(0.96575293,0.73235058,0.59242739);
  rgb(224pt)=(0.96592829,0.73844258,0.60053991);
  rgb(225pt)=(0.96612013,0.74451182,0.60871954);
  rgb(226pt)=(0.96632832,0.75055966,0.61696136);
  rgb(227pt)=(0.96656022,0.75658231,0.62527295);
  rgb(228pt)=(0.96681185,0.76258381,0.63364277);
  rgb(229pt)=(0.96709183,0.76855969,0.64207921);
  rgb(230pt)=(0.96739773,0.77451297,0.65057302);
  rgb(231pt)=(0.96773482,0.78044149,0.65912731);
  rgb(232pt)=(0.96810471,0.78634563,0.66773889);
  rgb(233pt)=(0.96850919,0.79222565,0.6764046);
  rgb(234pt)=(0.96893132,0.79809112,0.68512266);
  rgb(235pt)=(0.96935926,0.80395415,0.69383201);
  rgb(236pt)=(0.9698028,0.80981139,0.70252255);
  rgb(237pt)=(0.97025511,0.81566605,0.71120296);
  rgb(238pt)=(0.97071849,0.82151775,0.71987163);
  rgb(239pt)=(0.97120159,0.82736371,0.72851999);
  rgb(240pt)=(0.97169389,0.83320847,0.73716071);
  rgb(241pt)=(0.97220061,0.83905052,0.74578903);
  rgb(242pt)=(0.97272597,0.84488881,0.75440141);
  rgb(243pt)=(0.97327085,0.85072354,0.76299805);
  rgb(244pt)=(0.97383206,0.85655639,0.77158353);
  rgb(245pt)=(0.97441222,0.86238689,0.78015619);
  rgb(246pt)=(0.97501782,0.86821321,0.78871034);
  rgb(247pt)=(0.97564391,0.87403763,0.79725261);
  rgb(248pt)=(0.97628674,0.87986189,0.8057883);
  rgb(249pt)=(0.97696114,0.88568129,0.81430324);
  rgb(250pt)=(0.97765722,0.89149971,0.82280948);
  rgb(251pt)=(0.97837585,0.89731727,0.83130786);
  rgb(252pt)=(0.97912374,0.90313207,0.83979337);
  rgb(253pt)=(0.979891,0.90894778,0.84827858);
  rgb(254pt)=(0.98067764,0.91476465,0.85676611);
  rgb(255pt)=(0.98137749,0.92061729,0.86536915)
}]
table{%
x                      y
1 0.0117256572848317
1 0.00484731242928913
1 0.0214061659434444
1 0.0176980015353428
1 0.0389948209733882
1 0.0142798936462113
1 0.0163623201924089
1 0.0366806476596599
1 0.0150092308501674
1 0.0262225456426549
1 0.0196959953344804
1 0.015765392395867
2 0.0720224321686427
2 0.032232009878906
2 0.0117836589068514
2 0.0324705576352719
2 0.0151239965902505
2 0.0800881351516791
2 0.0522310155065477
2 0.0389958690053118
2 0.0656946079199793
2 0.0739687532962136
2 0.0371072366687383
2 0.0389439126101669
3 0.0327753410009603
3 0.129660865892371
3 0.101473008289126
3 0.0975077587592934
3 0.0641458981526171
3 0.0590537616543223
3 0.110768769216026
3 0.0602161190483435
3 0.0823043912721188
3 0.0311662459983359
3 0.0596030995905887
3 0.0619091565489956
4 0.28984104481721
4 0.167221962320939
4 0.208832372493364
4 0.216842778705326
4 0.304639908360357
4 0.0251287354772188
4 0.196717264310171
4 0.190469153330218
4 0.224936250968422
4 0.188402735638324
4 0.227431111206756
4 0.198537897374091
5 0.471783526278766
5 0.341699421226532
5 0.564313291600682
5 0.676474745595693
5 0.285024841483532
5 0.375643724307555
5 0.454103787152042
5 0.613823632245963
5 0.50895018462799
5 0.381900558636936
5 0.391633249506113
5 0.570533054982408
6 0.483626068835278
6 0.79393159013623
6 0.4933414552883
6 0.525573399340864
6 0.381572876031132
6 0.71256424351389
6 0.74769509336374
6 0.668909801731732
6 0.50580572628334
6 0.208178373454721
6 0.327127104761604
6 0.654592211356251
7 0.660754941815567
7 0.591164403369842
7 0.547549726087675
7 0.760778727670955
7 0.563768359506092
7 0.526066273432862
7 0.528069526226412
7 0.432979311267238
7 0.497778096384557
7 0.266792998296797
7 0.63798152867829
7 0.27227017381877
8 0.26338574172722
8 0.568318080564988
8 0.527190469384527
8 0.869481616884592
8 0.220645449189857
8 0.470677671553959
8 0.880584064649293
8 0.869857475930387
8 0.331899662600925
8 0.498059297405441
8 0.443327778138827
8 0.787358040817046
9 0.715706397477486
9 0.445089151989561
9 0.721837645319686
9 0.563084581355534
9 0.681780833622972
9 0.751063117249977
9 0.763343794079203
9 0.421225068831546
9 0.348947499018422
9 0.400514879905647
9 0.490290951051885
9 0.442163116052525
10 0.474868510403724
10 0.529828296163292
10 0.362651106540343
10 0.56594919400145
10 0.636157136807165
10 0.348936321284308
10 0.528482507671754
10 0.55200450204471
10 0.251257565982945
10 0.695745670885785
10 0.461954325496674
10 0.405317742032309
11 0.0653484261606403
11 0.0305122814585503
11 0.0478064455595864
11 0.0791449081811471
11 0.0391759809468478
11 0.0352209617086739
11 0.0329996734800697
11 0.0579936137567447
11 0.0564165057618885
11 0.0367334471542383
11 0.256957165103639
11 0.0346582216544619
12 0.079733779301463
12 0.0760876159115789
12 0.0605278552239381
12 0.0838006098438021
12 0.0696574121827838
12 0.0587019078938209
12 0.0802553446344433
12 0.0701124890136294
12 0.0233374933118904
12 0.0649001951936636
12 0.0743143851488472
12 0.0683153962862208
};
\addlegendentry{other -> [SEP]}
\legend{}
\addplot [only marks, draw=color1, fill=color1, colormap={mymap}{[1pt]
 rgb(0pt)=(0.01060815,0.01808215,0.10018654);
  rgb(1pt)=(0.01428972,0.02048237,0.10374486);
  rgb(2pt)=(0.01831941,0.0229766,0.10738511);
  rgb(3pt)=(0.02275049,0.02554464,0.11108639);
  rgb(4pt)=(0.02759119,0.02818316,0.11483751);
  rgb(5pt)=(0.03285175,0.03088792,0.11863035);
  rgb(6pt)=(0.03853466,0.03365771,0.12245873);
  rgb(7pt)=(0.04447016,0.03648425,0.12631831);
  rgb(8pt)=(0.05032105,0.03936808,0.13020508);
  rgb(9pt)=(0.05611171,0.04224835,0.13411624);
  rgb(10pt)=(0.0618531,0.04504866,0.13804929);
  rgb(11pt)=(0.06755457,0.04778179,0.14200206);
  rgb(12pt)=(0.0732236,0.05045047,0.14597263);
  rgb(13pt)=(0.0788708,0.05305461,0.14995981);
  rgb(14pt)=(0.08450105,0.05559631,0.15396203);
  rgb(15pt)=(0.09011319,0.05808059,0.15797687);
  rgb(16pt)=(0.09572396,0.06050127,0.16200507);
  rgb(17pt)=(0.10132312,0.06286782,0.16604287);
  rgb(18pt)=(0.10692823,0.06517224,0.17009175);
  rgb(19pt)=(0.1125315,0.06742194,0.17414848);
  rgb(20pt)=(0.11813947,0.06961499,0.17821272);
  rgb(21pt)=(0.12375803,0.07174938,0.18228425);
  rgb(22pt)=(0.12938228,0.07383015,0.18636053);
  rgb(23pt)=(0.13501631,0.07585609,0.19044109);
  rgb(24pt)=(0.14066867,0.0778224,0.19452676);
  rgb(25pt)=(0.14633406,0.07973393,0.1986151);
  rgb(26pt)=(0.15201338,0.08159108,0.20270523);
  rgb(27pt)=(0.15770877,0.08339312,0.20679668);
  rgb(28pt)=(0.16342174,0.0851396,0.21088893);
  rgb(29pt)=(0.16915387,0.08682996,0.21498104);
  rgb(30pt)=(0.17489524,0.08848235,0.2190294);
  rgb(31pt)=(0.18065495,0.09009031,0.22303512);
  rgb(32pt)=(0.18643324,0.09165431,0.22699705);
  rgb(33pt)=(0.19223028,0.09317479,0.23091409);
  rgb(34pt)=(0.19804623,0.09465217,0.23478512);
  rgb(35pt)=(0.20388117,0.09608689,0.23860907);
  rgb(36pt)=(0.20973515,0.09747934,0.24238489);
  rgb(37pt)=(0.21560818,0.09882993,0.24611154);
  rgb(38pt)=(0.22150014,0.10013944,0.2497868);
  rgb(39pt)=(0.22741085,0.10140876,0.25340813);
  rgb(40pt)=(0.23334047,0.10263737,0.25697736);
  rgb(41pt)=(0.23928891,0.10382562,0.2604936);
  rgb(42pt)=(0.24525608,0.10497384,0.26395596);
  rgb(43pt)=(0.25124182,0.10608236,0.26736359);
  rgb(44pt)=(0.25724602,0.10715148,0.27071569);
  rgb(45pt)=(0.26326851,0.1081815,0.27401148);
  rgb(46pt)=(0.26930915,0.1091727,0.2772502);
  rgb(47pt)=(0.27536766,0.11012568,0.28043021);
  rgb(48pt)=(0.28144375,0.11104133,0.2835489);
  rgb(49pt)=(0.2875374,0.11191896,0.28660853);
  rgb(50pt)=(0.29364846,0.11275876,0.2896085);
  rgb(51pt)=(0.29977678,0.11356089,0.29254823);
  rgb(52pt)=(0.30592213,0.11432553,0.29542718);
  rgb(53pt)=(0.31208435,0.11505284,0.29824485);
  rgb(54pt)=(0.31826327,0.1157429,0.30100076);
  rgb(55pt)=(0.32445869,0.11639585,0.30369448);
  rgb(56pt)=(0.33067031,0.11701189,0.30632563);
  rgb(57pt)=(0.33689808,0.11759095,0.3088938);
  rgb(58pt)=(0.34314168,0.11813362,0.31139721);
  rgb(59pt)=(0.34940101,0.11863987,0.3138355);
  rgb(60pt)=(0.355676,0.11910909,0.31620996);
  rgb(61pt)=(0.36196644,0.1195413,0.31852037);
  rgb(62pt)=(0.36827206,0.11993653,0.32076656);
  rgb(63pt)=(0.37459292,0.12029443,0.32294825);
  rgb(64pt)=(0.38092887,0.12061482,0.32506528);
  rgb(65pt)=(0.38727975,0.12089756,0.3271175);
  rgb(66pt)=(0.39364518,0.12114272,0.32910494);
  rgb(67pt)=(0.40002537,0.12134964,0.33102734);
  rgb(68pt)=(0.40642019,0.12151801,0.33288464);
  rgb(69pt)=(0.41282936,0.12164769,0.33467689);
  rgb(70pt)=(0.41925278,0.12173833,0.33640407);
  rgb(71pt)=(0.42569057,0.12178916,0.33806605);
  rgb(72pt)=(0.43214263,0.12179973,0.33966284);
  rgb(73pt)=(0.43860848,0.12177004,0.34119475);
  rgb(74pt)=(0.44508855,0.12169883,0.34266151);
  rgb(75pt)=(0.45158266,0.12158557,0.34406324);
  rgb(76pt)=(0.45809049,0.12142996,0.34540024);
  rgb(77pt)=(0.46461238,0.12123063,0.34667231);
  rgb(78pt)=(0.47114798,0.12098721,0.34787978);
  rgb(79pt)=(0.47769736,0.12069864,0.34902273);
  rgb(80pt)=(0.48426077,0.12036349,0.35010104);
  rgb(81pt)=(0.49083761,0.11998161,0.35111537);
  rgb(82pt)=(0.49742847,0.11955087,0.35206533);
  rgb(83pt)=(0.50403286,0.11907081,0.35295152);
  rgb(84pt)=(0.51065109,0.11853959,0.35377385);
  rgb(85pt)=(0.51728314,0.1179558,0.35453252);
  rgb(86pt)=(0.52392883,0.11731817,0.35522789);
  rgb(87pt)=(0.53058853,0.11662445,0.35585982);
  rgb(88pt)=(0.53726173,0.11587369,0.35642903);
  rgb(89pt)=(0.54394898,0.11506307,0.35693521);
  rgb(90pt)=(0.5506426,0.11420757,0.35737863);
  rgb(91pt)=(0.55734473,0.11330456,0.35775059);
  rgb(92pt)=(0.56405586,0.11235265,0.35804813);
  rgb(93pt)=(0.57077365,0.11135597,0.35827146);
  rgb(94pt)=(0.5774991,0.11031233,0.35841679);
  rgb(95pt)=(0.58422945,0.10922707,0.35848469);
  rgb(96pt)=(0.59096382,0.10810205,0.35847347);
  rgb(97pt)=(0.59770215,0.10693774,0.35838029);
  rgb(98pt)=(0.60444226,0.10573912,0.35820487);
  rgb(99pt)=(0.61118304,0.10450943,0.35794557);
  rgb(100pt)=(0.61792306,0.10325288,0.35760108);
  rgb(101pt)=(0.62466162,0.10197244,0.35716891);
  rgb(102pt)=(0.63139686,0.10067417,0.35664819);
  rgb(103pt)=(0.63812122,0.09938212,0.35603757);
  rgb(104pt)=(0.64483795,0.0980891,0.35533555);
  rgb(105pt)=(0.65154562,0.09680192,0.35454107);
  rgb(106pt)=(0.65824241,0.09552918,0.3536529);
  rgb(107pt)=(0.66492652,0.09428017,0.3526697);
  rgb(108pt)=(0.67159578,0.09306598,0.35159077);
  rgb(109pt)=(0.67824099,0.09192342,0.3504148);
  rgb(110pt)=(0.684863,0.09085633,0.34914061);
  rgb(111pt)=(0.69146268,0.0898675,0.34776864);
  rgb(112pt)=(0.69803757,0.08897226,0.3462986);
  rgb(113pt)=(0.70457834,0.0882129,0.34473046);
  rgb(114pt)=(0.71108138,0.08761223,0.3430635);
  rgb(115pt)=(0.7175507,0.08716212,0.34129974);
  rgb(116pt)=(0.72398193,0.08688725,0.33943958);
  rgb(117pt)=(0.73035829,0.0868623,0.33748452);
  rgb(118pt)=(0.73669146,0.08704683,0.33543669);
  rgb(119pt)=(0.74297501,0.08747196,0.33329799);
  rgb(120pt)=(0.74919318,0.08820542,0.33107204);
  rgb(121pt)=(0.75535825,0.08919792,0.32876184);
  rgb(122pt)=(0.76145589,0.09050716,0.32637117);
  rgb(123pt)=(0.76748424,0.09213602,0.32390525);
  rgb(124pt)=(0.77344838,0.09405684,0.32136808);
  rgb(125pt)=(0.77932641,0.09634794,0.31876642);
  rgb(126pt)=(0.78513609,0.09892473,0.31610488);
  rgb(127pt)=(0.79085854,0.10184672,0.313391);
  rgb(128pt)=(0.7965014,0.10506637,0.31063031);
  rgb(129pt)=(0.80205987,0.10858333,0.30783);
  rgb(130pt)=(0.80752799,0.11239964,0.30499738);
  rgb(131pt)=(0.81291606,0.11645784,0.30213802);
  rgb(132pt)=(0.81820481,0.12080606,0.29926105);
  rgb(133pt)=(0.82341472,0.12535343,0.2963705);
  rgb(134pt)=(0.82852822,0.13014118,0.29347474);
  rgb(135pt)=(0.83355779,0.13511035,0.29057852);
  rgb(136pt)=(0.83850183,0.14025098,0.2876878);
  rgb(137pt)=(0.84335441,0.14556683,0.28480819);
  rgb(138pt)=(0.84813096,0.15099892,0.281943);
  rgb(139pt)=(0.85281737,0.15657772,0.27909826);
  rgb(140pt)=(0.85742602,0.1622583,0.27627462);
  rgb(141pt)=(0.86196552,0.16801239,0.27346473);
  rgb(142pt)=(0.86641628,0.17387796,0.27070818);
  rgb(143pt)=(0.87079129,0.17982114,0.26797378);
  rgb(144pt)=(0.87507281,0.18587368,0.26529697);
  rgb(145pt)=(0.87925878,0.19203259,0.26268136);
  rgb(146pt)=(0.8833417,0.19830556,0.26014181);
  rgb(147pt)=(0.88731387,0.20469941,0.25769539);
  rgb(148pt)=(0.89116859,0.21121788,0.2553592);
  rgb(149pt)=(0.89490337,0.21785614,0.25314362);
  rgb(150pt)=(0.8985026,0.22463251,0.25108745);
  rgb(151pt)=(0.90197527,0.23152063,0.24918223);
  rgb(152pt)=(0.90530097,0.23854541,0.24748098);
  rgb(153pt)=(0.90848638,0.24568473,0.24598324);
  rgb(154pt)=(0.911533,0.25292623,0.24470258);
  rgb(155pt)=(0.9144225,0.26028902,0.24369359);
  rgb(156pt)=(0.91717106,0.26773821,0.24294137);
  rgb(157pt)=(0.91978131,0.27526191,0.24245973);
  rgb(158pt)=(0.92223947,0.28287251,0.24229568);
  rgb(159pt)=(0.92456587,0.29053388,0.24242622);
  rgb(160pt)=(0.92676657,0.29823282,0.24285536);
  rgb(161pt)=(0.92882964,0.30598085,0.24362274);
  rgb(162pt)=(0.93078135,0.31373977,0.24468803);
  rgb(163pt)=(0.93262051,0.3215093,0.24606461);
  rgb(164pt)=(0.93435067,0.32928362,0.24775328);
  rgb(165pt)=(0.93599076,0.33703942,0.24972157);
  rgb(166pt)=(0.93752831,0.34479177,0.25199928);
  rgb(167pt)=(0.93899289,0.35250734,0.25452808);
  rgb(168pt)=(0.94036561,0.36020899,0.25734661);
  rgb(169pt)=(0.94167588,0.36786594,0.2603949);
  rgb(170pt)=(0.94291042,0.37549479,0.26369821);
  rgb(171pt)=(0.94408513,0.3830811,0.26722004);
  rgb(172pt)=(0.94520419,0.39062329,0.27094924);
  rgb(173pt)=(0.94625977,0.39813168,0.27489742);
  rgb(174pt)=(0.94727016,0.4055909,0.27902322);
  rgb(175pt)=(0.94823505,0.41300424,0.28332283);
  rgb(176pt)=(0.94914549,0.42038251,0.28780969);
  rgb(177pt)=(0.95001704,0.42771398,0.29244728);
  rgb(178pt)=(0.95085121,0.43500005,0.29722817);
  rgb(179pt)=(0.95165009,0.44224144,0.30214494);
  rgb(180pt)=(0.9524044,0.44944853,0.3072105);
  rgb(181pt)=(0.95312556,0.45661389,0.31239776);
  rgb(182pt)=(0.95381595,0.46373781,0.31769923);
  rgb(183pt)=(0.95447591,0.47082238,0.32310953);
  rgb(184pt)=(0.95510255,0.47787236,0.32862553);
  rgb(185pt)=(0.95569679,0.48489115,0.33421404);
  rgb(186pt)=(0.95626788,0.49187351,0.33985601);
  rgb(187pt)=(0.95681685,0.49882008,0.34555431);
  rgb(188pt)=(0.9573439,0.50573243,0.35130912);
  rgb(189pt)=(0.95784842,0.51261283,0.35711942);
  rgb(190pt)=(0.95833051,0.51946267,0.36298589);
  rgb(191pt)=(0.95879054,0.52628305,0.36890904);
  rgb(192pt)=(0.95922872,0.53307513,0.3748895);
  rgb(193pt)=(0.95964538,0.53983991,0.38092784);
  rgb(194pt)=(0.96004345,0.54657593,0.3870292);
  rgb(195pt)=(0.96042097,0.55328624,0.39319057);
  rgb(196pt)=(0.96077819,0.55997184,0.39941173);
  rgb(197pt)=(0.9611152,0.5666337,0.40569343);
  rgb(198pt)=(0.96143273,0.57327231,0.41203603);
  rgb(199pt)=(0.96173392,0.57988594,0.41844491);
  rgb(200pt)=(0.96201757,0.58647675,0.42491751);
  rgb(201pt)=(0.96228344,0.59304598,0.43145271);
  rgb(202pt)=(0.96253168,0.5995944,0.43805131);
  rgb(203pt)=(0.96276513,0.60612062,0.44471698);
  rgb(204pt)=(0.96298491,0.6126247,0.45145074);
  rgb(205pt)=(0.96318967,0.61910879,0.45824902);
  rgb(206pt)=(0.96337949,0.6255736,0.46511271);
  rgb(207pt)=(0.96355923,0.63201624,0.47204746);
  rgb(208pt)=(0.96372785,0.63843852,0.47905028);
  rgb(209pt)=(0.96388426,0.64484214,0.4861196);
  rgb(210pt)=(0.96403203,0.65122535,0.4932578);
  rgb(211pt)=(0.96417332,0.65758729,0.50046894);
  rgb(212pt)=(0.9643063,0.66393045,0.5077467);
  rgb(213pt)=(0.96443322,0.67025402,0.51509334);
  rgb(214pt)=(0.96455845,0.67655564,0.52251447);
  rgb(215pt)=(0.96467922,0.68283846,0.53000231);
  rgb(216pt)=(0.96479861,0.68910113,0.53756026);
  rgb(217pt)=(0.96492035,0.69534192,0.5451917);
  rgb(218pt)=(0.96504223,0.7015636,0.5528892);
  rgb(219pt)=(0.96516917,0.70776351,0.5606593);
  rgb(220pt)=(0.96530224,0.71394212,0.56849894);
  rgb(221pt)=(0.96544032,0.72010124,0.57640375);
  rgb(222pt)=(0.96559206,0.72623592,0.58438387);
  rgb(223pt)=(0.96575293,0.73235058,0.59242739);
  rgb(224pt)=(0.96592829,0.73844258,0.60053991);
  rgb(225pt)=(0.96612013,0.74451182,0.60871954);
  rgb(226pt)=(0.96632832,0.75055966,0.61696136);
  rgb(227pt)=(0.96656022,0.75658231,0.62527295);
  rgb(228pt)=(0.96681185,0.76258381,0.63364277);
  rgb(229pt)=(0.96709183,0.76855969,0.64207921);
  rgb(230pt)=(0.96739773,0.77451297,0.65057302);
  rgb(231pt)=(0.96773482,0.78044149,0.65912731);
  rgb(232pt)=(0.96810471,0.78634563,0.66773889);
  rgb(233pt)=(0.96850919,0.79222565,0.6764046);
  rgb(234pt)=(0.96893132,0.79809112,0.68512266);
  rgb(235pt)=(0.96935926,0.80395415,0.69383201);
  rgb(236pt)=(0.9698028,0.80981139,0.70252255);
  rgb(237pt)=(0.97025511,0.81566605,0.71120296);
  rgb(238pt)=(0.97071849,0.82151775,0.71987163);
  rgb(239pt)=(0.97120159,0.82736371,0.72851999);
  rgb(240pt)=(0.97169389,0.83320847,0.73716071);
  rgb(241pt)=(0.97220061,0.83905052,0.74578903);
  rgb(242pt)=(0.97272597,0.84488881,0.75440141);
  rgb(243pt)=(0.97327085,0.85072354,0.76299805);
  rgb(244pt)=(0.97383206,0.85655639,0.77158353);
  rgb(245pt)=(0.97441222,0.86238689,0.78015619);
  rgb(246pt)=(0.97501782,0.86821321,0.78871034);
  rgb(247pt)=(0.97564391,0.87403763,0.79725261);
  rgb(248pt)=(0.97628674,0.87986189,0.8057883);
  rgb(249pt)=(0.97696114,0.88568129,0.81430324);
  rgb(250pt)=(0.97765722,0.89149971,0.82280948);
  rgb(251pt)=(0.97837585,0.89731727,0.83130786);
  rgb(252pt)=(0.97912374,0.90313207,0.83979337);
  rgb(253pt)=(0.979891,0.90894778,0.84827858);
  rgb(254pt)=(0.98067764,0.91476465,0.85676611);
  rgb(255pt)=(0.98137749,0.92061729,0.86536915)
}]
table{%
x                      y
1 0.0276078626782457
1 0.00279750015562103
1 0.131499650377238
1 0.321010979744904
1 0.836971312478719
1 0.0158612046122119
1 0.0539141364564373
1 0.01586944362404
1 0.0219273303874683
1 0.0138333676096673
1 0.268216506280927
1 0.0871804203190638
2 0.0124767146009558
2 0.2476906039214
2 0.0634802709796679
2 0.0202251957517517
2 0.122388205071057
2 0.166219459671288
2 0.513395830890064
2 0.128068704659522
2 0.0582814171427725
2 0.0392783725826209
2 0.183791766184874
2 0.0285688692623103
3 0.0559734722486391
3 0.149989661915536
3 0.117005062192836
3 0.0826633096946734
3 0.296570116729145
3 0.0457254661684396
3 0.0567055893695779
3 0.0792737081334756
3 0.178436875699141
3 0.0117806838586359
3 0.0742594266616394
3 0.0110418030449035
4 0.603609819370773
4 0.598557173559863
4 0.668151325846632
4 0.735028189817262
4 0.334701575902141
4 0.0331666576309925
4 0.53215500874625
4 0.56800825184872
4 0.329464584972049
4 0.0809086550781954
4 0.782023590508728
4 0.247847473440361
5 0.869407563981029
5 0.582999275208661
5 0.782811932461034
5 0.962382311091548
5 0.993430655434607
5 0.980662791869572
5 0.986136771378018
5 0.957179292901268
5 0.872572779550308
5 0.872271832063674
5 0.958525672645098
5 0.858775038933082
6 0.929296428818376
6 0.977320577738986
6 0.928477483546181
6 0.970175496768207
6 0.861218437702665
6 0.965107315024661
6 0.816464666682746
6 0.985895429076925
6 0.975187730750129
6 0.694728832721949
6 0.87187905421841
6 0.86466071123798
7 0.97953482283159
7 0.943497788551593
7 0.897417181682202
7 0.836986996894403
7 0.970722721919657
7 0.988034752663226
7 0.845309323916631
7 0.940960824279296
7 0.691924000038735
7 0.83368360045396
7 0.88742675774941
7 0.86543827057786
8 0.957634936402281
8 0.954730381107618
8 0.833979047489383
8 0.820015391755488
8 0.950843820648808
8 0.982079818439221
8 0.849783170487611
8 0.880537195942336
8 0.892501071546106
8 0.879644243196855
8 0.938556103711768
8 0.912164659464672
9 0.936382152275332
9 0.696296750492747
9 0.806033983150678
9 0.900084913467929
9 0.936334454589673
9 0.908659765001145
9 0.886334906662664
9 0.857572522419955
9 0.897150705733727
9 0.871888606088056
9 0.957676671835924
9 0.871690289703229
10 0.771666531345897
10 0.543398135965089
10 0.244700527141592
10 0.287806683118575
10 0.394508972523674
10 0.111382561904799
10 0.550156150366209
10 0.855790353785721
10 0.0439025908825927
10 0.515923846039862
10 0.147970951506831
10 0.0301323147075656
11 0.0911112017164405
11 0.0399704668236683
11 0.109452184247446
11 0.0525359587204988
11 0.0490969475571135
11 0.0464316396033433
11 0.0447480132223121
11 0.0549630134795617
11 0.105296741422799
11 0.0419506571093558
11 0.0769267332596487
11 0.0555365375867921
12 0.0831921079393837
12 0.0884098412012948
12 0.109545717750969
12 0.114132206640974
12 0.0967826643387203
12 0.109631632768322
12 0.0839664966161734
12 0.0838086622655342
12 0.10408199560542
12 0.095962242159423
12 0.0908210240188876
12 0.0775078395918245
};
\addlegendentry{[SEP] -> [SEP]}
\legend{}
\addplot [semithick, color0, forget plot]
table {%
1 0.0198906653239788
2 0.0458885154448799
3 0.0742153679519249
4 0.2032501012502
5 0.469657001470351
6 0.541909828674757
7 0.523829505546255
8 0.560898779070588
9 0.562087252996204
10 0.484429406609538
11 0.064413969243874
12 0.0674787069955068
};
\addplot [semithick, color1, forget plot]
table {%
1 0.149724142893712
2 0.131988784226524
3 0.0966187646430535
4 0.459468525560164
5 0.889762993126491
6 0.903367680357268
7 0.890078086796547
8 0.904372486682679
9 0.877175476785088
10 0.374778301607367
11 0.064001674562415
12 0.0948202025747439
};
\end{axis}

\end{tikzpicture}

%% file: token_classification_experiments/PredictTokens_AllLayers_NearestNeighbour_TestVsTrain.tex
% This file was created by tikzplotlib v0.8.2.
\begin{tikzpicture}

\begin{axis}[
legend cell align={left},
legend style={at={(0.09,0.5)}, anchor=west, draw=white!80.0!black},
tick align=outside,
tick pos=both,
x grid style={lightgray!92.02614379084967!black},
xlabel={Layer},
xmajorgrids,
width = 0.9\linewidth,
height = \tokenClassifierPlotHeight,
xmin=0.45, xmax=12.55,
xtick style={color=black},
y grid style={lightgray!92.02614379084967!black},
ylabel={1-NN Recovery Accuracy},
ymajorgrids,
ymin=0.719577257235672, ymax=1.01335346394116,
ytick style={color=black}
]
\addplot [thick, blue, dashed]
table {%
1 1
2 1
3 0.999161224893429
4 0.988471700593891
5 0.984162821488214
6 0.980674307354805
7 0.975394336122821
8 0.971017582887906
9 0.958194725475442
10 0.898643258729427
11 0.862571459526315
12 0.852749981264907
};
\addlegendentry{Linear Train}
\addplot [thick, blue]
table {%
1 1
2 1
3 0.999125258026798
4 0.988475486146777
5 0.981853030065317
6 0.969192708945533
7 0.945448901786302
8 0.923599170471342
9 0.893778659012765
10 0.817093700077433
11 0.769724237587454
12 0.73293072117683
};
\addlegendentry{Linear Test}
\addplot [thick, red, dashed]
table {%
1 1
2 1
3 0.999969772969571
4 0.99859985352983
5 0.991942961330643
6 0.989118104389958
7 0.992894943836259
8 0.99731302526962
9 0.994797904052378
10 0.974054525992238
11 0.937152443795104
12 0.956479354700536
};
\addlegendentry{MLP Train}
\addplot [thick, red]
table {%
1 1
2 1
3 0.999954022988506
4 0.998297522559177
5 0.988753483600016
6 0.978824743402777
7 0.96535017058847
8 0.952245663400465
9 0.930002052525579
10 0.880599672426699
11 0.82392226130497
12 0.819659406779177
};
\addlegendentry{MLP Test}
\end{axis}

\end{tikzpicture}

%% file: token_classification_experiments/rebuttal_plots/GeneralizingAcrossLayers/TokenIdentity_GeneralizeAcrossLayers_LINEAR_L2_Layers1-6-11-12_10FoldCV.tex
% This file was created by tikzplotlib v0.8.2.
\begin{tikzpicture}

\definecolor{color3}{rgb}{0.83921568627451,0.152941176470588,0.156862745098039}
\definecolor{color0}{rgb}{0.12156862745098,0.466666666666667,0.705882352941177}
\definecolor{color2}{rgb}{0.172549019607843,0.627450980392157,0.172549019607843}
\definecolor{color1}{rgb}{1,0.498039215686275,0.0549019607843137}

\begin{axis}[
legend cell align={left},
legend columns = 2,
legend style={at={(0.0,0.0)}, anchor=south west, draw=white!80.0!black},
tick align=outside,
tick pos=both,
x grid style={lightgray!92.02614379084967!black},
xlabel={Layer},
xmajorgrids,
width = 0.9\linewidth,
height = \tokenClassifierPlotHeightAppendix,
xmin=0.45, xmax=12.55,
xtick style={color=black},
y grid style={lightgray!92.02614379084967!black},
ylabel={Identifiability Rate},
ymajorgrids,
ymin=0.1, ymax=1.04002459268368,
ytick style={color=black}
]
\addplot [thick, color0]
table {%
1 1
2 0.999726814348332
3 0.989093844709141
4 0.975700927774565
5 0.957959105296981
6 0.922510134703363
7 0.852650716657109
8 0.659824843949406
9 0.484209097553923
10 0.362254122466812
11 0.209052902386952
12 0.19668822789023
};
\addlegendentry{l=1}
\addplot [thick, color1]
table {%
1 0.983249084858693
2 0.99853309164389
3 0.986219952888185
4 0.982850400783855
5 0.978249071053975
6 0.969408141541138
7 0.919313117411774
8 0.780124005176059
9 0.648464973236009
10 0.59641847602442
11 0.57675383105206
12 0.389117095636526
};
\addlegendentry{l=6}
\addplot [thick, color2]
table {%
1 0.958287529256427
2 0.984423806659397
3 0.965408687918805
4 0.9578804184915
5 0.960336249206783
6 0.950756935985801
7 0.940853031630971
8 0.897850159877526
9 0.837443347778777
10 0.783668869178028
11 0.769246666937309
12 0.491147115231167
};
\addlegendentry{l=11}
\addplot [thick, color3]
table {%
1 0.968196254455928
2 0.974226115480827
3 0.960936808447354
4 0.968928665673379
5 0.96462086899595
6 0.961419660223068
7 0.954781631614552
8 0.94083327650245
9 0.909421262886574
10 0.883029241253765
11 0.865762029765831
12 0.733504362659746
};
\addlegendentry{l=12}
\end{axis}

\end{tikzpicture}

%% file: token_classification_experiments/rebuttal_plots/GeneralizingAcrossLayers/TokenIdentity_GeneralizeAcrossLayers_MLP_L2_Layers1-6-11-12_10FoldCV.tex
% This file was created by tikzplotlib v0.8.2.
\begin{tikzpicture}

\definecolor{color3}{rgb}{0.83921568627451,0.152941176470588,0.156862745098039}
\definecolor{color0}{rgb}{0.12156862745098,0.466666666666667,0.705882352941177}
\definecolor{color2}{rgb}{0.172549019607843,0.627450980392157,0.172549019607843}
\definecolor{color1}{rgb}{1,0.498039215686275,0.0549019607843137}

\begin{axis}[
legend cell align={left},
legend columns = 2,
legend style={at={(0.0,0.0)}, anchor=south west, draw=white!80.0!black},
tick align=outside,
tick pos=both,
x grid style={lightgray!92.02614379084967!black},
xlabel={Layer},
xmajorgrids,
width = 0.9\linewidth,
height = \tokenClassifierPlotHeightAppendix,
xmin=0.45, xmax=12.55,
xtick style={color=black},
y grid style={lightgray!92.02614379084967!black},
ylabel={Identifiability Rate},
ymajorgrids,
ymin=0.1, ymax=1.04002459268368,
ytick style={color=black}
]
\addplot [thick, color0]
table {%
1 1
2 1
3 0.985910148581023
4 0.976396503579664
5 0.953589427343301
6 0.913010835913399
7 0.825907104482371
8 0.633233907287893
9 0.439533614166262
10 0.312296434946768
11 0.215492646657393
12 0.135300457001496
};
\addlegendentry{l=1}
\addplot [thick, color1]
table {%
1 0.980028814775755
2 0.984134997590195
3 0.986953573318726
4 0.986729059384
5 0.982414837650249
6 0.978333735263727
7 0.948912738603396
8 0.865564809378042
9 0.748055978035151
10 0.637029401413874
11 0.577605746347276
12 0.379312981071257
};
\addlegendentry{l=6}
\addplot [thick, color2]
table {%
1 0.891105445348839
2 0.934295374154055
3 0.92632648974526
4 0.938104159586653
5 0.94501250372943
6 0.946512818274902
7 0.941867822542513
8 0.931130605370299
9 0.904318134154977
10 0.84204839807607
11 0.824554749419401
12 0.565664521219615
};
\addlegendentry{l=11}
\addplot [thick, color3]
table {%
1 0.940891346460087
2 0.959542438325255
3 0.943592012192097
4 0.958540474709334
5 0.963826369002099
6 0.959675540291015
7 0.9560020417769
8 0.942280863122082
9 0.924548959171993
10 0.900520935548688
11 0.8882516646736
12 0.822846188668629
};
\addlegendentry{l=12}
\end{axis}

\end{tikzpicture}

%% file: token_classification_experiments/rebuttal_plots/GeneralizingAcrossLayers/TokenIdentity_GeneralizeAcrossLayers_MLP_COSINE_Layers1-6-11-12_10FoldCV.tex
% This file was created by tikzplotlib v0.8.2.
\begin{tikzpicture}

\definecolor{color3}{rgb}{0.83921568627451,0.152941176470588,0.156862745098039}
\definecolor{color0}{rgb}{0.12156862745098,0.466666666666667,0.705882352941177}
\definecolor{color2}{rgb}{0.172549019607843,0.627450980392157,0.172549019607843}
\definecolor{color1}{rgb}{1,0.498039215686275,0.0549019607843137}

\begin{axis}[
legend cell align={left},
legend columns = 2,
legend style={at={(0.0,0.0)}, anchor=south west, draw=white!80.0!black},
tick align=outside,
tick pos=both,
x grid style={lightgray!92.02614379084967!black},
xlabel={Layer},
xmajorgrids,
width = 0.9\linewidth,
height = \tokenClassifierPlotHeightAppendix,
xmin=0.45, xmax=12.55,
xtick style={color=black},
y grid style={lightgray!92.02614379084967!black},
ylabel={Identifiability Rate},
ymajorgrids,
ymin=0.4, ymax=1.04002459268368,
ytick style={color=black}
]
\addplot [thick, color0]
table {%
1 1
2 1
3 0.983895003127577
4 0.976827502966948
5 0.966528509868508
6 0.953251584362906
7 0.938834620369968
8 0.896512671099973
9 0.815522901101511
10 0.68915871014054
11 0.473397532096209
12 0.587079082232505
};
\addlegendentry{l=1}
\addplot [thick, color1]
table {%
1 0.997835359327088
2 0.998026163714948
3 0.99650522994931
4 0.997828197935477
5 0.997185465816785
6 0.9972688109081
7 0.991930328948876
8 0.979167909127594
9 0.940269430023697
10 0.855129445410211
11 0.765805377961985
12 0.706593187188366
};
\addlegendentry{l=6}
\addplot [thick, color2]
table {%
1 0.940155830381112
2 0.960644620847632
3 0.961584613452446
4 0.965455406583045
5 0.968532365143078
6 0.963228818743349
7 0.956643291482633
8 0.951434443975563
9 0.942840134011846
10 0.918567646389356
11 0.950810652608518
12 0.844885515860979
};
\addlegendentry{l=11}
\addplot [thick, color3]
table {%
1 0.934487222499449
2 0.949651009067318
3 0.951605932620895
4 0.961367770534158
5 0.961920857524043
6 0.957558084972196
7 0.952245011470002
8 0.945250995427592
9 0.935656080577058
10 0.904490642372089
11 0.899541741496457
12 0.93746665754491
};
\addlegendentry{l=12}
\end{axis}

\end{tikzpicture}

%% file: token_classification_experiments/rebuttal_plots/HiddenToHidden/TokenIdentity_HiddenToHidden_withBaselines_10FoldCV.tex
% This file was created by tikzplotlib v0.8.2.
\begin{tikzpicture}

\begin{axis}[
legend cell align={left},
legend columns = 3,
legend style={at={(0,0)},anchor=south west,font=\small, draw=white!80.0!black},
tick align=outside,
tick pos=both,
x grid style={lightgray!92.02614379084967!black},
xlabel={Layer},
xmajorgrids,
width = 0.98\linewidth,
height = \tokenClassifierPlotHeightAppendix,
xmin=1, xmax=12,
xtick style={color=black},
y grid style={lightgray!92.02614379084967!black},
ylabel={Identifiability Rate},
ymajorgrids,
ymin=0.92, ymax=1.01,
ytick style={color=black}
]
\addplot [thick, blue]
table {%
1 1
2 1
3 1
4 1
5 1
6 1
7 1
8 1
9 1
10 1
11 0.996034711063432
12 0.992069979489265
};
\addlegendentry{$\hat{g}_{cos,l}^{MLP}$}
\addplot [thick, blue, dotted]
table {%
1 1
2 1
3 1
4 1
5 1
6 1
7 1
8 1
9 1
10 1
11 0.993643235259517
12 0.991622901818517
};
\addlegendentry{$\hat{g}_{cos,l}^{lin}$}
\addplot [thick, blue, dashed]
table {%
1 1
2 1
3 0.981218063803342
4 0.983565805827924
5 0.998112599548865
6 0.999953965842655
7 1
8 0.989964553698845
9 0.999539658426552
10 0.964553698844543
11 0.946784514109469
12 0.970031763568568
};
\addlegendentry{$\hat{g}_{cos,l}^{naive}$}
\addplot [thick, black]
table {%
1 1
2 1
3 1
4 1
5 1
6 1
7 1
8 1
9 0.999953853253346
10 0.986599342687021
11 0.971882840456355
12 0.962364100282447
};
\addlegendentry{$\hat{g}_{L2,l}^{MLP}$}
\addplot [thick, black, dotted]
table {%
1 1
2 1
3 1
4 1
5 1
6 1
7 1
8 1
9 0.997472085571188
10 0.93836529788174
11 0.943682825894925
12 0.986905746024778
};
\addlegendentry{$\hat{g}_{L2,l}^{lin}$}
\addplot [thick, black, dashed]
table {%
1 1
2 1
3 0.981218063803342
4 0.989872485384155
5 0.995258481793491
6 1
7 1
8 0.999861897527966
9 0.971919164019703
10 0.939050775675551
11 0.941260415228099
12 0.927588270496709
};
\addlegendentry{$\hat{g}_{L2,l}^{naive}$}
\end{axis}

\end{tikzpicture}

%% file: token_classification_experiments/rebuttal_plots/HiddenToHidden/TokenIdentity_HiddenToHidden_mrpcFinetuned_baselines.tex
% This file was created by tikzplotlib v0.8.2.
\begin{tikzpicture}

\begin{axis}[
legend cell align={left},
legend columns = 2,
legend style={at={(0,0)},anchor=south west,font=\small, draw=white!80.0!black},
tick align=outside,
tick pos=both,
x grid style={lightgray!92.02614379084967!black},
xlabel={Layer},
xmajorgrids,
width = 0.98\linewidth,
height = \tokenClassifierPlotHeightAppendix,
xmin=1, xmax=12,
xtick style={color=black},
y grid style={lightgray!92.02614379084967!black},
ylabel={Identifiability Rate},
ymajorgrids,
ymin=0.55, ymax=1.01,
ytick style={color=black}
]
\addplot [thick, blue, dashed]
table {%
1 1
2 1
3 0.981218063803342
4 0.989872485384155
5 0.995258481793491
6 1
7 1
8 0.999861897527966
9 0.971919164019703
10 0.939050775675551
11 0.941260415228099
12 0.927588270496709
};
\addlegendentry{ $\hat{g}_{L2,l}^{naive}$}
\addplot [thick, blue]
table {%
1 1
2 1
3 0.981218063803342
4 0.998480872807623
5 0.996271233255075
6 1
7 1
8 1
9 0.936887170280348
10 0.915205082170971
11 0.934355291626387
12 0.592045297610827
};
\addlegendentry{$\hat{g}_{L2,l}^{naive,fine}$}
\addplot [thick, black, dashed]
table {%
1 1
2 1
3 0.981218063803342
4 0.983565805827924
5 0.998112599548865
6 0.999953965842655
7 1
8 0.989964553698845
9 0.999539658426552
10 0.964553698844543
11 0.946784514109469
12 0.970031763568568
};
\addlegendentry{$\hat{g}_{cos,l}^{naive}$}
\addplot [thick, black]
table {%
1 1
2 1
3 0.981218063803342
4 0.994245730331906
5 0.996501404041799
6 1
7 1
8 1
9 0.980251346499102
10 0.94747502646964
11 0.951894305574736
12 0.949960870966257
};
\addlegendentry{$\hat{g}_{cos,l}^{naive,fine}$}
\end{axis}

\end{tikzpicture}

%% file: token_classification_experiments/rebuttal_plots/HiddenToHidden/TokenIdentity_HiddenToHidden_colaFinetuned_baselines.tex
% This file was created by tikzplotlib v0.8.2.
\begin{tikzpicture}

\begin{axis}[
legend cell align={left},
legend columns = 2,
legend style={at={(0,0)},anchor=south west,font=\small, draw=white!80.0!black},
tick align=outside,
tick pos=both,
x grid style={lightgray!92.02614379084967!black},
xlabel={Layer},
xmajorgrids,
width = 0.98\linewidth,
height = \tokenClassifierPlotHeightAppendix,
xmin=1, xmax=12,
xtick style={color=black},
y grid style={lightgray!92.02614379084967!black},
ylabel={Identifiability Rate},
ymajorgrids,
ymin=0.85, ymax=1.01,
ytick style={color=black}
]
\addplot [thick, blue, dashed]
table {%
1 1
2 1
3 1
4 1
5 1
6 1
7 1
8 1
9 0.99301790701495
10 0.934368325940529
11 0.952932479053721
12 0.983160834565467
};
\addlegendentry{$\hat{g}_{L2,l}^{naive}$}
\addplot [thick, blue]
table {%
1 1
2 1
3 1
4 1
5 1
6 1
7 1
8 1
9 0.990553638902579
10 0.922375554460325
11 0.980614424182684
12 0.857072449482504
};
\addlegendentry{$\hat{g}_{L2,l}^{naive,fine}$}
\addplot [thick, black, dashed]
table {%
1 1
2 1
3 1
4 1
5 1
6 1
7 1
8 1
9 1
10 0.997042878265155
11 0.963528831936915
12 0.999096435025464
};
\addlegendentry{$\hat{g}_{cos,l}^{naive}$}
\addplot [thick, black]
table {%
1 1
2 1
3 1
4 1
5 1
6 1
7 1
8 1
9 1
10 0.996632166913094
11 0.9860358140299
12 0.963939543288976
};
\addlegendentry{$\hat{g}_{cos,l}^{naive,fine}$}
\end{axis}

\end{tikzpicture}

%% file: token_classification_experiments/rebuttal_plots/NeighbouringTokens/TokenIdentity_NeighbouringTokens_linear_cosine_mrpc.tex
% This file was created by tikzplotlib v0.8.2.
\begin{tikzpicture}

\begin{axis}[
legend cell align={left},
legend columns = 2,
legend style={at={(1,1)},anchor=north east,font=\small, draw=white!80.0!black},
tick align=outside,
tick pos=both,
x grid style={lightgray!92.02614379084967!black},
xlabel={Layer},
xmajorgrids,
width = 0.98\linewidth,
height = \tokenClassifierPlotHeightAppendix,
xmin=1, xmax=12,
xtick style={color=black},
y grid style={lightgray!92.02614379084967!black},
ylabel={Identifiability Rate},
ymajorgrids,
ymin=0.12011667668405, ymax=0.953553984353778,
ytick style={color=black}
]
\addplot [thick, red]
table {%
1 0.749420714892566
2 0.82911786967073
3 0.789067299836117
4 0.804065075712412
5 0.730003567903254
6 0.665171690170984
7 0.663801042713597
8 0.651325225394485
9 0.61672435908308
10 0.570979649015782
11 0.511707313915742
12 0.468296700792945
};
\addlegendentry{i-1}
\addplot [thick, red, dashed]
table {%
1 0.447175218717894
2 0.455956816311135
3 0.431231569097005
4 0.415994992603005
5 0.364523681874833
6 0.322919724495431
7 0.313352236414765
8 0.312907748323574
9 0.298965165511135
10 0.273703428632952
11 0.247843659449069
12 0.224936482295616
};
\addlegendentry{i-2}
\addplot [thick, red, dotted]
table {%
1 0.354262923368381
2 0.363981897759164
3 0.342740000113184
4 0.326535293564412
5 0.280838780387552
6 0.234327473242458
7 0.223844046744246
8 0.220437670820723
9 0.210205844103582
10 0.190275132385741
11 0.169239276147726
12 0.158000190669037
};
\addlegendentry{i-3}
\addplot [thick, blue]
table {%
1 0.757504970217499
2 0.823138006514783
3 0.91567047036879
4 0.871305352124488
5 0.818194509550191
6 0.759412241240207
7 0.714546720412003
8 0.67856957868019
9 0.665037141402644
10 0.618374840839739
11 0.559111764784784
12 0.506698942181329
};
\addlegendentry{i+1}
\addplot [thick, blue, dashed]
table {%
1 0.391982323608932
2 0.425294848374149
3 0.501332830548746
4 0.455383865022532
5 0.416730801895916
6 0.385635108552031
7 0.356247761711537
8 0.341188883616221
9 0.334478440557846
10 0.305308732492436
11 0.279694447605413
12 0.255655376775197
};
\addlegendentry{i+2}
\addplot [thick, blue, dotted]
table {%
1 0.345818279883445
2 0.350931989755419
3 0.361347005363682
4 0.335677869415274
5 0.306355214463519
6 0.271887543974883
7 0.25219585324366
8 0.243250456519087
9 0.235255466824295
10 0.217399966152302
11 0.196238439382056
12 0.186057311885104
};
\addlegendentry{i+3}
\end{axis}

\end{tikzpicture}

%% file: token_classification_experiments/rebuttal_plots/NeighbouringTokens/TokenIdentity_NeighbouringTokens_mlp_cosine_mrpc.tex
% This file was created by tikzplotlib v0.8.2.
\begin{tikzpicture}

\begin{axis}[
legend cell align={left},
legend columns = 2,
legend style={at={(1,1)},anchor=north east,font=\small, draw=white!80.0!black},
tick align=outside,
tick pos=both,
x grid style={lightgray!92.02614379084967!black},
xlabel={Layer},
xmajorgrids,
width = 0.98\linewidth,
height = \tokenClassifierPlotHeightAppendix,
xmin=1, xmax=12,
xtick style={color=black},
y grid style={lightgray!92.02614379084967!black},
ylabel={Identifiability Rate},
ymajorgrids,
]
\addplot [thick, red]
table {%
1 0.82089273797055
2 0.86483698847745
3 0.826822201797888
4 0.852506862498168
5 0.800410967866412
6 0.737505545575841
7 0.72995707625704
8 0.713648063652126
9 0.671657032385991
10 0.620678102173813
11 0.564387495450855
12 0.516606595551097
};
\addlegendentry{i-1}
\addplot [thick, red, dashed]
table {%
1 0.490358711076812
2 0.491278459503178
3 0.475256907504059
4 0.47377178078986
5 0.399383771548354
6 0.356426630605579
7 0.344512377506616
8 0.330255796441405
9 0.310888617683114
10 0.281327285342455
11 0.252041865963853
12 0.234007943870549
};
\addlegendentry{i-2}
\addplot [thick, red, dotted]
table {%
1 0.357367648585626
2 0.365827337858884
3 0.354683242882352
4 0.325167747055497
5 0.277084580368631
6 0.234529504692794
7 0.224839571309958
8 0.219689349351451
9 0.208852347645869
10 0.186380927819799
11 0.164526430084846
12 0.156657573566034
};
\addlegendentry{i-3}
\addplot [thick, blue]
table {%
1 0.795203633252317
2 0.850560904095573
3 0.932669053560608
4 0.895983591545306
5 0.864332514947734
6 0.82116073900652
7 0.771963374536978
8 0.729885611375805
9 0.714808068526939
10 0.662934646200566
11 0.60325797115207
12 0.550034530571411
};
\addlegendentry{i+1}
\addplot [thick, blue, dashed]
table {%
1 0.395544027431018
2 0.446447190914716
3 0.534193170372118
4 0.513802953482406
5 0.477916130492508
6 0.42892008777583
7 0.383925026300265
8 0.365794470630508
9 0.349772894572388
10 0.323528954697039
11 0.292217625618425
12 0.267624159139097
};
\addlegendentry{i+2}
\addplot [thick, blue, dotted]
table {%
1 0.342551440389014
2 0.347603208979221
3 0.357787047683944
4 0.338457828901
5 0.302748120610378
6 0.273992017115868
7 0.252440613094018
8 0.243003947582659
9 0.232594026777677
10 0.215138170691954
11 0.195873135982645
12 0.181942837169143
};
\addlegendentry{i+3}
\end{axis}

\end{tikzpicture}

%% file: token_classification_experiments/rebuttal_plots/NeighbouringTokens/TokenIdentity_NeighbouringTokens_mlp_l2_mrpc.tex
% This file was created by tikzplotlib v0.8.2.
\begin{tikzpicture}

\begin{axis}[
legend cell align={left},
legend columns = 2,
legend style={at={(1,1)},anchor=north east,font=\small, draw=white!80.0!black},
tick align=outside,
tick pos=both,
x grid style={lightgray!92.02614379084967!black},
xlabel={Layer},
xmajorgrids,
width = 0.98\linewidth,
height = \tokenClassifierPlotHeightAppendix,
xmin=1, xmax=12,
xtick style={color=black},
y grid style={lightgray!92.02614379084967!black},
ylabel={Identifiability Rate},
ymajorgrids,
]
\addplot [thick, red]
table {%
1 0.54957996179416
2 0.598096242828103
3 0.544919894098698
4 0.56131308849314
5 0.514590488276743
6 0.477020132687473
7 0.479196691280199
8 0.473885060157118
9 0.442761957331226
10 0.404714385584251
11 0.370886845730183
12 0.35286989268055
};
\addlegendentry{i-1}
\addplot [thick, red, dashed]
table {%
1 0.329246729894846
2 0.329156686358019
3 0.311876125927586
4 0.330163426262085
5 0.297517153557894
6 0.273098231550862
7 0.267788959370689
8 0.259389683008475
9 0.242131083322707
10 0.213618422847637
11 0.185680994257344
12 0.174806431590885
};
\addlegendentry{i-2}
\addplot [thick, red, dotted]
table {%
1 0.263697953530828
2 0.270154483844013
3 0.252544049532072
4 0.240695066510544
5 0.215434165011417
6 0.191593132708748
7 0.183412513167431
8 0.178314252479712
9 0.164892534130804
10 0.140700897558065
11 0.124408491008067
12 0.117072942295693
};
\addlegendentry{i-3}
\addplot [thick, blue]
table {%
1 0.517140667952794
2 0.570413619818327
3 0.654483786988117
4 0.614887407940961
5 0.572707608173994
6 0.539023284869309
7 0.504349970849158
8 0.482886878285531
9 0.468142356284828
10 0.412374010865561
11 0.375113426861886
12 0.359157393731397
};
\addlegendentry{i+1}
\addplot [thick, blue, dashed]
table {%
1 0.286058818196361
2 0.304949930066632
3 0.329493410065035
4 0.34069839788478
5 0.32404116170775
6 0.309737079103361
7 0.284118832238188
8 0.273554347257709
9 0.261581805925204
10 0.232229508491585
11 0.211532335272045
12 0.196248970395253
};
\addlegendentry{i+2}
\addplot [thick, blue, dotted]
table {%
1 0.241802334366384
2 0.244115264094209
3 0.248424890670497
4 0.247076462889866
5 0.2302799168215
6 0.208637573242437
7 0.192594204197453
8 0.186092607594404
9 0.176290608229353
10 0.155426864384896
11 0.141829561252293
12 0.135068821402667
};
\addlegendentry{i+3}
\end{axis}

\end{tikzpicture}

%% file: token_classification_experiments/rebuttal_plots/NeighbouringTokens/TokenIdentity_NeighbouringTokens_linear_l2_mrpc.tex
% This file was created by tikzplotlib v0.8.2.
\begin{tikzpicture}

\begin{axis}[
legend cell align={left},
legend columns = 2,
legend style={at={(1,1)},anchor=north east,font=\small, draw=white!80.0!black},
tick align=outside,
tick pos=both,
x grid style={lightgray!92.02614379084967!black},
xlabel={Layer},
xmajorgrids,
width = 0.98\linewidth,
height = \tokenClassifierPlotHeightAppendix,
xmin=1, xmax=12,
xtick style={color=black},
y grid style={lightgray!92.02614379084967!black},
ylabel={Identifiability Rate},
ymajorgrids,
]
\addplot [thick, red]
table {%
1 0.468048950587798
2 0.510209197239808
3 0.474987653970581
4 0.46934492044627
5 0.429058652627025
6 0.401327862007038
7 0.402436993203246
8 0.401685025880114
9 0.383734022828248
10 0.360079007108562
11 0.337354272395788
12 0.312393134543865
};
\addlegendentry{i-1}
\addplot [thick, red, dashed]
table {%
1 0.318967628030679
2 0.319709776711103
3 0.30386513014844
4 0.307832541672014
5 0.278379929907944
6 0.258677351083202
7 0.250172776100415
8 0.238564129052178
9 0.22210881506986
10 0.197106920558923
11 0.18339766672782
12 0.16844865118403
};
\addlegendentry{i-2}
\addplot [thick, red, dotted]
table {%
1 0.271717393051549
2 0.275515347711221
3 0.260328556369979
4 0.245851431428711
5 0.218780730245537
6 0.196784067719869
7 0.18400657087074
8 0.173242478031094
9 0.159504169144152
10 0.141753963916758
11 0.131771641297463
12 0.119308959946452
};
\addlegendentry{i-3}
\addplot [thick, blue]
table {%
1 0.452417008595546
2 0.499705020704613
3 0.562321333474956
4 0.531880606654424
5 0.495585982214638
6 0.467047465273416
7 0.442095729028825
8 0.424674927003696
9 0.407716690026134
10 0.359194089636099
11 0.332063406787739
12 0.308508240602214
};
\addlegendentry{i+1}
\addplot [thick, blue, dashed]
table {%
1 0.283750699911671
2 0.298040158611535
3 0.31902295053039
4 0.318230251874406
5 0.299608902339186
6 0.289142291043813
7 0.271014297139383
8 0.258408082021566
9 0.244544609812178
10 0.223235944415763
11 0.207548793929343
12 0.188238206058859
};
\addlegendentry{i+2}
\addplot [thick, blue, dotted]
table {%
1 0.248621285969511
2 0.251520935938616
3 0.257956416322176
4 0.253237818036712
5 0.233980999976053
6 0.217926378021778
7 0.197280727103812
8 0.187863807521281
9 0.17826630511785
10 0.163506463190848
11 0.151496663572081
12 0.138651729434074
};
\addlegendentry{i+3}
\end{axis}

\end{tikzpicture}

%% file: token_classification_experiments/rebuttal_plots/TokenIdentityColaMNLI/TokenIdentity_COLA.tex
% This file was created by tikzplotlib v0.8.2.
\begin{tikzpicture}

\begin{axis}[
legend cell align={left},
legend columns = 2,
legend style={at={(0,0)},anchor=south west,font=\small, draw=white!80.0!black},
tick align=outside,
tick pos=both,
x grid style={lightgray!92.02614379084967!black},
xlabel={Layer},
xmajorgrids,
width = 0.98\linewidth,
height = \tokenClassifierPlotHeightAppendix,
xmin=1, xmax=12,
xtick style={color=black},
y grid style={lightgray!92.02614379084967!black},
ylabel={Identifiability Rate},
ymajorgrids,
ymin=0.1, ymax=1.1,
ytick style={color=black}
]
\addplot [thick, blue]
table {%
1 0.999901185770751
2 0.999901185770751
3 0.999901185770751
4 0.999901185770751
5 0.999443536523989
6 0.998367382541182
7 0.995847731560149
8 0.993978862940988
9 0.989565388154948
10 0.980874489183246
11 0.971393450932618
12 0.961917111901124
};
\addlegendentry{$\hat{g}_{cos,l}^{MLP}$}
\addplot [thick, blue, dotted]
table {%
1 1
2 0.999901185770751
3 0.999901185770751
4 0.999767645940756
5 0.999382156186846
6 0.998649532324851
7 0.997701881115472
8 0.995706908214643
9 0.992807341909972
10 0.985032163858188
11 0.972368373614907
12 0.959763365058494
};
\addlegendentry{$\hat{g}_{cos,l}^{lin}$}
\addplot [blue, dashed]
table {%
1 1
2 0.999917857729588
3 0.999917857729588
4 0.994907179234434
5 0.89847215377033
6 0.790783637259734
7 0.706834236898308
8 0.623049121077707
9 0.620666995235748
10 0.383686545096106
11 0.22400197141449
12 0.442582552981764
};
\addlegendentry{$\hat{g}_{cos,l}^{naive}$}
\addplot [thick, black]
table {%
1 0.999901185770751
2 0.999901185770751
3 0.999809778092506
4 0.99926116378199
5 0.993046930511971
6 0.983713050155884
7 0.970273027855238
8 0.957625555539562
9 0.94435972475868
10 0.914769514665827
11 0.892233893860816
12 0.882365474240175
};
\addlegendentry{$\hat{g}_{L2,l}^{MLP}$}
\addplot [thick, black, dotted]
table {%
1 1
2 0.999901185770751
3 0.999901185770751
4 0.999710963863257
5 0.997633039498523
6 0.98982670000917
7 0.976139097798949
8 0.961533949969192
9 0.936583085729674
10 0.896146527433197
11 0.867792982963033
12 0.831582954594228
};
\addlegendentry{$\hat{g}_{L2,l}^{lin}$}
\addplot [black, dashed]
table {%
1 1
2 0.999899188467161
3 0.979988910731388
4 0.970764655476587
5 0.899591713292001
6 0.800141136145975
7 0.686324915570341
8 0.474167044709915
9 0.352437118806391
10 0.204042542466858
11 0.0994505771460255
12 0.149402691667927
};
\addlegendentry{$\hat{g}_{L2}^{naive}$}
\end{axis}

\end{tikzpicture}

%% file: token_classification_experiments/rebuttal_plots/TokenIdentityColaMNLI/TokenIdentity_MNLI_first500.tex
% This file was created by tikzplotlib v0.8.2.
\begin{tikzpicture}

\begin{axis}[
legend cell align={left},
legend columns = 2,
legend style={at={(0,0)},anchor=south west,font=\small, draw=white!80.0!black},
tick align=outside,
tick pos=both,
x grid style={lightgray!92.02614379084967!black},
xlabel={Layer},
xmajorgrids,
width = 0.98\linewidth,
height = \tokenClassifierPlotHeightAppendix,
xmin=1, xmax=12,
xtick style={color=black},
y grid style={lightgray!92.02614379084967!black},
ylabel={Identifiability Rate},
ymajorgrids,
ymin=0.1, ymax=1.1,
ytick style={color=black}
]
\addplot [thick, blue]
table {%
1 1
2 0.99994246260069
3 0.999850438450437
4 0.999408336708769
5 0.997911878141191
6 0.996118459361915
7 0.991055250544392
8 0.984284813066999
9 0.978309752063085
10 0.962687823352391
11 0.936154620164094
12 0.921345850545891
};
\addlegendentry{$\hat{g}_{cos,l}^{MLP}$}
\addplot [thick, blue, dotted]
table {%
1 1
2 0.99994246260069
3 0.9998413243509
4 0.99929458219883
5 0.998197962170393
6 0.994187456063352
7 0.990470418884223
8 0.986840459744
9 0.98025038771989
10 0.968858480734107
11 0.941171633458241
12 0.923890557688756
};
\addlegendentry{$\hat{g}_{cos,l}^{lin}$}
\addplot [blue, dashed]
table {%
1 1
2 0.999899188467161
3 0.976359695549171
4 0.970058974746711
5 0.906799737890015
6 0.822672513735571
7 0.729976309289783
8 0.572760723826806
9 0.512828267553808
10 0.319421341801502
11 0.154392862543475
12 0.392308080044357
};
\addlegendentry{$\hat{g}_{cos,l}^{naive}$}
\addplot [thick, black]
table {%
1 1
2 0.99994246260069
3 0.999755358743493
4 0.998171558304823
5 0.991152242541912
6 0.97713641301107
7 0.964999545342675
8 0.945712662509151
9 0.923089113297289
10 0.869079953897168
11 0.817107137751199
12 0.809260810802371
};
\addlegendentry{$\hat{g}_{L2,l}^{MLP}$}
\addplot [thick, black, dotted]
table {%
1 1
2 0.99994246260069
3 0.998855504876055
4 0.985924281385521
5 0.972945582551521
6 0.956892270853708
7 0.930157932463903
8 0.905485682507961
9 0.880842116241672
10 0.784128224637834
11 0.738166881919882
12 0.697228342397452
};
\addlegendentry{$\hat{g}_{L2,l}^{lin}$}
\addplot [black, dashed]
table {%
1 1
2 0.999899188467161
3 0.979988910731388
4 0.970764655476587
5 0.899591713292001
6 0.800141136145975
7 0.686324915570341
8 0.474167044709915
9 0.352437118806391
10 0.204042542466858
11 0.0994505771460255
12 0.149402691667927
};
\addlegendentry{$\hat{g}_{L2}^{naive}$}
\end{axis}

\end{tikzpicture}

%% file: gradient_notfinetuned_task_indep_figs/cola/contribution_all_task_independent_not_fine_tuned_cola.tex
% This file was created by tikzplotlib v0.8.2.
\begin{tikzpicture}

\definecolor{color0}{rgb}{1,0.498039215686275,0.0549019607843137}

\begin{axis}[
tick align=outside,
tick pos=both,
x grid style={white!69.01960784313725!black},
width=0.97\linewidth,
xlabel={Layer},
xmin=0.5, xmax=12.5,
height=\localKernelPlotsHeight,
xtick style={color=black},
xtick={1,2,3,4,5,6,7,8,9,10,11,12},
xticklabels={1,2,3,4,5,6,7,8,9,10,11,12},
y grid style={white!69.01960784313725!black},
ylabel={Contribution [\%]},
ytick={0,0.1,0.2,0.3, 0.4},
yticklabels={0,10, 20, 30, 40},
ymin=-0.00358753791078925, ymax=0.491626131813973,
ytick style={color=black},
xmajorgrids,
ymajorgrids
]

\addplot [black]
table {%
0.75 0.328322269022465
1.25 0.328322269022465
1.25 0.384652525186539
0.75 0.384652525186539
0.75 0.328322269022465
};
\addplot [black]
table {%
1 0.328322269022465
1 0.246842369437218
};
\addplot [black]
table {%
1 0.384652525186539
1 0.469116419553757
};
\addplot [black]
table {%
0.875 0.246842369437218
1.125 0.246842369437218
};
\addplot [black]
table {%
0.875 0.469116419553757
1.125 0.469116419553757
};
\addplot [black]
table {%
1.75 0.276858799159527
2.25 0.276858799159527
2.25 0.345082484185696
1.75 0.345082484185696
1.75 0.276858799159527
};
\addplot [black]
table {%
2 0.276858799159527
2 0.176074028015137
};
\addplot [black]
table {%
2 0.345082484185696
2 0.447359621524811
};
\addplot [black]
table {%
1.875 0.176074028015137
2.125 0.176074028015137
};
\addplot [black]
table {%
1.875 0.447359621524811
2.125 0.447359621524811
};
\addplot [black]
table {%
2.75 0.250459901988506
3.25 0.250459901988506
3.25 0.31762333959341
2.75 0.31762333959341
2.75 0.250459901988506
};
\addplot [black]
table {%
3 0.250459901988506
3 0.158121079206467
};
\addplot [black]
table {%
3 0.31762333959341
3 0.418242484331131
};
\addplot [black]
table {%
2.875 0.158121079206467
3.125 0.158121079206467
};
\addplot [black]
table {%
2.875 0.418242484331131
3.125 0.418242484331131
};
\addplot [black]
table {%
3.75 0.210307821631432
4.25 0.210307821631432
4.25 0.292475275695324
3.75 0.292475275695324
3.75 0.210307821631432
};
\addplot [black]
table {%
4 0.210307821631432
4 0.12497403472662
};
\addplot [black]
table {%
4 0.292475275695324
4 0.415652215480804
};
\addplot [black]
table {%
3.875 0.12497403472662
4.125 0.12497403472662
};
\addplot [black]
table {%
3.875 0.415652215480804
4.125 0.415652215480804
};
\addplot [black]
table {%
4.75 0.173801179975271
5.25 0.173801179975271
5.25 0.27562191337347
4.75 0.27562191337347
4.75 0.173801179975271
};
\addplot [black]
table {%
5 0.173801179975271
5 0.087195985019207
};
\addplot [black]
table {%
5 0.27562191337347
5 0.428209811449051
};
\addplot [black]
table {%
4.875 0.087195985019207
5.125 0.087195985019207
};
\addplot [black]
table {%
4.875 0.428209811449051
5.125 0.428209811449051
};
\addplot [black]
table {%
5.75 0.15407919511199
6.25 0.15407919511199
6.25 0.266919821500778
5.75 0.266919821500778
5.75 0.15407919511199
};
\addplot [black]
table {%
6 0.15407919511199
6 0.0655222237110138
};
\addplot [black]
table {%
6 0.266919821500778
6 0.4354048371315
};
\addplot [black]
table {%
5.875 0.0655222237110138
6.125 0.0655222237110138
};
\addplot [black]
table {%
5.875 0.4354048371315
6.125 0.4354048371315
};
\addplot [black]
table {%
6.75 0.150410428643227
7.25 0.150410428643227
7.25 0.26117105782032
6.75 0.26117105782032
6.75 0.150410428643227
};
\addplot [black]
table {%
7 0.150410428643227
7 0.067877434194088
};
\addplot [black]
table {%
7 0.26117105782032
7 0.427009493112564
};
\addplot [black]
table {%
6.875 0.067877434194088
7.125 0.067877434194088
};
\addplot [black]
table {%
6.875 0.427009493112564
7.125 0.427009493112564
};
\addplot [black]
table {%
7.75 0.132021028548479
8.25 0.132021028548479
8.25 0.255040004849434
7.75 0.255040004849434
7.75 0.132021028548479
};
\addplot [black]
table {%
8 0.132021028548479
8 0.0439329147338867
};
\addplot [black]
table {%
8 0.255040004849434
8 0.436995476484299
};
\addplot [black]
table {%
7.875 0.0439329147338867
8.125 0.0439329147338867
};
\addplot [black]
table {%
7.875 0.436995476484299
8.125 0.436995476484299
};
\addplot [black]
table {%
8.75 0.124202156439424
9.25 0.124202156439424
9.25 0.236375458538532
8.75 0.236375458538532
8.75 0.124202156439424
};
\addplot [black]
table {%
9 0.124202156439424
9 0.024015886709094
};
\addplot [black]
table {%
9 0.236375458538532
9 0.403824537992477
};
\addplot [black]
table {%
8.875 0.024015886709094
9.125 0.024015886709094
};
\addplot [black]
table {%
8.875 0.403824537992477
9.125 0.403824537992477
};
\addplot [black]
table {%
9.75 0.102202247828245
10.25 0.102202247828245
10.25 0.215688802301884
9.75 0.215688802301884
9.75 0.102202247828245
};
\addplot [black]
table {%
10 0.102202247828245
10 0.0238478910177946
};
\addplot [black]
table {%
10 0.215688802301884
10 0.384771466255188
};
\addplot [black]
table {%
9.875 0.0238478910177946
10.125 0.0238478910177946
};
\addplot [black]
table {%
9.875 0.384771466255188
10.125 0.384771466255188
};
\addplot [black]
table {%
10.75 0.101832427084446
11.25 0.101832427084446
11.25 0.209498669952154
10.75 0.209498669952154
10.75 0.101832427084446
};
\addplot [black]
table {%
11 0.101832427084446
11 0.0192743595689535
};
\addplot [black]
table {%
11 0.209498669952154
11 0.370822846889496
};
\addplot [black]
table {%
10.875 0.0192743595689535
11.125 0.0192743595689535
};
\addplot [black]
table {%
10.875 0.370822846889496
11.125 0.370822846889496
};
\addplot [black]
table {%
11.75 0.0941250789910555
12.25 0.0941250789910555
12.25 0.201523218303919
11.75 0.201523218303919
11.75 0.0941250789910555
};
\addplot [black]
table {%
12 0.0941250789910555
12 0.0189221743494272
};
\addplot [black]
table {%
12 0.201523218303919
12 0.362053841352463
};
\addplot [black]
table {%
11.875 0.0189221743494272
12.125 0.0189221743494272
};
\addplot [black]
table {%
11.875 0.362053841352463
12.125 0.362053841352463
};
\addplot [color0]
table {%
0.75 0.351834774017334
1.25 0.351834774017334
};
\addplot [color0]
table {%
1.75 0.309709250926971
2.25 0.309709250926971
};
\addplot [color0]
table {%
2.75 0.282665252685547
3.25 0.282665252685547
};
\addplot [color0]
table {%
3.75 0.248570322990417
4.25 0.248570322990417
};
\addplot [color0]
table {%
4.75 0.222661659121513
5.25 0.222661659121513
};
\addplot [color0]
table {%
5.75 0.204540282487869
6.25 0.204540282487869
};
\addplot [color0]
table {%
6.75 0.2020663022995
7.25 0.2020663022995
};
\addplot [color0]
table {%
7.75 0.18624434620142
8.25 0.18624434620142
};
\addplot [color0]
table {%
8.75 0.16958649456501
9.25 0.16958649456501
};
\addplot [color0]
table {%
9.75 0.15021301060915
10.25 0.15021301060915
};
\addplot [color0]
table {%
10.75 0.146259561181068
11.25 0.146259561181068
};
\addplot [color0]
table {%
11.75 0.137370854616165
12.25 0.137370854616165
};
\end{axis}

\end{tikzpicture}

%% file: gradient_notfinetuned_task_indep_figs/cola/rank_percent_task_independent_not_fine_tuned_cola.tex
% This file was created by tikzplotlib v0.8.2.
\begin{tikzpicture}

\definecolor{color0}{rgb}{0.12156862745098,0.466666666666667,0.705882352941177}

\begin{axis}[
tick align=outside,
tick pos=both,
x grid style={white!69.01960784313725!black},
xlabel={Layer},
height=\localKernelPlotsHeight,
width=0.97\linewidth,
xmin=-0.55, xmax=11.55,
xmajorgrids,
ymajorgrids,
xtick style={color=black},
xtick={0,1,2,3,4,5,6,7,8,9,10,11},
xticklabels={1,2,3,4,5,6,7,8,9,10,11,12},
y grid style={white!69.01960784313725!black},
ylabel={$\tilde{P}$ [\%]},
ymin=-0.0268523081977986, ymax=0.56389847215377,
ytick style={color=black},
ytick={0,0.1,0.2,0.3, 0.4, 0.5},
yticklabels={0,10, 20, 30, 40, 50}
]

\addplot [semithick, color0]
table {%
0 0
1 0.000410711352061771
2 0.007885657959586
3 0.102759980285855
4 0.179645145391819
5 0.261540988992936
6 0.227780515853458
7 0.322079842286841
8 0.356743880400854
9 0.47034664038114
10 0.477739444718252
11 0.537046163955972
};
\end{axis}

\end{tikzpicture}

%% file: gradient_notfinetuned_task_indep_figs/cola/relative_attr_per_layer.tex
% This file was created by tikzplotlib v0.8.2.
\begin{tikzpicture}

\definecolor{color0}{rgb}{0.12156862745098,0.466666666666667,0.705882352941177}
\definecolor{color1}{rgb}{1,0.498039215686275,0.0549019607843137}
\definecolor{color2}{rgb}{0.172549019607843,0.627450980392157,0.172549019607843}
\definecolor{color3}{rgb}{0.83921568627451,0.152941176470588,0.156862745098039}
\definecolor{color4}{rgb}{0.580392156862745,0.403921568627451,0.741176470588235}
\definecolor{color5}{rgb}{0.549019607843137,0.337254901960784,0.294117647058824}

\begin{axis}[
legend cell align={left},
legend columns = 2,
legend style={nodes={scale=0.6, transform shape},at={(0.5,0.01)}, anchor=south, draw=white!80.0!black},
width=0.97\linewidth,
height=\localKernelPlotsHeight,
tick align=outside,
tick pos=both,
ytick={0.06,0.08,0.1},
yticklabels={6,8,10},
x grid style={white!69.01960784313725!black},
xlabel={Layer},
xmin=-0.55, xmax=11.55,
xtick style={color=black},
xtick={0,1,2,3,4,5,6,7,8,9,10,11},
xticklabels={1,2,3,4,5,6,7,8,9,10,11,12},
y grid style={white!69.01960784313725!black},
ylabel={Rel. Contribution (\%)},
ymin=0.0492504438753602, ymax=0.104405610403338,
ytick style={color=black}
]

\addplot [very thick, color0]
table {%
0 0.0801405804430973
1 0.0839433395893612
2 0.0848374904937844
3 0.0857585302701932
4 0.0897124725349415
5 0.0850162141056881
6 0.0835158032222734
7 0.0849558853240764
8 0.0809557749115977
9 0.0817121890650777
10 0.0806920811352574
11 0.0787596389046518
};
\addlegendentry{1st}
\addplot [thin, color1]
table {%
0 0.0714438369895718
1 0.0681362913034536
2 0.0696718020725581
3 0.0764889762009918
4 0.0821417041522472
5 0.0851912169927058
6 0.0873217865105729
7 0.0890261057864144
8 0.090654553778389
9 0.093285298546456
10 0.0935086394011562
11 0.0931297882654833
};
\addlegendentry{2nd}
\addplot [thin, color2]
table {%
0 0.0630966330020989
1 0.0650441715609311
2 0.0683742418960019
3 0.0767282540023819
4 0.0826667533168565
5 0.0863569104494002
6 0.0885075029529869
7 0.0899716538231315
8 0.0920665924553173
9 0.0956029631427644
10 0.0960001662098855
11 0.0955841571882441
};
\addlegendentry{3rd}
\addplot [thin, color3]
table {%
0 0.0623702656254782
1 0.0691363872374885
2 0.0719864780894666
3 0.0760742325294512
4 0.0813731995845675
5 0.0843578068752509
6 0.0858997047723026
7 0.088005090696498
8 0.091483440227243
9 0.0949124450036601
10 0.0961986847966134
11 0.0982022645619799
};
\addlegendentry{4th and 5th}
\addplot [thin, color4]
table {%
0 0.0649652922247762
1 0.076182353169894
2 0.0798623587771428
3 0.0839582230246444
4 0.0835540394281137
5 0.0849843582121638
6 0.0834493117671693
7 0.0845050595111523
8 0.0863620201118708
9 0.0880834749857512
10 0.0898769411919785
11 0.0942165675953431
};
\addlegendentry{6th to 10th}
\addplot [very thick, color5]
table {%
0 0.0653056344423206
1 0.0803328749195577
2 0.0852306929427966
3 0.0897831244407688
4 0.0841776257451333
5 0.0853881494858688
6 0.0825458779045403
7 0.0830345771656921
8 0.0838409509259161
9 0.08354629266068
10 0.0854882278916779
11 0.0913259714750476
};
\addlegendentry{11th onwards}
\end{axis}

\end{tikzpicture}

%% file: gradient_notfinetuned_task_indep_figs/mnlim/contribution_all_task_independent_not_fine_tuned_mnlim.tex
% This file was created by tikzplotlib v0.8.2.
\begin{tikzpicture}

\definecolor{color0}{rgb}{1,0.498039215686275,0.0549019607843137}

\begin{axis}[
tick align=outside,
tick pos=both,
x grid style={white!69.01960784313725!black},
width=0.97\linewidth,
xlabel={Layer},
xmin=0.5, xmax=12.5,
height=\localKernelPlotsHeight,
xtick style={color=black},
xtick={1,2,3,4,5,6,7,8,9,10,11,12},
xticklabels={1,2,3,4,5,6,7,8,9,10,11,12},
y grid style={white!69.01960784313725!black},
ylabel={Contribution [\%]},
ytick={0,0.1,0.2,0.3, 0.4},
yticklabels={0,10, 20, 30, 40},
ymin=-0.0169826120720245, ymax=0.420615658734459,
ytick style={color=black},
xmajorgrids,
ymajorgrids
]

\addplot [black]
table {%
0.75 0.29502260684967
1.25 0.29502260684967
1.25 0.337305873632431
0.75 0.337305873632431
0.75 0.29502260684967
};
\addplot [black]
table {%
1 0.29502260684967
1 0.231600388884544
};
\addplot [black]
table {%
1 0.337305873632431
1 0.400724828243256
};
\addplot [black]
table {%
0.875 0.231600388884544
1.125 0.231600388884544
};
\addplot [black]
table {%
0.875 0.400724828243256
1.125 0.400724828243256
};
\addplot [black]
table {%
1.75 0.225383698940277
2.25 0.225383698940277
2.25 0.265414714813232
1.75 0.265414714813232
1.75 0.225383698940277
};
\addplot [black]
table {%
2 0.225383698940277
2 0.16548877954483
};
\addplot [black]
table {%
2 0.265414714813232
2 0.325446367263794
};
\addplot [black]
table {%
1.875 0.16548877954483
2.125 0.16548877954483
};
\addplot [black]
table {%
1.875 0.325446367263794
2.125 0.325446367263794
};
\addplot [black]
table {%
2.75 0.205920144915581
3.25 0.205920144915581
3.25 0.248141705989838
2.75 0.248141705989838
2.75 0.205920144915581
};
\addplot [black]
table {%
3 0.205920144915581
3 0.142747700214386
};
\addplot [black]
table {%
3 0.248141705989838
3 0.311435908079147
};
\addplot [black]
table {%
2.875 0.142747700214386
3.125 0.142747700214386
};
\addplot [black]
table {%
2.875 0.311435908079147
3.125 0.311435908079147
};
\addplot [black]
table {%
3.75 0.175932869315147
4.25 0.175932869315147
4.25 0.229718267917633
3.75 0.229718267917633
3.75 0.175932869315147
};
\addplot [black]
table {%
4 0.175932869315147
4 0.0998229160904884
};
\addplot [black]
table {%
4 0.229718267917633
4 0.310394018888474
};
\addplot [black]
table {%
3.875 0.0998229160904884
4.125 0.0998229160904884
};
\addplot [black]
table {%
3.875 0.310394018888474
4.125 0.310394018888474
};
\addplot [black]
table {%
4.75 0.137898862361908
5.25 0.137898862361908
5.25 0.208798184990883
4.75 0.208798184990883
4.75 0.137898862361908
};
\addplot [black]
table {%
5 0.137898862361908
5 0.0472134687006474
};
\addplot [black]
table {%
5 0.208798184990883
5 0.315112054347992
};
\addplot [black]
table {%
4.875 0.0472134687006474
5.125 0.0472134687006474
};
\addplot [black]
table {%
4.875 0.315112054347992
5.125 0.315112054347992
};
\addplot [black]
table {%
5.75 0.116684101521969
6.25 0.116684101521969
6.25 0.201047226786613
5.75 0.201047226786613
5.75 0.116684101521969
};
\addplot [black]
table {%
6 0.116684101521969
6 0.0385954231023788
};
\addplot [black]
table {%
6 0.201047226786613
6 0.327525466680527
};
\addplot [black]
table {%
5.875 0.0385954231023788
6.125 0.0385954231023788
};
\addplot [black]
table {%
5.875 0.327525466680527
6.125 0.327525466680527
};
\addplot [black]
table {%
6.75 0.108344323933125
7.25 0.108344323933125
7.25 0.190691217780113
6.75 0.190691217780113
6.75 0.108344323933125
};
\addplot [black]
table {%
7 0.108344323933125
7 0.0194594077765942
};
\addplot [black]
table {%
7 0.190691217780113
7 0.314193099737167
};
\addplot [black]
table {%
6.875 0.0194594077765942
7.125 0.0194594077765942
};
\addplot [black]
table {%
6.875 0.314193099737167
7.125 0.314193099737167
};
\addplot [black]
table {%
7.75 0.101253017783165
8.25 0.101253017783165
8.25 0.18143954873085
7.75 0.18143954873085
7.75 0.101253017783165
};
\addplot [black]
table {%
8 0.101253017783165
8 0.0285023786127567
};
\addplot [black]
table {%
8 0.18143954873085
8 0.301558792591095
};
\addplot [black]
table {%
7.875 0.0285023786127567
8.125 0.0285023786127567
};
\addplot [black]
table {%
7.875 0.301558792591095
8.125 0.301558792591095
};
\addplot [black]
table {%
8.75 0.0905241742730141
9.25 0.0905241742730141
9.25 0.168301939964294
8.75 0.168301939964294
8.75 0.0905241742730141
};
\addplot [black]
table {%
9 0.0905241742730141
9 0.0137981474399567
};
\addplot [black]
table {%
9 0.168301939964294
9 0.284955024719238
};
\addplot [black]
table {%
8.875 0.0137981474399567
9.125 0.0137981474399567
};
\addplot [black]
table {%
8.875 0.284955024719238
9.125 0.284955024719238
};
\addplot [black]
table {%
9.75 0.0730758756399155
10.25 0.0730758756399155
10.25 0.15289543569088
9.75 0.15289543569088
9.75 0.0730758756399155
};
\addplot [black]
table {%
10 0.0730758756399155
10 0.011562118306756
};
\addplot [black]
table {%
10 0.15289543569088
10 0.272557944059372
};
\addplot [black]
table {%
9.875 0.011562118306756
10.125 0.011562118306756
};
\addplot [black]
table {%
9.875 0.272557944059372
10.125 0.272557944059372
};
\addplot [black]
table {%
10.75 0.067252442240715
11.25 0.067252442240715
11.25 0.145602002739906
10.75 0.145602002739906
10.75 0.067252442240715
};
\addplot [black]
table {%
11 0.067252442240715
11 0.00589778833091259
};
\addplot [black]
table {%
11 0.145602002739906
11 0.263112097978592
};
\addplot [black]
table {%
10.875 0.00589778833091259
11.125 0.00589778833091259
};
\addplot [black]
table {%
10.875 0.263112097978592
11.125 0.263112097978592
};
\addplot [black]
table {%
11.75 0.0669852271676064
12.25 0.0669852271676064
12.25 0.144763737916946
11.75 0.144763737916946
11.75 0.0669852271676064
};
\addplot [black]
table {%
12 0.0669852271676064
12 0.00290821841917932
};
\addplot [black]
table {%
12 0.144763737916946
12 0.261410027742386
};
\addplot [black]
table {%
11.875 0.00290821841917932
12.125 0.00290821841917932
};
\addplot [black]
table {%
11.875 0.261410027742386
12.125 0.261410027742386
};
\addplot [color0]
table {%
0.75 0.316027969121933
1.25 0.316027969121933
};
\addplot [color0]
table {%
1.75 0.243075706064701
2.25 0.243075706064701
};
\addplot [color0]
table {%
2.75 0.22443750500679
3.25 0.22443750500679
};
\addplot [color0]
table {%
3.75 0.200894691050053
4.25 0.200894691050053
};
\addplot [color0]
table {%
4.75 0.170779451727867
5.25 0.170779451727867
};
\addplot [color0]
table {%
5.75 0.153328418731689
6.25 0.153328418731689
};
\addplot [color0]
table {%
6.75 0.144253276288509
7.25 0.144253276288509
};
\addplot [color0]
table {%
7.75 0.1357576623559
8.25 0.1357576623559
};
\addplot [color0]
table {%
8.75 0.124656230211258
9.25 0.124656230211258
};
\addplot [color0]
table {%
9.75 0.107691720128059
10.25 0.107691720128059
};
\addplot [color0]
table {%
10.75 0.0992663949728012
11.25 0.0992663949728012
};
\addplot [color0]
table {%
11.75 0.098657101392746
12.25 0.098657101392746
};
\end{axis}

\end{tikzpicture}

%% file: gradient_notfinetuned_task_indep_figs/mnlim/rank_percent_task_independent_not_fine_tuned_mnlim.tex
% This file was created by tikzplotlib v0.8.2.
\begin{tikzpicture}

\definecolor{color0}{rgb}{0.12156862745098,0.466666666666667,0.705882352941177}

\begin{axis}[
tick align=outside,
tick pos=both,
x grid style={white!69.01960784313725!black},
xlabel={Layer},
height=\localKernelPlotsHeight,
width=0.97\linewidth,
xmin=-0.55, xmax=11.55,
xmajorgrids,
ymajorgrids,
xtick style={color=black},
xtick={0,1,2,3,4,5,6,7,8,9,10,11},
xticklabels={1,2,3,4,5,6,7,8,9,10,11,12},
y grid style={white!69.01960784313725!black},
ylabel={$\tilde{P}$ [\%]},
ymin=-0.0199558872560334, ymax=0.419073632376702,
ytick style={color=black},
ytick style={color=black},
ytick={0,0.1,0.2,0.3, 0.4, 0.5},
yticklabels={0,10, 20, 30, 40, 50}
]

\addplot [semithick, color0]
table {%
0 0
1 0
2 0.00091046651791439
3 0.0437280398040573
4 0.103998358595573
5 0.174668513246647
6 0.174989100048729
7 0.219281372624452
8 0.248211125644379
9 0.352683952707035
10 0.377612782436973
11 0.399117745120669
};
\end{axis}

\end{tikzpicture}

%% file: gradient_notfinetuned_task_indep_figs/mnlim/relative_attr_per_layer.tex
% This file was created by tikzplotlib v0.8.2.
\begin{tikzpicture}

\definecolor{color0}{rgb}{0.12156862745098,0.466666666666667,0.705882352941177}
\definecolor{color1}{rgb}{1,0.498039215686275,0.0549019607843137}
\definecolor{color2}{rgb}{0.172549019607843,0.627450980392157,0.172549019607843}
\definecolor{color3}{rgb}{0.83921568627451,0.152941176470588,0.156862745098039}
\definecolor{color4}{rgb}{0.580392156862745,0.403921568627451,0.741176470588235}
\definecolor{color5}{rgb}{0.549019607843137,0.337254901960784,0.294117647058824}

\begin{axis}[
legend cell align={left},
legend columns = 2,
legend style={nodes={scale=0.6, transform shape},at={(0.5,0.01)}, anchor=south, draw=white!80.0!black},
width=0.97\linewidth,
height=\localKernelPlotsHeight,
tick align=outside,
tick pos=both,
ytick={0.06,0.08,0.1},
yticklabels={6,8,10},
x grid style={white!69.01960784313725!black},
xlabel={Layer},
xmin=-0.55, xmax=11.55,
xtick style={color=black},
xtick={0,1,2,3,4,5,6,7,8,9,10,11},
xticklabels={1,2,3,4,5,6,7,8,9,10,11,12},
y grid style={white!69.01960784313725!black},
ylabel={Rel. Contribution (\%)},
ymin=0.0529186648154596, ymax=0.102502793129767,
ytick style={color=black}
]

\addplot [very thick, color0]
table {%
0 0.0955119590785627
1 0.0911305652581356
2 0.0988387261054926
3 0.0946911365905828
4 0.0889816478634918
5 0.08547976722627
6 0.0833277528154862
7 0.0801111513985495
8 0.0756164875303088
9 0.0711158504422466
10 0.0676862215718627
11 0.0675087341190105
};
\addlegendentry{1st}
\addplot [semithick, color1]
table {%
0 0.0877611050018598
1 0.0782171962944239
2 0.0790518525116084
3 0.085147434517921
4 0.0843329766795687
5 0.0869012998840355
6 0.0870507090209714
7 0.0865289412321198
8 0.0842807974942437
9 0.0818604955275057
10 0.0792428788969625
11 0.0796243129387795
};
\addlegendentry{2nd}
\addplot [thin, color2]
table {%
0 0.0763710580358492
1 0.0731026883082837
2 0.0767119070552739
3 0.0840636403470449
4 0.0847695853835445
5 0.0881943068131491
6 0.0887644495730987
7 0.0887640515012462
8 0.0872811571547688
9 0.0858075603052433
10 0.0830748706774602
11 0.0830947248450374
};
\addlegendentry{3rd}
\addplot [thin, color3]
table {%
0 0.0693843271027206
1 0.0726747271289787
2 0.0734729359270733
3 0.0771133166035665
4 0.081023415955823
5 0.0856440487504336
6 0.0876565852399406
7 0.0894389265886062
8 0.0911788558242383
9 0.0920879276403967
10 0.0900772482201559
11 0.0902476850180666
};
\addlegendentry{4th and 5th}
\addplot [thin, color4]
table {%
0 0.0654747211479307
1 0.0746392536515034
2 0.0740401553331677
3 0.0751778460461516
4 0.0803673456865859
5 0.0834103282389505
6 0.0858345233575264
7 0.0881625419636541
8 0.0913667501835525
9 0.0937248746970056
10 0.0940052543406565
11 0.0937964053533152
};
\addlegendentry{6th to 10th}
\addplot [very thick, color5]
table {%
0 0.0551724888297463
1 0.0732353014026736
2 0.0747223026966636
3 0.0779198303795237
4 0.0830695036912072
5 0.0834553330926831
6 0.0840026343935399
7 0.086277714585994
8 0.089370801579515
9 0.093890010975741
10 0.0986351092572328
11 0.10024896911548
};
\addlegendentry{11th onwards}
\end{axis}

\end{tikzpicture}